2024-06

# MASTERING THE DIGITAL ART OF WAR: DEVELOPING INTELLIGENT COMBAT SIMULATION AGENTS FOR WARGAMING USING HIERARCHICAL REINFORCEMENT LEARNING

Black, Scotty E.

Monterey, CA; Naval Postgraduate School

# NAVAL POSTGRADUATE SCHOOL

**MONTEREY, CALIFORNIA**

# DISSERTATION

**MASTERING THE DIGITAL ART OF WAR: DEVELOPING INTELLIGENT COMBAT SIMULATION AGENTS FOR WARGAMING USING HIERARCHICAL REINFORCEMENT LEARNING**

by

Scotty E. Black

June 2024

Dissertation Supervisor: Christian J. Darken

THIS PAGE INTENTIONALLY LEFT BLANK

| **REPORT DOCUMENTATION PAGE** | | | *Form Approved OMB No. 0704-0188* |
|---|---|---|---|

| **1. AGENCY USE ONLY** ***(Leave blank)*** | **2. REPORT DATE** June 2024 | **3. REPORT TYPE AND DATES COVERED** Dissertation |
|---|---|---|

**4. TITLE AND SUBTITLE**
MASTERING THE DIGITAL ART OF WAR: DEVELOPING INTELLIGENT COMBAT SIMULATION AGENTS FOR WARGAMING USING HIERARCHICAL REINFORCEMENT LEARNING

**5. FUNDING NUMBERS**

**6. AUTHOR(S)** Scotty E. Black

**7. PERFORMING ORGANIZATION NAME(S) AND ADDRESS(ES)**
Naval Postgraduate School
Monterey, CA 93943-5000

**8. PERFORMING ORGANIZATION REPORT NUMBER**

**9. SPONSORING / MONITORING AGENCY NAME(S) AND ADDRESS(ES)**
N/A

**10. SPONSORING / MONITORING AGENCY REPORT NUMBER**

**11. SUPPLEMENTARY NOTES** The views expressed in this thesis are those of the author and do not reflect the official policy or position of the Department of Defense or the U.S. Government.

**12a. DISTRIBUTION / AVAILABILITY STATEMENT**
Distribution Statement A. Approved for public release: Distribution is unlimited.

**12b. DISTRIBUTION CODE**
A

**13. ABSTRACT (maximum 200 words)**

In today's rapidly evolving military landscape, advancing artificial intelligence (AI) in support of wargaming becomes essential. Despite reinforcement learning (RL) showing promise for developing intelligent agents, conventional RL faces limitations in handling the complexity inherent in combat simulations. This dissertation proposes a comprehensive approach, including targeted observation abstractions, multi-model integration, a hybrid AI framework, and an overarching hierarchical reinforcement learning (HRL) framework. Our localized observation abstraction using piecewise linear spatial decay simplifies the RL problem, enhancing computational efficiency and demonstrating superior efficacy over traditional global observation methods. Our multi-model framework combines various AI methodologies, optimizing performance while still enabling the use of diverse, specialized individual behavior models. Our hybrid AI framework synergizes RL with scripted agents, leveraging RL for high-level decisions and scripted agents for lower-level tasks, enhancing adaptability, reliability, and performance. Our HRL architecture and training framework decomposes complex problems into manageable subproblems, aligning with military decision-making structures. Although initial tests did not show improved performance, insights were gained to improve future iterations. This study underscores AI's potential to revolutionize wargaming, emphasizing the need for continued research in this domain.

**14. SUBJECT TERMS**
reinforcement learning, RL, hierarchical reinforcement learning, HRL, artificial intelligence, AI, combat simulation, abstraction, intelligent agents

**15. NUMBER OF PAGES**
259

**16. PRICE CODE**

| **17. SECURITY CLASSIFICATION OF REPORT** | **18. SECURITY CLASSIFICATION OF THIS PAGE** | **19. SECURITY CLASSIFICATION OF ABSTRACT** | **20. LIMITATION OF ABSTRACT** |
|---|---|---|---|
| Unclassified | Unclassified | Unclassified | UU |

NSN 7540-01-280-5500

Standard Form 298 (Rev. 2-89)
Prescribed by ANSI Std. 239-18

THIS PAGE INTENTIONALLY LEFT BLANK

**Distribution Statement A. Approved for public release: Distribution is unlimited.**

**MASTERING THE DIGITAL ART OF WAR: DEVELOPING INTELLIGENT COMBAT SIMULATION AGENTS FOR WARGAMING USING HIERARCHICAL REINFORCEMENT LEARNING**

Scotty E. Black
Lieutenant Colonel, United States Marine Corps
BSE, Missouri University of Science and Technology, 2005
MS, Modeling, Virtual Environments, and Simulation, Naval Postgraduate School, 2015

Submitted in partial fulfillment of the
requirements for the degree of

**DOCTOR OF PHILOSOPHY IN MODELING, VIRTUAL ENVIRONMENTS, AND SIMULATION**

from the

**NAVAL POSTGRADUATE SCHOOL**
**June 2024**

Approved by:
Christian J. Darken
Department of
Computer Science
Dissertation Supervisor
Dissertation Chair

Rudolph P. Darken
Department of
Computer Science

Brian Wade
TRADOC Analysis
Center Monterey

Imre L. Balogh
MOVES Institute

Douglas J. MacKinnon
Department of
Information Sciences

Approved by:
Charles P. Rowan
Chair, MOVES Institute

James B. Michael
Vice Provost of Academic Affairs

THIS PAGE INTENTIONALLY LEFT BLANK

# ABSTRACT

In today's rapidly evolving military landscape, advancing artificial intelligence (AI) in support of wargaming becomes essential. Despite reinforcement learning (RL) showing promise for developing intelligent agents, conventional RL faces limitations in handling the complexity inherent in combat simulations. This dissertation proposes a comprehensive approach, including targeted observation abstractions, multi-model integration, a hybrid AI framework, and an overarching hierarchical reinforcement learning (HRL) framework. Our localized observation abstraction using piecewise linear spatial decay simplifies the RL problem, enhancing computational efficiency and demonstrating superior efficacy over traditional global observation methods. Our multi-model framework combines various AI methodologies, optimizing performance while still enabling the use of diverse, specialized individual behavior models. Our hybrid AI framework synergizes RL with scripted agents, leveraging RL for high-level decisions and scripted agents for lower-level tasks, enhancing adaptability, reliability, and performance. Our HRL architecture and training framework decomposes complex problems into manageable subproblems, aligning with military decision-making structures. Although initial tests did not show improved performance, insights were gained to improve future iterations. This study underscores AI's potential to revolutionize wargaming, emphasizing the need for continued research in this domain.

THIS PAGE INTENTIONALLY LEFT BLANK

# Table of Contents

# List of Figures

# List of Tables

THIS PAGE INTENTIONALLY LEFT BLANK

# List of Acronyms and Abbreviations

**A2C** Advantage Actor-Critic

**A3C** Asynchronous Advantage Actor-Critic

**ADT** AlphaDogfight Trials

**AI** artificial intelligence

**BFM** basic fighter maneuvers

**BT** behavior trees

**CETaS** Centre for Emerging Technology and Security

**CMANO** Command: Modern Air Naval Operations

**CNN** convolutional neural network

**COA** course of action

**DADS** Dynamics-Aware Discovery of Skills

**DARPA** Defense Advanced Research Projects Agency

**DDPG** Deep Deterministic Policy Gradient

**DES** Discrete Event Simulation

**DIB** Defense Industrial Base

**DOD** Department of Defense

**DQN** Deep Q-Networks

**DRL** deep reinforcement learning

**DSB** Defense Science Board

**DSTL** Defence Science and Technology Laboratory

**DVAE** Dreaming Variational Autoencoder

**EDS** Ensemble Decision Systems

**FFRDC** Federally Funded Research and Development Center

**FMH** feudal multi-agent hierarchy

**FSM** finite state machine

**FuN** Feudal Network

**GCHRL** goal-conditioned hierarchical reinforcement learning

**GEMS** Gaming, Exercising, Modeling, and Simulation

**GVGAI** General Video Game AI

**GVGP** General Video Game Playing

**HAC** Hierarchical Actor-Critic

**HER** Hindsight Experience Replay

**HESS** Hierarchical Exploration approach with Stable Subgoal representation learning

**HIRO** HIerarchical Reinforcement learning with Off-policy correction

**HPC** high performance computer

**HRL** hierarchical reinforcement learning

**ICM** Intrinsic Curiosity Module

**KOG** King of Glory

**LESSON** LEarns the Subgoal representation with SlOw dyNamics

**LSTM** Long Short-Term Memory

| | |
|---|---|
| **MADDPG** | Multi-Agent Deep Deterministic Policy Gradient |
| **MAE** | Mean Absolute Error |
| **MCTS** | Monte Carlo tree search |
| **MDP** | Markov Decision Process |
| **ML** | machine learning |
| **MLP** | multi-layer perceptron |
| **MLSH** | metalearning shared hierarchies |
| **MOBA** | Multiplayer Online Battle Arena |
| **MoE** | Mixture of Experts |
| **MSE** | Mean Squared Error |
| **NATO** | North Atlantic Treaty Organization |
| **NPS** | Naval Postgraduate School |
| **NSCAI** | National Security Commission on Artificial Intelligence |
| **PE** | Professional Edition |
| **PHANG-MAN** | Policy Hierarchy for Adaptive Novel Generation of MANeuvers |
| **PLA** | People's Liberation Army |
| **PME** | Professional Military Education |
| **POMDP** | Partially Observable Markov Decision Process |
| **PPO** | Proximal Policy Optimization |
| **PRC** | People's Republic of China |
| **ReLU** | Rectified Linear Unit |
| **RL** | reinforcement learning |
| **RNN** | recurrent neural network |
| **RTS** | real-time strategy |
| **SAC** | Soft Actor Critic |
| **SB3** | Stable-Baselines 3 |
| **SEM** | Standard Error of the Mean |
| **SMDP** | Semi-Markov Decision Process |
| **SNN** | stochastic neural network |
| **SR** | Successor Representation |
| **STRAW** | STRategic Attentive Writer |
| **UARC** | University Affiliated Research Center |
| **UCAV** | unmanned combat aerial vehicle |
| **UCB** | Upper Confidence Bounds |
| **UK** | United Kingdom |

# Acknowledgments

I would like to express my deepest gratitude to my advisor, Chris Darken, for his invaluable guidance, mentorship, support, and encouragement throughout the course of my PhD journey. Your insights and feedback have been instrumental in shaping my research and helping me navigate the complexities of my work and the field of artificial intelligence at large. Without your trust and confidence in my abilities, I would not have been able to accomplish as much as I did. Thank you for believing in me and providing the support I needed to grow as a researcher.

I extend my sincere thanks to my committee members, Brian Wade, Rudy Darken, Imre Balogh, and Doug MacKinnon, for their mentorship, support, and invaluable feedback throughout the course of my research. Your encouragement, expertise, and guidance have been key to advancing my work.

I am grateful to my colleagues and friends, John Fischer, Charlie Rowan, and Charlie Timm, for their camaraderie, discussions, and support. I am also grateful to the many faculty and staff at NPS and at the MOVES Institute who have enabled my work along the way.

I also wish to acknowledge the travel funding provided by the Marine Corps Modeling and Simulation Office, the Office of Naval Research, and the Naval Warfare Studies Institute. Your resources have allowed me to participate in conferences, symposiums, and technical meetings across the world, which has significantly broadened my perspective, enhanced my research, and provided invaluable opportunities for networking and collaboration.

To my family, Jess, Eli, and Ellie, I could never thank you enough for your unwavering support, patience, and belief in me despite my many absences and long work days. Your love and encouragement have been my anchor throughout this rewarding journey. I am proud of you! You are the reason why anything I do is possible.

THIS PAGE INTENTIONALLY LEFT BLANK

# CHAPTER 1:
# Introduction

> War is the realm of uncertainty; three quarters of the factors on which action in war is based are wrapped in a fog of greater or lesser uncertainty. A sensitive and discriminating judgment is called for; a skilled intelligence to scent out the truth.
>
> —Carl von Clausewitz [1]

Portions from the core of this chapter have been published in the *Meeting Proceedings of the NATO Modeling and Simulation Group Symposium (NMSG) Simulation: Going Beyond the Limitations of the Real World* [2] and the *Proceedings of the 2023 Interservice/Industry Training, Simulation, and Education Conference (I/ITSEC)* [3]. These publications hold no copyrights over the presented material.

This dissertation focuses on scaling artificial intelligence (AI) through hierarchical reinforcement learning (HRL) to develop intelligent agents for combat simulations within a wargaming context. While existing AI systems have demonstrated success in simpler scenarios, they often fall short in the more complex environments typical of wargaming. To address this challenge, our research proposes advancements to current AI and RL methodologies. This includes the development of novel observation abstractions, the application of a distinct multi-model approach, and the employment of hybrid AI systems, all under a new HRL framework. These developments are aimed at improving both AI scalability and agent efficacy, particularly as we continue to increase complexity in our combat simulation environments. Our near-term goal is to enhance agent intelligence and adaptability that will ultimately contribute valuable insights and tools for wargaming applications that can aid in education, training, analysis, operational planning, and decision-making. Our ultimate vision is to harness the capabilities of AI to enhance the quality and accelerate the speed of our decision-making processes within the evolving complexities of the modern battlefield.

While there exist many different approaches to developing intelligent agents for combat simulations and commercial games, this development has predominantly been characterized by rule-based, scripted methodologies, with deep reinforcement learning (RL) approaches only fairly recently coming into play [4]. To date, *Scripted* methodologies—a term we use in this dissertation to generally refer to strategies governed by predefined sets of rules and behaviors—have been instrumental in creating effective, predictable, and logical agents for simulation and gaming environments. While these scripted methods have provided a foundational framework for agent behavior, their limitations—such as lack of adaptability, inability to handle unforeseen situations, and difficulty in scaling to complex scenarios—have become increasingly more apparent as we begin shifting to more complex and evolving combat scenario. This necessitates a transition towards more intelligent, flexible, and adaptive approaches that can provide more useful insights in military training, analysis, and strategic planning contexts.

The need for advanced methodologies, such as RL, has thus become clear, demonstrating the potential for improving adaptability and performance in agents through learning-based mechanisms. This recent shift towards machine learning (ML) marks a significant step forward in the development of intelligent agents—moving from rigid, rule-based frameworks to more dynamic, learning-oriented models. These models may be better suited to navigate the complexities and uncertainties inherent in modern simulations and gaming environments.

RL offers a framework for agents to continually learn and refine their behaviors over time by learning from past experiences, generalizing from these experiences, and adapting to changing conditions within the simulation. Despite its potential, however, applying RL in large combat simulations or complex games has introduced its own unique set of challenges. These include managing the complexity of such environments, overcoming learning inefficiencies in expansive state spaces, and ensuring that learned behaviors are generalizable across different scenarios or conditions. These challenges highlight the balance between the innovative potential of RL and the practical considerations required for its effective implementation in complex settings.

Thus, our research proposes a new approach using a scalable HRL framework to develop intelligent agents for combat simulations. These agents are designed and trained to manage and respond to the intricate dynamics of combat scenarios, exceeding the capabilities of previous models and approaches in this domain. Scalability, in this context, refers to the ability of our framework to effectively handle increasing levels of complexity and larger-scale scenarios without a loss of performance or the need for significant re-engineering. By improving scalability, this research advances the field of AI in gaming and military applications and provides a framework for developing intelligent agents specifically for use in combat simulations in support of wargaming.

## 1.1 Motivation

Despite significant advancements in other domains of AI, the application of advanced AI methodologies—particularly HRL—within the realm of combat simulations is still a relatively unexplored area. However, its potential impact cannot be overstated, especially considering the evolving landscape of modern combat and the need for better and faster decisions from our commanders and decision-makers. This research is predicated on the belief that leveraging HRL to enhance intelligent agents in wargaming can significantly contribute to future combat readiness. The ability to create more intelligent and adaptable agents within simulations will enable improved decision-making capabilities for military commanders, decision-makers, and planners alike. Given the growing complexities and rapidly evolving pace of modern warfare, developing and refining these AI-driven simulation tools is not just an academic endeavor but a critical component of national security and defense strategy.

### 1.1.1 Wargaming

While there does not exist a universally agreed upon definition of wargaming within the wargaming community, *Joint Publication 5-0 Joint Planning* [5] defines wargames as "representations of conflict or competition in a synthetic environment in which people make decisions and respond to the consequences of those decisions." Other definitions include McHugh's definition in *U.S. Navy Fundamentals of Wargaming* [6], "a simulation, in accordance with predetermined rules, data, and procedures, of selected aspects of a conflict situation." In a RAND study conducted for the U.S. Marine Corps, [7], Wong et al. stated that a "wargame involves human players or actors making decisions in an artificial contest environment and then living with the consequences of their actions." We concur with Caffrey [8] who stated, "[j]ust as in the old story of the blind men describing an elephant, definitions of wargaming tend to reflect who is doing the defining." Nevertheless, we think Perla's definition of wargame in *The Art of Wargaming* [9] encapsulates the various definitions in a comprehensive yet nuanced manner with, “a warfare model or simulation whose operation does not involve the activities of actual military forces, and whose sequence of events affects and is, in turn, affected by the decisions made by the players representing the opposing sides” [9]. Additionally, Perla [10] defines *wargaming* (as opposed to *wargame*) as “[a]n applied discipline encompassing the creation, use, synthesis and analysis of wargames to conduct research, explore concepts, develop and test hypotheses, and dynamically communicate insights to inform, educate, and train individuals and organizations.”

While wargames cannot predict nor fully replicate real-world scenarios, they provide a unique benefit not achievable without actual combat: insights into how we may make decisions in war [11]. Wargaming provides players with a unique opportunity to tackle challenges in a manner that is often hard for non-wargamers to replicate, resulting in significant and enduring educational benefits over time [12]. The variability in tactics, strategy, and combat outcomes throughout the game generates unpredictable threats and opportunities, compelling participants to grapple with classic, real-world dilemmas and make decisions accordingly [12]. Wargames provide a valuable platform for preparing decision-makers to navigate the complexities and uncertainties of the rapidly changing global dynamics we currently face [13]. Moreover, as Perla and McGrady point out in *Why Wargaming Works* [13], to maximize the benefits of wargames, we need to incorporate them with other resources such as military exercises, operational analysis, historical insights, and real-world experiences. This integration aids in forming clearer strategies and decisions for both current and future scenarios. Ultimately, as Perla highlights, “wargames can help us learn important things about uncertainty” [10]. They not only help distinguish between our true understanding and our assumptions, but also uncover insights about what might be missing and, potentially, reveal complete gaps in our understanding [10].

Given these diverse interpretations of wargames, as well as the broad applications for wargaming, it becomes apparent that the practice of wargaming can involve such disparate activities ranging from table-top games on a gameboard with physical pieces to completely simulated games within an elaborate virtual environment in the format of a video game [14]. Of note, although the term *simulation* itself may or may not involve a computer system, we use the term *simulation* in our research to refer specifically to *computer simulations*. While decisions in wargaming have typically been made by humans, we contend that, for some wargaming applications, decisions can be made by either a human or a comparable intelligent, adaptive computer AI and still meet the spirit and intent of wargaming.

### 1.1.2 The Need to Modernize Wargaming

Unfortunately, despite its centuries-long history [9], contemporary wargaming persists with outdated tools and techniques, reflecting a lack of evolution in the field. While traditional analog tools should undoubtedly continue to play a role, there is an increasing need to update wargaming practices for the 21st Century, incorporating modern advancements such as AI, "to evolve the current paradigm of wargaming—both in terms of technology and methodology" [7], [15]. The advancements in emerging technologies today hold the potential to enable breakthroughs that go beyond merely using machines as decision aids at strategic, operational, and tactical levels. They could also enable the automation of even the most intricate tasks and decision-making processes previously deemed beyond the capabilities of a machine.

In the last decade, the U.S. Department of Defense (DOD) has observed an increasing trend in its guidance, policies, and reports emphasizing the need to continue maintaining, advancing, and potentially reshaping our use of wargames and wargaming in support of Professional Military Education (PME), concept development, capability analysis, and force design [16]–[23]. Although former Deputy Secretary of Defense, Mr. Robert Work [22] advocated for a revitalization of wargaming throughout the DOD in 2015, minimal advancement, evolution, or transformation has occurred in how the various services have conducted or utilized wargames. In the 38th Commandant's Planning Guidance [17], General David Berger emphasized the urgent need for a revitalized approach to wargaming. He stressed the importance of significantly enhancing investments in wargaming, experimentation, and modeling and simulation (M&S) to foster improved thinking, innovation, and change [17]. Additionally, General Berger pointed out that despite considerable efforts over the past two decades to develop new concepts, there has been minimal effort to rigorously test these ideas through wargaming, analysis, and, ultimately, experimentation [17].

In 2022, a report issued by the Office of the Undersecretary of Defense for Research and Engineering (OUSD(R&E)) titled *Technology Vision for an Era of Competition* [24] called for the DOD to foster early research and uncover novel scientific advancements so as to prevent any technological surprises. Furthermore, this report underscored that the current era of strategic competition necessitates collaborative efforts between the Government, academic institutions, coalition partners, Federally Funded Research and Development Centers (FFRDCs), University Affiliated Research Centers (UARCs), the Defense Industrial Base (DIB), and small businesses. It highlighted that successful competition depends on the ability to quickly initiate new technology development, conduct rapid experimentation in relevant mission environments, and quickly transition technologies to our warfighters [24]. Lastly, the report emphasized the need to adopt technologies already existent in the commercial sector in areas such as AI, advanced computing, autonomy, and human-machine interfaces [24].

While the modernization of wargaming has long been acknowledged as necessary, a pivotal aspect of this transformation lies in the adoption and integration of technologies and scientific advancements from the commercial sector. The Final Report of the Defense Science Board (DSB) Task Force on Gaming, Exercising, Modeling, and Simulation (GEMS) [25] emphasized how improvements in emerging technologies have significantly enhanced GEMS capabilities—providing the DOD with economical and novel methods to explore new concepts, design and prototype future systems, simulate military campaigns, explore geopolitical issues, and deliver training that boosts warfighter performance and readiness. The DSB further highlighted the crucial role of harnessing GEMS technologies for the DOD to effectively tackle the emerging challenges in military training, advanced systems development, rapid acquisition, effective deterrence, and decisive warfighting [25]. Specifically, the report calls for the integration of digital engineering and scalable, adaptive GEMS tools across the DOD to ensure the effective and efficient development of new capabilities. Our research into scalable HRL frameworks directly addresses this need by offering a robust, flexible, and adaptive approach that can evolve with the complexity of modern combat scenarios, thereby enhancing GEMS capabilities in support of DOD requirements.

Given the increasing complexity of modern warfare, driven by the unprecedented amount of information available to our warfighters and the need for ever-increasingly more rapid decision-making, the imperative for wargaming modernization becomes clear. This urgency is reinforced by DOD and Service-level policies and guidance, showcasing a recognition of these evolving challenges. The ability of our forces to continue to be prepared for the future fight relies heavily on our commitment and capability to embed advanced technologies and methodologies into our wargaming process to ultimately improve and inform our warfighter's decision-making.

### 1.1.3 The People's Republic of China Wargaming Modernization

Although the DOD has been somewhat slow in significantly advancing our wargaming activities through the application of emerging technologies, our adversaries have quickly adapted and capitalized on this opportunity. They have utilized these advancements to reshape how they employ wargaming, ranging from enhancing military planning to refining command and control, and even improving professional military educationPME.

The history of wargaming within the People's Republic of China (PRC) as part of military operations is a longstanding tradition [26] with roots back to Sun Tzu [27]. This reliance on wargaming has continued in that it has allowed the PRC to bridge the experience gap given the PLA's self-acknowledged relative lack of recent real-world warfighting experience as compared to its global competitors [26]. Furthermore, as the PLA embarked on training and educating their officers to adapt to modern technology and the escalating complexity of modern warfare, they recognized that traditional classroom discussions were no longer adequate [27].

This led to the development of computerized wargaming and other M&S tools, a process that started as early as the 1990s [27]. Initially, these tools were met with skepticism by PLA leadership and were perceived as cumbersome and poorly coordinated [27]. However, in 2007, the People's Liberation Army (PLA) revitalized their initiatives to develop enhanced digital wargaming systems, aiming to expose and familiarize their commanders with decision-making in complex and dynamic conditions [27].

In this modern era, the PLA recognizes that the traditional focus on processes and procedures is quickly becoming obsolete. They have come to understand the importance of expediting the decision-making cycle to keep pace with the "informationized" battlefield, which penalizes delays and frequently demands immediate responses to evolving circumstances [27]. Ultimately, the PLA views digital wargames as fostering greater tactical flexibility and more comprehensive situational awareness, while also enabling their senior decision-makers to "think more completely, more precisely, more deeply, which will produce more effective levels of command stratagem" [27]. It should be of no surprise then that the PLA is now attempting to elevate the realism in wargames to help capture both the uncertainty and the time pressures present in modern combat [27].

In fact, unlike the U.S., the PLA has already begun to prioritize computerized approaches to wargaming over our more traditional forms [14]. As highlighted in the DSB's report [28] regarding our own imperative to utilize cost-effective training techniques, the PLA has already started employing similar methods in their training programs to tackle some of their recognized weaknesses, such as command decision-making. Additionally, they have expanded their wargaming initiatives at their PME institutions [14], [29].

As a result, wargaming has gained increasing popularity and prominence throughout the PRC [14]. The PLA has seized upon the commercialization of wargaming, as well as innovations and advancements from the modern gaming industry, to enhance the realism and quality of its games [14]. Furthermore, the PRC has relied on its national strategy for military-civil fusion to partner the PLA with high-tech companies to advance wargaming and military simulations in general [14]. Furthermore, wargaming has even transcended the PLA's PME system and has evolved into a pivotal component of their national defense education. Each year, thousands of participants consisting of both military and civilian students from universities across the nation engage in these annual wargaming competitions [14], [29].

Demonstrating the significance they attribute to this domain, the PLA is actively pursuing innovation in both the tools and methodologies employed in wargaming [14]. One such innovation is the employment of AI for wargaming [14], [26], [30], supporting the PLA's "intelligentization" priority. The PLA regarded the achievements of Google's AlphaGo in defeating world-champion Go Master Lee Sedol, and AI's triumph in Texas Hold 'em poker, as evidence that AI may indeed be relevant to the domain of wargaming [14]. One particular example of their advancing capabilities is an AI-driven wargaming simulation called *AlphaWar*, inspired by DeepMind's AlphaStar AI system that created a superhuman AI player for StarCraft II [26].

Additionally, the PLA is leveraging wargaming platforms to accelerate technological experimentation [14]. Recent progress includes PLA competitions that have focused on the advancement of AI systems for wargaming [14]. The PLA is leveraging this human-machine confrontation to find ways to improve military planning and develop decision-aid tools for future combat operations [14], [26], [29]. One such example of an AI-assisted system developed by the PRC is the *Mozi Joint Operations Deduction System* that they have begun using in training and education [14]. *Mozi* is a human-in-the-loop deduction system tailored to multi-domain operations that supports tactical to campaign-level joint operational scenarios. This platform aids in the planning of everything from combat organization to command and control [14]. In fact, this system drew inspiration from and aimed to emulate *Command: Modern Air Naval Operations (CMANO)*, a commercially available wargaming platform currently being used by the U.S. and numerous NATO allies [14]. Relative to previous PLA wargames, *Mozi* enables substantial enhancements to their planning process while also serving as a platform capable of developing and training intelligent agents [14].

These developments indicate a pivotal shift in the focus of our adversaries towards improving decision-making through advancements and integration of AI in wargaming. Particularly alarming is the fact that the PLA's investment in AI has now reached levels comparable to the U.S. DOD's spending in this same domain [31]. As the strategic landscape continues to shift, the need for the DOD to modernize wargaming becomes not merely an option but a critical requirement—essential for maintaining a competitive edge against our adversaries and ensuring warfighter decision readiness in the face of the emerging new, complex, and unknown challenges.

### 1.1.4 Leveraging AI for Wargaming

Considering these alarming activities and the recent advancements in AI's transformative potential, it becomes crucial that we vigorously pursue research and experimentation to establish a comprehensive grasp of AI's capabilities and limitations, as well as its applications in planning and wargaming. Only through this can the DOD enhance its readiness and adaptability to address strategic surprises and disruptions [32].

The National Security Commission on Artificial Intelligence (NSCAI) [33] emphasizes two key points: (1) "the rapidly improving ability of computer systems to solve problems and to perform tasks that would otherwise require human intelligence—and in some instances exceed human performance—is world altering;" and (2) "AI is expanding the window of vulnerability the United States has already entered." Thus, given these, the NSCAI asserts that "the United States must act now to field AI systems and invest substantially more resources in AI innovation to protect its security, promote prosperity, and safeguard the future of democracy" [33]. By developing new AI-enabled systems and integrating these with human decision-making and judgment, the NSCAI [33] contends that we will be able to enhance awareness across all domains, accelerate the speed of our decision cycles, improve the quality of our decisions, offer recommendations for different course of actions (COAs), and more quickly respond to our adversaries.

In a report funded by the United Kingdom (UK) Defence Science and Technology Laboratory (DSTL), the Centre for Emerging Technology and Security (CETaS) conducted a study, *Artificial Intelligence in Wargaming: An Evidence-Based Assessment of AI applications* [34], which involved a literature survey, case study analyses, expert interviews, and a workshop that brought together experts from both the defense and game AI communities. In their study, Knack and Rosamund highlight that a deeper understanding and proper use of AI could lead to a "potential revolutionary change in wargaming" [34]. The authors also highlight that, by investing in AI-enabling technologies, we could achieve decision superiority over our adversaries by introducing innovative data analytics techniques for decision-makers, while also facilitating the analysis of these decisions. They compiled a list of use cases gathered from a group of professional wargame designers, M&S experts, and non-defense AI specialists. These use cases covered various aspects of wargame design, execution, analysis, and logistics. The areas identified included those that could benefit from existing AI solutions currently used in other applications—such as automatic speech transcription—as well as more innovative, high-risk, high-reward AI solutions to support combat adjudication and COA generation. [34]. The report emphasized the need for AI systems to be trustworthy, explainable, and scalable, thus ensuring that AI-driven decisions can be understood and trusted by human operators using realistic wargaming scenarios, which is essential for effective integration and adoption of AI in wargaming [34].

One particularly notable area where AI could revolutionize wargaming is in COA development and comparison. COA analysis today mostly emphasizes assessing friendly plans with very little examination of how a thinking adversary might react based on their own objectives and capabilities [30]. Nevertheless, even if we increased our efforts to understand our adversary's thinking and predict their actions during a conflict, we would still be constrained by the limits of our own imagination. Thomas Schelling put it best in his *Impossibility Theorem*: "One thing a person cannot do, no matter how rigorous his analysis or heroic imagination, is to draw up a list of things that would not occur to him" [35].

Arguably, our collective experiences, training, and educational paradigms within the military can create a uniformity of thought that, while cohesive, may also limit our tactical and strategic creativity. This conformity is typically the result of deeply ingrained methodologies, developed and reinforced over years or even decades of service. However, the introduction of AI-enabled wargaming presents a potential opportunity to break free from this constraint. By employing intelligent agents that are unbound by our traditional military frameworks, we can challenge our conventional thought processes and possibly devise and explore new tactics and strategies that may have been once beyond our imagination.

Nevertheless, while substantial effort has gone into advancing this domain of intelligent agents, military planning and wargaming are markedly different compared to the conventional problems AI has typically addressed, such as image classification and natural language processing. To date, fortunately, the gaming domain has served as a good test-bed for investigating how to implement AI in support of wargaming. Early successes have included mastering checkers [36], backgammon [37], chess [38], and Go [39]. AI methods have also achieved success in other games such as Atari games [40] Super Mario Bros [41], Quake III [42], Dota 2 [43], StarCraft II [44], and No-Limit Texas Hold 'Em Poker [45]. Nevertheless, competitive games typically feature a set of fixed rules, defined parameters, consistent initial conditions, and outcomes that can be predicted based on known variables. While these games provide important insights into AI development in this domain, real-world wargaming scenarios are often more complex, featuring varying initial game states and significantly larger branching factors, which lead to more unpredictable outcomes. This complexity makes it challenging to translate AI successes in these games to military wargaming.

Thus, despite the achievements in applying AI to competitive games, further research is necessary to tackle the complexities inherent in military wargaming and simulations. The nuances of real-world strategic and operational scenarios demand innovations in AI that can scale effectively to meet these challenges.

### 1.1.5 The "Centaur" Concept for Wargaming

As detailed in the CeTAS report [34], there are numerous ways AI can be utilized to support wargaming. However, in this dissertation, we focus on the application of AI within the context of creating intelligent agents that are capable of making rational decisions amidst the large and complex state spaces typical of modern combat modeling and simulation M&S.

Creating an AI capable of winning games or surpassing human performance, however, is just the start of showing that AI can offer meaningful insights to wargamers, operational planners, and military leaders [46]. Nevertheless, we see these intelligent agents as the basis for developing modern decision-aids and support tools that can offer decision-makers greater accuracy, speed, and agility compared to traditional tools [32]. We believe that overlooking this step carries substantial risks as we advance into multi-domain operations [30] against an AI-equipped adversary.

The concept of human-computer collaboration, also referred to in the literature as human-machine teaming, was initially conceived by Licklider in 1960 [47], but it was former world champion chess player Garry Kasparov [48] who introduced the concept of "Centaur Chess"—where a human collaborates with a computer during play—following his defeat by IBM's Deep Blue in 1997. Despite his loss to an AI, Kasparov promoted the idea of viewing AI not as a threat, but as a tool that, when integrated with human capabilities, could achieve extraordinary results [48].

In his book *Deep Thinking: Where Machine Intelligence Ends and Human Creativity Begins* [48], Kasparov emphasized the importance of leveraging the distinct strengths of humans and machines. He noted that while computers are adept at brute-force calculations, analyzing millions of positions per second and calculating the best short-term tactical moves, humans bring a deeper strategic understanding, creativity, and the ability to discern long-term consequences of moves, primarily through intuition [48]. Kasparov argued that the combination of human intuition with machine computation often resulted in stronger chess play than either top grandmasters or computers could achieve alone. He observed that in many cases, even lower-ranked players assisted by computers could surpass top-tier grandmasters [48].

Kasparov [48] also discussed how the role of the human in this "centaur" partnership has evolved as chess AIs have improved. Initially, humans would focus on strategy while computers handled tactics. However, as chess AIs advanced, humans increasingly took on a "quality control" role, ensuring that the computer's suggested moves were in line with broader strategic objectives [48]. He speculated that the future of chess might not hinge on humans versus machines, but rather on which human-machine teams, using which interfaces, can perform the best. This collaboration, blending the machine's computational power with the human's ability to provide context, understanding, and intuition, has led to levels of play that surpass what either could achieve on their own [48].

Ultimately, we contend that developing intelligent agents is a foundational enabler to being able to fully leverage AI for wargaming, whether as an adversary force, intelligent teammates, tactical advisors, COA generators, COA analyzers, COA exploiters, future force design, combat adjudication, scenario planning, or simply to gain insights into potential outcomes. While scripted agents have proven useful thus far and will continue to be useful, the complexity and unpredictability of modern warfare require a new level of adaptability and learning capability that only ML can provide. By incorporating super-intelligent agents into combat simulations, we believe that wargaming can finally evolve from static and predictable to dynamic and insightful—mirroring the uncertainties of real-world operations.

## 1.2 Research Objective

This dissertation aims to extend the research into using RL to develop intelligent behaviors in combat simulations and enable the scaling of AI to deal with larger and more complex scenarios than has previously been possible, given reasonable computational and training budgets. Building on previous studies, we focus on an HRL framework to address the challenges of long-horizon tasks and complex decision-making environments in combat models. This approach is expected to enhance the scalability and applicability of RL by decomposing large problems into manageable sub-tasks, aligning with military decision-making structures. Our goal is to improve training efficiency and agent efficacy within the constraints of limited computational budgets.

In this dissertation, we use the term AI to refer to an agent's ability to think and act *rationally*, as opposed to *humanly*, as described by Russell and Norvig's [49]. Figure 1.1 depicts various definitions of AI, however, we focus on creating intelligent agents that align to the right side of this chart. Although we recognize the importance of thinking and acting human and we borrow concepts from the cognitive domain, the scope of this dissertation is to design agents that seek to maximize an *ideal* performance metric rather than an imitating *human* performance metric.

| **Thinking Humanly** | **Thinking Rationally** |
|---|---|
| "The exciting new effort to make computers think ... machines with minds, in the full and literal sense." (Haugeland, 1985)<br><br>"[The automation of] activities that we associate with human thinking, activities such as decision-making, problem solving, learning ..." (Bellman, 1978) | "The study of mental faculties through the use of computational models." (Charniak and McDermott, 1985)<br><br>"The study of the computations that make it possible to perceive, reason, and act." (Winston, 1992) |
| **Acting Humanly**<br><br>"The art of creating machines that perform functions that require intelligence when performed by people." (Kurzweil, 1990)<br><br>"The study of how to make computers do things at which, at the moment, people are better." (Rich and Knight, 1991) | **Acting Rationally**<br><br>"AI . . . is concerned with intelligent behavior in artifacts." (Nilsson, 1998) |

Figure 1.1. Definitions of Artificial Intelligence. Selected definitions of artificial intelligence, sorted into four groups. Source: [49].

To accomplish this objective, as part of developing a new HRL framework, we develop and apply a combination of action abstractions, temporal abstractions, and spatial abstractions that reduce the complexity inherent in large-scale combat simulations. Action abstractions simplify the decision-making process by condensing multiple actions into broader, more manageable categories. Temporal abstractions extend this concept by allowing higher-level agents to make decisions at a slower pace than their lower-echelon counterparts, thereby facilitating the execution of actions and strategies that unfold over longer time scales. Spatial abstractions help in breaking down large, intricate state spaces into smaller, more digestible observations, improving learning efficiency and agent efficacy.

### 1.2.1 Research Questions

The following research questions guide this research:

- Can an HRL approach enable agents to perform intelligently in large, complex scenarios?
- How do we develop an HRL architecture that can allow for scalability to larger scenarios than has previously been possible?
  - How can HRL be applied to enable scaling to complex state-action spaces that currently exceed our ability to compute with reasonable computing power available today via DOD high performance computers (HPCs)?
  - How can we build a scalable HRL method that can easily grow with complexity?
  - How do we best train each level of the hierarchy in a way that enables scalability but still provides for performance efficacy?
  - How can we abstract the observation space in a dimension-invariant manner that can work with any size scenario without the need to re-train the agents?
  - How can we abstract the observation space to balance training efficiency and performance efficacy?
  - How can we leverage a diverse collection of models, consisting of specialized models, generalized models, machine learning models, and scripted models, to produce agents that maintain high-performance in diverse scenarios?

### 1.2.2 Significance of the Study

This study advances the science of AI by creating a scalable framework that enables the development of intelligent agents designed to emulate, and ideally outperform, the nuances and complexities of human decision-making strategies. This advancement has the potential to transform the way we approach planning, wargaming, and real-time decision-making. As the character of war continues to evolve, the ability to rapidly adapt and learn from simulated scenarios could provide a significant strategic advantage.

Furthermore, this research has implications that extend beyond the military domain. It offers valuable insights into current AI technologies' capabilities and limitations, not just in combat simulations but also for gaming applications that share similar characteristics, such as real-time strategy games.

Lastly, this research contributes to the M&S and AI bodies of knowledge by developing new methods for scaling RL to more complex problems, such as state space abstraction, the use of specialized models over generalized models, and the integration of ML-trained and scripted models. By tackling these issues, this dissertation not only advances our understanding of AI's potential in military applications but also paves the way for future research and development in this critical area.

## 1.3 Research Methodology

This dissertation is divided into four research areas intended to enable the scaling of AI both individually and as a collective. After developing these new approaches, we validate each independently through experimentation and analysis and then integrate all four into our overall HRL framework. The following research areas are presented:

1. A localized observation abstraction that uses piecewise linear spatial decay to reduce the perceived complexity of the state space while still enabling agent efficacy.
2. A multi-model approach that employs a collection of unique and specialized AI models, designed to enhance performance by dynamically selecting the most effective model for the current game state.
3. A hybrid hierarchical AI framework that integrates RL models and scripted models, outperforming either model alone.
4. An HRL framework that integrates observation abstractions, multi-model approaches, and hybrid hierarchical agents into a scalable, self-similar architecture.

Each of these areas contributes to the overarching goal of creating AI systems that can handle increasingly more complex environments. By focusing on a self-similar design in our HRL training framework, we enable AI agents to manage larger maps containing more entities in a more structured and scalable manner.

## 1.4 Dissertation Roadmap

This dissertation is presented by expanding on various publications that have either been published in conference proceedings or accepted for publication. This section provides a roadmap for the dissertation.

### 1.4.1 Chapter 2

This chapter introduces essential background concepts for the development of intelligent agents, such as search methodologies, game theory, scripted agents, reinforcement learning, and hierarchical reinforcement learning. These core elements are crucial for a comprehensive understanding of the research explored in the subsequent chapters.

### 1.4.2 Chapter 3

In this chapter, the focus is on presenting and validating a novel approach to overcome the state space challenges faced by RL agents in larger scenarios by employing a localized observation abstraction using piecewise linear spatial decay. The core of this chapter has been accepted for publication at the 16th International MODSIM World Conference [50].

The introduced method simplifies the agents' perceived state space by abstracting their observations into a more compact and computationally manageable form while retaining critical spatial information. The study demonstrates through a series of experiments that a localized observation abstraction with piecewise linear spatial decay consistently outperforms traditional global observation approaches across different levels of scenario complexity. This suggests that these types of observation simplifications can provide a superior solution at reduced computational costs for scaling RL in complex environments, which has been a significant challenge in the field. The findings contribute to advancing the research in observation abstractions for RL and illustrate the potential of such techniques to facilitate the broader application of RL in complex, real-world settings, particularly in the domain of military simulations and wargaming.

### 1.4.3 Chapter 4

This chapter presents and validates our multi-model framework, which leverages a combination of scripted and reinforcement learning (RL) models to improve performance by dynamically employing the best model given the current state of the game. The core of this chapter has been accepted for publication at the 2024 SPIE Defense + Commercial Sensing Conference [51].

This multi-model framework revealed a significant performance increase, with the most comprehensive multi-model (i.e., the multi-model containing the most individual behavior models) outperforming all individual and simpler composite models. This suggests that even lower-performing individual models can contribute positively in specific contexts, underscoring the value of diversity and specialization within a model repository. The findings emphasize the potential of multi-model systems in enhancing decision-making in complex, dynamic environments typical of military simulations and beyond, advocating for a strategic blend of AI models and techniques to overcome the challenges inherent in training a single generalized model.

### 1.4.4 Chapter 5

This chapter presents and validates a hybrid hierarchical AI framework that integrates RL agents with scripted agents to optimize decision-making in larger combat simulation scenarios. Traditional scripted agents, while predictable and consistent, often fail in dynamic scenarios due to their rigidity. Conversely, RL agents offer adaptability and learning from interaction, though they struggle with large simulation environments and opaque decision-making processes.

We develop a novel approach where a hierarchical structure employs scripted agents for routine, tactical-level decisions and RL agents for strategic, higher-level decision-making. This synergy between the scripted model's consistency and the RL model's adaptability significantly improves performance, leveraging the strengths of both approaches while appearing to mitigate their weaknesses. This integration results in a more effective AI system that can handle a broader range of strategic and tactical challenges in military simulations.

### 1.4.5 Chapter 6

This chapter presents the culmination of our dissertation efforts, outlining the design, development, and integration of the methodologies discussed throughout into a novel HRL architecture and training framework. We explore the potential benefits and limitations of this HRL approach to modeling complex decision-making environments by integrating different levels of observation abstractions and our multi-model approach into the proposed framework. By evaluating the influence of these techniques on learning processes and decision-making efficacy compared to conventional scripted and RL approaches, we aim to gain further insights into the dynamics and challenges of architecting and training HRL systems.

### 1.4.6 Chapter 7

Our final chapter presents the core findings of our dissertation. We discuss the theoretical and practical implications of our research, highlight its strengths and limitations, outline our contributions to the field of AI and combat M&S, and answer the research questions posed in this chapter. Additionally, we leverage the findings from our overall HRL experiment to motivate and specify future work.

THIS PAGE INTENTIONALLY LEFT BLANK

# CHAPTER 2:
# Background

Portions from this chapter have been published in the *Meeting Proceedings of the NATO Modeling and Simulation Group Symposium (NMSG) Simulation: Going Beyond the Limitations of the Real World* [2] and the *Proceedings of the 2023 Interservice/Industry Training, Simulation, and Education Conference (I/ITSEC)* [3]. These publications hold no copyrights over the presented material.

This chapter provides the background information needed to understand our approach of using HRL to scale AI to deal with the complexities of large combat simulation scenarios. While there exist various methods, algorithms, and techniques theoretically capable of computing optimal behaviors to maximize agent efficacy, their inability to scale to more complex environments, assuming reasonable computational resources and time available, render these approaches less suitable for the development of intelligent agents within combat simulations. We will explore several foundational approaches including game-tree search algorithms, game-theoretic strategies, commonly employed scripted methodologies, and relevant ML techniques.

## 2.1 Search

The concept of *search* plays a central and prominent role in ML and AI as it forms the foundation for algorithms to explore and solve problems [49]. Unlike statistical inference, which focuses on drawing conclusions and making predictions from data samples, *search* involves algorithmically exploring data structures like trees or graphs to find solutions or optimize outcomes. This distinction is crucial, as search enables AI systems to systematically process state or game information, aiding in a range of tasks from data retrieval and pathfinding to computing optimal sequences of actions that adhere to specific objective or loss functions [49].

A *game tree* is often utilized to depict the structure of a game, illustrating all possible decisions and outcomes that emerge from gameplay [49]. Similarly, in the context of search problems, this concept extends to the representation of a search space as a growing tree of nodes, each representing a potential state or decision point [52]. A simple example for the game *Tic-Tac-Toe* is shown in Figure 2.1.

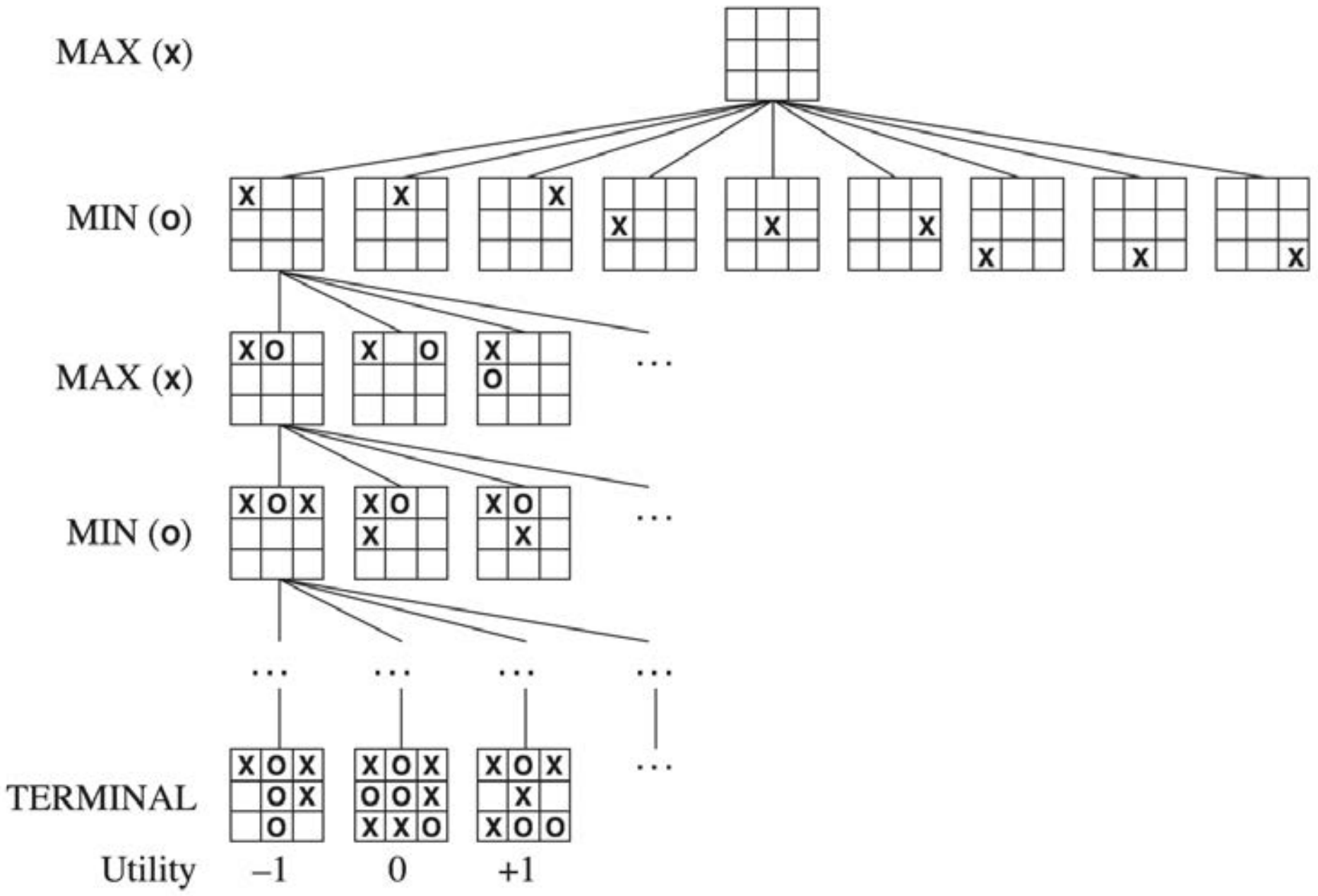

Figure 2.1. Partial Game Tree of Tic-Tac-Toe. The top node is the initial state. Only part of the tree is showing with alternating Xs and Os moves. Source: [49].

A *solution* to this *search problem* is essentially an action sequence leading to the *goal* state, or a winning terminal state [49]. The essence of *search*, therefore, is to find this *solution* by applying each legal action to the current state and generating the resultant set of states. This pattern continues down each branch until the goal state is reached and the final solution is returned. While all search algorithms adhere to this fundamental framework, their divergence lies in their method of selecting the next state or node for expansion [49].

This methodology employed in choosing the next node to expand is called the *search strategy* [49]. *Search strategies* can be broadly categorized into *uninformed* (or blind) and *informed* (or heuristic) search. *Uninformed search strategies* do not use any problem-specific knowledge. Instead, they explore the search space systematically without any guidance towards the goal state. Examples include breadth-first and depth-first search. In contrast, *informed search strategies* utilize problem-specific knowledge through heuristics or models of the environment to estimate the cost of a path from the current state to the goal, aiming to more efficiently navigate the search space [49]. Algorithms like $A^*$ and greedy best-first search are examples of *informed search strategies* [49].

Nevertheless, the complexity and efficiency of a search algorithm is significantly affected by factors such as the state space size, branching factor, and depth of the solution [49]. While optimizations and techniques such as pruning, iterative deepening, and Monte Carlo tree search (MCTS) can be applied to mitigate these challenges, search can still quickly become an intractable problem [49].

Thus, although applying classical search techniques to a problem can theoretically achieve optimal solutions, it can come at prohibitive costs. Combat simulations often involve a large number of agents, many possible actions, and extensive sequences of turns, leading to both large branching factors and depths in their game trees. This makes the search space exponentially large, rendering traditional search methods like alpha-beta pruning [49] impractical due to the enormous amount of computation still required despite best-case pruning. While MCTS has shown promise in some games, it involves extensive sampling of the action space to estimate the value of actions, which is still computationally intensive in these applications.

As a result, using conventional search algorithms becomes less viable for use in combat simulations. This exponential growth in potential states, combined with the need for rapid decision-making in dynamic environments, highlights the need for more advanced, scalable approaches. Therefore, new approaches that can manage extensive search spaces, even if it means at a slight compromise on decision quality, are crucial for advancing the development of highly effective and adaptable agent behaviors within combat simulations.

## 2.2 Game Theory

The discussion of *game theory* is also essential in the study of agent behavior development. While search algorithms help us navigate complex data structures and decision spaces, *game theory* shifts the focus toward strategic interactions among multiple decision-makers. This theoretical framework addresses how rational agents should behave in environments where their actions directly affect and are affected by the actions of others. This concept is foundational for developing intelligent agents that can operate effectively in environments marked by competition, collaboration, and/or negotiation.

Mathematical *game theory* is a branch of economics that conceptualizes environments consisting of multiple agents as games [49] and studies the interaction and subsequent consequences of these self-interested agents [53]. Von Neumann describes a *game* as "simply the totality of the rules which describe it" [54]. *Game theory* essentially analyzes these games with either simultaneous or sequential moves and other sources of partial observability—perfect or imperfect information—to determine the best strategy that maximizes the expected return for each player [49].

A common approach used to model an agent's goals or interests is through *utility theory*. This approach measures an agent's preference for actions among its available alternatives using a *utility function*. This function maps states from the agents' environment to numeric values, which are then used as a measure of the agent's "happiness" [53]. With these utility functions, taking an optimal action in an environment simply involves maximizing this numerical value. However, complexity arises when multiple utility-maximizing agents are present. In these cases, the actions of one of the agents can affect the other's utility [53].

Games are normally represented in *normal form*, which is a matrix form that contains each player's utility for each possible state in the game. Although more complex representations exist in Bayesian games and extensive-form games, most representations can be simplified to a *normal form*, which is the most foundational form in game theory [53]. Figure 2.2 shows the *normal form* representation of *Rock, Paper, Scissors*. An example of an extensive-form representation of a generic game involving imperfect information is shown in Figure 2.3.

| | Rock | Paper | Scissors |
|---|---|---|---|
| Rock | 0, 0 | −1, 1 | 1, −1 |
| Paper | 1, −1 | 0, 0 | −1, 1 |
| Scissors | −1, 1 | 1, −1 | 0, 0 |

Figure 2.2. Normal Form Game Representation. Matrix representation of the game *Rock, Paper, Scissors*. Source: [53].

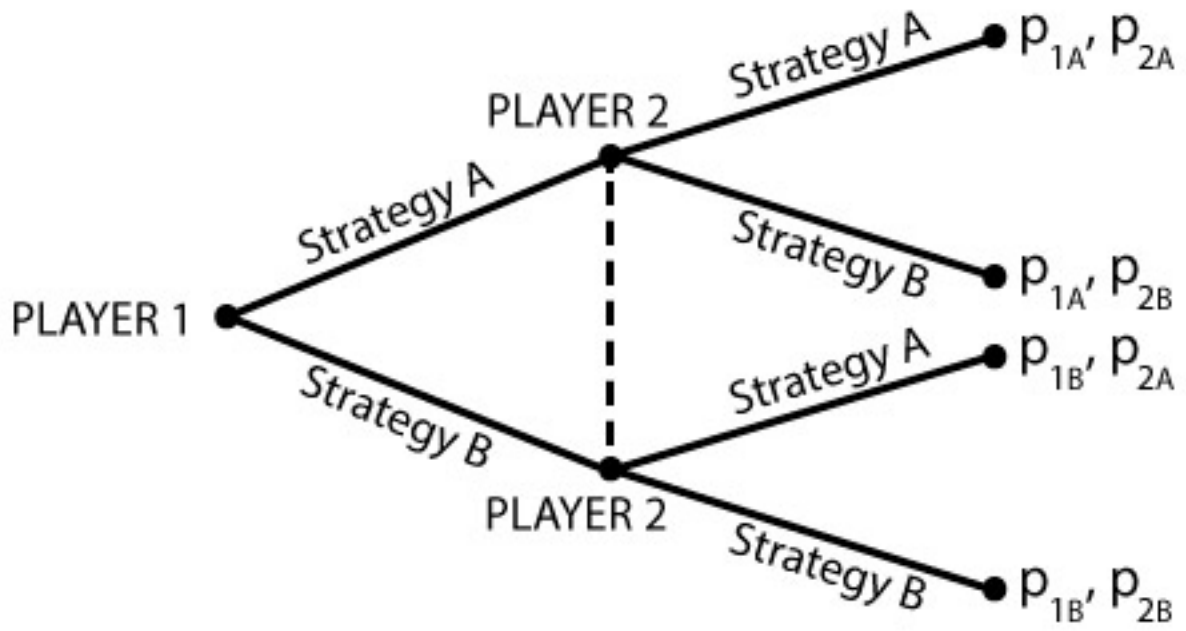

Figure 2.3. Extensive Form Game Representation. Extensive form game of a generic imperfect-information game. The dashed line connecting two nodes means that the player does not know which node they occupy. Source: [55].

For normal form games, there exist two general *strategies* that select the appropriate action from the set of possible choices. One is called a *pure strategy*, which involves simply selecting a single action (or best action) to play. The other is called a *mixed strategy*, which involves using a probability distribution to randomize over a set of available actions [53].

A *pure strategy* can be viewed as a particular instance of a *mixed strategy* in which the probability distribution is simply collapsed to 1.

The *expected utility* is used to calculate the payoffs for mixed strategy profiles. First, the probability of achieving each result, given a strategy or policy, is calculated, followed by the average of the outcome payoffs weighted by the probability of each possibility [53]. To reason over these games, an agent will then seek to maximize its expected payoff given its environment through an *optimal strategy* [53]. There are two fundamental solution concepts to obtain an optimal strategy: *Pareto optimality* and *Nash equilibrium*.

An outcome is considered *Pareto optimal* "if there is no other outcome that all players would prefer" [49]. Every game must possess one or more optima, and there must be at least one optimum where all players can adopt pure strategies [53]. If all players prefer one outcome over any others, then these non-preferred outcomes would be called *Pareto dominated* [49]. *Pareto domination* enables the establishment of partial ordering among various strategy profiles, particularly when pinpointing a single "best" outcome might be challenging, and instead, a collection of incomparable optima exists [53].

While *Pareto optimality* is evaluated from an external observer's perspective, *Nash equilibrium* is assessed from the standpoint of an individual agent [53]. A *Nash equilibrium* represents a stable strategy profile where no agent has an incentive to alter their strategy, regardless of the strategy adopted by the other agents. It is theorized that in any game, provided there are a limited number of players and defined action profiles, there exists at least one *Nash equilibrium* [49].

To date, game theory has proven useful in creating intelligent agents. Different approaches leveraging game theory have long been explored for addressing adversary-style problems, predictions, and recommendations [56]. Although the field of AI has historically focused on turn-taking, two-player, zero-sum, perfect-information games [49], recently, more research has been conducted in simultaneous moving, imperfect-information, and cooperative games. One example is the recent success of this game-theoretical approach that was behind the AI that defeated top professional poker players in competition, reasoning with incomplete and imperfect information [45].

While game theory and the idea of a Nash equilibrium can provide crucial insights into agent decision-making and strategy formulation, their practical application in the complex scenarios inherent in combat simulations presents substantial challenges. Implementing traditional game theory approaches in more complex environments is not straightforward. Transforming intricate multi-agent interactions from these environments into matrix form becomes a daunting task due to the exponential increase in possible game states and outcomes. Additionally, as scenarios become more complex, calculating a best-response strategy with a reasonable timeframe becomes computationally infeasible. This challenge is exacerbated when trying to represent dynamic interactions within the rigid framework of normal or extensive-form games. In fact, extensive-form games, which attempt to map out every possible move in a scenario, closely resemble search problems due to their tree-like structures of decisions and outcomes, and thus suffer from the same exponential increase in complexity as seen in traditional search methods.

Furthermore, just as in the field of economics, where game theory was born, the aspiration for a universal, all-encompassing game-theoretic solution appears to remain elusive. Arguably, no single model or approach can fully capture or predict the intricate dynamics of warfare, where conditions are constantly shifting, and the array of influencing factors—from logistical elements to technological advancements to human decision-making processes—is vast and challenging to model accurately. These realities highlight the inherent limitations of applying game theory in its classical form to modern combat simulations and underscore the need for more adaptive and scalable approaches.

## 2.3 Scripted Agents

Despite the theoretical promise of search and game theory in offering optimal solutions, the practical development of intelligent agent behaviors often leans towards scripted methodologies. Traditional decision-making algorithms such as rule-based systems, behavior trees, and finite state machines (FSMs) are examples of approaches central to agent design in games and simulations [57]. Most strategy games and combat simulations implementing intelligent agents employ these types of hard-coded methodologies due to their reliability, predictability, and ease of implementation. In this dissertation, we refer to these types of approaches that do not rely on exhaustive search, game theory, or machine learning as *Scripted Agents*.

### 2.3.1 Rule-Based Systems

*Rule-based systems* are a class of AI systems that apply a set of pre-defined rules to a particular problem so as to generate a solution [58]. The system evaluates data or inputs against these rules in a sequential or prioritized manner to determine outcomes or actions. These systems are extensively used in expert systems, diagnostic tools, and data processing applications due to their simplicity, explainability, and transparency. They enable straightforward reasoning processes and are easy to modify and understand. However, they can become cumbersome and difficult to manage with an increasing number of rules and can struggle with situations not covered by existing rules, leading to inflexible and sometimes brittle decision-making frameworks. Additionally, rule-based systems do not learn from new data; instead, they operate solely within the confines of their predefined rule sets, limiting their adaptability to dynamic or complex environments [59].

The heart of a rule-based system lies in its rules, which are typically expressed in *if-then* logic [59]. The "if" part (antecedent condition) specifies a situation or a set of conditions, while the "then" part (consequent action) specifies the response or action to be taken when the conditions are met. Each rule in the system encapsulates a piece of knowledge or a specific guideline, making the system highly modular. This modularity facilitates ease of updates and maintenance since changes can typically be made to individual rules without affecting the entire system [59]. Figure 2.4 shows a simple example of a rule-based system schematic.

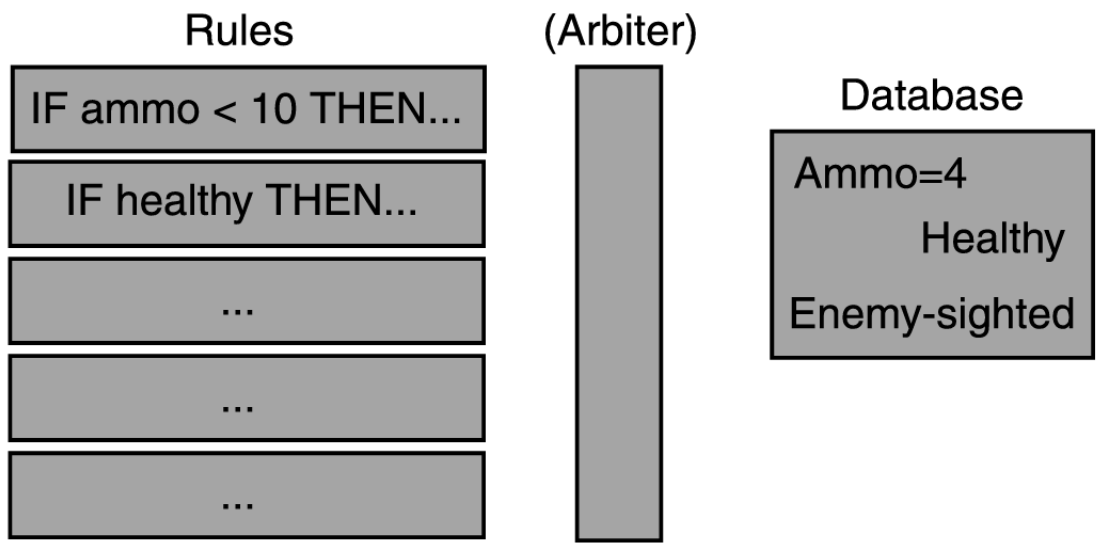

Figure 2.4. Schematic of a Rule-Based System. Source: [57].

Another core component of a rule-based system is the *inference engine*, which is responsible for applying the rules to the knowledge base to deduce new information or make decisions. It performs two primary tasks: matching rules against known facts or data and executing the actions specified by those rules. The effectiveness and efficiency of a rule-based system are significantly influenced by the design and implementation of its *inference engine* [59]. The *inference engine* continuously scans the knowledge base to identify which rules' conditions are met. This process, known as *pattern matching*, can be computationally intensive, especially in systems with a large number of rules or when the knowledge base is extensive [59]. Optimization techniques can help improve the efficiency of this process by minimizing the need to reevaluate unchanged parts of the knowledge base. Once the applicable rules are identified, the inference engine must determine the order in which to execute them. This decision is governed by the system's *execution strategy*, which can vary significantly between different rule-based systems.

Common *execution strategies* include *forward chaining* and *backward chaining*. *Forward chaining* can be viewed as data-driven, as the inference engine initiates with the known data and applies rules to infer more data until a goal is reached or no more rules apply [49]. *Backward chaining*, on the other hand is goal-driven in that the engine starts with a goal and works backward, looking for rules that could result in the goal being achieved (i.e., it identifies what facts must be true for the goal to be satisfied) [49]. In situations where multiple rules are applicable simultaneously, the inference engine employs a conflict resolution strategy to determine which rule to execute. Typical strategies include prioritizing rules based on specificity, recency of the facts involved, or explicit priorities assigned to each rule.

Rule-based systems are inherently transparent because the decision-making process is explicitly defined through understandable rules [58]. This transparency allows users to follow the system's reasoning, which can be crucial in domains where understanding the basis for decisions is important, such as medical or legal fields. A rule-based system's actions are deterministic and directly derived from its set of rules. Given the same input and set of rules, the system will always produce the same output. This predictability is advantageous in applications where consistency is critical.

However, the effectiveness of a rule-based system heavily depends on the quality and comprehensiveness of its rules. In the context of combat simulations, crafting these rules requires in-depth knowledge of military tactics and the ability to anticipate potential scenarios that might arise in a simulation. This can be a complex and time-consuming process, as it involves encoding detailed knowledge into the system and ensuring that the rules interact as intended without creating conflicts or unintended behavior. Thus, developing an exhaustive set of rules that can cover all possible scenarios is a significant challenge, especially in complex or dynamic domains, such as wargaming.

Additionally, these systems lack the ability to learn from new data or experiences unless manually updated by adding new rules or modifying existing ones. This can lead to agents acting inappropriately or failing to respond to new threats, tactics, or operational environments, thus limiting the usefulness of this approach. Consequently, while rule-based systems are powerful tools for codifying and applying expert knowledge, they are not inherently adaptable or flexible to changing conditions without expert human intervention.

### 2.3.2 Behavior Trees

*Behavior trees*, also referred to as *Decision Trees*, extend the rule-based structure and have emerged as a flexible and modular approach for organizing decisions in a hierarchical manner that mirrors natural decision-making processes, thereby enhancing system flexibility while providing a distinct delineation of decision pathways. Originating in the domain of control theory and robotics, *behavior trees* have been increasingly adopted in AI as they provide a clear, hierarchical structure that can be easily understood, modified, and extended [57].

As the name implies, behavior trees represent behaviors as trees where each node represents a decision or a task. These trees dictate the sequence in which behaviors are executed, making them particularly useful for game AI, where agents need to perform a wide range of actions in response to changing environments. An example is shown in Figure 2.5. The structure of a behavior tree comprises leaf nodes, which execute actions or evaluate conditions, and composite nodes, which regulate the execution flow. The execution flow can be sequential, parallel, or conditional, depending on the composite nodes used, such as sequences, selectors, and parallel nodes [57].

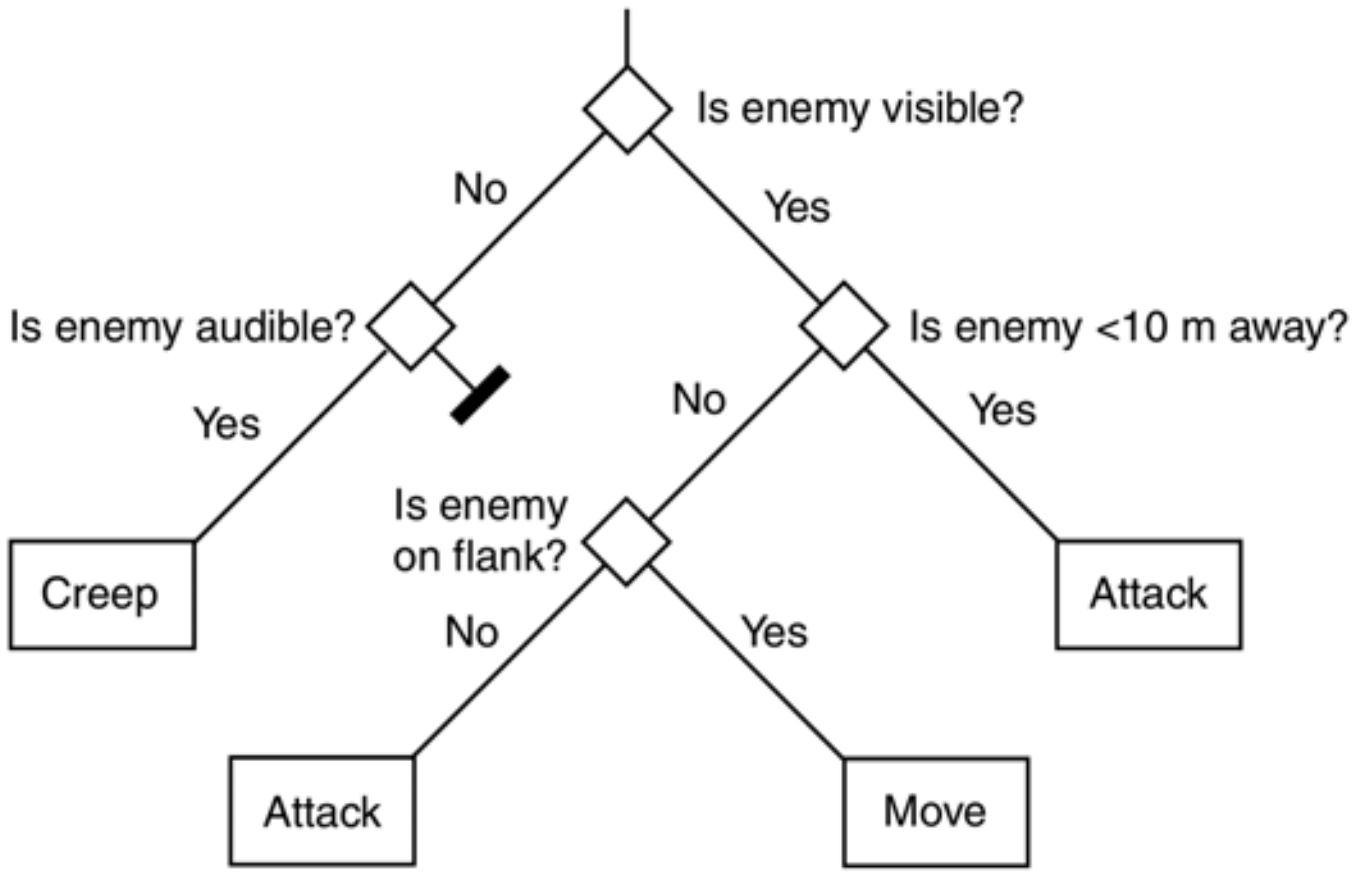

Figure 2.5. Behavior Tree. A basic behavior tree where decisions are contingent on the enemy's position. Source: [57].

At the core of behavior trees are two types of nodes: *leaf nodes* and *composite nodes*. *Leaf nodes*, which include *action nodes* and *condition nodes*, are the fundamental units of tasks. *Action nodes* represent tangible actions the agent performs, such as moving, attacking, or defending, while *condition nodes* check for specific conditions or states within the environment [57]. *Composite nodes*, on the other hand, dictate the control flow of the tree, determining the order and manner in which child nodes are executed. These include *sequence nodes*, which execute their child nodes in a fixed order; *selector nodes*, which try each child in turn until one succeeds; and *parallel nodes*, which execute multiple children simultaneously. By combining these nodes, developers can create intricate behavior patterns ranging from simple decision-making to complex, multi-step strategies [57].

A behavior tree is typically traversed from top to bottom and from left to right, evaluating and executing nodes according to their type and the result of their child nodes. This approach provides a modular and hierarchical structure, enabling the reuse of sub-trees and facilitating the organization and maintenance of complex behavior logic [57].

Due to their clarity and modular nature, behavior trees have gained popularity in the game development community as a tool for implementing AI that can adapt to player actions and changing game states [57]. These advantages allow game developers to create more complex, adaptive AI behaviors while maintaining clarity and manageability in code. The hierarchical structure of behavior trees makes them particularly useful for decomposing intricate tasks into simpler, more manageable subtasks, facilitating the development and implementation of AI behaviors.

However, behavior trees are not without their drawbacks. Designing and maintaining large behavior trees can become cumbersome and may require specialized tools to manage effectively [57]. Like with rule-based methods, designing behavior trees requires in-depth knowledge of the domain and the ability to anticipate what scenarios the agent may encounter. Moreover, while behavior trees offer a clear structure for decision-making, they can sometimes lead to overly deterministic AI behaviors that lack the unpredictability and variation seen in more advanced AI techniques. Additionally, the static nature of traditional behavior trees can limit their flexibility, making it challenging to dynamically adapt AI behaviors to new situations or player actions without human expertise, manual adjustments, and possibly extensions to the tree structure.

### 2.3.3 Finite State Machines

*FSMs* are another fundamental concept used to model agent behaviors. *FSMs* simplify the representation of agent states, breaking down behaviors into clear, discrete stages with defined transitions, thus facilitating targeted problem-solving. An *FSM* comprises of a finite number of *states*, *transitions* linking those states, and *actions* associated with each state [57].

A *state* represents a particular condition or mode of behavior of an agent [57]. Each *state* encapsulates the behavior of the agent during a specific phase or scenario, such as patrolling, movement to contact, attacking, or retreating. These *states* dictate how the agent behaves and interacts with its environment [57]. *Transitions* delineate the circumstances under which an FSM transitions from one *state* to another. These conditions can be based on various factors, such as player proximity, health levels, time, or external events. *Transitions* play a crucial role in enabling the agent's behavior to dynamically adjust to changes in the game environment or the agent's own status. Associated with each *state* are *actions* that the agent performs while in that *state* [57]. *Actions* can include moving towards an objective, attacking an enemy, or any other activity relevant to the *state's* role. These *actions* are carried out as long as the agent remains in the corresponding *state* [57].

FSMs are often represented visually using state diagrams, where states are depicted as nodes, and transitions are depicted as directed edges connecting the nodes [57]. This allows for a clear and intuitive way to understand and design the behavior of agents by showing all possible states and transitions in one schematic [57]. An example of a simple state machine is shown in Figure 2.6.

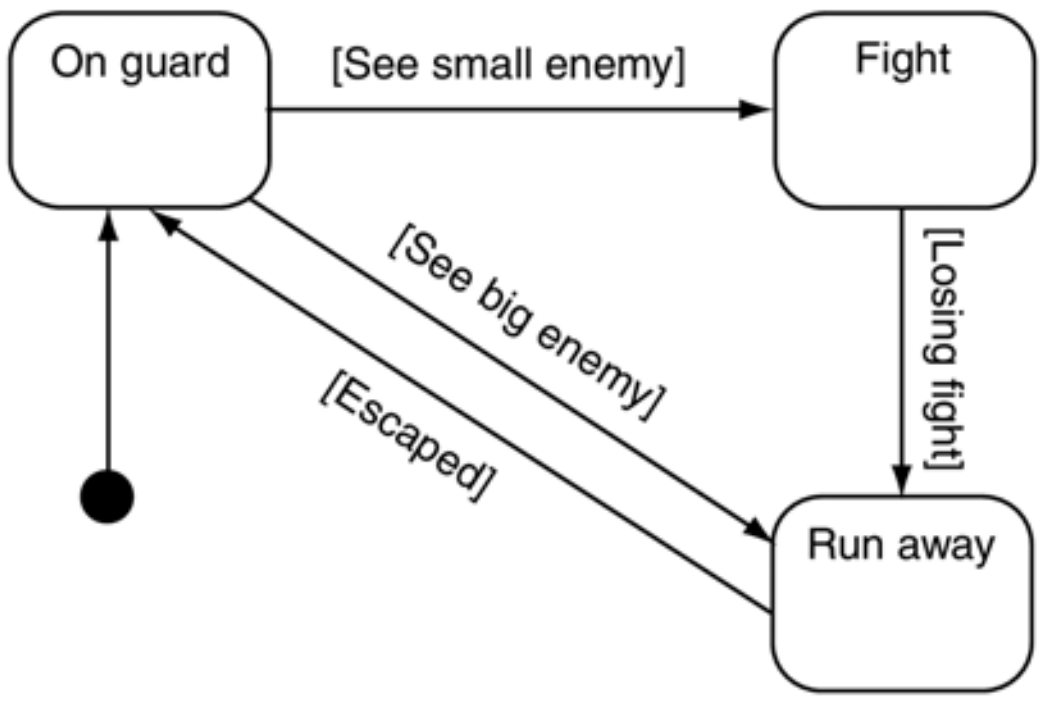

Figure 2.6. A Simple State Machine. The three states of On Guard, Fight, and Run Away are shown with the corresponding transitions. Source: [57].

FSMs are typically implemented using programming constructs such as *switch-case* statements, *if-then* blocks, or more sophisticated state management systems [57]. These structures control the logic for transitioning between states and executing the corresponding actions based on the current state. Furthermore, to manage complexity and allow for more nuanced behavior, FSMs can be hierarchical or nested. In hierarchical FSMs, states themselves can contain sub-FSMs, enabling a layered approach to agent behavior. This allows for the decomposition of complex behaviors into simpler, more manageable components [57].

FSMs have been a classic tool in game AI development, offering a straightforward way to manage agent behavior through a series of states and transitions [57]. They provide a clear and straightforward method for representing state transitions, making them particularly useful for systems where clarity and predictability are key. Their simplicity aids in easy understanding, communication, and implementation, especially in simple to moderately complex scenarios. The modular nature of FSMs allows for easier debugging and updating parts of the system without impacting the whole [57]. This characteristic is especially beneficial in applications with a well-defined and finite number of states, such as certain types of games, user interface flows, and protocols, where predictability and precision are paramount.

However, like with the other scripted approaches, the application of FSMs has its own challenges, particularly as system complexity increases. The notorious "state explosion problem" [60] describes how FSMs can become unwieldy and difficult to manage due to exponential growth in state variables, making them impractical for more intricate games or applications. This lack of scalability and flexibility is a major drawback, as FSMs struggle to adapt to unexpected scenarios or changes in dynamic environments. Additionally, the maintenance overhead grows with complexity, and the inherently limited contextual awareness of FSMs can hinder decision-making processes in more dynamic environments [57].

### 2.3.4 Summary of Scripted Methodologies

While this section has covered three core scripted methodologies commonly used in developing agent behaviors, there exist many other approaches. Each comes with its own set of rules, strategies, and structures tailored to tackle different types of decision-making processes and environmental challenges.

For example, *Production Systems* and *Expert Systems* are similar to, and often synonymous with, Rule-Based Systems in that they apply a set of rules (productions) to a database (working memory), where each rule specifies an action to be taken if certain conditions are met [59]. *Blackboard Systems* are collaborative problem-solving models where different components (knowledge sources) work together by reading from and writing to a common data structure (the blackboard) to solve a problem [61]. *Goal-Oriented Systems* dynamically create action sequences to achieve a specific goal, allowing agents to make complex decisions and adapt to changes in the environment [62]. *Utility Systems* evaluate choices based on a utility score, which combines various factors, enabling agents to choose the most appropriate action based on current context [63]. *Statecharts* are an extension of FSMs that supports hierarchy, concurrency, and communication among states, making them particularly suited for designing complex interactive systems [64]. Although these are defined separately, it is not uncommon to see a combination of these scripted approaches employed in conjunction.

Driven by well-defined rules and logical frameworks, these methodologies create a strong base for crafting agents that perform reliably and consistently in familiar situations [57], [65]. However, this predictability comes with inherent inflexibility, making them less effective in unexpected or novel scenarios [66]. Furthermore, common to all scripted methods is the requirement for an in-depth understanding of unit capabilities, military tactics, and the operational environment. This specialized knowledge is essential to encode the decision-making processes and actions governed by whichever methodology is chosen, ensuring that AI-controlled units act in a realistic and tactically sound manner.

This requirement for extensive domain-specific knowledge, however, presents a challenge. Creating, maintaining, and updating the method's knowledge representations to keep pace with evolving tactics, new technologies, and changing operational environments can be labor-intensive, time-consuming, and often subjective for complex domains, such as warfighting. This is especially true for large-scale simulations involving many units and complex interactions within dynamic simulation environments.

Furthermore, the nature of scripted methods restricts the ability to optimize behaviors. While these frameworks allow for the programming of behaviors based on best practices or established strategies, they inherently lack the dynamic adaptability found in learning-based models. This means that while we can encode what we believe to be the best course of action, the system cannot autonomously improve or adapt beyond its pre-programmed knowledge. This lack of adaptability is a major drawback if the intent is to create intelligent agents in a rapidly evolving world where unexpected situations and novel strategies continually emerge. Thus, while scripted approaches provide a foundation for constructing believable and doctrinal behaviors, their rigidity and the requirement for continual manual updates limit their effectiveness in environments where optimization and adaptability may be critical for success.

## 2.4 Reinforcement Learning

Given these limitations of classical search, game theory, and scripted methodologies, we shift to reinforcement learning (RL), an approach that allows agents to learn and adapt through experience. RL provides a framework for agents to optimize their decision-making process in dynamic environments, offering a practical tool for developing more responsive strategies than would be possible using the previously described approaches.

The concept of RL is one that should be familiar to us. When children engage in play, they might not have a formal instructor, but they possess a direct sensorimotor connection to their surroundings [67]. This connection provides the child with a wealth of information regarding its environment that eventually leads the child to infer cause and effect, comprehend the consequences of their actions, and discern the actions necessary to achieve a goal [67].

The concept of RL emerged from a combination of two research areas: (1) the investigation of animal behavior in reaction to stimuli; and (2) the study of effective methodologies to optimal control problems [68]. In other words, whether it's Pavlov's discovery of classical conditioning [69], Bellman's introduction of dynamic programming [70], Minsky's discussion of RL [71], Williams introduction of REINFORCE algorithms [72], Sutton's introduction of temporal-difference methods [73], Watkins and Dayan's introduction of the Q-learning algorithm [74], RL provides us with a computational approach to learning from interactions with the environment.

Although RL is a subset of machine learning, this approach diverges significantly from supervised learning, which relies heavily on pre-labeled datasets. For example, using supervised learning to teach a machine to play chess requires explicit examples of optimal moves for each position it might encounter. However, obtaining a dataset that accurately and comprehensively represents all possible situations is often infeasible. RL, by contrast, does not depend on labeled data or predefined solutions. Instead, it leverages trial and error, allowing agents to evaluate the outcomes of their actions directly. This method enables agents to not only adapt and learn in environments where outcomes are uncertain or complex but also to discover novel solutions that may not have been encoded in any dataset or anticipated by human experts, thereby expanding the range of potential strategies and responses.

The task in RL is to leverage aggregated rewards to learn the best or optimal strategy [67]. Besides an *agent* and the *environment*, four primary components constitute an RL system: a *policy*, a *reward signal*, a *value function*, and optionally, a *model* of the environment [67]. An RL problem, shown in Figure 2.7, typically consists of an *agent*, and an *environment*. The environment is represented by states $s \in S$. Within this environment, the *agent* selects an *action* $a_t$ based on the current *state* $s_t$ such that $a_t = A(s_t)$. Once the *agent* has selected an *action* at time $t$ $(a_t)$, a *reward* $r_{t+1}$ is received and the *agent* now finds itself in the subsequent *state* $s_{t+1}$. The selected action $a_t$ is output from *policy* (or strategy) $\pi$ is essentially a mapping from *states* $s \in S$ to *action* $\pi(s, a)$ using a probability function. As the *agent* tries new actions within their *environment*, it will eventually learn, or converge on, the optimal *policy*. This policy does not just maximize short-term *rewards*, but also long-term *rewards*. Lastly, when used in model-based RL, a *model* emulates the dynamics of the *environment*, enabling the *agent* to make inferences about how the *environment* might respond [67].

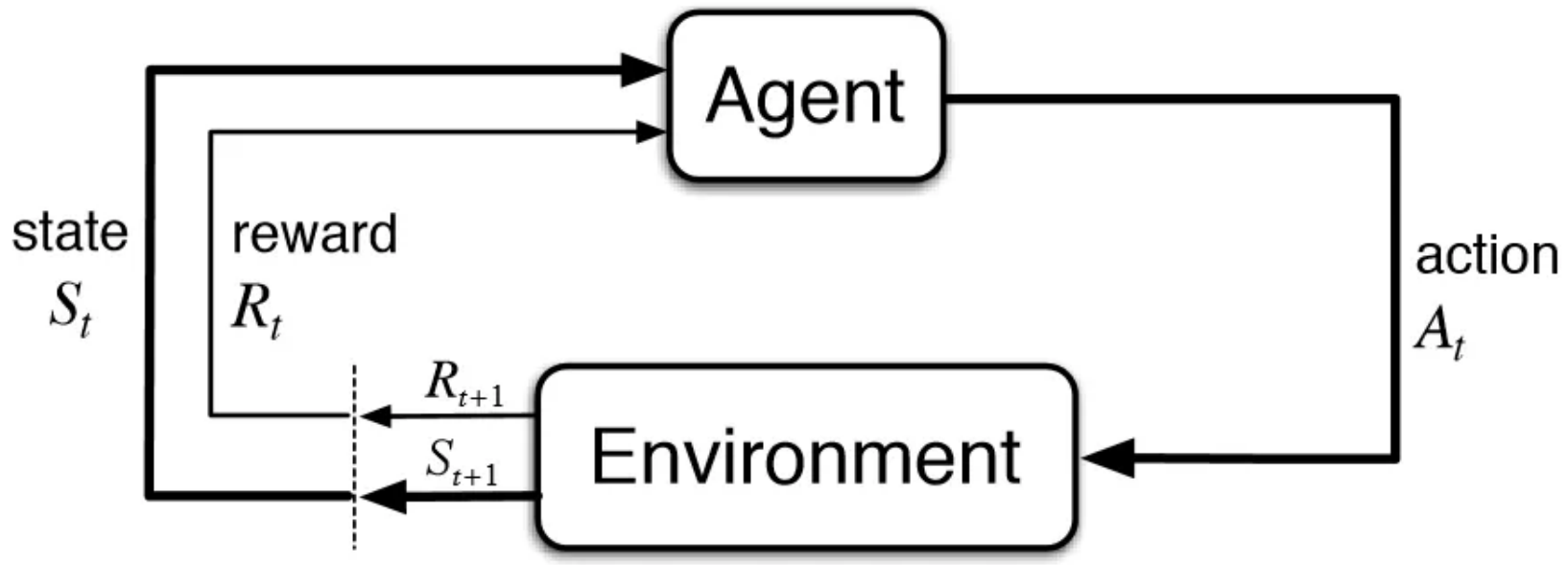

Figure 2.7. The Reinforcement Learning Problem. Source: [67].

Essentially, RL enables an agent to explore its environment and acquire knowledge on how to act by associating situations with actions, aiming to maximize a reward signal [67]. What is more, actions taken by the learner may not only affect near-term rewards but also long-term rewards down the line. This leads to two important characteristics that differentiate RL from other ML approaches: *trial-and-error search*, and *delayed rewards* [67].

What makes reinforcement challenging, however, is that an RL agent must effectively solve three classical machine learning problems in combination: *generalization*, *exploration*, and *credit assignment* [75]. *Generalization* refers to how an agent can use previous experiences to take the best action in a scenario it has not yet seen. *Exploration* refers to the exploration versus exploitation dilemma faced in RL where the agent must balance trying new actions and maximizing rewards from known actions. *Credit assignment* refers to how the agent must attribute delayed rewards received with the most relevant decisions made previously. On their own, each of these problems is difficult; however, in RL, the learning agent must address all three of these simultaneously [75].

### 2.4.1 Markov Decision Process

The Markov Decision Process (MDP) plays a key role in RL as it allows for a mathematical representation of the RL problem [67], as well as provides an underlying theory in how to maximize expected rewards over time. MDPs, which can also be referred to as stochastic dynamic programs or stochastic control programs, are essentially models for sequential decision-making. Because sequential decision-making is fundamentally characterized by MDPs, RL typically operates under the assumption of an MDP framework.

Core to the concept of MDP is the *Markov* property. This property simply states that the future is influenced solely by the present state, independent of the past. Thus, a state signal that retains all relevant information needed for a decision is said to be *Markov* [67]. For example, the current layout of a chess board would be enough to summarize all relevant information of the sequence of moves that may have led to it, and thus the game of chess can be said to have the *Markov* property.

More formally, an MDP consists of a tuple $M = (S, A, P, R, \gamma)$ where $S$ is a set of a finite state that describes the environment; $A$ is a set of finite actions that represent the actions an agent can take; $P$ is a transition function or probability matrix that describe the probability transitioning between states; $R$ is the scalar reward function that depends on the state observed and the action that was taken; and $\gamma$ is a discount factor between 0 and 1 that defines the value of future rewards [67].

To illustrate the benefits of assuming the Markov property, we can first consider the complexities of a non-Markovian scenario, where future states depend on an exhaustive history of past states, actions, and rewards. In such cases, the probability distribution required to predict future outcomes, as shown in Equation 2.1, involves a comprehensive account of all events leading up to the current state. This requirement for historical context results in a significantly increased computational burden, as the system must maintain and process a potentially vast amount of information to make accurate predictions [67].

$$p(s', r|s, a) = Pr\{R_{t+1} = r, S_{t+1} = s'|S_0, A_0, R_1, ..., S_{t-1}, A_{t-1}, R_t, S_t, A_t\}, \tag{2.1}$$

for all $r$, $s'$, and all values of the prior events $S_0$, $A_0$, $R_1$, ..., $s_{t-1}$, $A_{t-1}$, $R_t$, $S_t$, $A_t$, where $S_t$ is the state representation at time $t$, $R_t$ is the reward at time $t$, $s$ is the state, $a$ is an action, and $A_t$ is the action at time $t$ [67].

If we assume the Markov property, however, then the outcome at $t + 1$ relies solely on the current state and actions at time $t$, resulting in:

$$p(s', r|s, a) = Pr\{R_{t+1} = r, S_{t+1} = s'|S_t, A_t\}, \tag{2.2}$$

for all $r$, $s'$, $S_t$, and $A_t$ where $p(s', r|s, a)$ is the probability of transitioning to state $s'$, with reward $r$, from $s$, $a$. Essentially, the *Markov property* would make Equation 2.1 and Equation 2.2 equivalent [67].

Adopting the Markov property, thus, simplifies the decision-making calculations by assuming that the future only depends on the present state and not the past [67]. Therefore, the system needs to consider only the current state and action to predict the next state and reward, thereby dramatically reducing the complexity of the problem. This simplification allows for more efficient computation and more manageable state representations, making it more feasible to develop and implement practical algorithms for decision-making in dynamic environments. Even when state signals are non-Markov, however, it may still be appropriate to think of them as approximations of a Markov state when using RL [67].

For applications where the state and action spaces are finite, we can specify a *finite MDP*. Assuming a state $s$ and action $a$, the transition probability of reaching the next state $s'$ and reward $r$ is given by:

$$p(s', r|s, a) = Pr\{R_{t+1} = r, S_{t+1} = s'|S_t = s, A_t = a\}. \tag{2.3}$$

In the learning process, the ultimate aim is to uncover the *optimal policy* $\pi^*(s)$, guiding the best action selection from each state to maximize cumulative future rewards [67]. Typically, this is accomplished through methodologies like value iteration or policy iteration. Here, the agent iteratively enhances its policy by considering received rewards and updated estimates of future rewards. Value functions are pivotal in this process, as they quantify the anticipated long-term return from a specific state under a given policy [67].

Two main types of value functions are commonly used in RL: the *state-value function* $V(s)$ and the *action-value function* $Q(s,a)$ [67]. The *state-value function* $V(s)$ estimates the expected total accumulated reward for being in state $s$ and following policy $\pi$. This expectation reflects the sum of rewards the agent anticipates receiving over time starting from state $s$ [67]. This is mathematically represented by the Bellman equation for $V(S)$:

$$V^{\pi}(s) = \mathbb{E}_{\pi}[R_{t+1} + \gamma V^{\pi}(S_{t+1})|S_t = s] \tag{2.4}$$

where $R_{t+1}$ is the reward after transition, $\gamma$ is the discount factor that quantifies the difference in importance between near-term and long-term rewards, and $\mathbb{E}_{\pi}$ is the expected value of the sum of the near-term reward and the discounted value of future rewards [67].

Likewise, the *action-value function* $Q(s,a)$ provides an estimate of the anticipated return obtained by executing action $a$ in state $s$, and subsequently adhering to policy $\pi$ [67]. The Bellman equation for $Q(s,a)$ is expressed as follows:

$$Q^{\pi}(s,a) = \mathbb{E}[R_{t+1} + \gamma \sum_{a'} \pi(a'|S_{t+1}) Q^{\pi}(S_{t+1}, a')|S_t = s, A_t = a] \tag{2.5}$$

where $\pi(a'|S_{t+1})$ represents the probability of choosing action $a'$ in state $S_{t+1}$ as dictated by policy $\pi$ [67].

These equations highlight the recursive nature of the value functions, where the value associated with a current state (or state-action pair) is determined by combining the immediate reward with the discounted value of future rewards [67]. This recursive evaluation, in turn, forms the foundation for many policy evaluation and improvement algorithms in RL, allowing agents to optimize their decision-making so as to maximize their long-term reward. As the agent iterates through various actions and observes their outcomes, it refines its value estimates and policy decisions, gradually converging toward optimal behavior [67].

This approach enables the agent to balance exploring new actions to enhance future potential and exploiting known actions to optimize immediate rewards. By applying RL within the MDP framework, agents can learn from their environment and adapt their strategies over time, leading to robust and flexible decision-making capabilities that are essential for navigating complex and dynamic environments.

### 2.4.2 Reinforcement Learning Algorithms

There exist a variety RL algorithms through which an agent attempts to learn to make optimal decisions. Key learning algorithms include value iteration, policy iteration, Q-learning, State-Action-Reward-State-Action (SARSA), Proximal Policy Optimization (PPO), and Deep Q-Networks (DQN). Although many more exist, they are typically improvements, variants, or combinations of these core learning algorithms.

#### Value Iteration

*Value iteration* is a dynamic programming method used in RL to calculate the optimal policy by iteratively improving the value functions [67]. It is based on the Bellman Equation [70] and aims to calculate the *utility* (or maximum value) achievable from each state when adhering to an optimal policy.

The key concept behind *value iteration* is that it iteratively updates the value of each state under the assumption of taking the action that maximizes the expected reward plus the discounted future reward [67]. This process repeats until the value function converges. *Value iteration* is particularly powerful because it guarantees value and policy convergence, assuming finite action state and action spaces and a reasonable discount factor. However, its computational expense can be high for large state spaces, as it requires updating the value of all states in each iteration. Despite this, *value iteration* remains one of the fundamental algorithms used to solve deterministic and stochastic planning problems in RL [67].

### Policy Iteration

*Policy iteration* is another dynamic programming technique used in RL to find the optimal policy [67]. Unlike value iteration, which relies on updating the value function until it converges, *policy iteration* iterates between evaluating and improving policies until it reaches the optimal policy. This alternation between evaluating a fixed policy and improving it ensures a monotonic improvement towards the optimal policy [67].

While the policy evaluation step can be computationally intensive, especially for large state spaces, *policy iteration* typically converges faster than value iteration in terms of number of iterations needed [67]. This makes *policy iteration* particularly suitable for problems where the policy evaluation step can be efficiently implemented or when the state and action spaces are reasonably sized [67].

### Q-Learning

*Q-Learning* [76] stands as one of the most significant advancements in RL. *Q-Learning* is model-free, off-policy RL algorithm used to find the optimal policy for a finite MDP. It allows an agent to learn the value of an action in a particular state without requiring a model that represents the environment. Q-Learning functions by learning the action-value function $Q(s, a)$ which gives the value of selecting action $a$ in state $s$. This streamlines the analysis of the Q-Learning algorithm and enables faster convergence [67].

As the agent interacts with the environment, it needs to balance exploration with exploitation, which is typically managed by an exploration policy, such as $\epsilon$-greedy [67]. After each action is taken, the Q-value for the state-action pair is recomputed. The agent then continually updates its Q-values based on the outcomes of its actions and the corresponding rewards from the environment. This process is repeated for each step or episode of interaction with the environment. After sufficient learning, or once the Q-values have converged, the optimal policy is derived by choosing the maximum Q-value action for each specific state. Assuming each and every state-action pair can be visited and subsequently updated, Q-Learning is guaranteed to converge to an optimal policy [67].

*Q-Learning* has been applied successfully to various domains, including robotics, game playing, and control systems. Its simplicity and effectiveness make it one of the foundational algorithms in RL.

### SARSA

*SARSA*, or *State-Action-Reward-State-Action*, is an on-policy RL algorithm used to learn the value of actions and to subsequently find the optimal policy for a given problem [67]. Unlike off-policy algorithms, which ascertain the value of the best policy regardless of the actions taken, *SARSA* learns the value of the policy being carried out by the agent. The core of the *SARSA* algorithm is to compute the Q-value based on the observed transitions and rewards. This approach allows SARSA to directly adapt and optimize its strategy based on the outcomes of the agent's actions, ensuring safer learning in environments where actions have significant consequences.

By updating Q-values based on the trajectory followed by the current policy, *SARSA* takes into account the actual policy's behavior, including exploration steps. This can lead to safer learning policies compared to Q-learning, as the policy evaluation is done with respect to the actual policy followed rather than independently, as in off-policy algorithms. However, this can also lead to slower convergence to the optimal policy since it is evaluating the exploratory actions as well [67].

### Proximal Policy Optimization

*Proximal Policy Optimization (PPO)* [77] is a policy gradient method for RL that addresses the efficiency and complexity issues associated with earlier algorithms. A policy gradient method directly modifies the parameters of a policy in order to maximize the expected return. Unlike value-based methods, which first estimate the value of actions and then derive a policy, policy gradient methods work directly with the policy itself, which defines the probabilities of selecting each action given a state. This policy is typically represented as a parametrized function, and the parameters are adjusted using gradients of a performance measure, usually the expected return. The objective is to adjust the policy parameters incrementally, aiming to enhance the likelihood of actions associated with greater returns. Through this approach, *PPO* aims to balance the exploration-exploitation dilemma by limiting the size of policy updates, making the training process more stable and reliable [77].

*PPO* introduces an objective function that prevents the policy from changing too drastically at each update [77]. This is achieved by "clipping" the probability ratio between the updated and previous policies, ensuring that the updates remain within a predefined range. This clipped objective encourages the policy to improve while maintaining a controlled exploration rate. Unlike other algorithms that update policies based on sampled data only once, *PPO* reuses each batch of data for multiple rounds of stochastic gradient ascent, thus improving sample efficiency by extracting more information from each data batch. Additionally, *PPO* utilizes advantage estimation, a technique that evaluates how much better an action is relative to the average action at a given state [77]. This helps in accurately guiding the policy towards more rewarding actions.

Because *PPO* does not require complex computations for step size adjustment, it is more practical for real-world applications and has been demonstrated to be effective in a wide variety of environments, from discrete to continuous state spaces. *PPO* is particularly noted for its robustness and ease of use as compared to other on-policy learning methods. Due to its balance between performance and simplicity, *PPO* has become one of the standard algorithms for training RL agents, especially in environments where sampling efficiency and stability are crucial.

**Deep Q-Network**

*Deep Q-Networks (DQN)* [78] represent a significant advancement in RL, merging the traditional Q-learning framework with the power of deep neural networks to effectively navigate complex, high-dimensional state spaces. By integrating these two methodologies, *DQN* overcomes some of the limitations of conventional Q-learning, providing a robust solution for environments that were previously deemed intractable due to their vastness and complexity.

In DQN, a deep neural network is used to estimate the Q-value function. The network receives the state as input and produces the estimated Q-values for all possible actions within that state. This enables the management of high-dimensional state spaces, which traditional Q-learning struggles to handle efficiently. DQN also utilizes an experience replay buffer mechanism to break the correlations between consecutive samples. Experiences (composed of *state*, *action*, *reward*, and *next state*) are stored in a replay buffer, and during the training phase, mini-batches of these experiences are randomly sampled and used to update the network. This approach improves learning stability and efficiency.

To further stabilize learning, *DQN* employs fixed Q-targets. Instead of updating the Q-value with estimates from the constantly changing network, DQN maintains a separate, fixed network to generate the Q-value targets. The weights of this target network are updated less frequently (every few thousand steps) than the primary network, reducing the risk of divergence.

By addressing challenges such as overfitting to recent experiences and instability due to highly correlated inputs, *DQN* dramatically improves the practical applicability of RL to complex problems. The *DQN* framework has been effectively applied to various tasks, demonstrating the capability to learn efficient policies from high-dimensional observation inputs.

### 2.4.3 Reinforcement Learning Summary

The introduction of RL marks a significant shift away from traditional methodologies of developing intelligent agent behaviors, such as classical search, game theory, and scripting approaches, enabling agents to learn optimal or near-optimal behaviors solely through direct interaction with their environments. By applying a trial-and-error strategy, these agents can discover effective policies that maximize long-term rewards, making RL a powerful tool for addressing complex decision-making challenges across various domains.

To date, RL agents have shown to be particularly effective in uncertain environments where the optimal strategy is not known a priori. Notable successes include RL achieving expert or even superhuman performance levels in the games of StarCraft [44], Dota 2 [43], Go [79], and Atari [40].

However, RL's application in the more complex domain of combat simulations in support of wargaming has yet to surpass the performance of human or scripted agents. Research in RL for combat simulations reveals that, although effective in smaller contexts, scaling up to larger scenarios often leads to poor performance [11], [80]–[82]. This issue primarily stems from the exponential increase in state space complexity [70] and RL's sample inefficiency problem [40], which becomes more pronounced as the complexity of the observation space grows.

## 2.5 Hierarchical Reinforcement Learning

Building on the foundation of RL, this section introduces Hierarchical Reinforcement Learning (HRL). To effectively use AI in combat simulations for wargaming, RL agents need the capacity to scale up to accommodate the demands of large-scale wargames that feature extensive state spaces and hundreds or thousands of entities. However, as the size, number, and diversity of elements—such as weapons, units, and terrain—increase within a scenario, so too does the volume of information (both action and state spaces). This escalation can rapidly lead to the scenario becoming computationally intractable. Managing this complexity remains a significant challenge in developing AI-driven wargaming tools that can realistically simulate such extensive and dynamic environments. Fortunately, HRL algorithms have been shown to outperform standard RL algorithms on long-horizon and complex problems [83].

In their chapter *Behavioral Hierarchy: Exploration and Representation* [84], Barto et al. explore how behavioral modules can act as reusable components that can be arranged hierarchically to produce a diverse array of behaviors. On the other hand, traditional RL requires mastering hundreds or even thousands of small tasks independently and from scratch. While this is feasible for simpler games, it becomes computationally expensive and often impractical for more complex games. Despite extensive research focused on improving sample efficiency in RL, the absence of prior knowledge still imposes a significant limit on the speed of learning [85]. Therefore, RL on its own is likely insufficient for scaling effectively in more complex applications like wargaming.

Employing a hierarchical strategy to tackle this exponential problem holds promise as breaking down complex tasks into hierarchical components and reusing old or similar skills is a method already used by humans [85]. This hierarchical breakdown could aid in creating complex behaviors by addressing the problem through multiple levels of abstraction [49].

HRL "decomposes a reinforcement learning problem into a hierarchy of subproblems or subtasks such that higher-level parent-tasks invoke lower-level child tasks as if they were primitive actions" [83]. This decomposition could consist of several hierarchical levels. Additionally, the subproblems generated from this decomposition can be RL problems themselves. The benefit of using a hierarchical decomposition approach is that it can reduce the overall computational complexity, provided that the problem can be structured in such a manner [83].

Instead of a single policy, HRL can employ a hierarchy of policies to collectively determine the behavior of an agent [86]. This task decomposition can reduce what may have been a long-horizon problem into a series of shorter-horizon problems, allowing higher-level tasks to persist for longer timescales as compared to lower-level actions [86]. Furthermore, this type of approach may make it easier for more structured exploration and easier-to-learn subtasks.

In a traditional RL problem, an agent receives an observation and produces an action at regular intervals. However, in an HRL setup, an agent either receives or must uncover background knowledge that enables it to decompose the problem, either explicitly or implicitly [83]. The agent then leverages this knowledge to solve the problem more efficiently by developing a policy that maximizes future rewards [83].

A classic problem used to illustrate some of the benefits of HRL is the *four-room problem* [83], depicted in Figure 2.8. This diagram shows a facility with four rooms each connected by doorways. The agent is represented as a black oval, and the goal is the exit at the top of the facility labeled “Goal.” Each cell, or square, represents a possible position for the agent. The agent can sense its location and move one step at a time either north, south, east, or west. The objective for the agent is to reach the “Goal” door at the top.

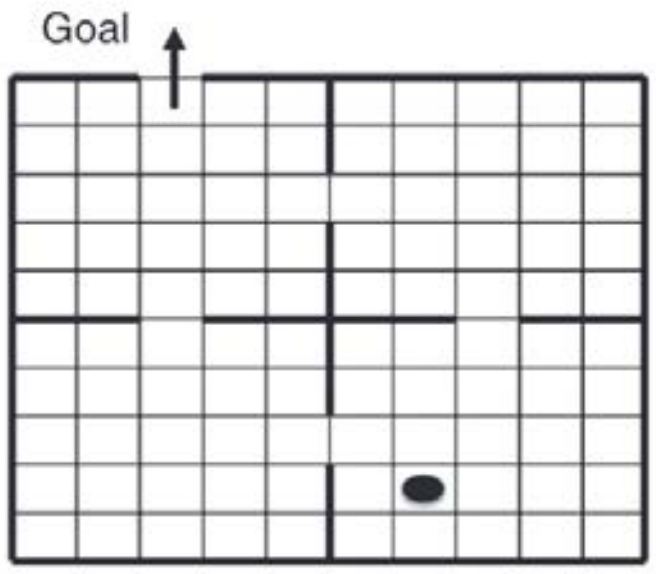

Figure 2.8. Four Room Problem. The classical four-room problem with an agent, shown as a solid black oval, is tasked to find the goal, shown as the exit at the top. Source: [83].

If we were to use a straightforward RL method to solve this, we would need to store 400 Q-values (100 states x 4 possible actions per state) [83]. Using an HRL approach, however, we can decompose this problem into four identical rooms, as shown in Figure 2.9. First, we would only have to learn to tackle two subproblems—one that optimizes reaching a room exit to the north, and another for reaching a room exit to the west. This approach would require learning a total of 200 Q-values (25 states $\times$ 4 possible actions per state $\times$ 2 subproblems). We can then formulate a higher-level subproblem consisting of only 4 abstracted rooms as its states. This would require storage of only 8 Q-values (4 states $\times$ 2 possible actions per state) [83].

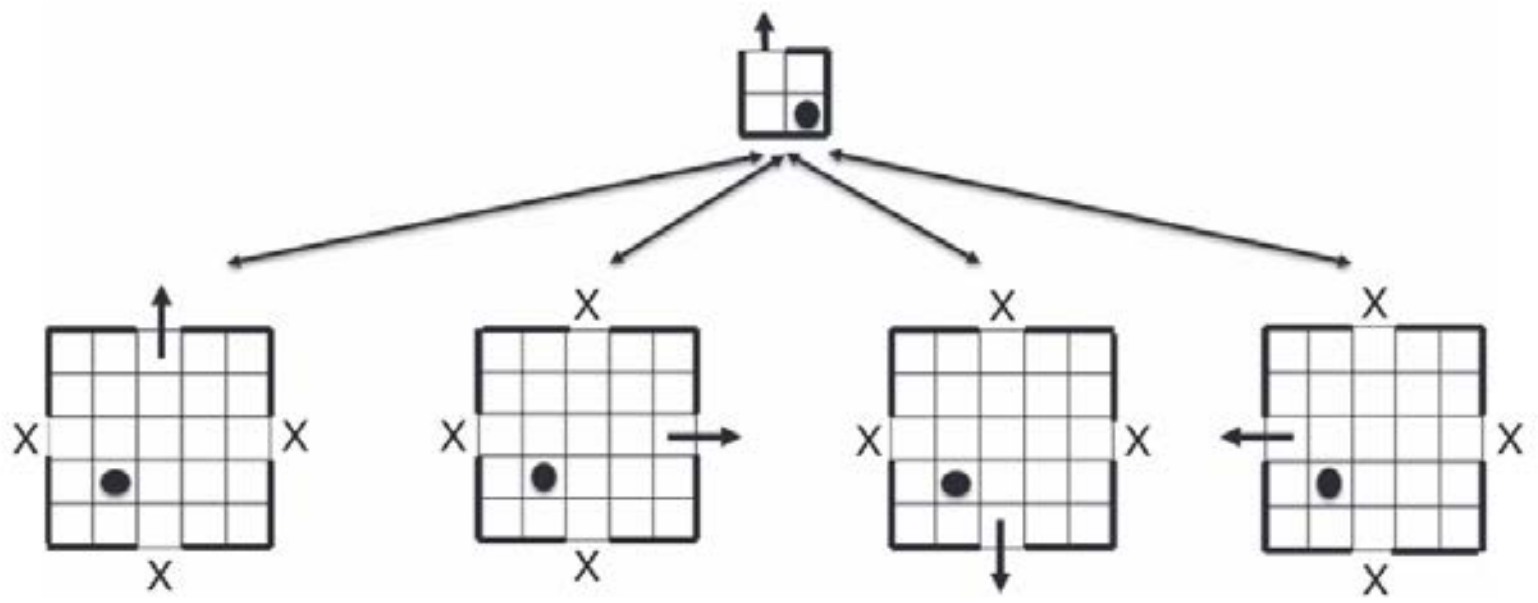

Figure 2.9. Task Decomposition for Four Room Problem. A task-hierarchical breakdown of the four-room problem presents four subordinate subtasks, each representing generic sub-MDPs for exiting a room. Source: [83].

Once a higher—level policy is established—for example, exiting the first room through the west doorway—control can be transferred to a lower-level policy designed to handle the subtask of exiting a room to the west. After completing this subtask, control is returned to the higher-level policy and the process repeats. In this particular scenario, by implementing a two-level hierarchy, the number of Q-values that need to be searched and stored has been reduced from the original 400 to now 208 [83].

As one can see, HRL has the potential to extend the traditional RL framework to better address complex, multi-tiered environments, like those found in combat simulations. By decomposing tasks into a structured hierarchy of subtasks, HRL has shown promise in reducing computational demands and improving learning efficiency. However, the im-plementation of effective HRL systems requires careful consideration of state abstraction, reward integration, solution optimality, and hierarchical structure learning [83].

### 2.5.1 Semi-Markov Decision Process

Enabling HRL are Semi-Markov Decision Processes (SMDPs) [87]. SMDPs provide a mathematical framework that underpins the temporal and hierarchical structure within HRL, allowing for the integration of decision-making at different time scales and levels of detail. Although SMDPs were initially developed for time-stepped systems, they can still be applied to discrete event simulations (DES) to account for the differences in action duration across multiple levels of decision-making.

In human decision-making, actions and their consequences span a variety of time scales, presenting a challenge not adequately addressed by traditional MDP frameworks. Human decision-making often entails selecting from options that extend over time, across a wide spectrum of durations [87]. Consider the planning involved in accomplishing a military objective. For example, a commander's decision to initiate an attack involves not only the immediate action but also a series of subsequent decisions—from a broad strategy of coordinating fires among multiple units all the way down to the discrete action of a Marine firing their weapon. This complexity requires a shift from the conventional, time-step-focused MDP approach, which lacks the mechanism to handle such temporal depth, to a framework that can more accurately represent the hierarchical nature of decision-making in complex environments.

The concept of extending MDPs and applying SMDPs to RL specifically was proposed by Sutton et al. [87], enabling a continuous-time decision problem to be approached as a discrete-time system. Unlike in MDPs, where each action is considered to occur over a single time step, SMDPs allow for actions (or decisions) that span multiple time steps—providing a more nuanced and realistic representation of temporal dynamics in sequential decision-making problems.

The key element of SMDPs is the introduction of the *semi-Markov* property, which dictates that the system's future state of the system is influenced not only by the current state and action but also by the time elapsed since entering that state [87]. This characteristic enables SMDPs to model a wide variety of real-world processes more effectively than traditional

MDPs, from queueing systems and maintenance models to more complex decision-making scenarios found in robotics and automated control systems.

In SMDPs, the decision-making process involves a sequence of *states*, *actions*, and *rewards*, with the addition that the *time* between decisions (decision epochs) can vary. Each *action* results in a *transition* to a new *state* according to a probability distribution, and the *time* until the next decision, or *action*, is determined by a holding time distribution. The objective in SMDPs, similar to MDPs, is to compute a *policy* that optimizes the expected cumulative *reward*, considering both immediate and future *rewards* [87].

SMDPs enable the idea of using temporal abstractions within the RL framework and MDPs [87]. As discussed in the previous section, MDPs do not typically involve any temporal abstractions—MDPs are grounded in discrete time steps where an action performed at time $t$ impacts the state and reward at time $t + 1$. Therefore, conventional MDP methods do not allow for the simplifications and potential efficiencies available using higher levels of temporal abstractions [87].

In the context of SMDPs, the value functions extend the principles of MDPs to accommodate scenarios where actions can have variable durations [87]. Similar to MDPs, SMDPs utilize *state-value* and *action-value functions* to evaluate the expected returns. However, these functions must now account for the varying time intervals associated with each action.

The *state-value function* in SMDPs is given by:

$$V^{\pi}(s) = \mathbb{E}_{\pi}[R_{t+\tau} + \gamma^{\tau} V^{\pi}(S_{t+\tau}) | S_t = s] \tag{2.6}$$

where $t$ represents the time until the next decision point, which is a random variable dependent on the action taken and the state transition; $R_{t+\tau}$ is the reward received after $\tau$ time steps; and $\gamma^{\tau}$ is the discount factor raised to the power of $\tau$, reflecting the discounted value of future rewards over variable time steps.

The *action-value function*, or Q-function, in SMDPs incorporates the variable time steps explicitly, estimating the expected return of choosing an action $a$ in state $s$ per policy $\pi$:

$$Q^{\pi}(s,a) = \mathbb{E}_{\pi}[R_{t+\tau} + \gamma^{\tau} V^{\pi}(S_{t+\tau}) | S_t = s, A_t = a] \tag{2.7}$$

In both expressions, the expectation considers the time dynamics introduced by $\tau$, making these value functions suitable for guiding decision-making in environments where actions are temporally extended.

By incorporating the time element explicitly, SMDPs offer a nuanced approach to evaluating and optimizing policies in environments where actions have durations [87]. This accommodates a broader range of real-world decision-making scenarios, from robotic control, where actions take measurable time to complete, to strategic planning in combat simulations, where decisions unfold over varying time intervals. Iteratively updating policies based on these value functions enables agents to not only select the most rewarding actions but also to strategically consider the timing of each action, optimizing both the sequence and duration of actions to achieve the best long-term outcomes.

### 2.5.2 Options

The term *options* in HRL refers to a framework that provides agents with the ability to execute actions over extended periods [87]. This introduces a structured approach to complex decision-making, enabling agents to more efficiently learn from and more effectively perform in vast and dynamic environments. *Options* facilitate breaking down complex tasks into simpler subtasks, which are easier to solve and can be reused across different parts of the task space. By incorporating *options* into the learning process, agents gain the capacity to operate within and across different scales of time and space, enabling more sophisticated and adaptable behavioral strategies.

*Options* encompass both simple, immediate actions as well as more complex, extended sequences of actions. When a specific set of *options* is employed, it results in a discrete-time SMDP that operates within the confines of the original MDP, as illustrated in Figure 2.10 [87]. The figure's top panel depicts the trajectory of states across discrete intervals typical of an MDP; the middle panel depicts longer-range state transitions of an SMDP's timeline; and the bottom panel depicts how utilizing *options* allows these two approaches to be combined. In essence, the foundational system is still an MDP characterized by straightforward, single-step transitions. However, the introduction of *options* allows for the creation of more substantial transitions that can span several discrete time steps [87].

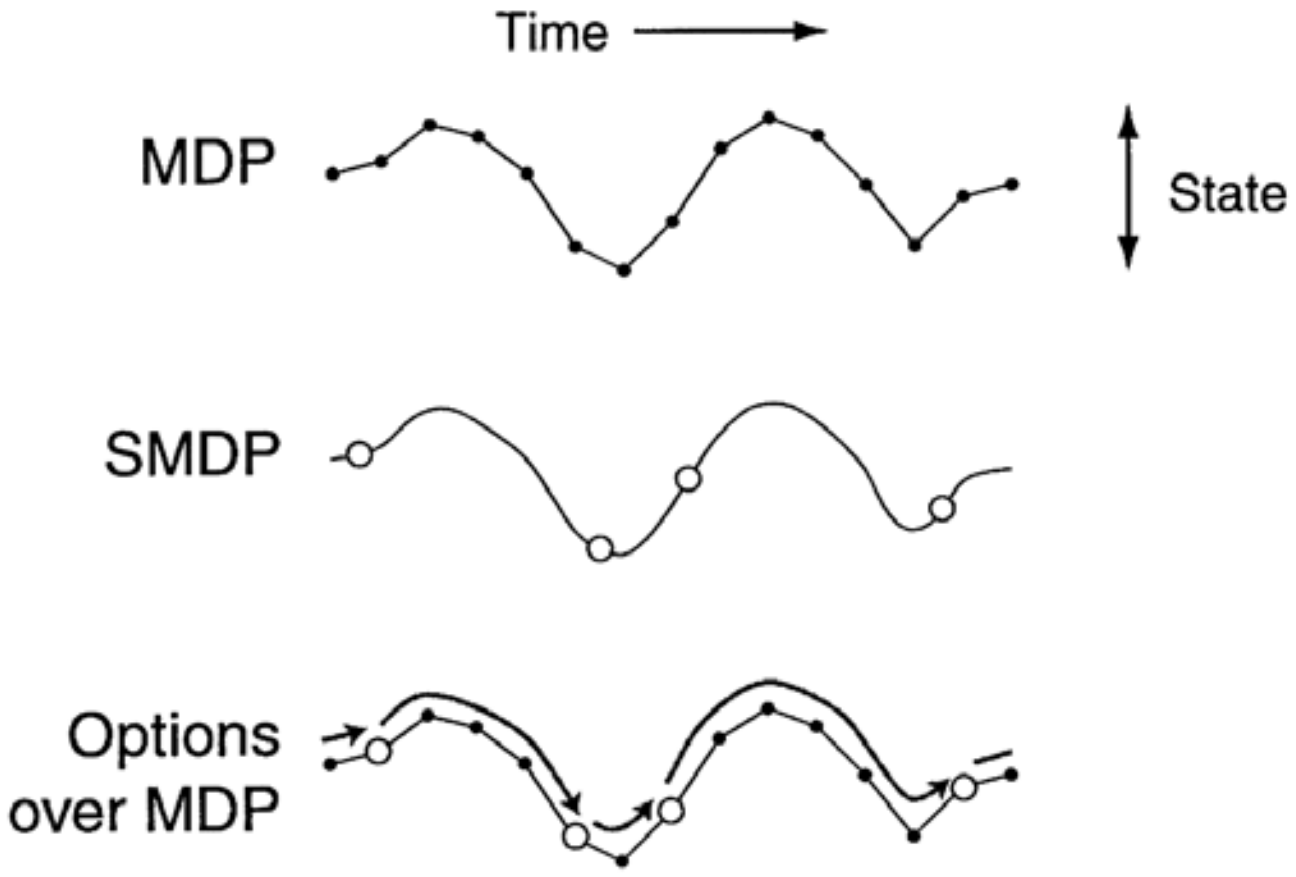

Figure 2.10. In an MDP, the state trajectory consists of short, discrete-time steps, while in an SMDP, it encompasses broader, continuous-time transitions. Source: [87].

There are three components that make up an *option*. The first is an *initiation set* $I$ which is a subset of states from which the *option* can be initiated. The second is a *policy* $\pi$ that specifies the behavior of the agent assuming the current state. The last one is a *termination condition* $\beta$ that dictates when an *option* should terminate. An *option* functions in that first an action $a_t$ is selected from the probability distribution $\pi(s_t)$, resulting in a transition to state $s_{t+1}$. The *option* will then either terminate according to probability $p(s_{t+1})$ or continue selecting the next action $a_{t+1}$, and so on [87]. Once the *option* does terminate, the agent can then select another *option* [87]. The initiation set $I$ and termination condition $\beta$ will restrict the boundaries within which the *option's* policy $\pi$ must be defined. It is often safe to assume that any states $s \in S$ where an *option* could be allowed to persist are also states where the same *option* might be allowed to be chosen [87].

The primary insight from this approach is that by grouping sequences of actions into *options*, we can manage larger state spaces and more complex decisions than with traditional RL using MDPs. This allows for more structured exploration and can reduce the complexity of the learning task. When integrating *options* into HRL, we may not only enhance the ability to handle complex scenarios with a hierarchical decision framework but also potentially increase the learning speed as the agent can leverage past experiences more efficiently.

### 2.5.3 Hierarchical Reinforcement Learning Summary

SMDPs and the concept of options significantly enhance the traditional RL framework and enable HRL. SMDPs introduce a temporal dimension to the decision-making process, where actions have variable durations, leading to transitions between states and associated rewards. This temporal flexibility allows for a more nuanced approach to modeling complex, dynamic environments, particularly when actions have long-lasting effects.

The integration of options within HRL and SMDPs further advances this framework by introducing high-level decision-making strategies. Options, representing temporally extended sequences of actions, allow agents to plan over longer horizons, breaking down complex tasks into more manageable subtasks. This hierarchical structuring enables agents to focus on overarching goals as well as immediate, individual actions.

By accommodating the variability in action durations and allowing for hierarchical task decomposition, HRL offers a more comprehensive and flexible approach to decision-making than traditional RL. This makes HRL particularly applicable to domains involving higher levels of complexities, such as combat simulations, where they can enable more sophisticated decision-making systems.

## 2.6 Summary

This chapter thoroughly explores a range of computational decision-making strategies. It begins with an overview of game-tree search algorithms, emphasizing their role in a systematic exploration of well-defined problem spaces. While these methods are highly effective in deterministic environments, they encounter significant limitations in handling the computational complexity and dynamic changes present in more uncertain scenarios.

We then covered the foundations of Game Theory, which offers a method to model strategic decision-making within competitive and cooperative contexts. Despite its profound theoretical underpinnings and ability to model equilibria among rational agents, game theory's practical application can be challenging in complex domains such as warfighting without significant computational resources being applied.

Next, we covered scripted methodologies such as rule-based systems, behavior trees, and finite state machines. These approaches provide a structured way to dictate agent actions through predefined logic, offering predictability and ease of implementation. However, their inherent rigidity and need for expert domain knowledge make them less feasible for domains characterized by complexity and constant changes to tactics and strategies.

We introduce RL, which represents a significant paradigm shift that emphasizes learning through interaction with the environment. RL empowers agents to optimize their decision-making through a process of trial, error, and reward-based learning, without the need for a pre-defined model of the environment. This approach, however, is not without its challenges, especially when applied to environments characterized by large state and action spaces, where scalability and efficiency become extremely challenging.

Finally, we discuss how HRL emerges as an advanced iteration of RL, aiming to overcome these scalability issues by structuring complex problems into hierarchies of simpler subproblems. Through the employment SMDPs and the options framework, HRL enable multi-level learning and planning, drawing parallels with human cognitive strategies for problem-solving by breaking down complexity into more digestible components.

This chapter highlights the progression from exhaustive search, to hard-coded approaches, to finally more flexible, learning-based methods in computational decision-making. Each strategy offers unique advantages but also faces specific limitations, pointing to the ongoing need for integrated approaches that leverage the strengths of various methods to address the complexities of wargaming applications.

# CHAPTER 3:
# Localized Observation Abstraction Using Piecewise Linear Spatial Decay

This chapter's core material has been accepted for publication and presentation at the *MODSIM World 2024 Conference* [50] and is pending publication. The material is extended in this chapter to provide additional elaboration and detail. This publication holds no copyright over the presented material.

As discussed in Chapter 2, traditional RL agents struggle to learn in large and complex environments due to the amount of state space information necessary to properly represent the number of entities and detailed terrains characteristic of wargaming. This complexity, compounded by RL's sample inefficiency, vastly increases the computational resources and time-to-train needed to achieve satisfactory agent performance outcomes—often rendering the process impractically costly and time-consuming.

This chapter presents a novel approach to overcome this RL challenge by abstracting the agent's observation space while preserving sufficient detail of the relevant portions of the environment. By abstracting the state space into a more compact and computationally manageable observation while still maintaining critical spatial information, we aim to enhance training efficiency while significantly reducing the computational load needed. Specifically, our investigation delves into optimizing training efficacy against the backdrop of limited computational budgets—a common constraint in applying RL to combat simulations.

Through our analysis, we demonstrate that a localized observation strategy consistently outperforms a global observation method across increasing levels of complexity. This finding emphasizes the superiority of this approach when training agents in complex scenarios where spatial relationships are essential—offering a way to help RL scale to still produce acceptable levels of performance in larger, more dynamic environments than have previously been possible in the domain of combat simulations.

## 3.1 Objective

In this study, we develop, implement, and test a localized observation abstraction approach using piecewise linear spatial decay. Our goal is to compare the tradeoffs between the traditional global observation approach and our approach using localized observation abstraction with piecewise linear spatial decay.

The results of this study inform when and how we should use this abstraction method in our overall HRL framework.

## 3.2 Background

The notion of *abstraction* for AI is not new and has been used since the beginning of AI and logic, dating back to the work by Whitt in approximating dynamic programs [88], [89]. Giunchiglia and Walsh [90] informally define *abstraction* as "the process of mapping a representation of a problem onto a new representation." Russel and Norvig [49] define *abstraction* as "the process of removing detail from a representation" [49]. Ho et al. [91] emphasize the critical role of abstraction in decision-making for both humans and machines, emphasizing its need for managing the growing costs associated with decision-theoretic calculations. They outline how psychology and neuroscience recognize abstraction as vital in organizing motor sequences, habits, and planning, thereby highlighting its importance in AI and RL for providing solutions to human-level problems. In essence, *abstraction* allows us to consider what is relevant and to disregard what is irrelevant based on the specific task at hand [90].

The process of abstraction is simply accomplished by mapping a *ground* representation of a problem to a new *abstract* representation of the same problem [90]. Figure 3.1 shows an example scenario of a hiker in the woods. Through this mapping, we can remove unnecessary details and preserve only certain desirable properties relevant to the task to be performed. In this case, the *abstract* representation of the hiker's environment allows the hiker to navigate more efficiently and effectively than the original *ground* representation.

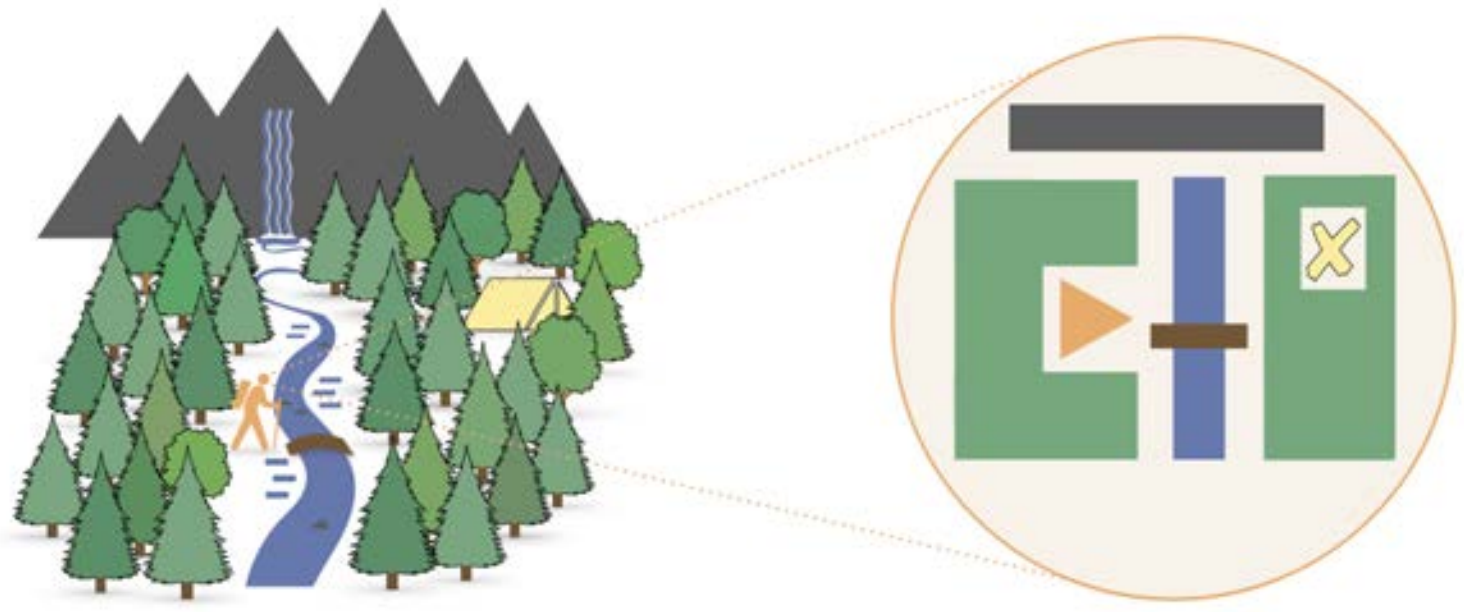

Figure 3.1. Abstraction Process Example. An example of the abstraction process for a hiker in the woods. Source: [75].

Within the context of RL, Abel [75] formalizes this notion and comprehensively investigates the role of abstraction in RL in detail, particularly focusing on *state abstraction*. However, he goes further and discusses how the concept of abstraction can be split into two categories: *state abstraction* and *action abstraction*. From these categories, he defines two additional types of abstraction: *state-action abstraction* and *hierarchical abstraction*. These four types of abstractions are shown in Figure 3.2. In this chapter, we are only concerned with and cover *state abstraction* and leave the other types of abstraction for discussion in Chapter 6.

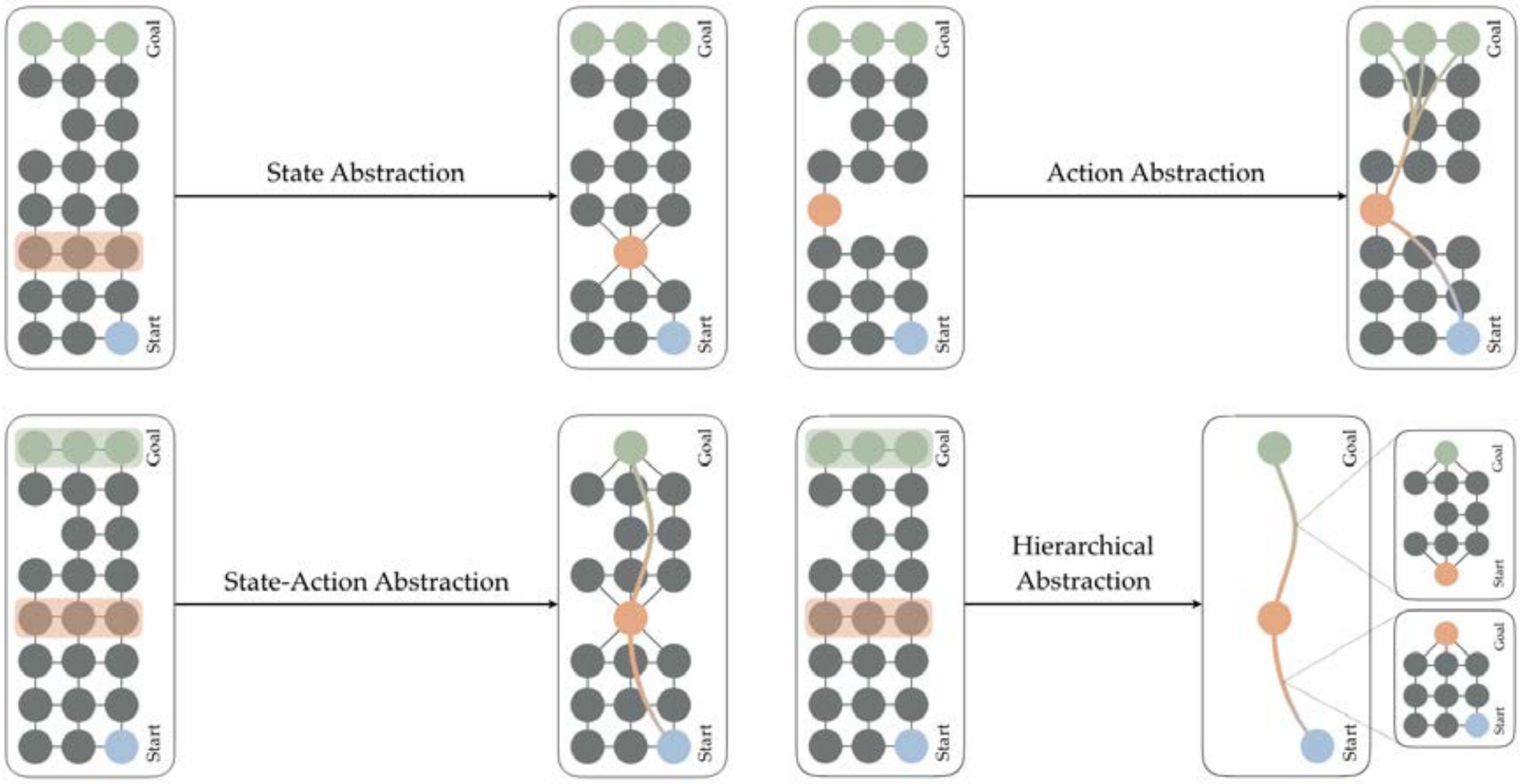

Figure 3.2. Overview of Different Types of Abstraction. The four different types of abstractions for Markov decision processes are shown. Source: [75].

A *state abstraction* involves capturing only the most relevant aspects of the environment, aiming to determine which changes to the environment are actually substantive [75]. The goal of this type of abstraction is to simplify the state space by clustering like states together, ensuring that no crucial aspects of the problem are altered. This abstract state then becomes the agent's understanding of the current environment. Thus, determining what information to discard becomes a central question in formulating abstractions [75].

To leverage state abstractions, as shown in Figure 3.3, once an MDP produces a state $s$, we can then pass it through a function $\phi$ that then yields an abstracted state $s_\phi$, which becomes the new input to an RL agent [75]. This way, an agent does not need to know the actual full state space representing the environment. Furthermore, this allows for a division between $\phi$ and the agent which results in the ability to examine $\phi$ separately from the RL algorithm itself [75].

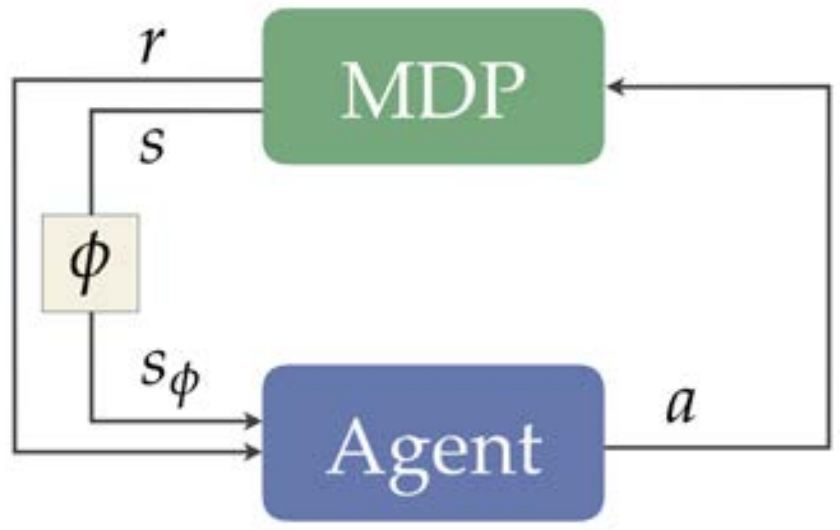

Figure 3.3. Reinforcement Learning with State Abstraction. The state output $s$ by a Markov decision process is fed through a function $\phi$ prior to becoming an input $s_\phi$ to the agent. Source: [75].

It should be evident then that abstraction can play a critical role in simplifying complex decision-making processes [91]. Given limited computational power and a complex enough environment, agents in a simulation cannot model everything in their environment and still learn appropriate or optimal behaviors within a reasonable time. As the complexity of the environment increases—and assuming that a minimum level of performance is desired—agents may have to discard some information and focus only on relevant information to solve a specific problem. This form of abstraction allows for a more manageable representation of intricate environments, enhancing the learning efficacy of AI agents [91]. Moreover, this approach not only has the potential to reduce computational demands but may also improve the adaptability and performance of AI-trained agents in scenarios that may be significantly different from the scenarios for which the agents were trained [89].

As Shanahan and Mitchell [92] explore in depth in their research, for abstraction to be most useful, "the domain of a concept's application must be larger than the domain of its acquisition." We contend that abstraction is critical to transferring concepts learned from one setting to another that differs from which it acquired said concept. Nevertheless, because abstractions inherently involve discarding information—which could compromise the effectiveness of decisions made based on these abstractions—it's crucial to recognize and balance the trade-off between simplifying learning (or making it tractable) and retaining sufficient information to enable the discovery of an optimal policy [75]. The more information we abstract from the state space, the more information we lose, making it increasingly difficult to guarantee an optimal or near-optimal solution [93]. Nevertheless, researchers agree that a tradeoff exists in that, while coarser abstractions might lead to sub-optimal actions, they facilitate improved planning and value iteration [93].

In this dissertation, we use Abel's [75] definition of *State Abstraction* as a function $\phi : S \rightarrow S_\phi$, which maps each true environmental state $s \in S$ into an abstract state $s_\phi \in S_\phi$. In other words, an abstracted state serves as the agent's interpretation of the current environment, which will discard or simplify some information.

## 3.3 Related Works

The field of abstraction in RL is still a relatively underexplored area. Nevertheless, research in this area shows a wide variety of approaches, each addressing different aspects that might be relevant to the decision-making processes. Understanding these works contextualizes our research and highlights the gaps this study aims to fill.

As discussed in detail in Section 2.5.1, Sutton et al. [87] pioneered the concept of temporal abstraction in RL by extending MDPs [94] and proposing Semi-Markov Decision Processes (SMDPs). This foundational work emphasized understanding temporal factors in decision-making, focusing on the abstraction of actions rather than states. While crucial in developing our overall HRL framework, it differs from this specific part of our approach, which instead explores state abstraction in the spatial context.

Further exploring abstraction in games with large state spaces, Sandholm [95] introduces a broad overview of game abstraction techniques, particularly focusing on solving large incomplete-information games, such as poker. His work involves creating simpler models of games that maintain strategic similarity with the original models. Sandholm's process works in three stages, as illustrated in Figure 3.4. First, the game undergoes abstraction to create a simplified version that maintains strategic similarities with the original game. Second, this abstract version is then analyzed to find an equilibrium or near-equilibrium strategy. Lastly, the strategy derived from the abstract game is adapted and applied back to the context of the original game. While Sandholm's work also diverges from our focus in that it primarily addresses abstraction through the lens of game theory, its principles and methodologies offer valuable insights into our work on using localized observation abstractions within an RL framework.

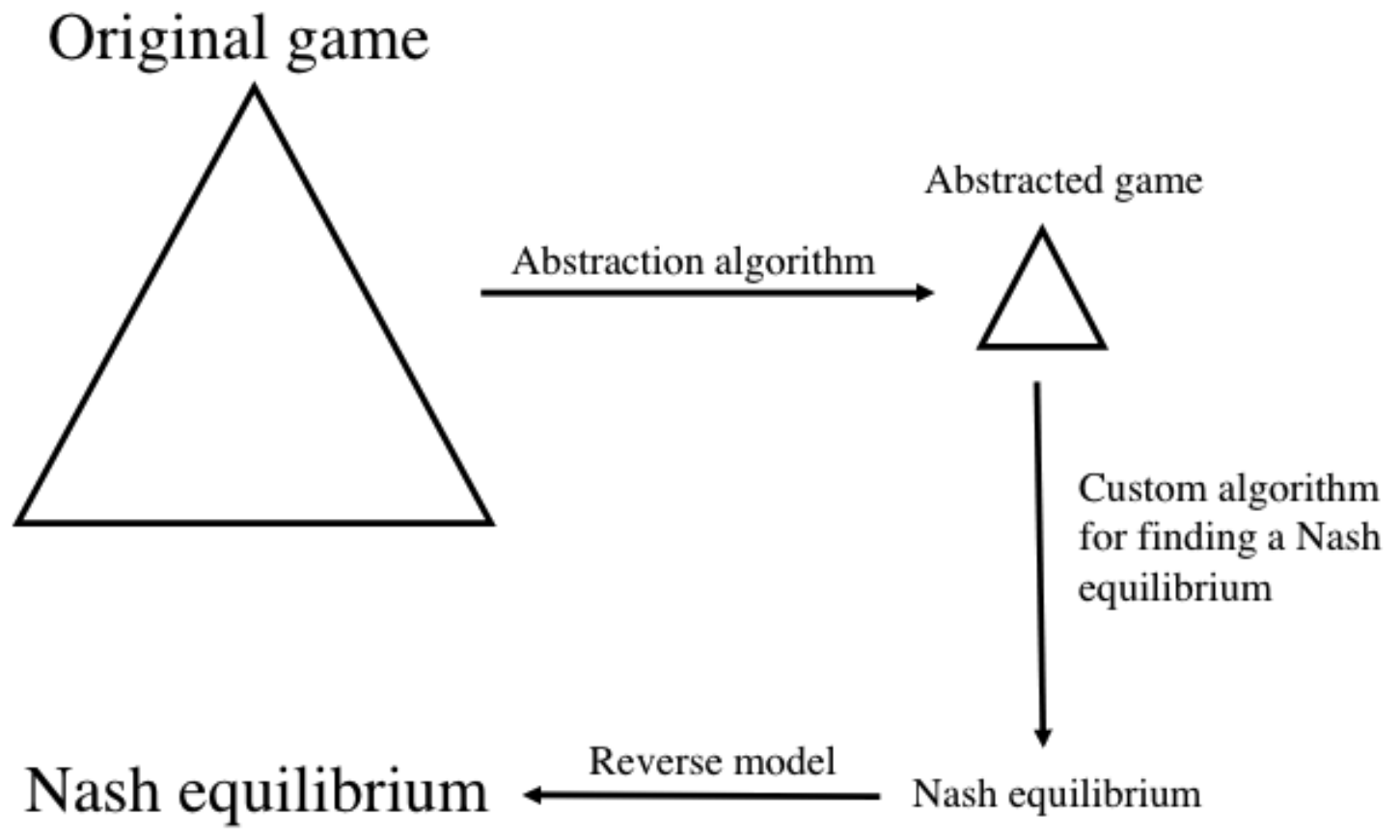

Figure 3.4. Framework for Abstracting Games. Source: [95].

Andersen et al.'s study [96] investigates enhancing exploration in RL through the development of a Dreaming Variational Autoencoder (DVAE). DVAE abstracts complex environments into a manageable representation. By learning to encode and decode the environmental states, it abstracts the essential features of these states into a lower-dimensional space. This helps the RL agent better handle and interpret large or complex state spaces more effectively, focusing on the most relevant features for decision-making. Thus, this approach is designed to address challenges in RL, particularly in environments with sparse feedback, by generating probable future states for improved exploration and decision-making. The authors introduce *Deep Maze*, a versatile maze environment, to test DVAE, offering various scenarios to challenge the algorithm across different complexities and observability conditions. [96].

The DVAE algorithm is illustrated in Figure 3.5. Initially, the agent accumulates experiences to leverage the experience-replay mechanism during its *Run-Agent* function [96]. In this phase, the agent navigates the state-space using a policy shaped by Gaussian distribution. The agent then takes action, makes observations, and saves these observations in its experience replay buffer. Once the agent attains a terminal state, the DVAE algorithm processes state-action pairs from the replay buffer, encoding them to predict potential future states, which are subsequently preserved in a dedicated buffer [96].

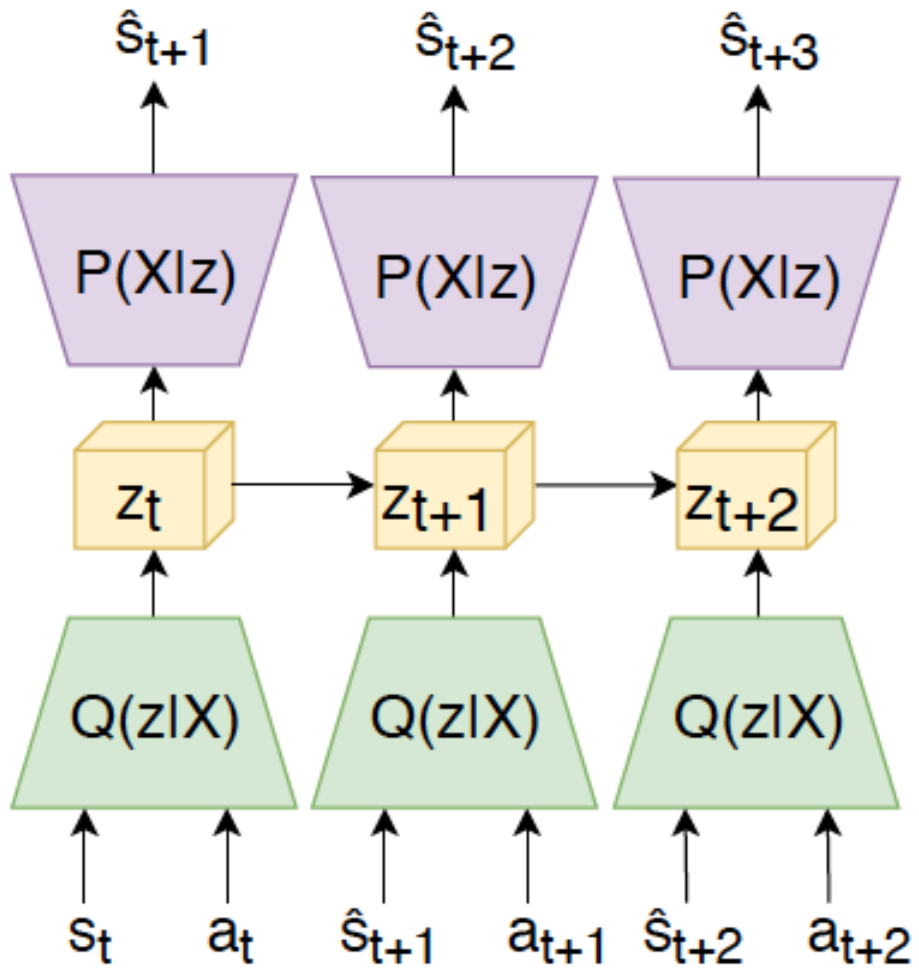

Figure 3.5. Dreaming Variational Autoencoder (DVAE) Model. The model takes in state and action pairs, encodes into latent space, and then decodes to a likely future state. $Q(z|X)$ represents the encoder, $z_t$ represents the latent-space, and $P(X|z)$ represents the decoder. Source: [96].

Their experiments show promising initial results, highlighting DVAE's potential in environment modeling for better exploration in RL. Their emphasis, however, is on probabilistic latent space encoding as opposed to our deterministic spatial abstraction representation. Additionally, while DVAEs offer valuable insights into data encoding and the potential for improved exploration, their approach requires exhaustive exploration of the environment for effective learning. This extensive exploration requirement challenges scalability to larger and more complex environments, diverging from our objective of scaling to these larger scenarios.

In a more focused application, Allen et al. [97] present an approach for learning state abstractions that preserve the Markov property. The core contribution of this work lies in defining theoretical conditions necessary for an abstraction to be able to retain the Markov property and proposing a feasible training method that integrates inverse model estimation with temporal contrastive learning. This method allows for learning Markov abstract state representations without relying on reward information or the need to predict observations, thereby overcoming practical sparse reward and high-dimensional challenges. Allen et al. demonstrate the effectiveness of their approach through empirical evaluation in a visual gridworld domain alongside a collection of continuous control benchmarks. Their methodology not only captures the underlying structure of these domains but also leads to improved sample efficiency over existing RL methods [97].

Although Allen et al.'s [97] methodology also focuses on state abstraction and seeks to improve learning efficiency, their study contributes more so in the theoretical sense focusing on maintaining MDP integrity. Our approach, on the other hand, involves using tailored abstractions to improve scalability to high-dimensional state spaces.

Lastly, Jergeus et al. [98] present an approach to synthesizing efficient communication schemes in multi-agent systems by combining symbolic methods and ML into a neuro-symbolic system. The core idea is to enable agents to extend their initial primitive language with novel higher-level concepts through interaction and RL, thus facilitating generalization and more informative communication through shorter messages. This method shows potential in allowing agents to converge more quickly on collaborative tasks. Although they are conducting a form of abstraction, Jergeus et al.'s [98] focus is on probabilistic abstractions of linguistic communications while our abstraction approach is specifically designed to handle spatial information in a nuanced way.

Each of these works illustrates the broader application of abstraction in RL. Collectively, they demonstrate the diverse methods of tackling complexity in the decision-making process. Our research builds upon these foundations and, because we contend that spatial relationships are critical to effective decision-making in most wargaming scenarios, we explicitly address the underexplored area of spatial state abstraction in the complex and intricate domain of combat simulations.

As discussed by Abel [75] in his Ph.D. dissertation, *A Theory of State Abstraction for Reinforcement Learning*, RL agents currently face significant challenges in generalizing experiences, exploring environments, and learning from delayed and sparse feedback, all within limited computational constraints. Abel [75] highlights the necessity of abstraction in these processes, focusing on state abstraction, to improve sample efficiency for RL. Furthermore, he outlines three desiderata for useful state abstraction—preserving near-optimal behavior, being learnable and computable efficiently, and reducing the time or data needed for effective decision-making—all of which we also seek to achieve in this study.

## 3.4 Atlatl Simulation Environment

In this section, we describe the Atlatl Combat Simulation [99] environment we use throughout this dissertation to develop, implement, experiment, and validate our research methodology. This simulation environment has been used in [11], [80]–[82], [100], [101]. Atlatl is a simple yet efficient combat model developed at the Naval Postgraduate School. Atlatl comprises a deliberately simplistic core combat model in addition to a surrounding Gymnasium [102] infrastructure that supports rapid AI experimentation. The environment is equipped with hooks that facilitate integration with standard RL and algorithms and codebase, such as Stable-Baselines 3 (SB3) [103]. Such a foundational environment allows researchers to efficiently and effectively develop, apply, and evaluate state-of-the-art AI solutions to operational and tactical challenges that would not be possible using operational or high-fidelity simulation systems.

An example of a simple scenario in Atlatl is depicted in Figure 3.6. Units are visually represented by their corresponding military operational symbols and graphics, while terrain features are depicted with colors (e.g., urban areas are gray, mountainous terrain is brown, water is blue).

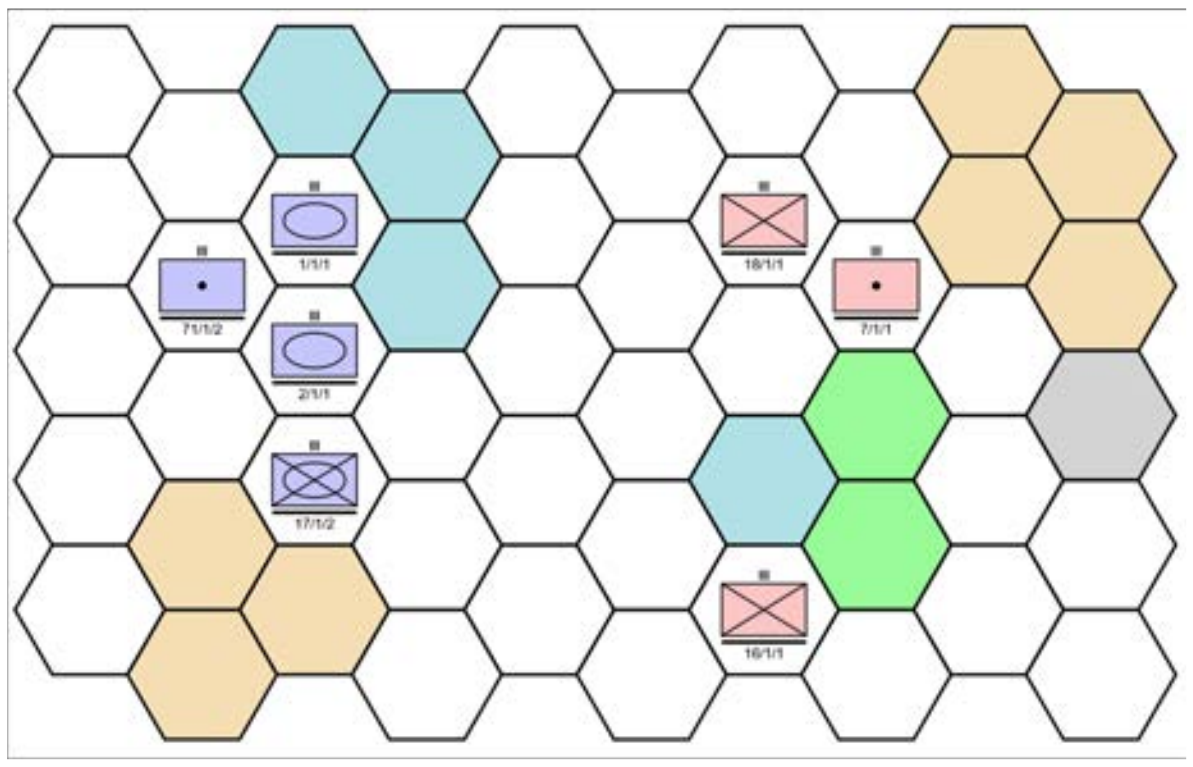

Figure 3.6. Example of an Atlatl Scenario Gameboard. Source: [99].

### 3.4.1 Gymnasium

The Atlatl simulation environment uses the *Farama Foundation Gymnasium* [102] (formerly OpenAI Gym) framework to enable RL. *Gymnasium* offers a standard API for a wide range of RL environments, crucial for developing, testing, and comparing RL algorithms. The API contains four main functions: `make()`, `reset()`, `step()`, and `render()`. *Gymnasium* also contains a core `Env` high-level Python class that represents an MDP. Within *Gymnasium*, MDPs are implemented as `Env` classes, along with `Wrappers` which allow for environment modification without the need to alter the underlying code directly [102].

The framework's structure, with clearly defined observation and action spaces, helps in developing models that can interpret and act within the Atlatl simulation environment. For our action space (`env.action_space`) we use a `Discrete` space that allows for 0 to $n$ possible values for actions. For our observation space (`observation_space`), we use a `Box` space that allows an $n$-dimensional continuous space bounded by user-defined upper and lower limits, which describe the valid values the observations can take.

### 3.4.2 Stable-Baselines 3

Atlatl implements SB3 [103] as its default RL library. Built on Pytorch [104], SB3 is open-source and actively maintained and used by the RL community for both research purposes and real-world applications. The library provides implementations of many standard and emerging RL algorithms, including Advantage Actor-Critic (A2C), Proximal Policy Optimization (PPO), Deep Q-Networks (DQN), and many others. SB3 offers a simple, consistent API for training, evaluating, and deploying models. It also includes features that allow for saving/loading models, logging, and environment wrappers for preprocessing. The library focuses on providing implementations that are both efficient and reliable, with a strong emphasis on code quality and reproducibility.

Figure 3.7 depicts the relationship between Atlatl, Gymnasium, and SB3. Atlatl is registered and implemented following the Gymnasium API standards. This enables integration with SB3 and allows us to leverage its RL algorithms for the training and deployment of RL models within Atlatl.

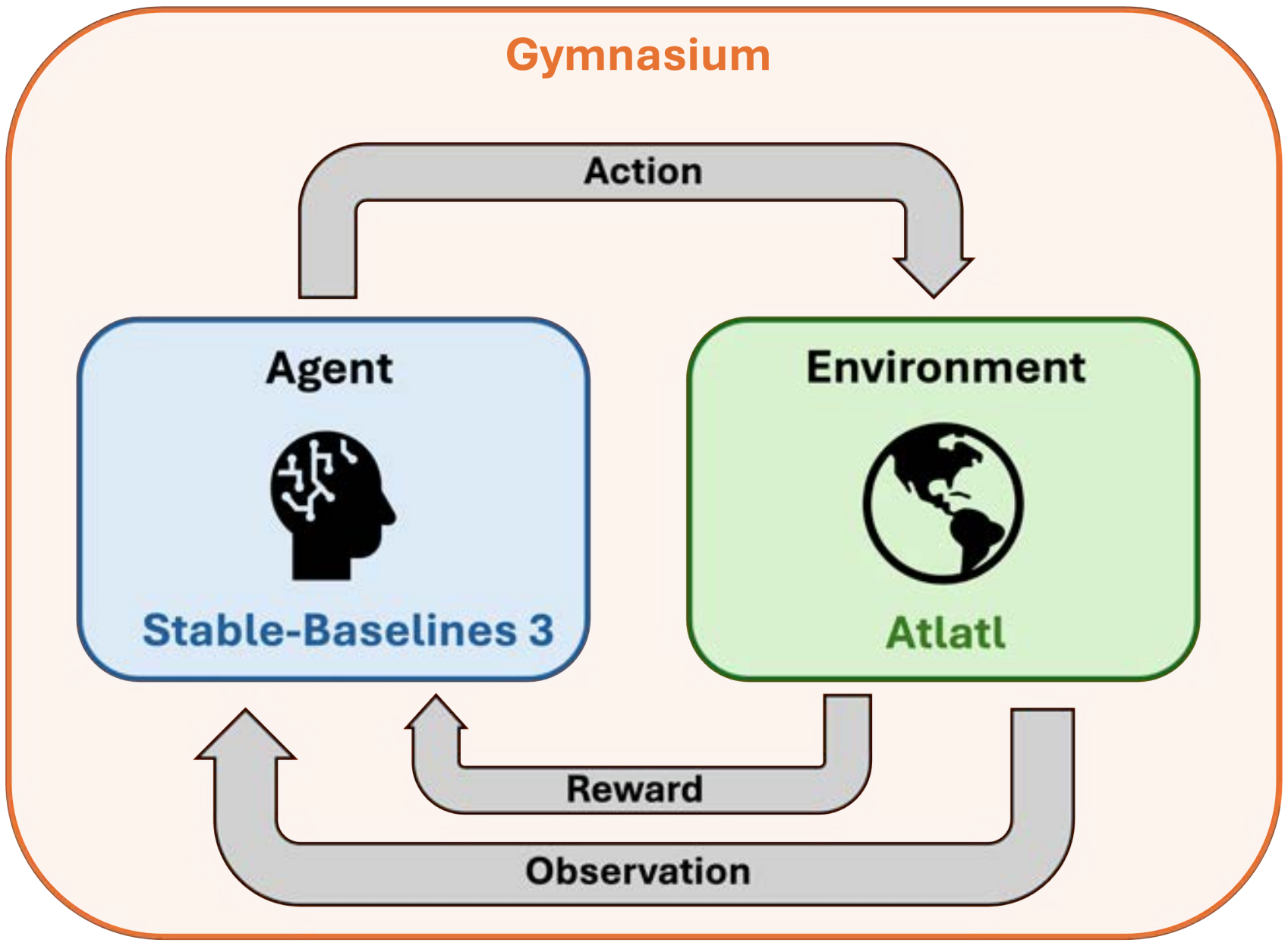

Figure 3.7. Relationship Between Gymnasium, Stable-Baselines 3, and Atlatl.

Figure 3.8 depicts the relationship between Gymnasium and SB3 specifically. *Observations* are obtained from the Gymnasium environment, passed through a *Features Extractor*, and input into some form of a *Fully-Connected Network* for inference. This results in a value that is passed back to the Gymnasium environment, which is then converted into an *Action*.

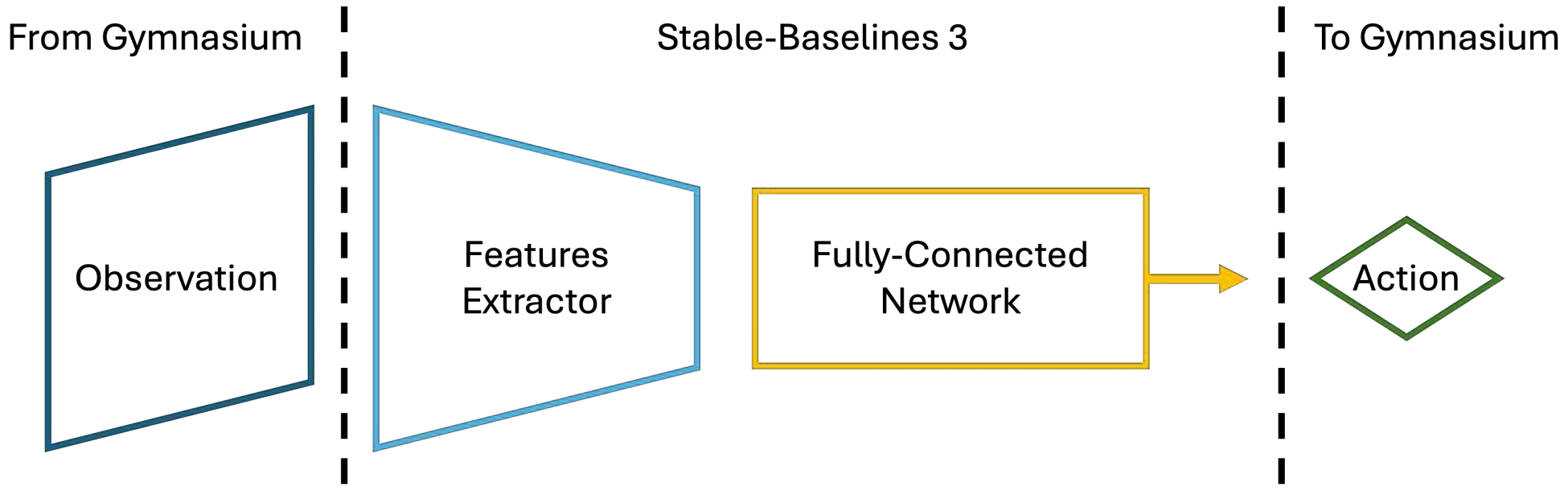

Figure 3.8. Relationship Between Gymnasium and Stable-Baselines 3. Adapted from [105] and [11].

### 3.4.3 Simulation Execution in Atlatl

Atlatl can be run either headless or via a web browser interface. Running Atlatl headless (i.e., without any graphical output) allows an AI to play against another AI autonomously without any human input. Using the web browser interface, a human player can engage in matches against either an AI opponent or another human player. Moreover, the interface features a browser-based replay functionality, enabling the replay of AI versus AI engagements.

The simulation environment is run using a Python script and can take in a number of arguments which include the AI class, AI model, random seed, number of repetitions, etc. Of interest to this dissertation is the `scenarioCycle` and the `scenarioSeed` arguments.

The `scenarioCycle` argument determines how many unique scenarios are generated and reused throughout the simulation. For instance, setting `scenarioCycle` = 5 results in the simulation creating five distinct scenarios at the start, which will then be repeatedly used in sequence for each new game iteration. Setting `scenarioCycle` = 0, however, results in each new game featuring a scenario that is entirely new and randomly generated, ensuring no deliberate repetition of scenarios across games. This approach allows for controlled variability in the simulation, facilitating the study of agent performance across a specific set of conditions or across a broader range of unpredictable environments.

The `scenarioSeed` argument specifies the random seed used for initializing the environment's state. This feature enables consistent replication of experiments, ensuring that the initial starting scenario conditions under which agents operate remain identical across different experiments. By controlling the randomness in this manner, we can systematically compare the performance of various strategies and algorithms under the same environmental conditions, providing a fair and controlled basis for evaluating improvements or changes in agent behavior.

### 3.4.4 Scoring

The scoring system within Atlatl is fully customizable. Currently, performance within Atlatl, scoring is determined by factors such as kills, losses, and control of urban areas, although these metrics can be readily adjusted according to preferences. Game scores are given from the perspective of the blue player. For this dissertation, scoring is primarily computed based on two factors: combat effectiveness and control of urban areas (used to represent cities).

We use the following scoring function:

$$S_{total} = S_{blue_city} + S_{blue_combat} - (S_{red_city} + S_{red_combat}) \tag{3.1}$$

where $S_{blue_city}$ is the score per city owned by the blue faction; $S_{blue_combat}$ is the score for each red agent damaged in combat by the blue faction; $S_{red_city}$ is the score per city owned by the red faction, and $S_{red_combat}$ is the score for each blue agent damaged in combat by the red faction.

As shown, control of urban hexagons plays a significant role in calculating the player's score. At the start of each scenario, unoccupied urban hexagons are not controlled by any faction. Control shifts only when an entity occupies the urban hexagon, with the controlling faction awarded a score of $\frac{24}{n_{cities}}$ points per phase controlled, where $n_{cities}$ is the number of cities present in a given scenario. Of note, once an urban hexagon is occupied, it remains under that faction's control even if the entity vacates the hexagon, up until an entity of the opposing faction occupies the same urban hexagon.

Regarding combat, each entity begins with an initial 100 strength points. Each damage point inflicted on a red entity translates into a positive point for the blue faction, while each damage point inflicted on a blue entity translates into a negative point for the blue faction. If an entity's strength drops below 50 points, it is removed from the game (i.e., deemed ineffective) and the remaining strength points are awarded to the opposing faction.

### 3.4.5 Scripted Agents

Atlatl contains a scripted behavior model called *Pass-Agg* that we use throughout our experiments as a baseline model. *Pass-Agg* uses a behavior tree structure where it first assesses its posture as either "passive" (defensive) or "aggressive" (offensive) based on the relative combat power (or strength) of its faction as compared to its opponent. This combat power is determined during each agent's turn by summing up the health remaining of each available unit for each faction. If the agent's faction has greater or equal combat power than its opposing faction, it adopts an offensive behavior model. Otherwise, it adopts a defensive behavior model.

For each moveable unit, the *Pass-Agg* algorithm first checks if there is at least one enemy unit within its attack range (i.e., a target); if so, it uses a uniform distribution to select a unit to attack. Of note, the infantry units we use in this dissertation all have an attack range of one hexagon (i.e., they can only attack opponent units that are adjacent to their own position).

If no targets are available, the *Pass-Agg* algorithm calculates movement based on a scoring system that is influenced by its posture. The scoring function first retrieves a list of hexagons the unit on-move can legally move to, including the hexagon it currently occupies. The infantry units all have a movement range of one hexagon (i.e., they can move to an unoccupied adjacent hexagon). It then calculates the Euclidean distance from each of these candidate hexagons to each opponent unit and to each urban hexagon.

If the unit is in an offensive posture, the score of each candidate hexagon is calculated by summing the distance to the nearest opponent unit with the distance to the nearest urban hexagon. If one of these does not exist (e.g., all enemy units are defeated, or no urban hexagons exist in the scenario), it is ignored in the scoring. The candidate hexagon with the lowest score is chosen as the hexagon to move to. This essentially results in the agent moving toward the hexagon that best achieves the objective of both defeating the opponent and capturing the urban hexagon(s).

If the unit is in a defensive posture, the score of each candidate hexagon is simply the calculated Euclidean distance to its nearest urban hexagon. Likewise, the hexagon with the lowest score is chosen as the hexagon to move to. This essentially results in the agent moving towards the nearest urban hexagon.

Overall, this system allows for rapid, deterministic responses to changing game dynamics, adhering to a set of predefined rules and scripts. Furthermore, *Pass-Agg* exhibits goal-oriented behavior, albeit in a basic form, aiming to achieve specific objectives such as capturing key terrain or defeating enemy units. While not leveraging complex planning algorithms, this goal-oriented approach ensures that the agent's actions are consistently directed toward achieving its overarching strategic objectives of maximizing the game score. While a simple model, it has proven to be an effective agent that takes human-like actions and regularly achieves reasonably good scores.

### 3.4.6 Global Observation

The default global observation in Atlatl consists of an $18 \times n \times m$ tensor, where $n$ and $m$ are the height and width respectively of the gameboard. For this specific study, we use square gameboards (e.g., a $5 \times 5$ scenario consists of an observation space of $18 \times 5 \times 5$). Each channel of the observation represents one specific type of information to be captured, as shown in Figure 3.9. Specifically, channel 0 is a binary (specifically one-hot encoded) matrix depicting where the blue unit to be moved (or on-turn) is located; channel 1 a is a binary matrix depicting all blue units that still have the ability to move during the current phase; channel 2 is a binary matrix depicting all legal moves available for the unit on-move; channels 3 and 4 are matrices that depict the health level (scaled from 0 to 1.0) of each respective unit on the gameboard based on factions; channels 5 through 8 are binary matrices depicting unit types; channels 9 through 13 are binary matrices representing terrain; channels 14 and 15 are binary matrices depicting the city owner (i.e., which faction was the last to pass through an urban hexagon); channel 16 is a matrix filled with a single number, a phase indicator value representing the current phase of the game; and channel 17 is a matrix similarly filled with the normalized game score. While we recognize that these last two features can be represented more compactly as vectors or scalars rather than matrices, we maintain the matrix construct to keep the neural net structures simple.

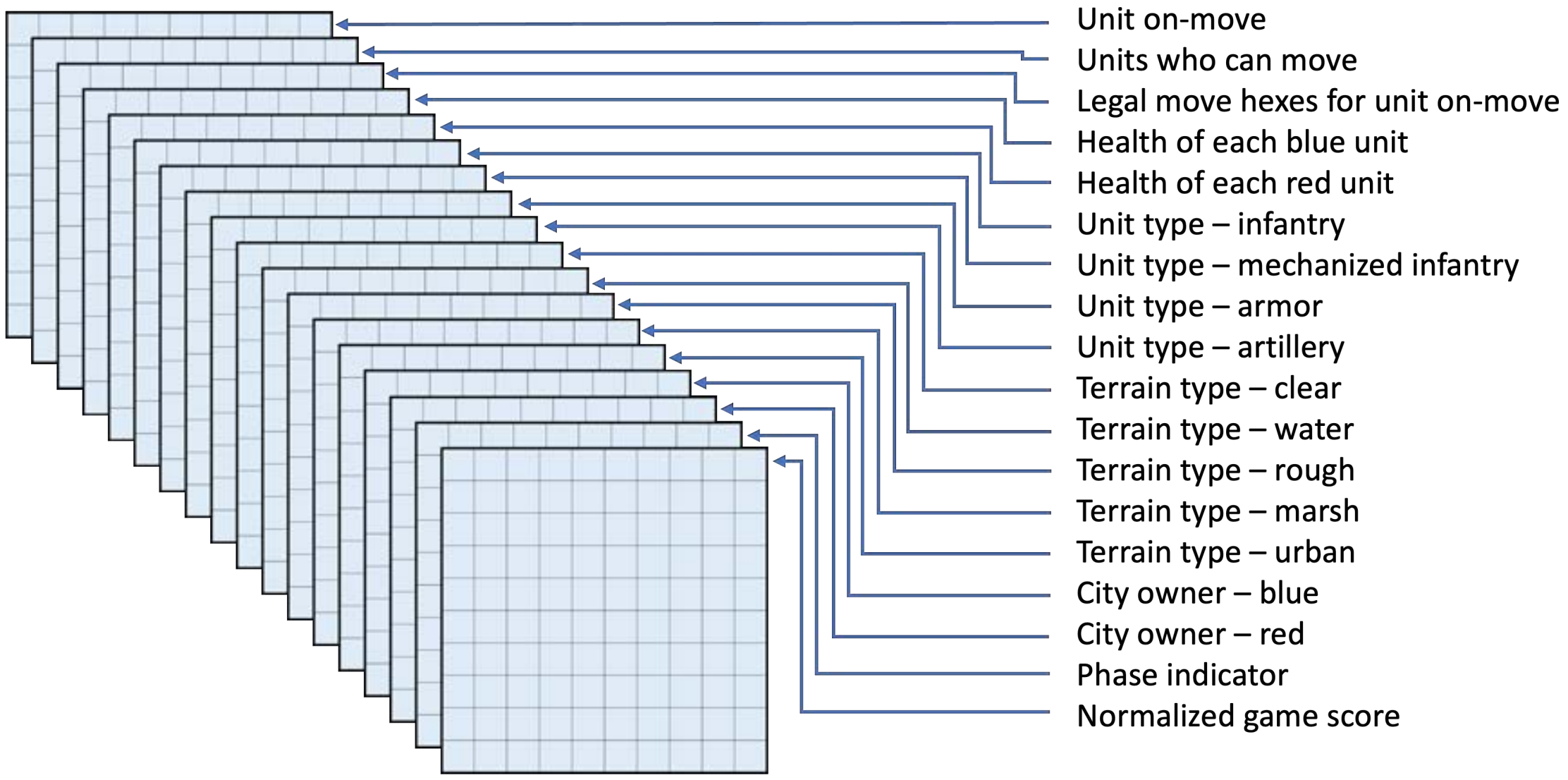

Figure 3.9. Agent Global Observation in Atlatl. The agent's observation consists of 18 channels, with each channel representing the information depicted.

This type of observation has been used in previous research using Atlatl [11], [80]–[82], [100], [101] and has proven effective in smaller-scale scenarios, up to a $5 \times 5$ assuming a `scenarioCycle` of around five or less. Scaling to larger map sizes or increasing the `scenarioCycle`, however, typically has resulted in relatively poor performance without substantial additional techniques being applied. Nevertheless, we conduct our own training and evaluation using this observation type in this study.

## 3.5 Methodology

We employ the following methodology to compare the tradeoffs between the traditional global observation approach and our approach using localized observation abstraction with piecewise linear spatial decay. We also compare the performance of each of these to our baseline *Pass-Agg* scripted agent.

### 3.5.1 Localized Observation Abstraction Using Piecewise Linear Spatial Decay

Our localized observation space takes in the game's global observation space described above and performs additional processing to compress the information into an $18 \times 7 \times 7$ observation, regardless of actual gameboard size. Even gameboards smaller than $7 \times 7$ are represented as a $7 \times 7$ with the area outside of the gameboard simply represented with zeros. To construct the localized $7 \times 7$ matrix, we first center the global matrix on the agent on-move. We then divide the entire area into 24 equal sectors (due to the outer perimeter of this matrix consisting of 24 total squares) of 15° each. Finally, we multiply each entry from the original observation by a weight $w$ calculated as a function of Euclidean distance $d$, determined by the following piecewise linear function, and visually depicted in Figure 3.10:

$$w(d) = \begin{cases} 1, & \text{for } d \leq 3 \\ 1 - 0.9(\dfrac{d - n_d}{m_d - n_d}), & \text{for } n_d < d < m_d \\ 0.1 - 0.9(\dfrac{d - m_d}{f_d - m_d}), & \text{for } m_d \leq d < f_d \\ 0.01, & \text{for } \geq f_d \end{cases} \tag{3.2}$$

where $n_d$ is the *near distance* inside which information should remain to the original scale; $m_d$ is the *middle distance* inside which information should linearly decrease in importance; $f_d$ is the *far distance* inside which information will scale at a lesser rate; and 0.01 represents the minimum weight for distances greater than $f_d$. An example of this process is shown in Figure 3.11.

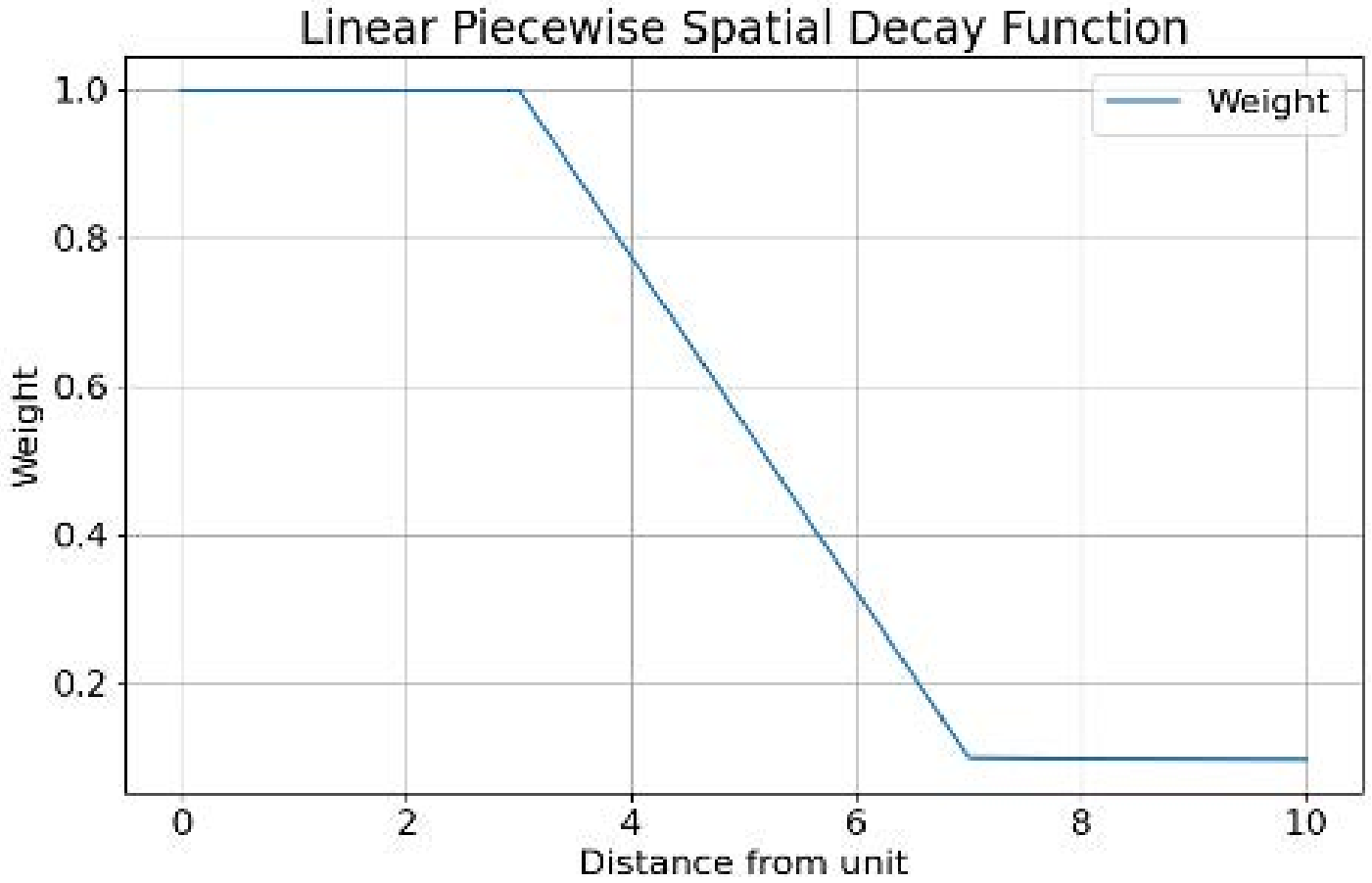

Figure 3.10. Visual Depiction of Piecewise Linear Decay Function. Not visually apparent is a very gradual negative slope that extends from $x = 7$ to $x = 100$.

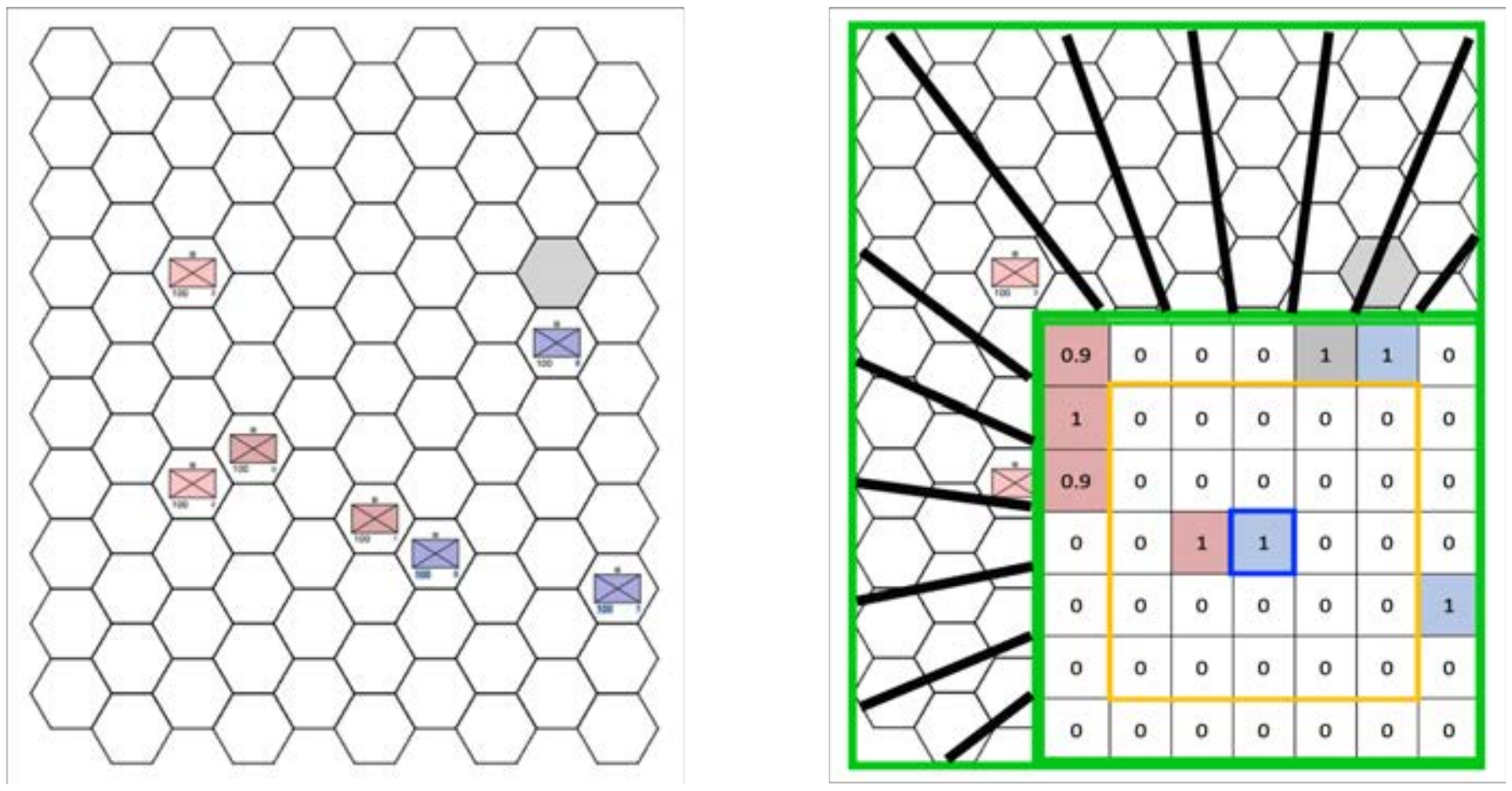

Figure 3.11. Localized Abstraction Radial Partition. The left image shows an example gameboard. The right image shows the localized abstraction of that gameboard. The area around the inner $5 \times 5$ is divided into $15^\circ$ radial sectors. Values in each radial sector are multiplied by a weight determined by the piecewise linear decay function, summed, and then inserted in the outer perimeter of the localized observation matrix.

In our experiment, we set $n_d = 3$, $m_d = 7$, and $f_d = 100$. This configuration ensures that within a radius of approximately three hexagons from the agent, the information is preserved at full resolution. We contend that this area is critical for immediate tactical decision-making, providing the agent with precise data where it is most impactful for short-term actions. For distances beyond the immediate three-hexagon radius up to approximately seven hexagons, we apply a linearly decaying factor, reducing the resolution of information to model the decrease in tactical significance with distance, but still allowing for the agent to maintain awareness of this information. From approximately a radius of seven to 100 hexagons, we decay information at a slower rate to reflect the diminishing importance of distant entities while still considering their potential strategic value. Outside of 100 hexagons, we use a minimum weight of 0.01. Using this equation, we balance the agent's focus between detailed local understanding of the areas closer to the agent, and a broader, albeit less detailed awareness of the areas distant to the agent.

A conceptual illustration of this method is shown in Figure 5. Moving from left to right, the first image in Figure 5 is an example 10×10 gameboard with a single urban hexagon, three blue units, and four red units. The second image shows the inner 5×5 grid overlay in blue. Everything within this 5×5 will remain to scale due to its multiplication by a weight of 1.0. The third image depicts the area in which each element is first multiplied by a linearly decaying weight, then summed with the other values found within the respective 15° radial sector, and finally inserted into the respective location in the outer perimeter layer of the 7×7 localized grid. For each channel in the observation, this results in a 7×7 grid where the inner 5×5 is the two layers of adjacent hexagons surrounding the unit on-move to scale and at full resolution, and the outer perimeter of the 7×7 grid represents all the information from the rest of the gameboard compressed by radial sector into a single value. To prevent the convolutional kernels from being distorted by disproportionately high values on the perimeter, we allow a maximum value of 1.0. This measure ensures that the data across both inner and perimeter hexagons is kept within a uniform scale, allowing the convolutional network to accurately interpret spatial relationships and maintain consistent performance across the gameboard. Nevertheless, due to the spatial transformation of the observation, we recognize that it may not be optimal that the same set of convolution kernels are used across both the perimeter and the inner hexagons.

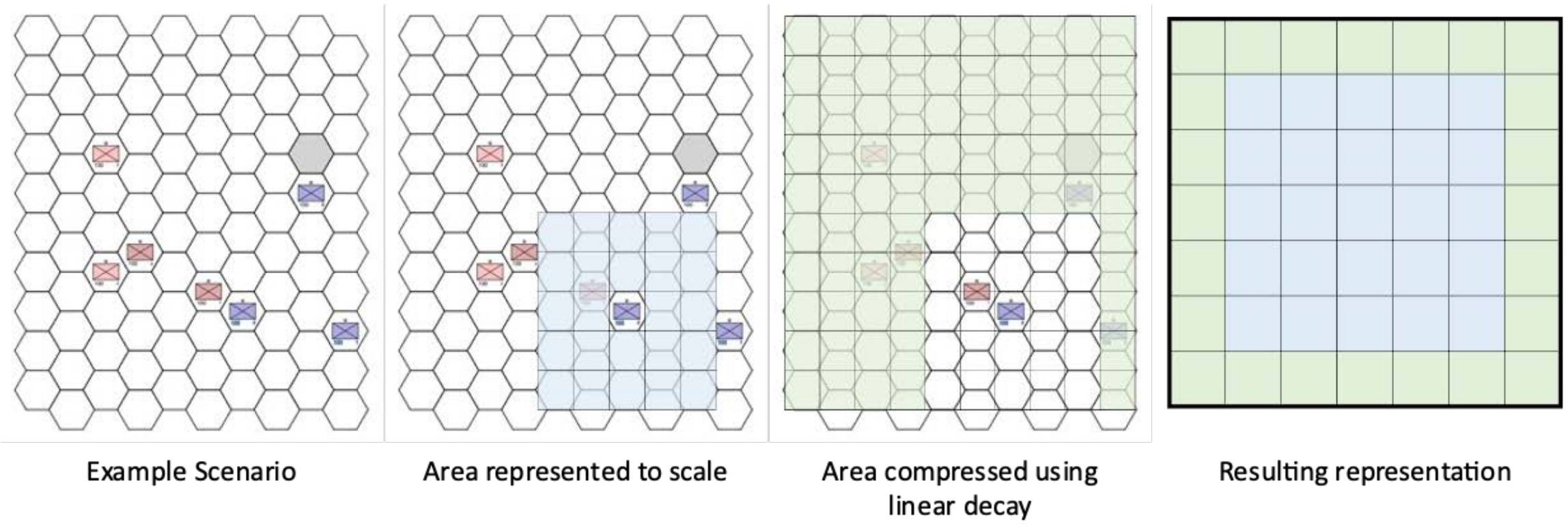

Figure 3.12. Graphical Depiction of Localized Observation.

It must be noted that fidelity of information is lost in this process. For example, a unit at 100 strength points 10 hexagons away would be represented with a value of 0.7, whereas a unit at 50 strength points that is four hexagons away would be represented with a value of 0.775. Furthermore, because we are also summing up by radial sector, the resulting value can be misleading or ambiguous as it conflates separate data points (though of the same type of information) into a single value—potentially confounding the interpretation of individual unit strengths. Nevertheless, we postulate that this approach still provides the agent with sufficient information to make optimal or near-optimal near-term decisions, while still maintaining awareness of the entire gameboard.

Within the context of Abel's [75] definition of State Abstraction $\phi : S \rightarrow S_\phi$, $\phi$ is our function that takes the global state $s \in S$, applies our localized piecewise linear spatial decay method described above, and converts it into an abstracted state $s_\phi \in S_\phi$.

### 3.5.2 Experiment

Using the Altatl simulation environment, we design and conduct the following experiment to evaluate and assess tradeoffs between the traditional global observation and our proposed localized observation abstraction approach using piecewise linear spatial decay.

#### Gymnasium Environment

The action space for our RL agent is defined as seven discrete actions, one for each adjacent hexagon, plus the option to "pass" (i.e., take no action). Legal moves are defined as either moving to an unoccupied adjacent hexagon or engaging in combat by selecting a hexagon occupied by a unit of the opposing faction.

The state space for our RL agent using the localized observation is the resultant $18 \times 7 \times 7$ observation space from taking our original global state space $s$ and passing it through $\phi$ to produce the abstracted observation space $s_\phi$. The state space for the RL agent using the global observation is the $18 \times n \times n$ observation space where $n$ is the complexity level (i.e., the length of one side of the scenario's square gameboard).

### Neural Network Architecture

For both RL-trained agents, we use a residual convolutional neural network (CNN) specifically designed to process a hexagonally structured input observation space of any size which, for this study, is either $18 \times 7 \times 7$ or $18 \times n \times n$ depending on whether we are training the agent with localized observation or global observation, respectively. The architecture uses convolutional layers to transform the input observation tensor into 64 output channels. This is followed by seven additional layers of 64 channels each. Each layer features *HexagDLy* [106] hexagonal convolutions with a kernel size of $1 \times 1$ and a stride of 1. Additionally, in each layer, we include a Rectified Linear Unit (ReLU) activation function and a residual connection. After seven layers, the resulting multi-dimensional tensor is then flattened into a one-dimensional tensor and is passed through a final linear layer. This layer maps the flattened tensor to a 512-dimensional feature vector, which is then passed through a final ReLU activation function.

### Reinforcement Learning Algorithm

We employ the DQN algorithm [103] for both RL-trained agents. The hyperparameters used were optimized through extensive hyperparameter tuning in similar scenarios, though not specific to this experiment. The final configuration includes a learning rate of 0.0002, a buffer size of 1,000,000, learning starting at 10,000 steps, a batch size of 64, and a discount factor ($\gamma$) of 0.93. The target network update interval was set to 1,000 steps. For exploration, we employed an initial epsilon ($\epsilon_i$) of 1.0, decaying linearly to a final epsilon ($\epsilon_f$) of 0.01, with an exploration fraction of 1.0. The training frequency was set to every 4 steps, with a gradient step of 1 per training update.

### Scenarios

We use randomly generated scenarios consisting of square hexagonal gameboards with one urban hexagon and no other terrain. We use scenario gameboard sizes from $3 \times 3$ up to $12 \times 12$ in increments of one, with each representing an increase in complexity level, where complexity level 3 is represented by a $3 \times 3$, complexity level 4 by a $4 \times 4$, and so on. Examples are shown in Figure 3.13. Each game begins with a random number of entities per faction with a minimum and maximum number computed as a factor of the length of

the gameboard, where:

$$num_units_{min} = \frac{gameboard_length}{2} \quad (3.3)$$

$$num_units_{max} = gameboard_length \quad (3.4)$$

For example, for a $5 \times 5$ gameboard, the scenario would start with a random number of units per faction between three and five; whereas for a $10 \times 10$ scenario, the random number of starting units per faction would be between five and 10. This allows for a reasonable maximum and minimum number of units per faction on the gameboard at the start of each scenario.

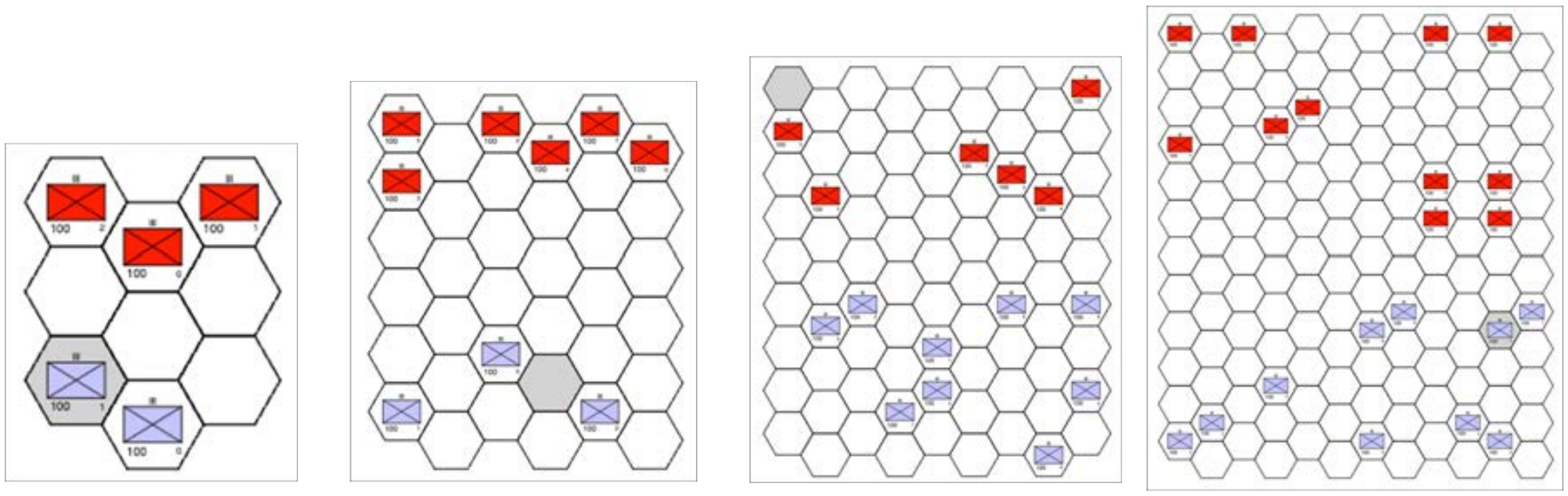

Figure 3.13. Example Scenarios for Observation Abstraction Experiment. From left to right, illustrated are Complexity Levels 3, 6, 9, 12.

Each scenario also includes one urban hexagon randomly placed according to force ratio. If one faction has a smaller force ratio (i.e., fewer units as compared to the opposing faction), the urban hexagon is placed on their side of the gameboard. If the force ratios are equal (i.e., both factions have an equal number of units), the urban hexagon is placed in a neutral location along the middle axis of the board.

We set the number of phases in the game as:

$$num_phases = 4 * gameboard_length \quad (3.5)$$

where each phase is one entire turn for one faction (i.e., one faction is allowed to make one legal move for each of its available entities). Setting the number of phases to this value provides enough turns for a unit to go from one end of the gameboard to the opposite end and return, likely giving them enough turns to execute complex maneuvering if warranted.

**Training**

We use *Pass-Agg* as the adversary behavior model and train each of our models for 10 million steps using `scenarioCycle` = 0 (i.e., scenarios generated independently at random without any intentional repetition).

To learn effective behaviors, we employ the Boron reward system that balances defeating the opposing faction and occupying urban hexagons with preserving its own force. Our rewards are computed at each time step using the following equation:

$$R_{engineered} = \max(R_{raw}, 0)\frac{S_c}{S_o} + B_t I_t \qquad (3.6)$$

where $R_{raw}$ is the difference in game score between the current time step and the previous time step; $S_c$ is the current total friendly strength; $S_o$ is the original total friendly strength; $B_t$ is a terminal bonus reward of 25 points that research shows discourages units from moving into the adversary units' attack range during the last turn of the game; and $I_t$ is a terminal game state indicator that takes on a value of 1 if the game is terminal or 0 otherwise.

The learning-curve graphs for each of our RL-trained models are depicted in Figure 3.14. These graphs show the mean rewards obtained through evaluation over the course of the training process. Model evaluation during training was conducted every 100,000 training steps for 100 episodes each.

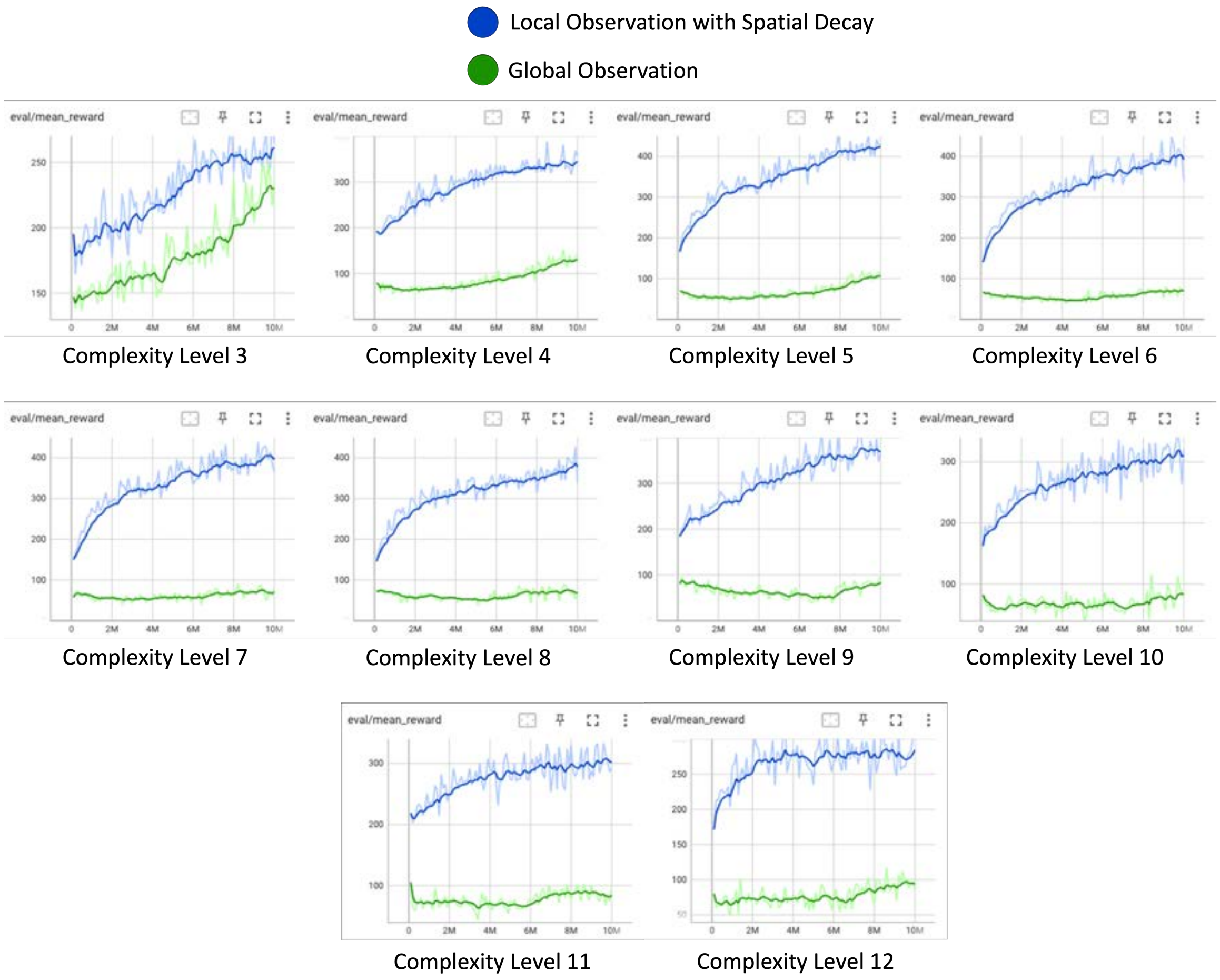

Figure 3.14. Learning Curves for RL-Trained Agents. The y-axis is the mean reward obtained during evaluation. The x-axis is the number of training steps performed. The blue line depicts the RL-trained model using the localized observation abstraction with piecewise linear spatial decay; the green line depicts the RL-trained model using the global observations

As the graphs clearly illustrate, our proposed method clearly outperforms the global observation method by a large margin in training. Even in the simplest scenario (complexity level 3) where we expected the global observation to outperform our method, we find that the local observation with spatial decay approach was superior.

**Evaluation**

While our RL-training learning-curves suggest that our method is superior to the global observation approach, we confirm this by performing separate and distinct evaluations. These evaluation runs are specifically aimed at validating model performance using the actual game scores as the benchmark, rather than mean rewards during training.

Of note, for conciseness, in the rest of this chapter we use the term *Local* to refer to our RL-trained model utilizing the localized observation abstraction with piecewise linear spatial decay; the term *Global* to refer to our RL-trained model utilizing the global observation space; the term *Scripted* to refer to the *Pass-Agg* agent; and the term *Random* to refer random-acting agent (i.e., an agent that takes random actions).

In this study, we also utilize the *Scripted* behavior model as the adversary model in our evaluations. Additionally, we evaluate *Scripted* vs. *Scripted* as our scripted model baseline. Given the anticipated superiority of *Scripted* agents over RL-trained models at higher complexity levels—attributable to the exponential growth in state space complexity that RL models tend to struggle with—we aim to assess the extent of improvement an RL-trained agent offers over a scripted agent while also seeking to determine the specific complexity threshold when this relationship reverses. The performance of the *Scripted* model thus serves as a benchmark to ascertain the point where RL-trained models no longer exceed the effectiveness of a scripted approach. However, it is important to acknowledge that despite the anticipated efficiency of the *Scripted* agents in more structured environments, they are not inherently optimal across all scenarios. *Scripted* agents operate under predefined rules that may not adapt well to unexpected changes or novel situations encountered in more dynamic or complex environments. Lastly, we also include an evaluation against a random-actions model to assess the point where our RL-trained models perform no better than, or converge to, the behavior of a random actor.

We conduct a total of four evaluations against the baseline *Scripted* adversary to assess performance:

- *Local* vs. *Scripted*
- *Global* vs. *Scripted*
- *Scripted* vs. *Scripted*
- *Random* vs. *Scripted*

Table 3.1. Mean Scores ($\bar{x}$) across 100,000 Games for Each Model at Each Level of Complexity

| | | Complexity | | | | | | | | | |
|---|---|---|---|---|---|---|---|---|---|---|---|
| | | **3** | **4** | **5** | **6** | **7** | **8** | **9** | **10** | **11** | **12** |
| **Model** | **Local** | 181.4 | 227.7 | 225.3 | 88.4 | -154.0 | -394.6 | -491.4 | -620.2 | -860.2 | -908.6 |
| | **Global** | 31.5 | -203.9 | -488.6 | -653.8 | -808.8 | -885.5 | -1039.0 | -1116.0 | -1229.9 | -1307.1 |
| | **Scripted** | 50.0 | 97.4 | 124.3 | 124.5 | 133.2 | 130.6 | 133.1 | 134.9 | 141.9 | 128.7 |
| | **Random** | -339.5 | -465.7 | -628.8 | -724.2 | -865.4 | -948.0 | -1078.6 | -1158.4 | -1284.7 | -1364.6 |

Table 3.2. Raw Standard Error of the Mean (SEM) of Each Model's Scores across 100,000 Games

| | | Complexity | | | | | | | | | |
|---|---|---|---|---|---|---|---|---|---|---|---|
| | | **3** | **4** | **5** | **6** | **7** | **8** | **9** | **10** | **11** | **12** |
| **Model** | **Local** | 0.9 | 1.2 | 1.7 | 2.0 | 2.2 | 2.1 | 2.2 | 1.8 | 1.3 | 1.3 |
| | **Global** | 1.0 | 1.0 | 0.8 | 0.8 | 0.7 | 0.8 | 0.9 | 0.9 | 1.0 | 1.0 |
| | **Scripted** | 1.0 | 1.3 | 1.7 | 1.9 | 2.3 | 2.6 | 3.0 | 3.3 | 3.7 | 3.9 |
| | **Random** | 0.5 | 0.6 | 0.6 | 0.6 | 0.7 | 0.7 | 0.7 | 0.8 | 0.8 | 0.8 |

For each of our evaluation runs, we use the scenario parameters specified in Section 3.5.2.

## 3.6 Results and Discussion

With each model, we run an evaluation consisting of 100,000 games using `scenarioCycle` = 0 for each behavior model at each complexity level against our baseline *Pass-Agg* adversary model. The mean scores are presented in Table 3.1. Overall, we see that *Local* outperforms *Global* across all complexity levels by a large margin. Furthermore, we also see that *Local* outperforms *Scripted* in complexity levels 3 through 5 by a large margin and, as expected, begins to fall off as complexity increases.

To verify that these differences in mean scores between our models are statistically significant, we set $\alpha = 0.05$ and run the Tukey-Kramer Honest Significant Difference (HSD) test. This test conducts pairwise comparisons between all possible pairs and details which specific groups' means are significantly different from each of the other groups. We find statistical significance between mean scores for each model across all complexity levels used in our experiment, each producing a p-value of $< .0001$. We show the Standard Errors of the Mean (SEM) for each model based on complexity levels in Table 3.2.

Figure 3.15 depicts the line plot of *Mean Score* $\bar{x}$ by *Complexity* for each model evaluated. Figure 3.16 shows the same data normalized to the random-actions behavior model (*Random*) using the standard normalization formula:

$$x_{Norm} = \frac{x - \bar{x}_{Random}}{\sigma_{Random}} \tag{3.7}$$

We normalize to *Random* for visualization purposes as we contend that an untrained neural network should produce random actions and can, therefore, be representative of the reasonably worst-case score. However, we acknowledge that even with training, in extreme cases, it is still possible for a neural network to do worse than a random actor. Nevertheless, we use *Random* as our baseline for our zero line for graphical purposes to better observe how and when *Global* and *Local* converge to *Random* (i.e., when our models perform no better than a random actor).

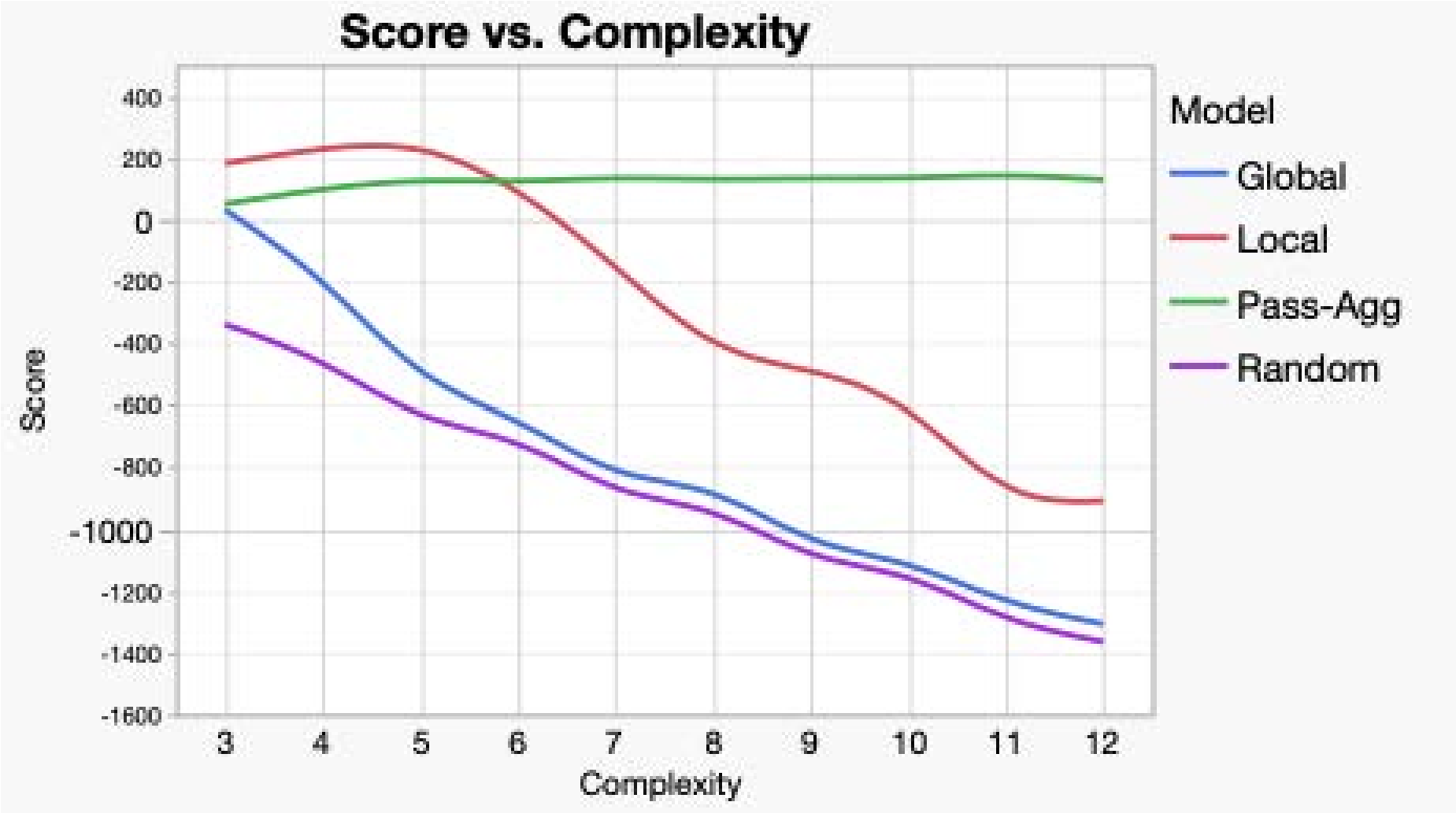

Figure 3.15. Mean Score vs. Complexity Graph.

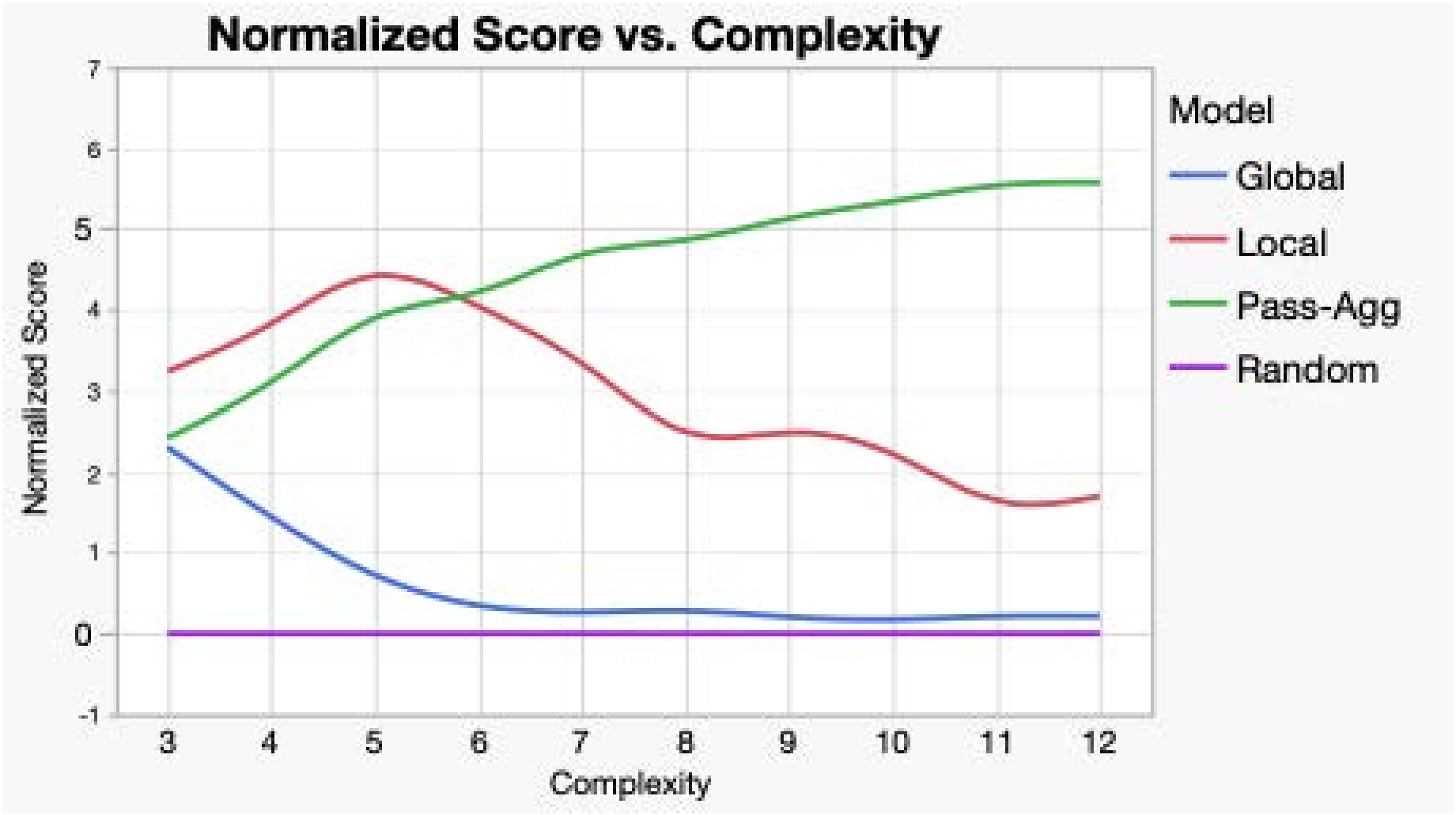

Figure 3.16. Normalized Mean Score vs. Complexity Graph.

As shown in both Figures 3.15 and 3.16, *Local* significantly outperforms *Global* across all complexity levels. It is clearly evident that *Global's* performance begins to decrease from the beginning and then converges to our theoretical zero of *Random*. *Local*, on the other hand, maintains improved performance over *Scripted* until complexity level 5, after which it begins to decline in performance until it converges at a level still considerably above *Global*.

Surprisingly, while we expected *Global* to outperform *Local* in the smaller scenarios (e.g., in complexity levels 3 and 4), we found that *Local* outperformed *Global* by a large margin (149.940 points or 475.975% even in the simplest of scenarios). This increased performance of *Local* over *Global*, even in the smaller scenarios, may be due to the localized observation always being centered on the agent on-move. This could facilitate learning as this consistent agent-centric perspective could allow for quicker generalization. Visual replays of these scenarios confirm the superior performance of *Local* over *Global* across all scenarios.

## 3.7 Conclusion

Overall, this research presents a compelling case for implementing a localized observation abstraction when training models using RL, specifically within environments where spatial relationships may be crucial. Whereas we hypothesized a trade-off space between the global and localized observation approaches, we find that a localized observation with spatial decay consistently outperforms a global observation approach across all levels of complexity examined. The superior performance of using localized observation is particularly striking in the smaller-scale scenarios, as it was anticipated that the global observation approach would be at least as good as the localized approach, if not better. We posit that the efficacy of the localized abstraction approach is likely due to the agent's improved ability to generalize when centered in the observation space, significantly enhancing the learning process and decision-making ability. This approach balanced reducing state-space complexity with the retention of relevant information, thereby better optimizing the agent's performance.

Revisiting Abel's [75] three desiderata for useful state abstraction (preserving near-optimal behavior, being learnable and computable efficiently, and reducing the time or data needed for effective decision-making), we find our observation abstraction clearly accomplishes all three. Our agents performed better than agents using global observations given a set training budget; our abstraction of the state space proved more efficient than training the agent to reach the same performance threshold using the global observation space; and our abstraction method reduced the time needed for training to reach a desired performance threshold.

Regarding our overall HRL architecture, this study suggests that limiting the board size (or *Objective Area*) of our lowest-level (*Operator*) agent to a $5 \times 5$ space may result in superior, yet manageable, results over our baseline scripted *Pass-Agg* agent.

Furthermore, the outcomes of this study underscore the potential of localized observation abstractions to become a pivotal component in the application of RL in complex, dynamic environments in general. This study, specifically, paves the way for future investigations into more sophisticated observation abstraction methods to better enable RL scalability by demonstrating the limitations of a global observation approach and the advantages of a localized observation approach with a spatial decay component.

THIS PAGE INTENTIONALLY LEFT BLANK

# CHAPTER 4: A Multi-Model Approach

This chapter's core material has been accepted for publication and presentation at the *SPIE Defense + Commercial Sensing Conference* [51] and is pending publication. It was honored with the *Artificial Intelligence and Machine Learning Best Student Paper Award* in *Artificial Intelligence and Machine Learning for Multi-Domain Operations Applications VI*. The material is extended here to provide additional elaboration and detail.

In this chapter, we examine the application and efficacy of a multi-model approach in modeling intelligent agent combat behaviors for wargaming, focusing on a comparative analysis between single-models (i.e., a traditional individual behavior model) and multi-models. The core of this research lies in developing a multi-model framework that incorporates various AI methodologies, which include supervised learning, RL, and scripted behaviors. This framework is then evaluated through an experimental design comparing the performance outcomes of single-models against those of composite multi-models within the Atlatl simulation environment.

The findings presented here aim to demonstrate the benefits of a multi-model over a single-model approach in enhancing agent performance, indicating a promising direction for future AI development in the complex decision-making environments characteristic of wargaming. Through our experiment and analysis, we seek to validate this approach and gain insights into the advantages and implications of incorporating this framework within our overall HRL framework. Furthermore, this chapter also individually contributes to the ongoing research on the scalability and effectiveness of AI within the domain of combat simulations.

 Paper titled "A Multi-Model Approach to Modeling Intelligent Combat Behaviors for Wargaming" has been accepted and is pending publication in *Artificial Intelligence and Machine Learning for Multi-Domain Operations Applications VI*.

## 4.1 Background

Like with abstraction from the previous chapter, the concept of leveraging multiple models in systems to enhance problem-solving is not new. Research into using multiple models to address complexity goes back decades [107]–[109].

The prevalent approach used to implement the idea of combining many models is through ensemble modeling, a methodology that has been supported by a growing body of research within the ML community [110]. In essence, ensemble modeling combines multiple predictive models to improve accuracy, reduce variance, and enhance decision-making by leveraging the strengths of each constituent model. Figure 4.1 depicts a generic ensemble architecture where input is provided to a collection of learners (or models), who each provide an output. These outputs are then combined in some form by the system, generally through a weighted voting mechanism, to finally provide an output or prediction. While there have been many nuanced approaches to date, there are three general areas that have led to the current implementation of ensemble methods: *combining classifiers*, *ensemble of weak learners*, and *Mixture of Experts (MoE)*.

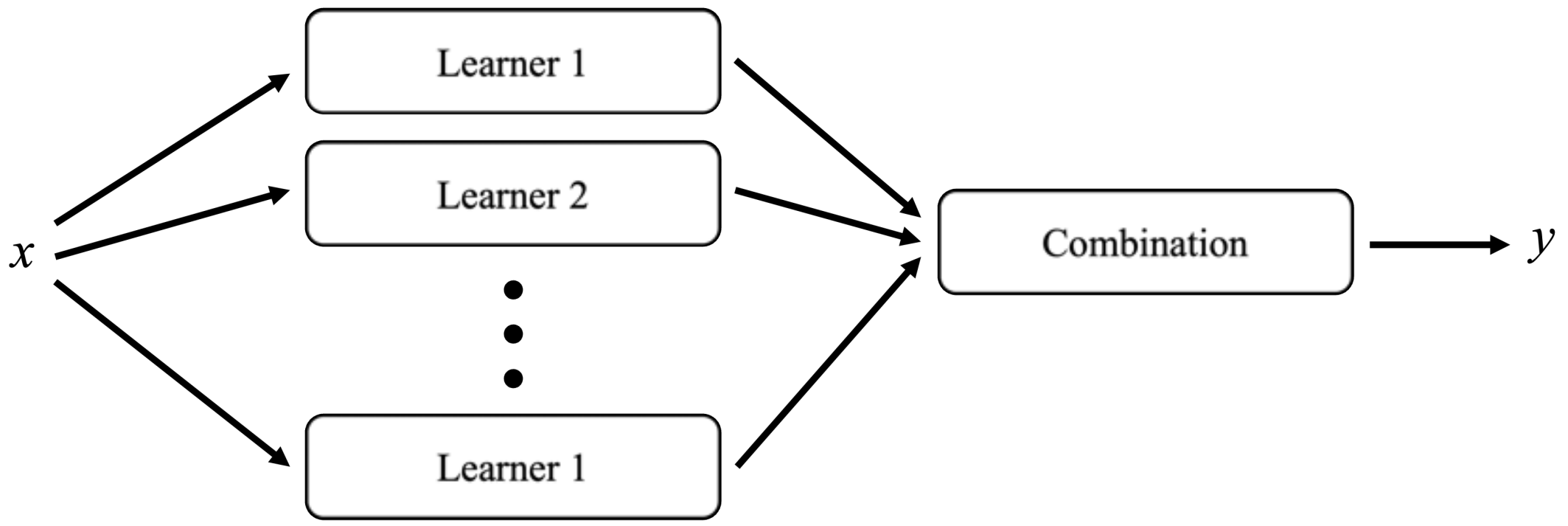

Figure 4.1. Common Ensemble Architecture. Adapted from [110].

*Combining classifiers* refers to a strategy in ML where multiple classification models (classifiers) are used together to improve the overall accuracy and robustness of predictions [110]. This approach is based on the premise that different classifiers, when combined, can complement each other's strengths and mitigate individual weaknesses, leading to better performance than any single classifier could achieve on its own. *Combining classifiers* can be done in various ways, but some common methodologies include *voting*, *bagging*, and *boosting*. *Voting* simply allows each classifier to vote, sometimes weighted, and the classification with the most votes is then chosen as the final prediction. *Bagging*, or *bootstrap aggregating*, involves training different classifiers on different subsets of the data and then aggregating their predictions, often by voting, to make a final decision. *Random Forests* is an example of this approach. Lastly, *boosting* involves sequentially training classifiers in such a way that each classifier focuses on correctly classifying instances that were misclassified by previous classifiers. The final prediction is then made by a weighted vote of all classifiers, where weights are determined by each classifier's accuracy. *AdaBoost* is an example of this type of methodology.

Whereas the *combining classifiers* approach seeks to develop strong classifiers by focusing on the integration of models through enhancing stability, reducing variance, and improving accuracy, the *ensemble of weak learners* method is a strategy that combines multiple "weak" classifiers in situations where it is advantageous or necessary to use simple models (due to computational constraints or the need for model transparency) that together can better capture complex patterns more effectively than a single complex model. By "weak" classifier, it is typically implied that each classifier may only be slightly better than random guessing, though they can be stronger as well. However, even weak classifiers, when combined with the other classifiers can create a model with much higher accuracy. The key principle behind this approach is that by aggregating the predictions of several weak learners, the ensemble can correct for the mistakes of individual learners and enhance overall prediction performance [110]. Ensemble methods that leverage weak learners also include *boosting* and *bagging* as described above. The main advantage of using ensembles of weak learners is that they can achieve high accuracy and robustness, which might not be possible with any single weak learner. These methods are particularly useful when dealing with complex data sets where modeling the underlying patterns with a single model is challenging.

The third area of ensemble modeling is *MoEs*. This method is an ensemble learning technique that divides a complex problem into smaller, more manageable parts, each handled by a specialized model known as an *expert*. The decision on which expert to favor for a given input is made by a *gating network* that learns to weigh the input of each expert based on the current context or characteristics of the data [110]. Each expert is trained to become proficient in a specific segment of the problem space, effectively becoming a specialist for certain types of inputs or scenarios. The gating network, on the other hand, is trained to analyze the inputs and allocate varying levels of trust or weights to the predictions made by each expert, thereby dynamically selecting the most appropriate experts for each given input. The final output of the MoE model is a weighted combination of the predictions from all experts, where the weights are determined by the gating network. This approach leads to a divide-and-conquer strategy where the overall model tries to learn a mixture of parametric models and then employ combining rules to obtain an output.

In essence, the idea of using multiple different models draws on the principle that no single algorithm excels at all types of tasks or data conditions. This diversity in model performance has led to the development of systems that combine models to improve overall efficacy. The necessity for these ensemble approaches arises from the intricate nature of many real-world problems, which often contain varied and complex data structures that challenge single-model solutions. By aggregating predictions from multiple models, ensemble methods have been shown to reduce error variance and avoid overfitting, leading to more accurate predictions [111], [112].

Although these conventional ensemble methods leverage the strengths of diverse models to generally improve performance and reduce variance and bias, they are not without limitations [113]. Because ensemble methods generally involve averaging or some form of weighted combinations of results, these methods require careful recalibration of weights or retraining when adding new models, which can be complex and time-consuming. As the number of models in the ensemble increases, the complexity of managing and dynamically adjusting the weights can scale poorly. Furthermore, ensemble modeling typically requires all sub-models to run concurrently, often leading to longer training times and slower prediction speeds.

While ensemble modeling techniques aim to mitigate overfitting by pooling predictions, they do not inherently address the issue of selecting the most appropriate model for a given context or state, potentially limiting their adaptability and efficiency in dynamic or even more complex environments where a single model may be vastly superior to the other models in the collection. This is especially true in tasks involving discrete actions or decisions, such as in the domain of intelligent agents, where the output of a model often corresponds to specific, distinct actions. Averaging outputs may not apply well to these instances, where a clear, decisive action is often required. In these cases, averaging behaviors or outputs may instead dilute the effectiveness of any individual behavior, potentially leading to indecisiveness or suboptimal actions. This dilution occurs as the unique strengths and expert capabilities of individual models become less impactful in the aggregated decision-making process, making it difficult to maintain the high performance that might be achieved by a more targeted approach. For example, in a navigation task, averaging the directions suggested by multiple models might result in a path that is neither optimal nor practical. Selecting the best single model for a given context, however, might better ensure that decisions are clear and based on the best available strategy. Although it would theoretically be possible to train an MoE model to learn to assign zero weights to all but one behavior model, it would likely require extensive engineering and training—thus becoming more prone to overfitting and effectively negating the main reasons for using ensemble methods.

The highlighted challenges emphasize the need for a framework that not only leverages multiple specialized models but can also, by design, dynamically select the single most appropriate model given the current environment and context. In addition to eliminating the overhead of running all models within an ensemble simultaneously (as is necessary by ensembling methods in order to combine outputs), this proposed strategy preserves the specialized expertise of each model while potentially also improving performance over single-model approaches. Moreover, this proposed approach allows for the employment of well-validated behavior models in a way that each operates under conditions for which they were specifically designed, ensuring that outputs are validated responses to inputs rather than a conflation of multiple model outputs. This not only enhances performance but also increases the transparency of the predictive process as compared to conventional ensemble methods.

## 4.2 Related Works

The idea of "many-model thinking" is explored in detail in *The Model Thinker: What You Need to Know to Make Data Work for You* [114]. In this book [114] and related paper [115], Page argues that an ensemble of models allows us to make up for shortcomings in any one of the models. This is shown in experiments by Hong and Page [116], where a group composed of random intelligent problem solvers outperformed a group of the top-performing problem solvers—highlighting the trade-off between diversity and ability.

Jacobs and Jordan [117] introduce the concept of adaptive mixture of local experts, which served as the foundation for the concept of MoE [118]. In their paper, the authors investigate various error functions to improve the expert networks learning process within the mixture of local experts methods. In this model, as discussed above, the system is composed of multiple networks (or experts), each specialized in handling subsets of the training data, in addition to a gating network that determines which experts should be active for a given input. This paper demonstrates the effectiveness of the approach with a vowel discrimination task, where the system effectively divides the problem among experts, each focusing on different aspects of the task. The division into experts allows the model to handle complex tasks more efficiently than a single, monolithic network [117].

While Jacobs and Jordan [117] use separate expert networks, the gating network still uses weights to decide how each expert should contribute to the final decision, as is characteristic in ensemble learning. Essentially, the gating network assigns weights to the experts' outputs based on the current input, and then these weighted outputs are combined to form the final output. Our work diverges from this methodology in that we seek to differentiate between models, dynamically selecting the single best model rather than weighting and combing outputs so as to not dilute the specialization benefits of any given individual model.

Results extending MoE include studies by Tang et al. [119], Eigen et al. [120], Dobre and Lascarides [121], and Krishnamurthy et al. [122]. Tang et al. [119] enhance the MoE architecture by partitioning input data to assign specific local regions of the input space to individual experts utilizing a self-organizing feature map. Eigen et al. [120] extend MoEs by creating a stacked *Deep Mixture of Experts* model that improves performance on the MNIST dataset through the use of multiple sets of a combination of gating and experts. Dobre and Lascarides [121] show success applying a combination of MoE and transfer learning to the board game *Settlers of Catan*. Krishnamurthy et al. [122] introduce a new MoE model architecture and training method that uses attentive gating rather than the traditional MoE gating. While these all contribute to advancing MoE approaches, they still rely on weighted averaging to produce a decision.

More recently, and most relevant, Helfenstein [123] explored the integration of the MoE approach with MCTS for the game of chess, leveraging deep learning to better address the complexity of the game. They introduce a framework where multiple specialized models, or experts, are deployed to respond to specific phases of the game—opening, middlegame, and endgame—guided by a strategic gating mechanism. This results in a sparsely activated model architecture that offers computational efficiencies. Their findings show a notable improvement in playing strength as compared to traditional single-model systems, underscoring the potential of leveraging expert knowledge and strategic principles in neural network designs for board games. This work highlights not just the value of using multiple experts, but value of combining multiple types of AI approaches as well.

In their study, Rodgers et al. [124] develop an ensemble agent in the arcade game *Ms. Pac-Man*. The paper applies ensemble principles to real-time decision-making, resulting in complex agent behaviors composed of simpler elements or *voices*. Each *voice* focuses on a single task or goal, contributing to the final decision made by an arbiter. They accomplish this by building an Ensemble Decision Systems (EDS), illustrated in Figure 4.2. The *pre-filter* uses the game emulator to remove the moves that lead to certain death. The *voices* are simply each agent trained on a simple task. The *arbiter*, which is the heart of the system, takes the outputs of each *voice*, combines them, and then generates a final decision [124]. The Ensemble Agent set a new AI world record in performance for Ms. Pac-Man, beating a monolithic agent based on MCTS that used the same forward model. This approach not only allowed for flexibility in adding or removing components without affecting the overall system but also achieved a level of efficiency by separating behaviors into reactive or deliberative actions, depending on the task [124].

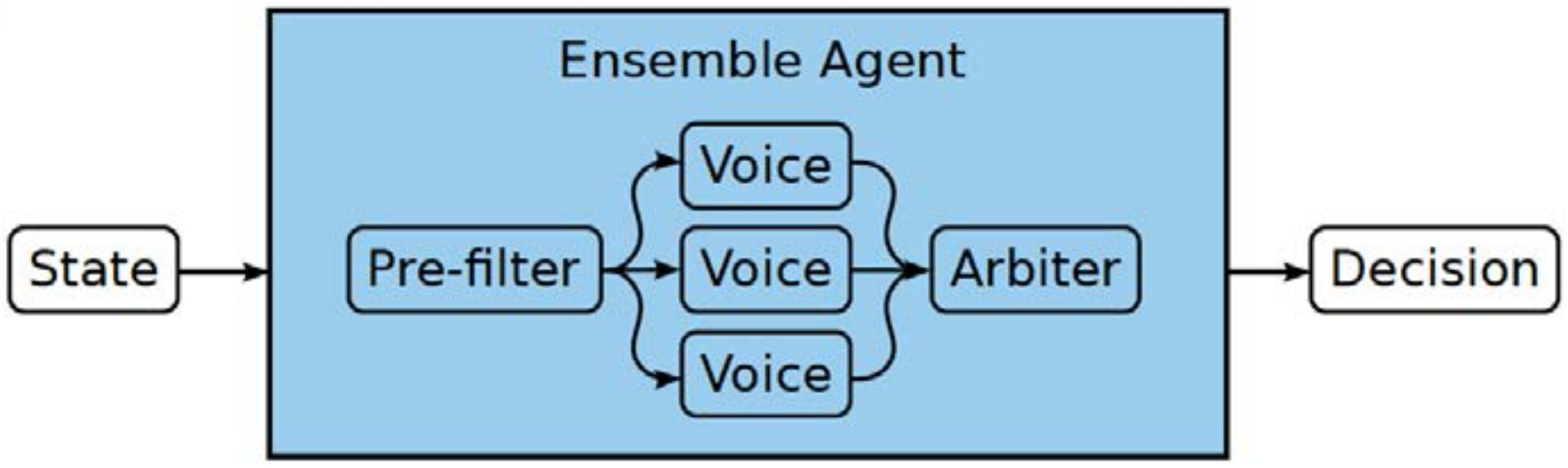

Figure 4.2. Ensemble Decision System. Source: [124].

This study is relevant and insightful to our work due to its focus on ensemble methods for decomposing complex problems into manageable sub-goals. While our approach emphasizes selecting the best model given the current state of the game, rather than combining outputs, the underlying concept of utilizing a diverse set of specialized agents or models to address different aspects of a problem is shared. However, our specific method diverges in that we focus on evaluating and selecting the most suitable model for the current game state, potentially bypassing the need for extensive validation required for each model in traditional ensemble approaches. This could offer an advantage in terms of efficiency and adaptability in rapidly changing game environments.

Anderson et al. [125] continue their work in EDS for General Video Game Playing (GVGP) and seek to create agents capable of performing well across a diverse array of video games. This study extends their previous study [124], continuing to combine multiple algorithms—each with its own strengths and weaknesses—to enhance overall performance and adaptability. By implementing different configurations of an EDS and analyzing its performance against individual algorithms within a General Video Game AI (GVGAI) competition framework, Anderson et al. demonstrate the potential of this approach to increase the generality and effectiveness of game-playing agents [125].

Overall, although the EDS did not outperform all individual algorithms, Anderson et al. [125] showed the value of an ensemble approach as it maintained individual algorithm strengths and added some generality to the agents by allowing it to succeed at more types of games. This approach is shown to indeed be more effective than individual algorithms alone in generalizing across various games by leveraging the complementary strengths of different individual gaming AIs rather than modifying individual algorithms to overcome their inherent limitations. However, in games with homogeneous dynamics, such as the singular domain of combat simulations, EDS may not offer any significant advantages. Furthermore, in EDS, as in most ensemble systems that combine predictions, the decision-making process of different algorithms may conflict with each other, leading to suboptimal performance as compared to that which could potentially be obtained by using the single best algorithm for the given task. Lastly, while EDS aims to improve generalization across different games, this approach may sometimes be at odds with the need for specialized strategies in certain games or applications. While our work is motivated by their insights and conclusions, we approach the problem from the perspective of dynamically differentiating between models rather than combining models.

In this section, we have reviewed various ensemble methods and their applications. Traditional ensemble techniques typically aggregate model outputs, which can obscure individual strengths and complicate model validation. In contrast, our approach focuses on selecting the most suitable model for the current state of the game, avoiding the need for extensive validation of each model. This methodology simplifies model management while potentially enhancing overall adaptability in more dynamic environments. By prioritizing model selection over model output aggregation, we maintain the unique advantages of each model, allowing for more targeted and effective responses in complex scenarios. This approach represents a significant departure from conventional ensemble methods, emphasizing efficiency and specificity in model usage.

## 4.3 Research Objective

While we borrow the idea of using many different models to solve a complex problem, rather than taking a divide-and-conquer approach as is typically associated with MoE [118] or using a voting mechanism to combine model outputs as is typical of ensemble learning [126], in this study, we aim to leverage a collection of models by dynamically selecting the best model at each time step given the current state of the game.

Because ensemble methods combine predictions in one way or another, they require careful selection of models based on their performance, as ensemble methods can sometimes fail to improve predictions or even perform worse than individual models if not properly managed [113]. Furthermore, ensemble methods can depend significantly on how the diversity and quality of models used in the ensemble are managed [113]. Using an approach that seeks to differentiate between the models and only select the best model given a specific situation, however, we may be able to leverage a diversity of models (which include a combination of RL-trained and scripted agents) without concern for how an individual model may impact the collection as a whole. Furthermore, instead of using heuristics or mental simulations to evaluate how each model in our collection would perform in a given situation—as is typically done by humans [127]—we instead employ pre-trained neural networks to accomplish this evaluation. This helps overcome the known difficulties of creating heuristics for complex problem spaces, as well as the inefficiencies of running complete simulations—which often require computationally-demanding Monte Carlo runs to account for stochastic policies. Nevertheless, while we do not claim in this study that the proposed approach outperforms traditional ensemble approaches in general, we postulate that this approach may be better suited for the development of intelligent agents for combat simulations.

Of note, in this dissertation, we use the term *multi-model* to refer to a model composed of a collection of individual behavior models (or policies) as is presented in this chapter. We use the term *single-model* to refer to an individual behavior model which may be used as a stand-alone model or within a *multi-model* framework.

Through this study, we seek to investigate the efficacy of an agent that employs a *multi-model* framework over an agent that employs the traditional *single-model* framework. Furthermore, we seek to determine if a *multi-model* composed of more *single-models* outperforms one composed of fewer *single-models*.

### 4.3.1 Methodology

To accomplish this research objective, first, we construct several *Multi-Models* composed of different combinations of *Single-Models*. Next, we generate the data to train a neural network that serves as a *Score Prediction Model* for each *Single-Model* within the *Multi-Model*. We then develop a simple evaluation function for our *Multi-Model* to select the appropriate *Single-Model* based on the current game state. With these, we conduct an experiment using the Atlatl simulation environment and measure the mean game score of each *Single-Model* and each *Multi-Model*. Lastly, we analyze the results and provide our conclusions.

## 4.4 Design of Experiment

We designed our experiment to conduct pairwise comparisons between each of our behavior models, which include eight single-models and four multi-models, for a total of 66 comparisons. Our objective is to investigate whether there exists a statistically significant difference in the mean scores across 100,000 games between the single-models and each of the multi-models. Additionally, we aim to gain an understanding of whether a multi-model composed of more single-models exhibits better performance over a multi-model comprised of fewer single-models.

### 4.4.1 Behavior Models

For this study, we use four scripted models, four RL models, and four multi-models, for a total of 12 models.

#### Scripted Models

The scripted behavior models include the following:

- *Pass-Agg*: implemented as described in Section 3.4.5.
- *Pass*: employs the *Pass-Agg* algorithm, but always with a defensive posture.
- *Agg*: employs the *Pass-Agg* algorithm, but always with an offensive posture.
- *Burt-Plus*: this model is similar to *Pass-Agg* in that it adopts either an offensive or defensive posture based on the combat power ratio, but it includes two key changes. First, instead of using a uniform distribution to select a target, it attacks the target with the least strength first. Second, the scoring system that evaluates potential hexes for unit positioning prioritizes less surrounded locations and incorporates weighted distances to enemies, cities, and friendly units. Like *Pass-Agg*, this method aims to strategically guide units toward optimal positions by taking into account enemy locations and city locations, but also adds the consideration of friendly force locations. It then uses weighted scores to select its target move hexagon.

#### Reinforcement Learning Models

The RL models used in this study are all trained in the Atlatl simulation environment using an observation space consisting of 18 channels of the $5 \times 5$ hexagonal gameboard (as described in Section 3.4.6 and shown in Figure 3.9). This input is then passed through the residual CNN described in Section 3.5.2. Of note, although we describe our localized observation abstraction in the previous chapter, for this experiment we use the conventional global observation. This decision is made to avoid confounding results, ensuring that the outcomes of each of our experiments remain distinct and attributable to their respective variables for clearer assessment and analysis.

We train these RL models from scratch using the DQN learning algorithm and a training budget of 2 million steps. Parameter values include a learning rate of 0.0001; batch size of 64; buffer size of 1,000,000; learning starting at 50,000 steps; discount factor ($\gamma$) of 0.99;

target network update of 1,000 steps; training frequency of 4 steps; target update interval of 1,000; maximum gradient norm of 10; gradient step per update of 1; and initial exploration ($\epsilon_i$) of 1.0 which linearly decays to a final exploration rate ($\epsilon_f$) of 0.01 over the entire training period (exploration fraction of 1). The intent of training these RL algorithms is not simply to develop the best single RL model, but to develop a collection of RL models that include generalized as well as specialized behaviors. The following RL models are trained and employed:

- *RL-Boron*: an RL-trained agent that uses an engineered reward system that primarily relies on the actual game score while seeking to minimize losses (as calculated in Equation 3.6 in Section 3.5.2).
- *RL-Scotty*: an RL-trained agent that is indifferent to the game score and is trained with a reward system that encourages remaining close together near cities while also massing attacks on a single opponent entity when numerically superior.
- *RL-S-Killer*: a modification of *RL-Scotty* in that it is rewarded during training only for inflicting damage on its opponent, irrespective of whether it is an unsound decision to do so.
- *RL-S-City*: a modification of *RL-Scotty*, however, instead of favoring careless attacks, it is rewarded for occupying cities and remaining grouped together around the city—only attacking when an opponent entity comes within range.

**Multi-Models**

The multi-models we use in this study are composed of different combinations of the above 8 models, depicted in Table 4.1, and include:

- *Multi-Model (3)*: consists of only the three scripted agents *Pass-Agg*, *Pass*, and *Agg*
- *Multi-Model (4)*: consists of all four scripted agents (*Pass-Agg*, *Pass*, *Agg*, and *Burt-Plus*)
- *Multi-Model (6)*: consists of all four scripted agents as well as the two general RL-trained agents *RL-Boron* and *RL-Scotty*
- *Multi-Model (8)*: consists of all four scripted agents, two general RL-trained agents (*RL-Boron* and *RL-Scotty*) and two specialized RL-trained agents (*RL-S-Killer* and *RL-S-City*)

Table 4.1. Composition of Multi-Models

| Multi-Model (3) | Multi-Model (4) | Multi-Model (6) | Multi-Model (8) |
|---|---|---|---|
| Pass-Agg | Pass-Agg | Pass-Agg | Pass-Agg |
| Pass | Pass | Pass | Pass |
| Agg | Agg | Agg | Agg |
| | Burt-Plus | Burt-Plus | Burt-Plus |
| | | RL-Boron | RL-Boron |
| | | RL-Scotty | RL-Scotty |
| | | | RL-S-Killer |
| | | | RL-S-City |

### 4.4.2 Score Prediction Models

To feed the evaluation function that determines which behavior model to choose, we first train a separate *Score Prediction Model* for each of our single-models. Our *Score Prediction Model* is a CNN that takes in the state observation and outputs a predicted game score. This prediction assumes that the blue faction plays the rest of the game according to the *Score Prediction Model*'s corresponding behavior model, and the red faction plays according to a pre-determined behavior model. In this study, we use *Pass-Agg* as the opponent behavior model for all training and testing. This *Score Prediction Model* is called each time a blue entity needs to take an action. As Atlatl operates on a turn-based system rather than a time-step simulation, we designate each instance when an entity on the gameboard is prompted to take action as an *action-selection step*. At each *action-selection step*, each *Score Prediction Model* gives us its estimate of the final game score:

$$\hat{y} = \text{P_Model}_i(s) \tag{4.1}$$

where $\hat{y}$ is the predicted score of the game; $\text{P_Model}_i$ is a score prediction model $i$; and $s$ is the observation of the current state space provided to the model.

#### Neural Network Architecture

The score prediction model's neural network architecture is similar to that of our RL model's CNN as described in Section 3.5.2. However, through extensive hyperparameter tuning, we modified the network to use six residual hexagonal convolutional layers of 128 convolution units each. The output of the final convolutional layer is then flattened and passed through two fully-connected linear layers of 8,000 units each. Between each convolutional and linear layer, we use a Rectified Linear Unit (ReLU) activation function. The loss criterion we employ is the mean squared error (MSE) function. For model optimization, we select the Adam optimizer. The learning rate is set to 0.0001, with a weight decay parameter of 0.00001.

**Training**

For the dataset to train each score prediction model, we use Atlatl to generate 6 million samples of synthetic data using the respective single model and the scenario described in Section 4.4.4. Each exemplar consists of a tensor representation of the observation as the input and a final game score as the label. We repeat this process to train a separate prediction model for each of the eight single-models that we employ in this study, for a total of 48 million samples.

From the 6 million samples for each single-model, we use 4 million for training, 1 million for training validation, and 1 million for final testing of our model prior to incorporating each into our multi-model. We find training our model over five epochs minimized our MSE training validation loss before beginning to overfit to our training data. For each of the eight models for which we train a score prediction model, we observe mean absolute errors (MAEs) ranging from approximately 65 to 80 points. For reference, in the games used for this experiment, the maximum score achieved was 640, the minimum score was −640, and the mean score was 146.688. In a separate analysis, we found that the models' errors were generally highest in the early phases of the game but lower in the later phases of the game. In fact, the models were typically about twice as accurate in the last third of the game than in the beginning third.

### 4.4.3 Multi-Model Framework

We design our multi-model to take in the same observation that our RL agents use for both training and inference. This observation is then passed to the respective score prediction model for each of the behavior models available in our model repository, as depicted in Fig. 4.3. To determine which behavior model to use at each action-selection step, we run an inference on each of our score prediction models and select the behavior model corresponding to the prediction model that outputs the highest game score. More formally, we design a simple model-selection function called at each action-selection step:

$$\text{Index}_{\text{B_Model}} = \underset{i \in [i,\ldots,n]}{\arg\max}(\text{P_Model}_i(s)) \tag{4.2}$$

where $\text{Index}_{\text{B_Model}}$ is the index of the behavior model selected to infer the next action; $\text{P_Model}_i$ is a score prediction model; and $s$ is the observation passed to the score prediction model.

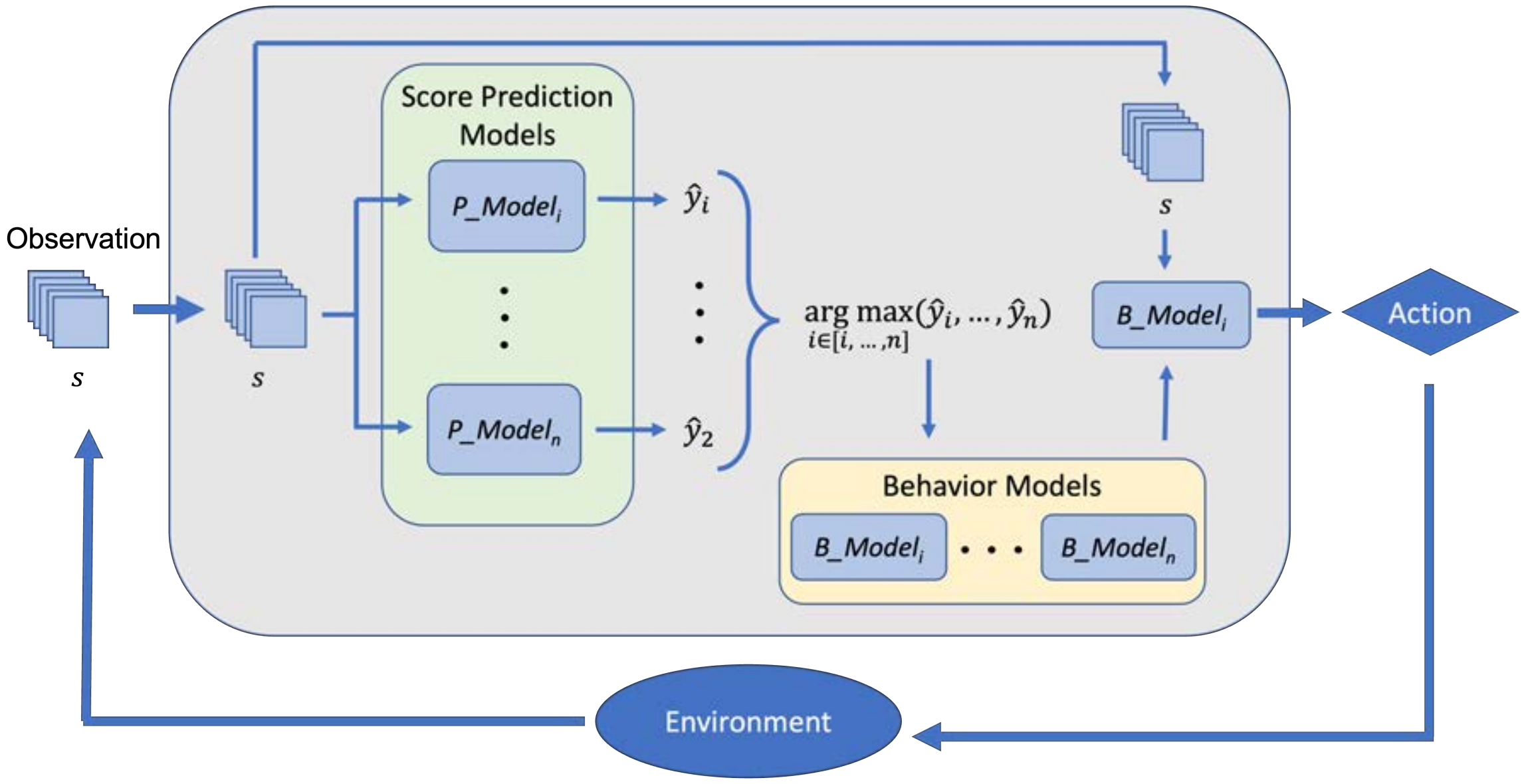

Figure 4.3. Multi-Model Framework. At each step where an agent is called to take an action, the framework receives a state observation and forwards this to each of its score prediction models. Each of these models then predicts a game score, which is subsequently passed to an evaluation function. Using the evaluation function, a behavior model is chosen. Subsequently, the state observation is transmitted to the appropriate behavior model and an action is selected.

### 4.4.4 Experiment

For our experiment, we build each of the four multi-models listed in Section 4.4.1 using our multi-model framework. We then design and run the following experiment to assess the performance of each of our multi-models as compared to each of the single-models.

**Scenario**
We use an Atlatl scenario generator to create $5 \times 5$ gameboards, where a random number of units and one city are randomly placed at the start of each game. The number of units per faction is chosen randomly between two and four. To structure the gameplay environment, at the start of each game, the board is divided either along a north-south or east-west axis. Initially, a random side (north, south, east, or west) is chosen for the first unit; subsequently, all additional units from the same faction are positioned in random, non-overlapping positions on the same side, while units from the opposing faction are placed on the opposite side.

The city is positioned according to force ratio. If both factions are of equal starting strength, the city is randomly placed in a location along the middle axis of the board (i.e., in a neutral location). If one faction is weaker, the city is randomly placed on the side of the board of the weaker faction (i.e., in a location generally favorable to the weaker faction).

We use a maximum of 10 phases, where each phase is one entire turn for one faction (i.e., one faction is allowed to make one legal move for each of its available entities). Within each phase, the AI rotates through its available entities in a fixed order when selecting an action for each entity. We set `scenarioCycle` = 0 (as described in 3.4.3), resulting in an infinite number of random starting conditions with no deliberate repetition. An example of this scenario is shown in Figure 4.4.

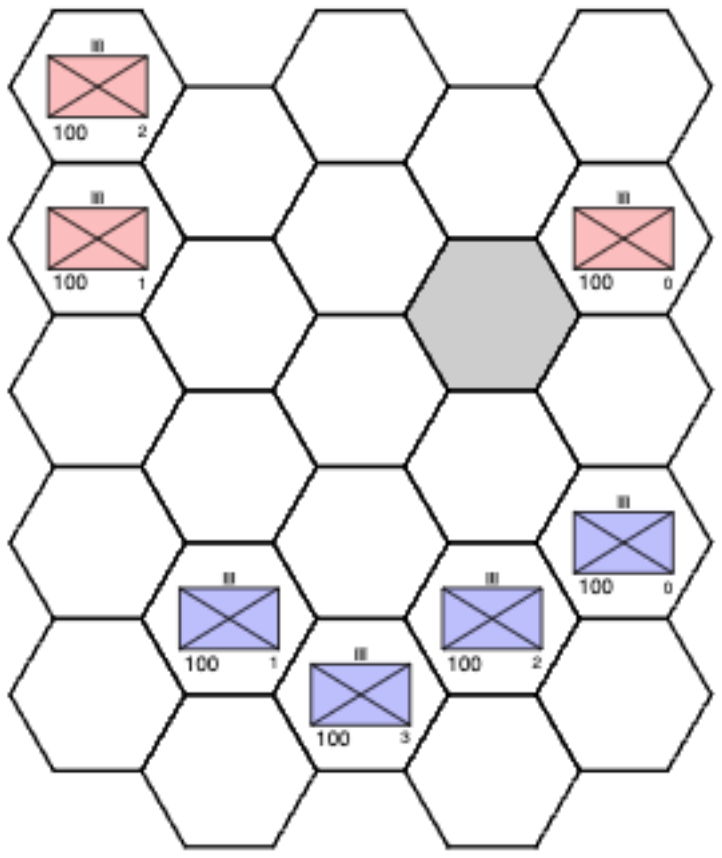

Figure 4.4. Example Experiment Scenario. A 5 × 5 hexagonal gameboard with randomly initialized blue and red forces. The city hexagon is also randomly initialized and placed according to force ratio. In this case, the red is the weaker faction, and thus, the city is positioned on their side of the gameboard.

**Model Evaluation**

We run each of our single-models and multi-models for 100,000 games in Atlatl using the random scenario generator above, resulting in a combined 1.2 million games. We use *Pass-Agg* as the adversary behavior model. For each game iteration, we capture the final game score, calculated as described in Section 3.4.4.

## 4.5 Results and Discussion

The final mean scores and Standard Errors of the Mean (SEMs) of the 100,000 runs for each model are shown in Table 4.2. A box and whisker plot is shown in Figure 4.5. To better depict the differences in mean scores achieved by each model, a bar chart is depicted in Figure 4.6. Of the scripted agents, *Burt-Plus* performs the best with a mean score of 80.281. Of the RL agents, *RL-Boron* performs the best, though with a mean score of 21.823—only about half as good as the worst-performing scripted agent. It is worth noting that the other three RL agents (*RL-Scotty*, *RL-S-Killer*, and *RL-S-City*) all have negative mean scores with *RL-S-City* having the worst score of −103.280. This is not unexpected as these agents were trained to exhibit specific combat behaviors rather than to maximize the game score.

Table 4.2. Mean Scores ± Standard Error of the Mean of 100,000 games

| Scripted Models | | | |
|---|---|---|---|
| Pass-Agg | Pass | Agg | Burt-Plus |
| 68.962 ± 0.907 | 74.964 ± 0.924 | 42.995 ± 0.879 | 80.281 ± 0.910 |
| RL Models | | | |
| RL-Boron | RL-Scotty | RL-S-Killer | RL-S-City |
| 21.823 ± 0.820 | −11.472 ± 0.803 | −18.223 ± 0.892 | −103.280 ± 0.807 |
| Multi-Models | | | |
| Multi (3) | Multi (4) | Multi (6) | Multi (8) |
| 82.812 ± 0.933 | 93.831 ± 0.939 | 125.487 ± 0.868 | 130.540 ± 0.866 |

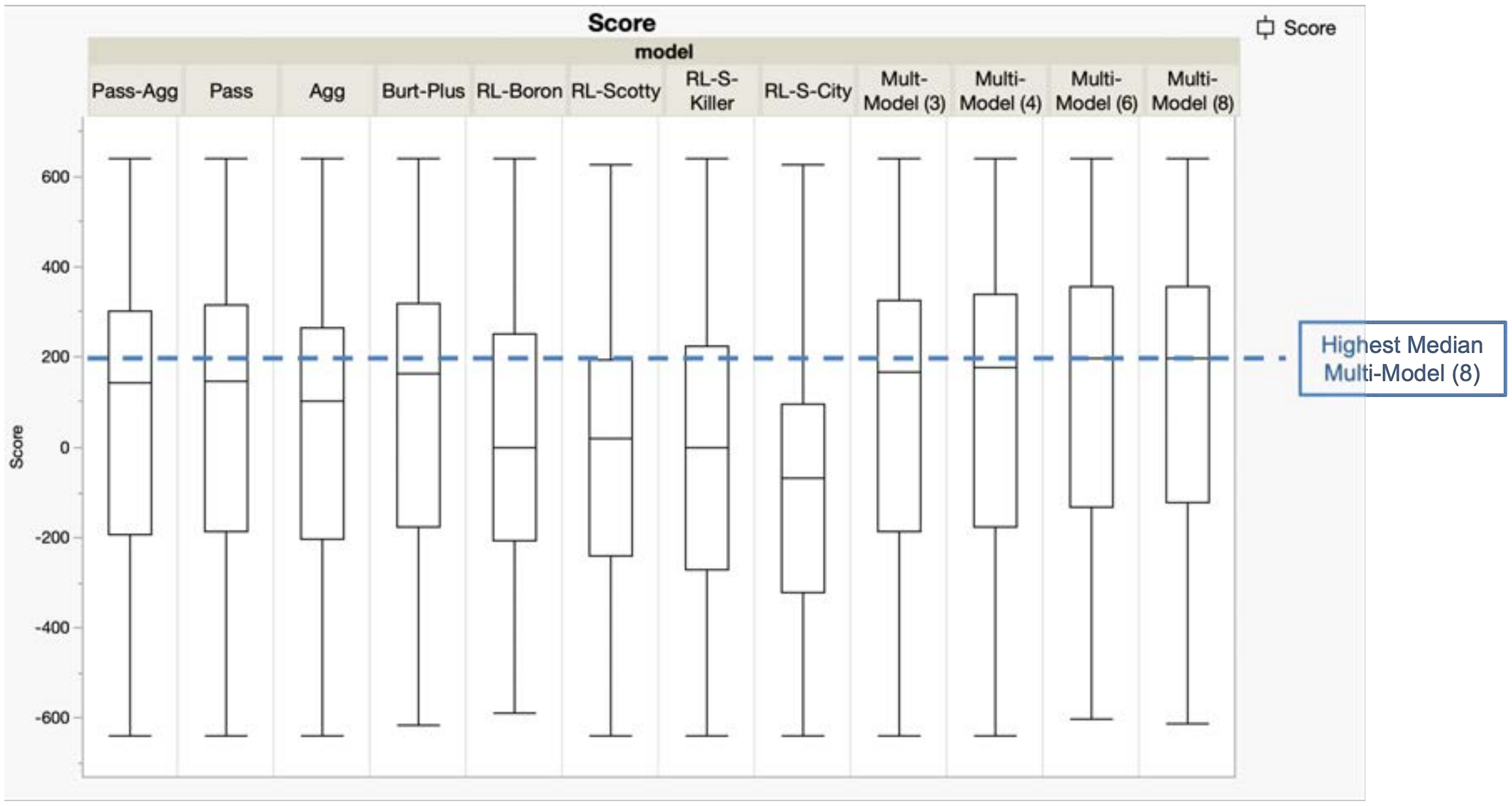

Figure 4.5. Box and Whisker Plot of the Multi-Model Experiment. The wide variance in this chart is due to the variety of randomly-generated scenarios played, ranging from very offensive to very defensive.

The multi-models produce the highest of all scores, with the *Multi-Model (8)* producing a mean score of 130.540—a 62.6% improvement over the best-performing single-model (*Burt-Plus*). This is a very interesting result in that one may intuit that by taking a model like *Multi-Model (4)*, composed of the top four performing single-models, and adding lower-performing models, that the overall performance of the multi-model would either stay constant or decrease. However, we show that even by adding the two worst models with negative mean scores (*RL-S-Killer* and *RL-S-City*) to *Multi-Model (6)*, we get a model (*Multi-Model (8)*) that outperforms all other models in terms of mean score.

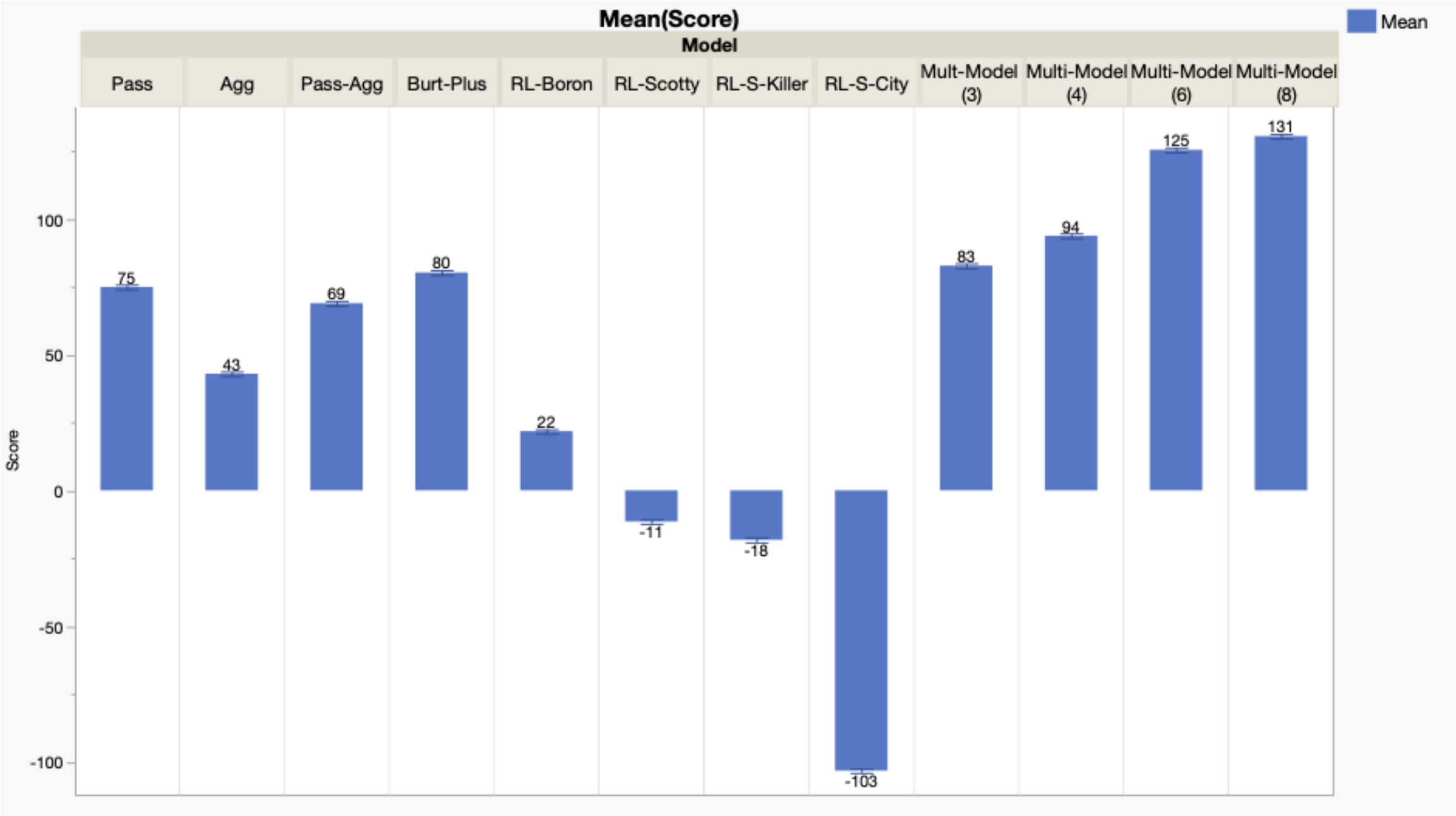

Figure 4.6. Bar Chart Depicting Mean Scores in Multi-Model Experiment. Depiction of the mean score achieved by each model.

This suggests that although some of our single-models performed poorly overall, they probably succeeded in certain, very specific situations—which seemed to be accurately captured by our score prediction models. To verify this, we capture how many times each single-model is selected by the multi-model. For *Multi-Model (8)*, as depicted in Table 4.3, we see that even the worst-performing single-models are still selected at some points over the course of the evaluation runs. However, further investigation is needed to better understand this dynamic and determine if this selection is simply due to errors in the prediction model or if it is due to the score prediction model recognizing specific game states in which these models excel.

Table 4.3. Percent Selection of Each Single-Model in Multi-Model (8)

| | Scripted Models | | | | RL Models | | | |
|---|---|---|---|---|---|---|---|---|
| Model | Pass-Agg | Pass | Agg | Burt-Plus | RL-Boron | RL-Scotty | RL-S-Killer | RL-S-City |
| % Chosen | 7.3 | 31.3 | 26.0 | 11.0 | 18.3 | 1.3 | 2.9 | 1.9 |
| % Total | 75.6 | | | | 24.4 | | | |

To determine whether the differences in mean scores between the single-models and the multi-models are statistically significant, we set $\alpha = 0.05$ and run the Tukey-Kramer Honest Significant Difference (HSD) test. As noted in the previous chapter, this test goes beyond the standard One-Way Analysis of Variance (ANOVA) test in that it conducts pairwise comparisons between all possible pairs and details which specific groups' means are significantly different from each of the other groups.

Our analysis results in an overall standard error (SE) (i.e., an estimate of the variability of the difference between any two group means) of 1.245 and confirms statistical significance between each and every pair with the exception of the pairs (*Multi-Model (8)*, *Multi-Model (6)*) and (*Multi-Model (3), Burt-Plus*). A summary of the test is shown in Fig. 4.7.

**Connecting Letters Report**

| Level | | Mean |
|---|---|---|
| Multi-Model (8) | A | 129.6016 |
| Multi-Model (6) | A | 125.4874 |
| Multi-Model (4) | B | 93.8305 |
| Mult-Model (3) | C | 82.8122 |
| Burt-Plus | C | 80.2810 |
| Pass | D | 74.9642 |
| Pass-Agg | E | 68.9622 |
| Agg | F | 42.9952 |
| RL-Boron | G | 21.8235 |
| RL-Scotty | H | -11.4681 |
| RL-S-Killer | I | -18.2228 |
| RL-S-City | J | -103.2804 |

Levels not connected by same letter are significantly different.

Figure 4.7. Tukey-Kramer HSD test connecting letters report. With the exception of the pairs (*Multi-Model (8)*, *Multi-Model (6)*) and (*Multi-Model (3), Burt-Plus*), all means are significantly different from each other.

From these results, we find that all four multi-models outperform each of the scripted models from which they are composed. We also show that adding more models to the multi-model can result in even better performance, even if those additional models perform worse overall. Although we cannot claim statistical significance between the means for *Multi-Model (6)* and *Multi-Model (8)*, we note an overall increasing trend as we add more models to each of the multi-models. This is an insightful result as it can lead to more efficient training of behavior models since training specialized models can typically be done with fewer samples or training steps than training more generalized models. Furthermore, this finding suggests that diversity in specialized individual models can enhance overall system performance to handle complex scenarios more effectively than any single model could on its own.

## 4.6 Conclusion

Our results suggest that a multi-model approach consisting of a broad range of individual behavior models that uses neural networks to predict scores, and a simple evaluation function to select the best model given the current state space, can lead to significantly better performance as compared to a single-model approach. We also conclude that a multi-model composed of more models (up to eight in our experiment) generally outperforms multi-models composed of fewer models (down to three in our experiment). Lastly, we show that a multi-model containing specialized single-models, even when these models do not generally perform well individually, can still outperform a multi-model containing fewer, but better-performing single-models.

This proposed multi-model framework not only enhances our overall HRL framework, but also contributes to the advancement of intelligent agents on its own in several ways. First, rather than attempting to develop a single RL agent that performs well across a broad spectrum of all possible scenarios, we can instead combine many different models consisting of both scripted and RL agents, and both generalized and specialized models. Second, allowing for a model to contain a diverse set of single models can ensure that the overall multi-model is effective even if one model underperforms at a specific task. Third, by selecting the best-performing model for specific situations rather than diluting the specialization effect by averaging out the unique strengths of each model (as is done in ensemble modeling), we ensure that the most effective pure strategies or tactics are used only when needed, enhancing overall system performance. Lastly, the selection mechanism in this multi-model framework supports quicker development and deployment by allowing models to be added, removed, re-trained, or fine-tuned independently, without necessitating adjustments to any weighted voting mechanisms or other models in the ensemble. Thus, the multi-model approach offers a flexible and efficient alternative to traditional ensemble techniques, providing a methodology for leveraging the best of specialized models while maintaining robustness through a diverse model library.

In summary, this multi-model approach has shown significant improvement over its embedded single-models. This study, on its own, furthers the ongoing research in utilizing machine learning to model intelligent combat behaviors. Furthermore, our results suggest that a combination of approaches that incorporate various AI techniques and leverage diverse models could enable us to overcome some of the limitations of relying on either an RL or scripted approach alone.

# CHAPTER 5:
# A Hierarchical Hybrid AI Approach

This chapter describes how we construct and validate a hierarchical hybrid AI approach that integrates RL and scripted agents in a way that improves performance beyond either approach alone. It differs from the previous multi-model chapter in that this specific method seeks to establish a hierarchical relationship between models, rather than a flat repository of many different models. By structuring the AI system hierarchically, we aim to leverage the strengths of both scripted and RL methods for different levels of decision-making (higher-level operational decisions vs. lower-level tactical decisions) while mitigating their respective weaknesses. In this study, RL agents are utilized for longer-term decision-making and adaptation in response to evolving circumstances at the operational level, while scripted agents are employed to handle well-defined, routine tasks and to provide a consistent baseline behavior at the tactical level. This hybrid approach aims to create a more robust and effective AI system overall.

While we address scalability in Chapter 3 via observation abstraction, and in Chapter 4 via the multi-model framework, in this chapter, we address and validate another form of scalability that involves employing a hierarchical hybrid AI approach using both RL and scripted agents. This method facilitates problem decomposition, enabling more adaptive RL agents to tackle complex operational-level challenges while allowing scripted agents to handle simpler tactical-level problems. Through our experiment and analysis, we seek to validate this approach and gain insights into how to best incorporate this methodology within our overall HRL framework. Furthermore, like the previous chapters, this chapter also individually contributes to ongoing research on the scalability and effectiveness of AI within the domain of combat simulations.

## 5.1 Background

As discussed in Chapter 2, the development of intelligent agents for combat simulations in support of wargaming has predominantly been characterized by rule-based, scripted methodologies with RL approaches only recently introduced.

Detailed in Section 2.3, rule-based systems, grounded in if-then logic, offer a methodical framework for agent decision-making [57]. They provide a structured approach, allowing for decisions to be made based on specific, predefined criteria, which results in clear and consistent actions. Behavior trees extend this structure, organizing decisions in a hierarchical manner that mirrors natural decision-making processes, thereby enhancing system flexibility while providing a distinct delineation of decision pathways. This structure not only facilitates easier updates and modifications but also supports varied behavioral patterns. FSMs simplify the representation of agent states, breaking down behaviors into clear, discrete stages with defined transitions, thus facilitating targeted problem-solving. Incorporating goal-based systems into this framework allows agents to pursue specific objectives, allowing them to align their actions and strategies toward achieving these goals. These methodologies together create a strong base for crafting agents that perform reliably and consistently—driven by well-defined rules and logical frameworks—thus allowing for reasonable decision-making across most situations.

Although these scripted designs based on specific domain knowledge have been instrumental in creating effective, predictable, and logical agents for familiar situations, relying on predefined rules, heuristics, and algorithms often comes with inherent inflexibility and rigidity—making them less effective in unexpected or novel situations [66]. The scripted agent's reliance on fixed logic and predetermined pathways often limits the agent's ability to adapt to larger, more dynamic environments [57], [65]. This underscores the need for more advanced, adaptable AI approaches capable of learning and responding to novel challenges, while still leveraging the consistency and predictability of scripted methodologies.

Addressing this need, as described in detail in Section 2.4, RL provides a framework for agents to learn and adapt through interactions with their environment, allowing agents to adapt to changing conditions within the simulation by learning from past experiences, generalizing from these experiences, and improving their behaviors over time. RL agents are particularly effective in uncertain environments where the optimal strategy is not known a priori, achieving notable success in games such as StarCraft [44], Dota 2 [43], Go [79], and Atari [40].

Nevertheless, the application of RL in large combat simulations has met substantial challenges, primarily due to the complexity of these environments and the inefficiencies associ-ated with learning in large state spaces. Research in RL for combat simulations reveals that, although effective in smaller contexts, scaling up often leads to performance that has yet to consistently surpass the performance of human or scripted agents [11], [80]–[82]. This issue primarily stems from the exponential increase in state space complexity [70] and RL's sample inefficiency problem [40], which becomes more pronounced as the complexity of the observation space grows. Additionally, the black-box nature of RL models can result in opaque decision-making processes, making it difficult to trust and interpret the actions taken by RL agents.

These described limitations of both scripted and RL approaches highlight the need to explore more sophisticated methodologies that can bridge the gap between rigid, rule-based systems and adaptable, learning frameworks. This background section sets the stage for advancing our understanding of how a hybrid approach, leveraging the strengths of both scripted systems and RL, can potentially overcome their individual limitations. In this chapter, we explore the integration of both of these approaches, striving to create intelligent agents that are not only effective in predictable environments but also capable of learning and adapting to more complex scenarios.

## 5.2 Related Works

In this section, we review literature that explores the integration of multiple AI methodologies to enhance intelligent agent development, covering areas from RL-enhanced behavior tree development to tactical applications in military simulations and game development where RL is combined with some other form of AI. The following review describes current advancements in hybrid AI systems and identifies research gaps, framing this unique contribution to the field of AI behavior development.

Ulam et al. [128] propose combining model-based meta-reasoning with RL to improve their game-playing agent's adaptability in the game of *FreeCiv*. Meta-reasoning is a technique where an agent is equipped with a self-model, which represents its own knowledge and reasoning capabilities. When the agent fails to complete a task, it uses this self-model along with traces of its reasoning processes to determine the cause of the failure. By identifying what went wrong, the agent can then adjust its knowledge and reasoning strategies to improve future performance [128]. Their approach involves the application of meta-reasoning to guide RL, helping reduce the learning space and learning time by focusing on relevant sub-tasks, leading to improved performance and efficiency as compared to conventional RL agents. While an improvement over more traditional RL approaches, it relies on the need for a hand-crafted, accurate, and comprehensive meta-reasoning model which, like scripted models themselves, may be difficult to construct in games involving complex strategies [128].

Li et al. [129] explore the integration of RL and behavior trees to enhance intelligent agents. Their approach leverages the strengths of both RL and behavior trees, synchronizing their execution and defining appropriate reward functions to guide effective decision-making. While our research differs in that we do not seek to incorporate RL nodes within our scripted agents' behavior trees, their research offers valuable insights into integrating RL with rule-based agents, particularly in managing complex and hierarchical task assignments [129].

Zhao et al. [130] propose a method for generating interpretable behavior trees from RL policies, aiming to enhance the interpretability of RL in applications such as robotics and gaming. This involves translating RL policies directly into behavior trees, addressing the challenge of making RL decisions understandable and manageable. While this approach is indeed valuable in informing our research, our specific hybrid approach in this study involves the incorporation of RL and scripted agents during gameplay, rather than relying on RL to generate a scripted agent [130].

Yu et al. [131] present an enhanced decision-making method for combat simulations through the integration of RL and supervised learning by employing an improved Multi-Agent Deep Deterministic Policy Gradient (MADDPG) framework. It showcases significant performance improvements, underlining the benefits of combining RL with additional AI techniques to help address large action spaces and sparse rewards. While this is an improved approach to RL, it differs from our approach in that we use a hybrid approach both in the training and control of our agents, rather than just in the training of our agents [131].

Further, research by Hu et al. [132], Bignold et al. [133], and Kallstrom and Heintz [134] explore the tactical applications of RL in military simulations, demonstrating the practical implications for air combat maneuver planning and persistent rule-based interactions. On the game development front, Zhao et al. [135], Noblega et al. [136], and Kim and Ahn [137] explore AI's role in game balancing and development, while Vazquez-Nunez et al. [138], Joppen et al. [139], and Ponsen [140] introduce hybrid computational intelligence, showcasing the efficacy of combining various methodologies for adaptive game tactics.

While the foundational insights from the literature offer a broad spectrum of approaches to advancing AI in domains relevant to combat simulations, this study distinguishes itself by advancing a hierarchical hybrid model that combines the adaptability of RL for higher-level decisions with the precision of scripted systems for lower-level decisions. This approach not only helps make RL-based decision-making in complex environments more tractable but also proposes a unique solution to the integration of two different AI systems, aiming to enhance both the strategic flexibility and tactical execution of AI agents in combat simulations.

## 5.3 Methodology

To assess the efficacy of using a hierarchical hybrid AI approach that integrates an RL manager agent with scripted subordinate agents, we employ the following methodology using the Atlatl simulation environment described in Section 3.4. We conduct an experiment where we evaluate the performance of this framework as compared to scripted and RL agents alone. We then provide our results and conclusion.

### 5.3.1 Scripted Agent

As in previous experiments, we use the *Pass-Agg* behavior model, described in Section 3.4.5, as our *Scripted Agent*. This agent allows for rapid, deterministic, and reasonable responses that adhere to a set of predefined, handcrafted rules and scripts.

### 5.3.2 Integration with Reinforcement Learning

To address the limitations of both scripted and RL agents, our approach introduces an *RL Manager Agent* that operates in conjunction with scripted units, which we refer to as *Scripted Subordinate Agents* in this chapter. We designate this integrated system consisting of an *RL Manager Agent* and its assigned *Scripted Subordinate Agents* as a *Hybrid Agent*. This *Hybrid Agent* allows for the combination of strengths of both approaches: the reliability and known quantity of rule-based decision-making at the tactical level, and the adaptability and optimization capabilities of RL at the strategic level. Of note, while the *Scripted Subordinate*

*Agents* are actual entities on the gameboard, the *RL Manager Agent* is not an actual entity on the gameboard, but rather an abstract agent that issues objective areas to guide its *Scripted Subordinate Agents*.

When using the *Hybrid Agent* model, at scenario initiation, after units are randomly created per a given `scenarioSeed` and `scenarioCycle` (described in Section 3.4.3), each unit is assigned to their respective *RL Manager Agent*. For this experiment, each *RL Manager Agent* is initially assigned and controls three units, which are positioned in close proximity to each other. For example, if the Blue faction starts with six units, a total of two individual *RL Manager Agents* are created to control three units each. Figure 5.1 shows an example of the hierarchical structure of two *Hybrid Agents*, each consisting of an *RL Manager Agent* and its three *Scripted Subordinate Agents*.

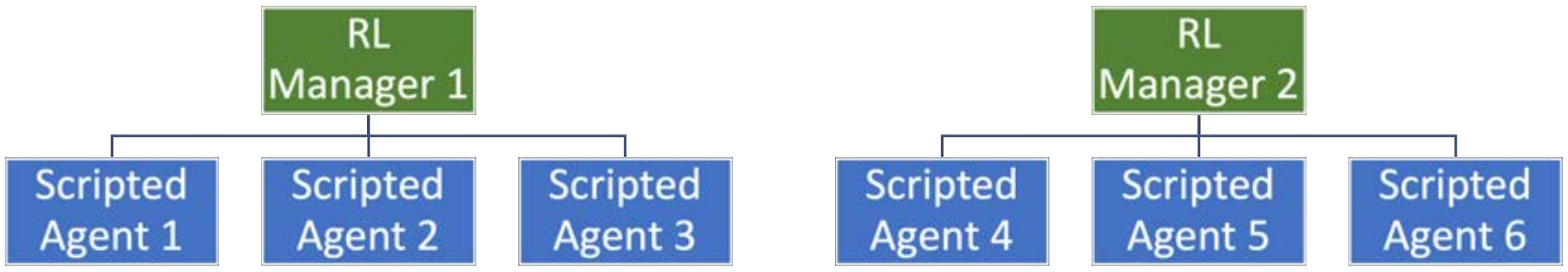

Figure 5.1. Hierarchical Hybrid AI. Each scripted agent on the board is assigned to an RL Manager agent.

During gameplay, the *RL Manager Agent* operates by selecting an objective area and passing this to each of its *Scripted Subordinate Agents*. Each *RL Manager Agent's* objective area is essentially a super "hexagon" in that it contains a center hexagon and two surrounding layers of hexagons. An example is shown in Figure 5.2, where the super hexagon is shaded and outlined in blue. Also shown in this figure is an example of the Blue faction being divided into two *RL Manager Agents*, each controlling three *Scripted Subordinate Agents*. One manager's *Scripted Subordinate Agents* are color-coded blue, and the other manager's *Scripted Subordinate Agents* are color-coded green.

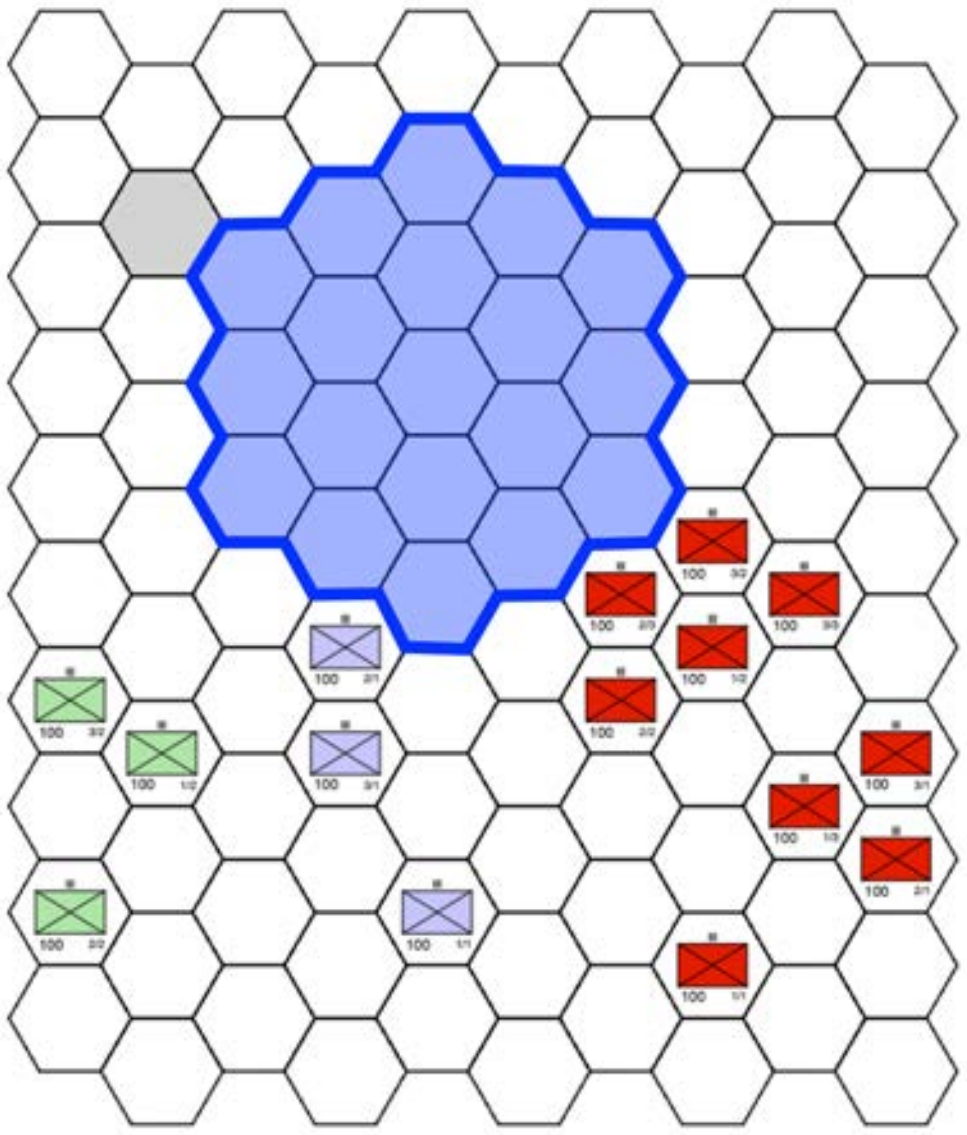

Figure 5.2. RL Manager's Objective Area is denoted by the blue area referred to here as a super "hexagon." This area consists of a center hexagon and the next two surrounding layers of hexagons.

Unlike the *Scripted Subordinate Agents*, *RL Manager Agents* do not make decisions at each time step in the game. Instead, the *RL Manager Agent's* temporal abstraction is variable and event-driven. Each *RL Manager Agent* is only called to make a decision when all of its units (that are not ineffective) arrive within its objective area. At this point, the respective *RL Manager Agent* generates its own objective area independent of the other *RL Manager Agents*. This mirrors the "options" framework [87] where decisions span over multiple time steps, focusing on achieving broader objectives, rather than responding to every single environmental change. Figure 5.3 depicts the manager's meta-model, illustrating the relationship between the *RL Manager Agent* and its *Scripted Subordinate Agents*.

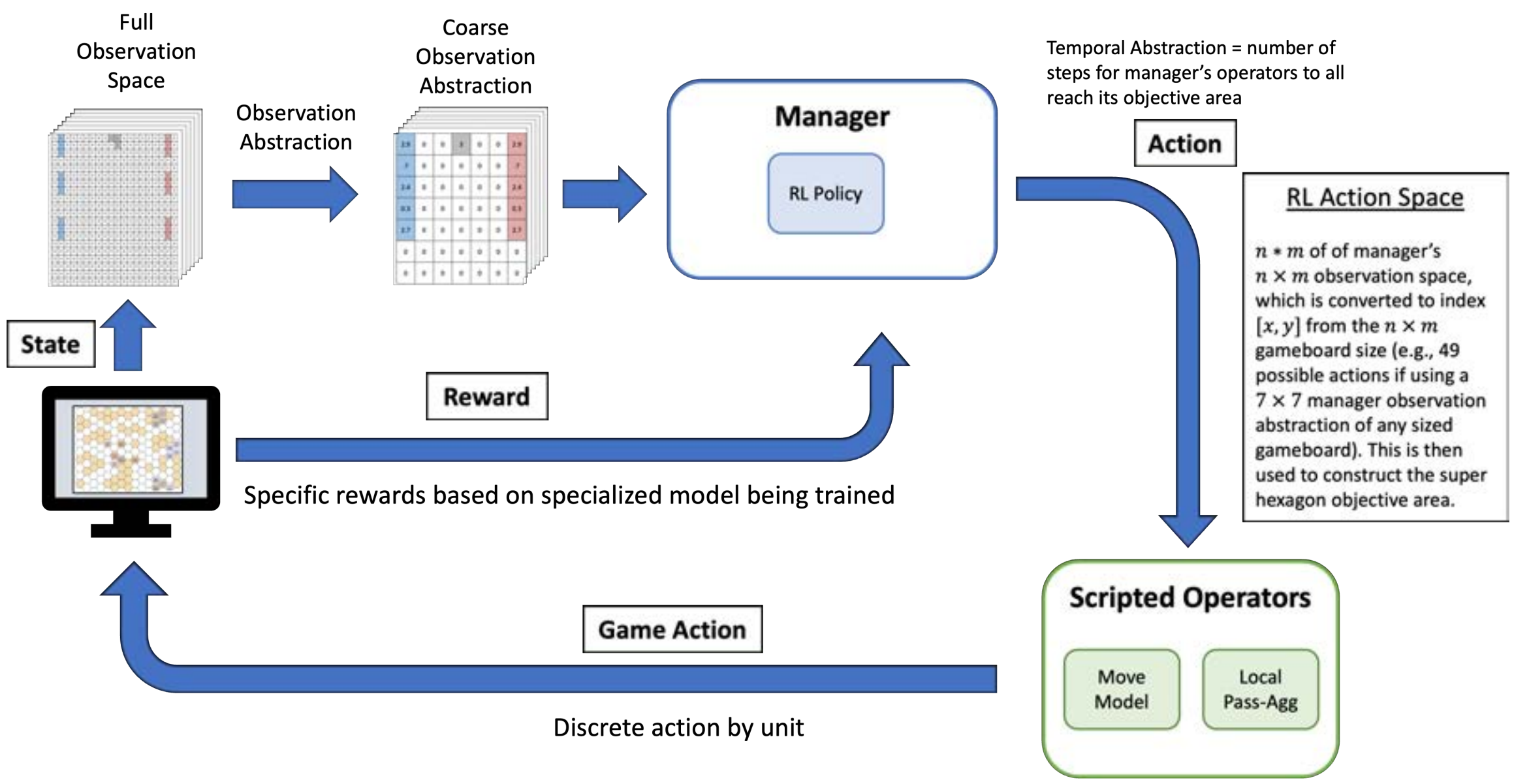

Figure 5.3. Manager Meta Model. The full observation space is abstracted into a coarse observation space. This is passed to the Manager, which takes an action. In this study, the action consists of selecting the center of one of 49 grids. The center of the selected grid is then mapped to the original gameboard map, and an objective area is constructed, which is then passed to the subordinate agents. These subordinate agents then execute one of two behavior modules based on whether or not they are within the objective area. The subordinate agent then takes an action that alters the environment, restarting the loop.

Conversely, *Scripted Subordinate Agents* make decisions at each time step using one of two behavior modules. First, the *Scripted Subordinate Agent* checks to see if it is within its assigned objective area. If the *Scripted Subordinate Agent* is not within its objective area, it activates its *Move Module* where it simply selects the hexagon to move to that is closest to its objective area. However, if an attack opportunity exists (i.e., an adversary unit is within its attack range), the will attack the opposing faction's unit instead of moving. Once the *Scripted Subordinate Agent* detects that it is within its objective area, it activates its *Fight Module* where the unit executes the *Pass-Agg* behavior algorithm described in Section 5.3.1. Of note, in this implementation, we restrict the information available to the *Scripted*

*Subordinate Agent* to that of its objective area. In other words, the game's global state space is culled to only include the information within the *Scripted Subordinate Agent's* objective area. Figure 5.4 describes each of the behavior modules used by the *Scripted Subordinate Agents*.

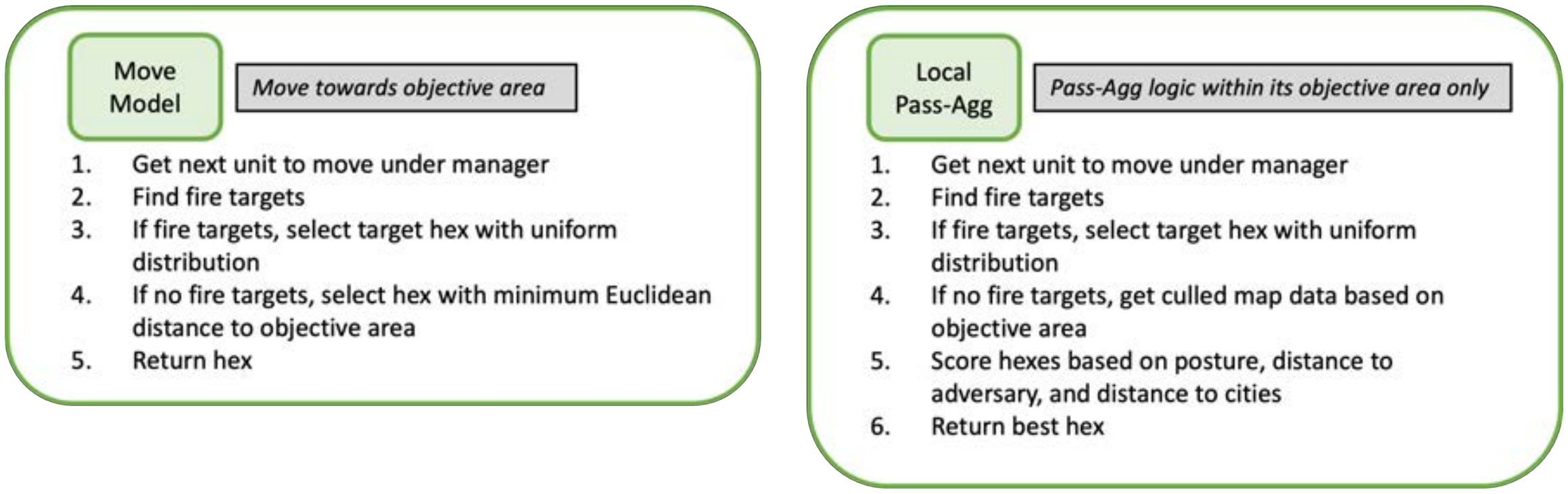

Figure 5.4. Subordinate Agent's Scripted Modules. The scripted subordinate agents contain two modules. If the agent is not within its manager's objective area, it executes its Move Model. If the agent is within its manager's objective area, it executes the Pass-Agg behavior model but uses only information within its objective area. The exception to this is for finding fire targets, for which the agent will examine all adjacent hexagons even if they are located outside of its objective area.

### 5.3.3 Manager's Observation Abstraction

As discussed in Section 3.4.6, the default RL agent's global observation space in Atlatl consists of an $18 \times n \times m$ tensor, where $n$ and $m$ are the height and width of the gameboard. Rather than providing the *RL Manager Agent* with the entire observation space, we modify a few of the default feature extractors as well as abstract the resulting observation in a map size-invariant way to provide the *RL Manager Agent* with a more tractable, compressed observation space. We do this by taking the full three-dimensional observation tensor and applying a coarse abstraction, which results in a final observation space of $17 \times 7 \times 7$.

This reduction involves reducing the full tensor into a smaller grid by comparing the original and reduced grid sizes to ensure proportional representation. Each cell in the original grid is then mapped to the reduced grid based on the overlap, with values adjusted to reflect the proportionate overlapped area, preserving the original spatial information. An example of this process is shown in Figure 5.5. Although the figure only shows an example using the channel depicting red forces, this same process is repeated for all channels within the *RL Manager Agent's* observation. This method provides the *RL Manager Agent* with a simplified yet accurate state space representation, enhancing its processing efficiency without significant loss of detail. In other words, the entire $10 \times 10$ grid is uniformly reduced into a $7 \times 7$ grid by summing values based on the overlapping sections.

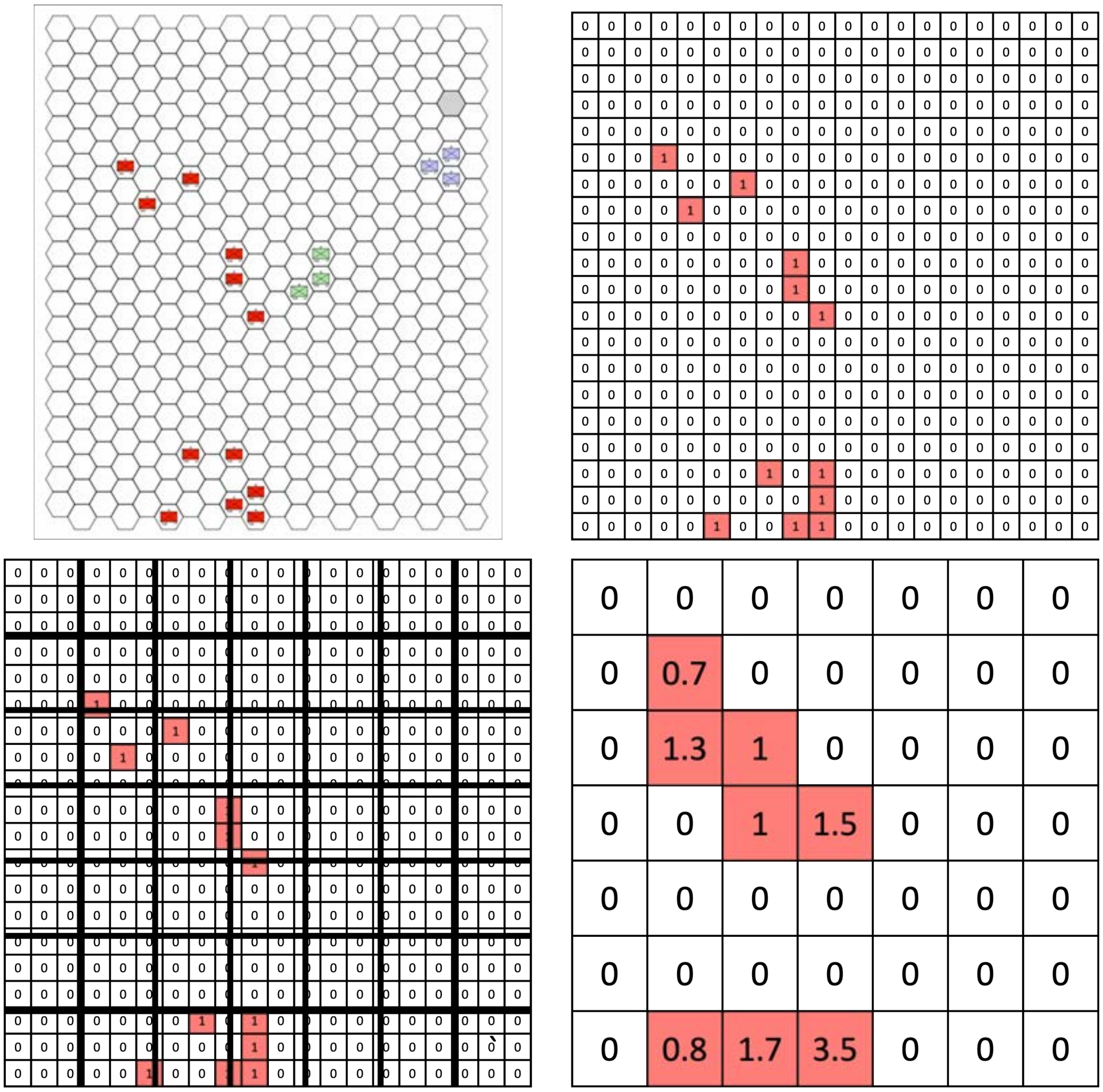

Figure 5.5. Coarse Observation Abstraction. The top left image depicts an example $20 \times 20$ gameboard. The top right image depicts the matrix representation of the red channel in the observation. The bottom left image depicts a $7 \times 7$ grid superimposed on the original $20 \times 20$ matrix. The bottom right image depicts how the values within each grid are summed up based on the area within each specific square of the $7 \times 7$ grid.

As shown in Figure 5.6, the manager's observation space is similar to that of the agent's global observation discussed in Section 3.4.6 with a few key changes. In the manager's observation, channel 0 depicts the blue units that belong to the current *RL Manager Agent* on-move, and channel 1 depicts the other *RL Manager Agents'* objective areas. The rest of the channels remain the same as previously described for the individual global RL agents.

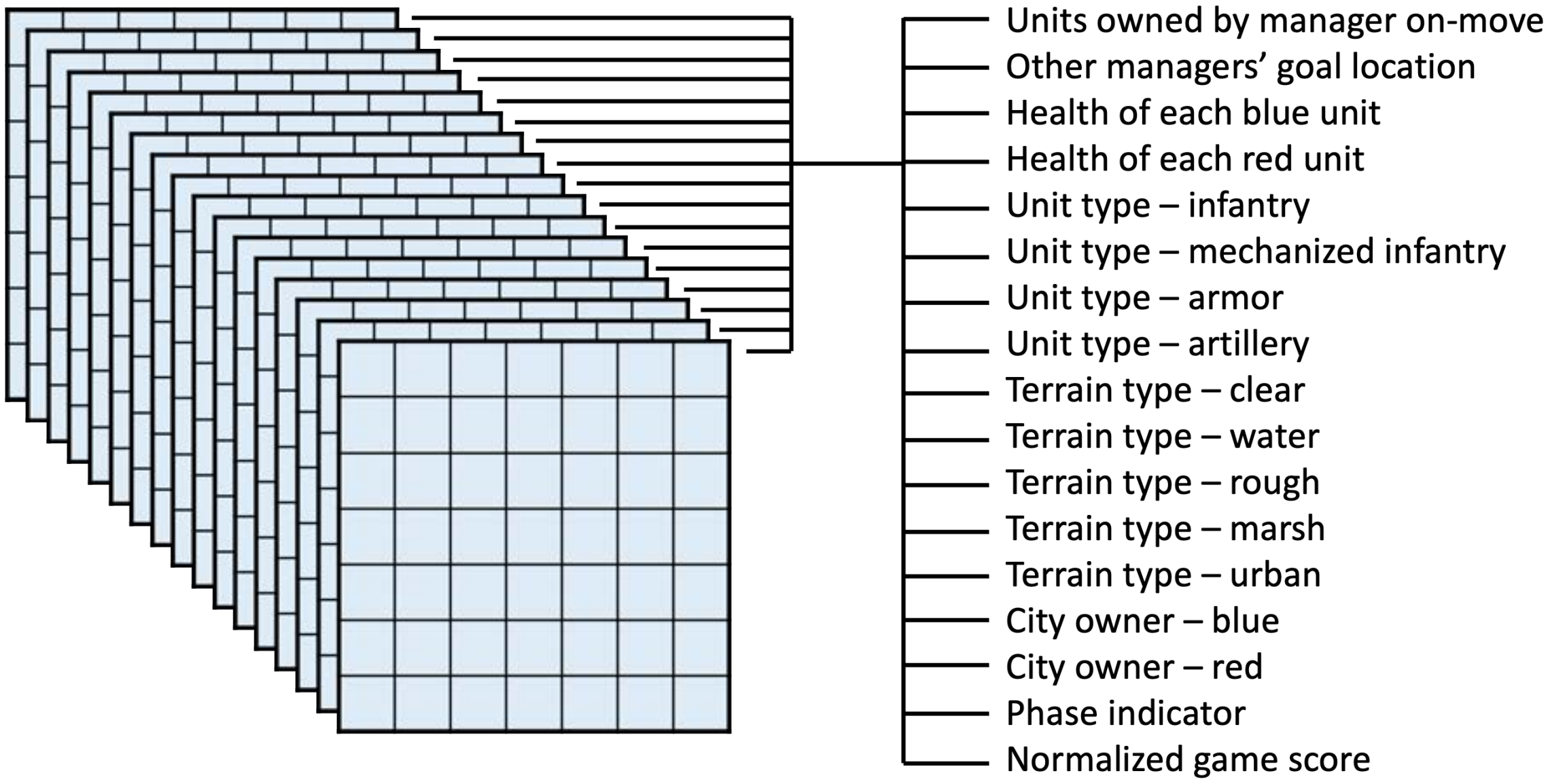

Figure 5.6. Manager Observation. A 17×7×7 tensor represents the information as denoted.

### 5.3.4 Gymnasium Environment

We create a separate Gymnasium environment to train our *RL Manager Agents*. The action space for these agents is defined as the set of super hexagons whose center hexagons contain the centers of a uniform $7 \times 7$ square grid superimposed over the scenario map, resulting in a total of 49 discrete actions. Essentially, each action corresponds to the center of one of the squares in the $7 \times 7$ grid depicted in the bottom left image of Figure 5.3.3.

The state space for our RL agent is the resultant $17 \times 7 \times 7$ observation space from taking our original global state space $s$ and passing it through this new manager observation function $\phi$ to produce the abstracted observation space $s_{\phi}$.

### 5.3.5 Neural Network Architecture

We use the same residual CNN architecture described in 3.5.2; however, for our manager, we use rectangular rather than HexagDLy convolutions since we uniformly abstract the hexagonal representation into an $18 \times 7 \times 7$ square grid representation. Then, as in the previously described residual CNN architecture, this input is converted into 64 output channels, followed by 7 additional layers of 64 channels each, which is then flattened to a 512-dimensional feature vector. Each layer features a ReLU activation function, a residual connection, and pointwise convolutions with a kernel size of $1 \times 1$ and a stride of 1.

### 5.3.6 Reinforcement Learning Algorithm

Based on initial experiments, we employ the DQN algorithm as it has shown better performance over the PPO algorithm for this task. We use the same hyperparameter values as denoted in Section 3.5.2.

### 5.3.7 Scenarios

For this experiment, we use randomly generated scenarios consisting of $10 \times 10$ hexagonal gameboards. Each game initiates with a randomly assigned number of either six or nine units per faction. This results in either two or three managers per faction, respectively. Each scenario also includes either one or two urban hexagons randomly placed according to force ratio as described in Section 3.5.2. The number of phases in the game is set to 40, where, again, each phase is one entire turn for one faction. Two examples of scenario initial starting conditions are shown in Figure 5.7.

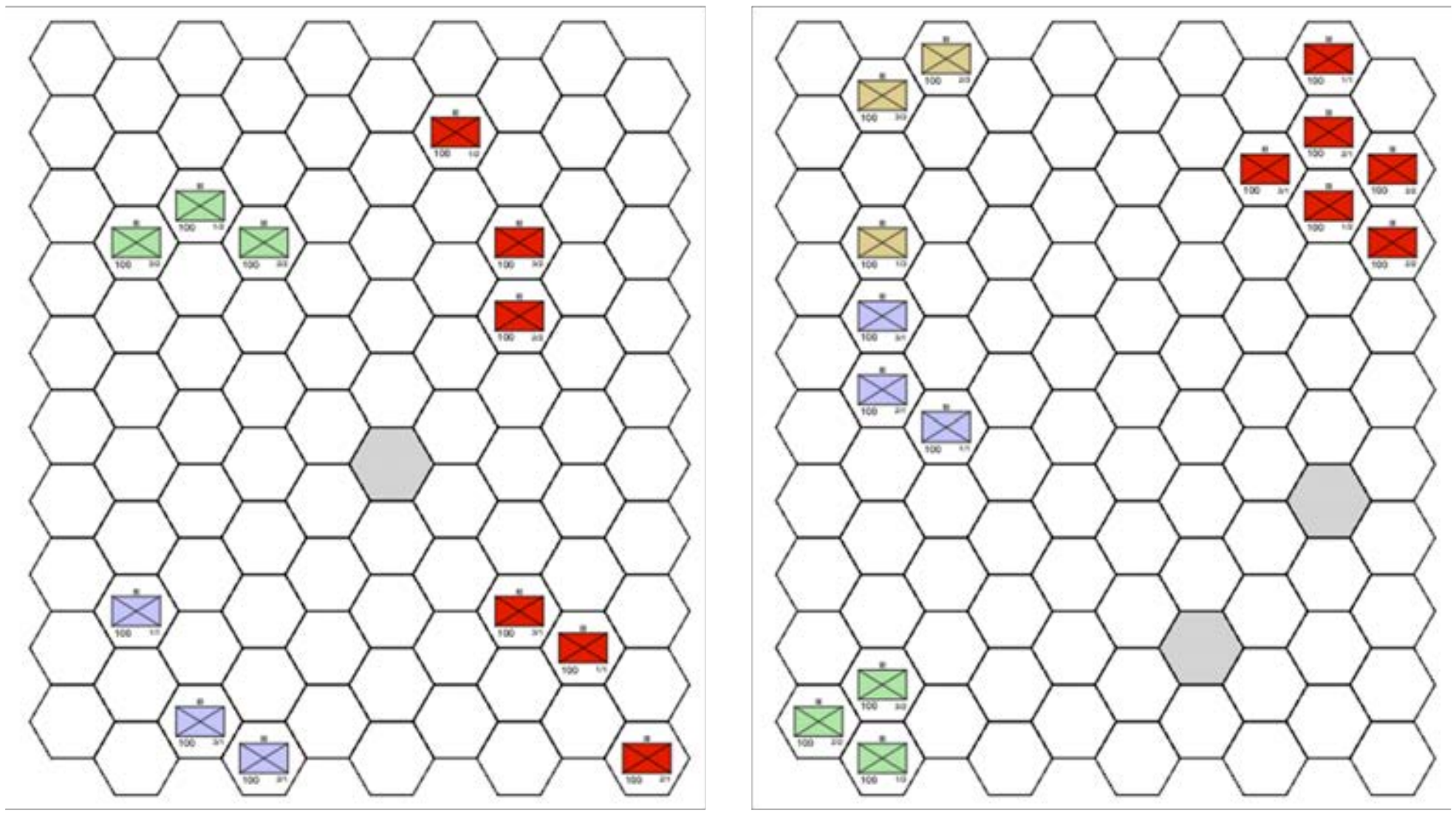

Figure 5.7. Example Scenarios. Two examples of randomly generated scenarios. The Blue faction is color-coded blue, green, and yellow to indicate they belong to separate Managers. The Red faction is color-coded red. Urban hexagons are color-coded grey.

### 5.3.8 Reinforcement Learning Training

Due to the complexity of the task, we use a `scenarioCycle` of 10 rather than 0 to determine if the *RL Manager Agent* is able to learn this task within a reasonable timeframe. Initial experiments revealed that, although using a `scenarioCycle` of 0 showed promise, it would require much larger training budgets to achieve reasonable performance for this study. Therefore, we choose a `scenarioCycle` of 10 which still results in a very challenging but feasible task. This `scenarioCycle` results in the simulation generating 10 random scenarios and then continuously cycling through these 10 scenarios, thus introducing a level of variability and unpredictability into the training process without overwhelming the *RL Manager Agent* with infinite possibilities—ensuring that the agent encounters a variety of situations but also gets the chance to learn from repeated exposure to the same scenarios. We believe that a `scenarioCycle` of 10 provides a reasonable balance between novelty and repetition, aiding in the deeper learning of strategies that are effective across different scenarios while also adapting to specific challenges presented within each scenario.

We use a training budget of 10 million steps for each of our *RL Manager Agents* and train each model against *Pass-Agg* as the adversary agent. Additionally, to ensure that our results are not overly dependent on a particular set of starting conditions, we use a separate `scenarioSeed` to train each of 5 different *RL Manager Agents*. These varying random seeds result in a different set of scenarios starting conditions as well as variations in learning experiences, which, when aggregated, offer a more comprehensive understanding of the model's performance.

To learn effective behaviors, we adapt the Boron reward system described in Section 3.5.2 which balances defeating the opposing faction and occupying urban hexagons with preserving its own force. This adapted reward function adheres to the SMDP framework in that it ensures that a manager's individual unit rewards are appropriately accumulated over the manager's temporal abstraction. This reward is computed each time a manager is called to make a decision (at the beginning of the game and any time all of the manager's units are within their assigned objective area):

$$R_{engineered,} = max(R_m - P_g, 0)\frac{S_{m,c}}{S_{m,o}} + B_t I_t \tag{5.1}$$

where $m$ represents a specific manager; $R_m$ is the accumulated manager $m$'s score based on its units' kills, losses, and urban hexagons held (as described in Section 3.4.3); $P_g$ is a penalty of 25 points for assigning the same objective area as one or more of the other managers (which discourages managers from assigning the same objective area unless it is overwhelmingly advantageous to do so); $S_{m,c}$ is the current total unit strength for manager $m$; $S_{m,o}$ is the original total unit strength for manager $m$; $B_t$ is a terminal bonus reward of 25 points that our research shows discourages units from moving into the adversary units' attack range during the last turn of the game; and $I_t$ is a terminal game state indicator that takes on a value of 1 if the game is terminal and a value of 0 otherwise.

The learning curves for each of the five seeds used are shown in Figure 5.8. As depicted, all of the *RL Manager Agents* still appear to be learning at training termination, indicating that if we afforded these agents a larger training budget, performance would likely continue to improve.

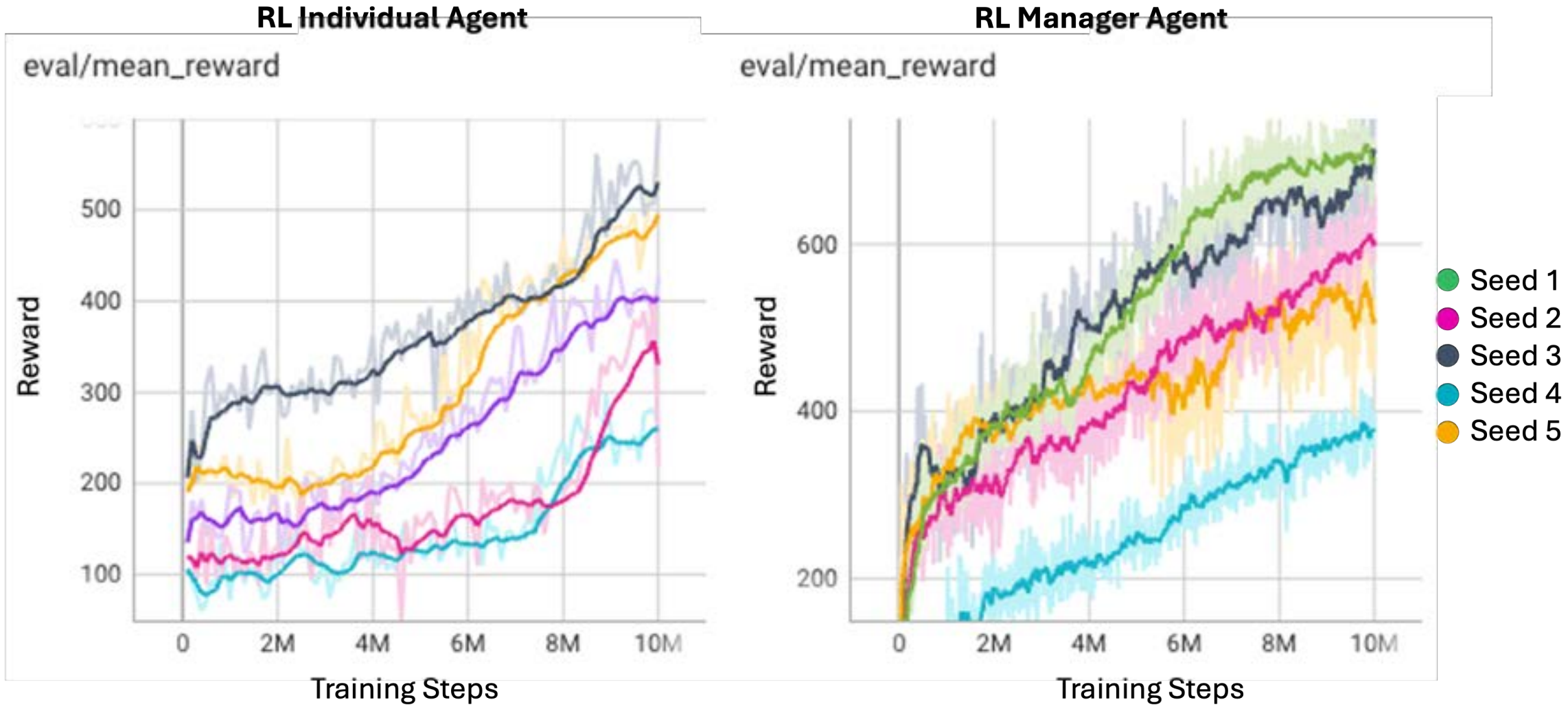

Figure 5.8. Learning Curves for RL Agent Training. The graph on the left depicts Mean Rewards vs. Training Steps for the RL Individual Agents, and the graph on the right is for the RL Manager Agents. The graph shows that all agents have yet to plateau when reaching our training budget of 10 million training steps. Of note, the RL Individual Agents are evaluated every 100, 000 training steps, while the RL Manager Agents are evaluated every 10, 000 training steps. This does not affect learning, but it is the reason for the reduced number of data points on the left graph as compared to the right graph.

We also train *RL Individual Agents* to include as part of our analysis. Unlike the *RL Manager Agent* which controls groups of units via objective areas, *RL Individual Agents* control the game actions of individual units (i.e., instead of a script delineating the unit's action, the RL model selects the action). Including this model as part of our evaluation allows us to compare our hybrid approach to the two models we integrate and look to improve upon.

We use the same methodology from Chapter 3 to develop and train these *RL Individual Agents* as described in Section 3.5.2. We use the global observation shown in Figure 3.9, the neural network architecture described in Section 3.5.2, and the reward equation depicted in Equation 3.6.

### 5.3.9 Model Evaluation

We evaluate each of our trained *RL Manager Agent* models using *Pass-Agg* as the adversary. We run 100, 000 games for each trained manager using the same respective `scenarioSeed` that was used for training, recording each score. We conduct three distinct evaluations for each seed used: *Hybrid Agent* vs. *Scripted Agent*; *RL Individual Agent* vs. *Scripted Agent*; and *Scripted Agent* vs. *Scripted Agent*. The performance of the *Scripted Agent* and the *RL Individual Agent* serve as our benchmarks to assess if our *Hybrid Agent* approach is more effective than either or both of these more traditional approaches.

## 5.4 Results and Discussion

The results of our evaluation are summarized in Table 5.1 and illustrated in Figure 5.9. Table 5.1 depicts the Mean Scores ± the Standard Error of the Mean (SEM) across the 100, 000 evaluation games for each model and each random seed used. Figure 5.9 shows the box plots for each model across the different seeds.

Table 5.1. Model Performance Results

| | Mean Scores ± SEM | | |
|---|---|---|---|
| Random Seed | Scripted | RL Individual | Hybrid |
| 1 | −19.9 ± 2.5 | −572.3 ± 1.2 | 692.1 ± 3.0 |
| 2 | −63.7 ± 2.7 | −366.2 ± 2.2 | 671.1 ± 2.4 |
| 3 | 162.1 ± 3.1 | −334.0 ± 2.3 | 580.4 ± 3.4 |
| 4 | −117.2 ± 2.1 | −693.2 ± 0 | 13.3 ± 2.8 |
| 5 | 262.9 ± 2.5 | −1051.1 ± 0.9 | 493.1 ± 3.3 |
| Overall Mean | −0.904 | −603.371 | 489.998 |

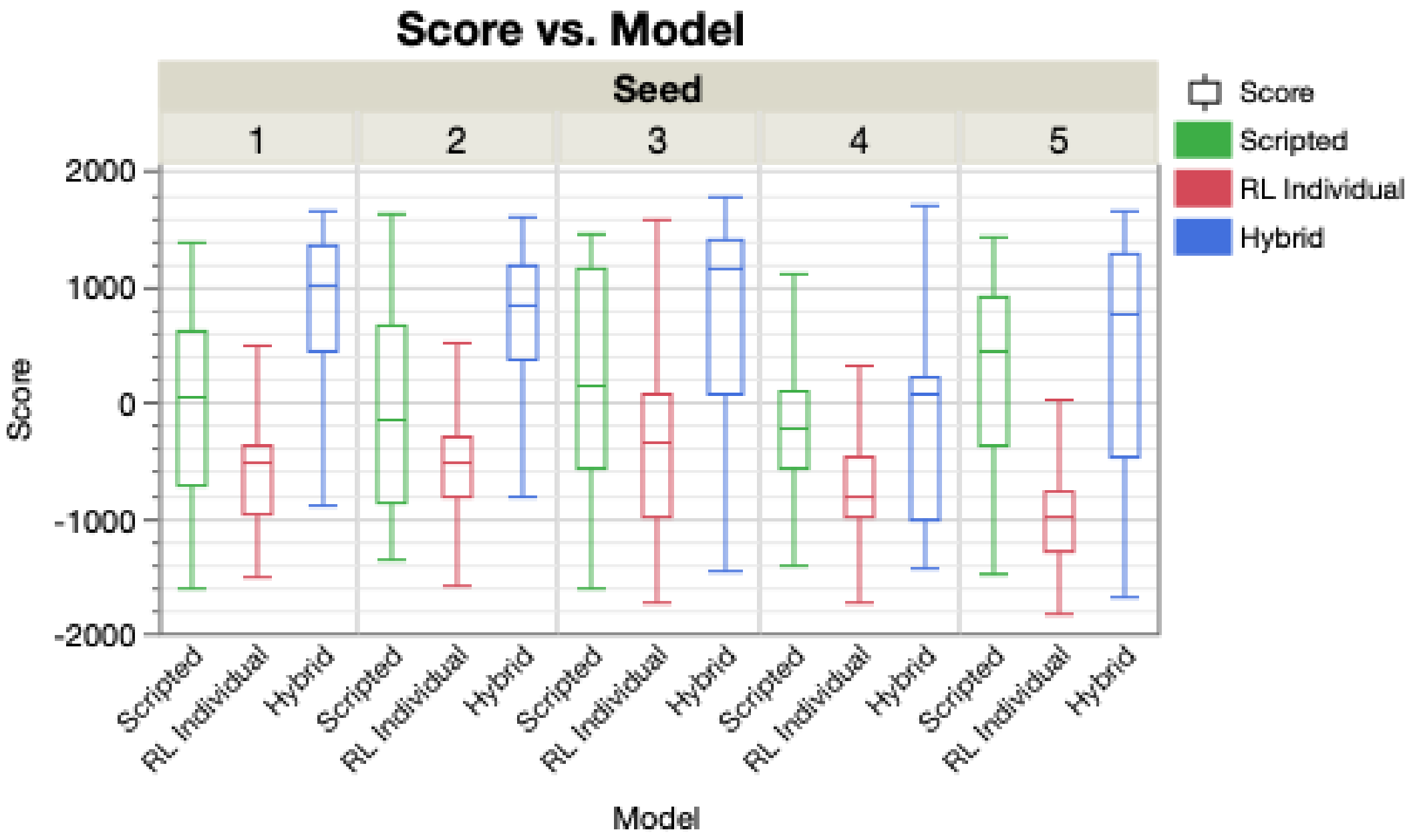

Figure 5.9. Box Plot of Scores. Box plots of Score vs. Model grouped by Seed are shown.

To confirm statistical significance, we conduct a Paired-Sample t-Test, setting $\alpha = 0.05$, and comparing the mean scores between *Hybrid Agent* vs. *Scripted Agent*; *Scripted Agent* vs. *Scripted Agent*; and *RL Individual Agent* vs. *Scripted Agent* across each of the 5 random seeds, for a total of 10 pairwise comparisons. Each of these returned a p-value of < 0.001, indicating statistical significance across all comparisons.

Our analysis shows that the *Hybrid Agent* outperforms both the *Scripted Agent* and the *RL Individual Agent* by large margins across the board. Visual replays of all scenarios also confirmed superior performance of the *Hybrid Agent*. This outcome validates the potential efficacy of integrating the tactical reliability of scripted agents with the strategic adaptiveness of RL.

These results demonstrate the *Hybrid Agent's* ability to learn and apply strategies across a number of diverse scenarios, though we have not been able to generalize to a `scenarioCycle` of 0 using the described architecture. However, the learning curves of the *RL Manager Agents* show a consistent improvement in performance, suggesting that with extended training or increased computational resources, better performance and further generalizability are still achievable.

Nevertheless, our analysis suggests that by delegating specific tactical decisions to scripted agents, the RL component is freed to concentrate on higher-level strategic considerations, thus reducing the computational burden and enhancing overall system performance. This division of labor not only streamlines the decision-making process but also improves the scalability of the model to handle more complex and larger-scale scenarios.

Lastly, these results also show that the integration of RL with scripted agents can lead to a more nuanced and flexible response mechanism, allowing the *Hybrid Agent* to effectively adapt to changing conditions and unforeseen challenges more quickly by allowing developers to independently alter either the scripted behavior algorithms or the RL training mechanisms. This blend of predictability and adaptability is particularly valuable in the context of military simulations, where both qualities are essential for realistic and effective strategy formulation.

## 5.5 Conclusion

This study confirms the benefits of employing a hierarchical hybrid AI model for combat simulation applications. Given a limited number of starting conditions, our evaluation and analysis demonstrate that this hybrid approach surpasses the capabilities of traditional RL and scripted strategies when applied independently.

We find that the integration of RL with scripted agents effectively mitigates the inherent limitations of each method when used separately. Specifically, employing RL using a higher level of abstraction is essential, as directly applying RL at the unit level is impractical due to RL's sample inefficiency problem. Additionally, scripted agents, while providing consistency and reliability for routine tasks, struggle with executing complex maneuvering and adapting to dynamic changes in the environment due to their rigid, predefined rules. By employing a hierarchical structure, this approach utilizes RL for higher-level strategic decision-making while employing scripted agents to manage routine, lower-level tactical decisions. This two-level approach not only improves overall performance but also offers a scalable solution capable of adapting to increasingly complex scenarios.

Nevertheless, this study is not without its challenges and limitations. We acknowledge existing hurdles, particularly concerning computational efficiency and the scalability of RL algorithms in general and within our hybrid model. As we integrate this approach into our overall HRL framework, we will simplify the problem by further reducing the observation space and the action space, as well as providing for more efficient computations of the objective areas by using a square box rather than a super hexagon.

Overall, the insights gained from this study offer a promising path forward for the development of AI-driven agents in strategic combat simulations, real-time strategy games, and other similar domains. By combining the predictable precision of scripted systems with the dynamic adaptability of RL, this method advances the research of intelligent agents in complex simulation environments and embodies a novel integration of established and emerging AI methodologies—offering an alternative to improving the capabilities and performance of intelligent systems in this challenging domain.

# CHAPTER 6:
# Hierarchical Reinforcement Learning Framework

Portions from this chapter have been published in the *Meeting Proceedings of the NATO Modeling and Simulation Group Symposium (NMSG) Simulation: Going Beyond the Limitations of the Real World* [2] and the *Proceedings of the 2023 Interservice/Industry Training, Simulation, and Education Conference (I/ITSEC)* [3]. These publications hold no copyrights over the presented material.

As discussed in Chapter 2, RL has shown great promise in developing intelligent agents but encounters significant challenges when tasked with learning optimal policies in environments characterized by long task horizons and extensive state and action spaces, such as those found in realistic combat simulations. These complex environments typically create a vast exploration space, subject to Bellman's "curse of dimensionality" [83]. Standard RL algorithms often cannot perform at or above human levels in such environments or tend to exhibit brittleness when overspecialized to perform specific tasks but then transferred to even slightly varied scenarios [86]. HRL has been shown to address these issues by structuring the RL problem into smaller, manageable sub-tasks that may render complex problems more tractable. However, while HRL may be able to more efficiently address high-dimensional problems, it does not ensure that the solutions to these decomposed problems are optimal compared to the original RL problem [83]. Nonetheless, a hierarchical approach is well-aligned to military organizational and decision structures, offering potential benefits in explainability and validation by subject-matter experts.

This chapter integrates the methodologies discussed in previous chapters into a new HRL framework, designed to enhance intelligent agent behaviors in more complex combat simulation scenarios. As shown in Chapter 5, by structuring the AI system hierarchically, we aim to leverage the strengths of various AI methods for different levels of decision-making. Specifically, in this chapter, we explore the integration of observation abstractions and multi-model frameworks within a three-level hierarchical structure.

## 6.1 Background

We cover much of the background information for this study in Chapter 2. Sections 2.4 and 2.5 specifically discuss RL and HRL where we highlight why and how RL has been particularly effective in uncertain environments where the optimal strategy is not known a priori, such as in the games of StarCraft [44], Dota 2 [43], Go [79], and Atari [40]. However, we also highlight how, despite showing success in smaller contexts [80]–[82], the application of RL in the more complex domain of combat simulations in support of wargaming has yet to surpass the performance of human or scripted agents [11]. This issue primarily stems from the exponential increase in state space complexity [70] and RL's sample inefficiency problem [40], which becomes more pronounced as the complexity of the observation space grows. HRL shows promise in addressing this problem in that it breaks down an RL problem into a hierarchical structure of sub-problems, where higher-level parent tasks call upon lower-level child tasks as primitive, or basic, actions [83]. This hierarchical decomposition approach offers the advantage of reducing overall computational complexity, provided the problem can be structured accordingly [83]. HRL algorithms have been shown to outperform standard RL algorithms on long-horizon or complex problems [83].

In essence, HRL leverages abstraction as a fundamental component which can take form as *state abstraction*, *action abstraction*, or a combination of both in *state-action abstraction*. Furthermore, implicit in action abstraction is the notion of *temporal abstraction* in that higher-level actions will usually involve a timespan overlapping multiple lower-level actions. In Chapter 3, we highlight the role state abstraction plays in simplifying complex decision-making processes [91] and, given limited computational power and a complex enough simulation environment, agents cannot model everything in their environment and still learn effective behaviors within a reasonable time. Assuming that a minimum level of performance is desired, as the complexity of the environment increases, so will agents need to discard some information and focus only on information relevant to a specific problem [91]. Abstractions enable both humans and machines to make quicker and often more effective decisions by focusing only on the relevant details and ignoring less relevant information [141]. Therefore, implementing abstraction techniques could not only lessen computational requirements but also enhance the adaptability and efficacy of AI-trained agents in environments that diverge from those in which they were originally trained [75].

In this chapter, we introduce the idea of action abstractions in more detail. This concept helps address one of the fundamental challenges in RL: the explore-exploit dilemma. This problem encapsulates the dual need for RL agents to explore their environment to gather new information while also exploiting their existing knowledge to make optimal decisions. By using action abstractions, agents can narrow down the state space, enabling them to more efficiently identify promising future states that would be difficult to find if they had to search the entire state space. This balancing act becomes particularly difficult in settings where the majority of rewards may be zero, with occasional signals that distinguish good actions from bad actions [75].

Initially described in Section 2.5.2, *options* represent a form of action abstraction in RL [87]. *Options* extend the usual notion of single-step actions to multi-step strategies that achieve specific sub-goals. By using *options*, an RL agent can perform complex tasks more efficiently by focusing on higher-level objectives rather than individual actions. This allows for more structured exploration and decision-making in environments where planning over longer horizons may be beneficial [87]. *Options* can simplify the action space by allowing the agent to make decisions at higher levels of abstraction, reducing the need for detailed planning at every step. Furthermore, *options* are known to aid in knowledge transfer across different tasks, enhance exploration by directing the agent toward significant states, and make the planning process more efficient by reducing the complexity and depth of the decision tree an agent must consider [75], [142], thus simplifying the planning process. This efficiency is particularly beneficial in environments where decisions have long-term implications, and simplifying the complexity of planning can lead to faster and more effective decision-making.

To illustrate the concept of options, we use the *Four Rooms* example presented by Sutton et al. [87] illustrated in Figure 6.1. In this gridworld environment, each cell represents a possible location to which the agent can move. The agent can only move up, down, left, or right. In each of the four rooms, there exist two specific hallway options ($o_1$ and $o_2$). These options are designed to navigate the agent from any point within a room to one of two designated hallway cells that exit the room. Each hallway option employs a policy, $\pi_o$, that maps the shortest path to its intended hallway cell while avoiding routes that could accidentally lead to the other hallway. An example of such a policy for one hallway option is shown in Figure 6.2. Thus, when the agent is given the goal $G_1$ and chooses option $o_1$, it will then employ the underlying policy $\pi$ to go from its current location to option $o_1$.

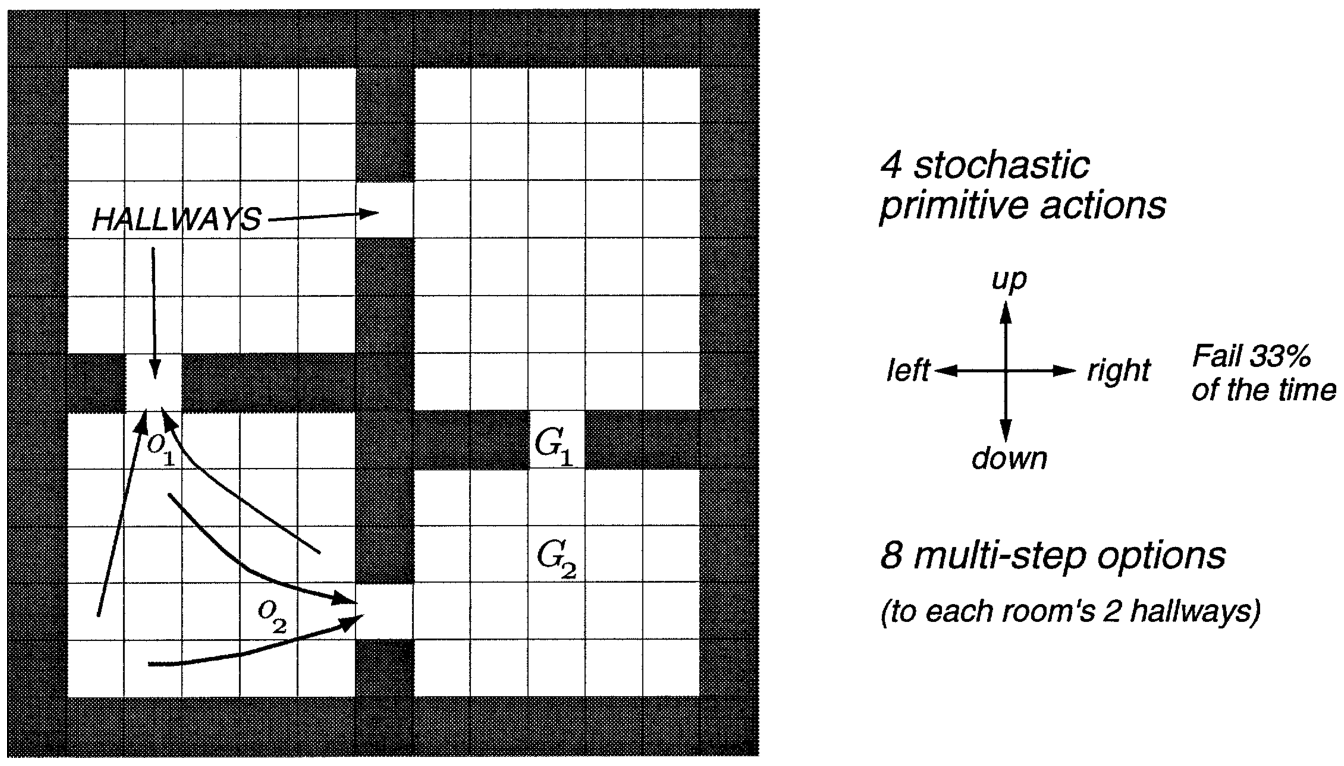

Figure 6.1. Four Rooms Problem. The four rooms problem presented by Sutton et al. [87] with cell-to-cell actions and room-to-room hallway options. Two of the hallway options are shown with arrows labeled $o_1$ and $o_2$. Goals are labeled $G_1$ and $G_2$. Source: [87].

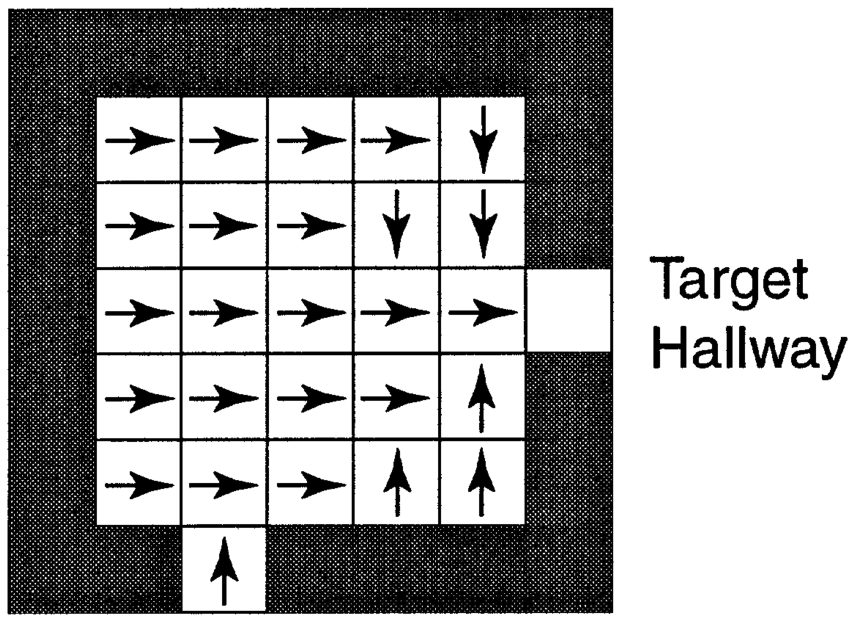

Figure 6.2. Underlying Policy for Hallway Options. An example of the underlying policy for one of the eight hallway options. Source: [87].

The value of using the concept of options is shown in Figure 6.3. This figure illustrates the difference between planning using original actions ($O = A$) and planning with the inclusion of hallway options ($O = H$). The top part of the figure displays the value function after the first two iterations of synchronous value iteration (SVI) using only primitive actions. In each iteration, the valued states expand by one cell and, as shown, after two iterations, most states retain their initial arbitrary zero value. The bottom part shows the value functions for SVI when incorporating hallway options. After the first iteration, accurate values are achieved for all states in rooms next to the goal state due to the existing underlying policy, and by the second iteration, accurate values are achieved by all states within the gridworld. While the values underwent minor changes in further iterations, a complete and optimal policy was already established. Unlike the step-by-step approach, the hallway options facilitated a more rapid, room-by-room planning process.

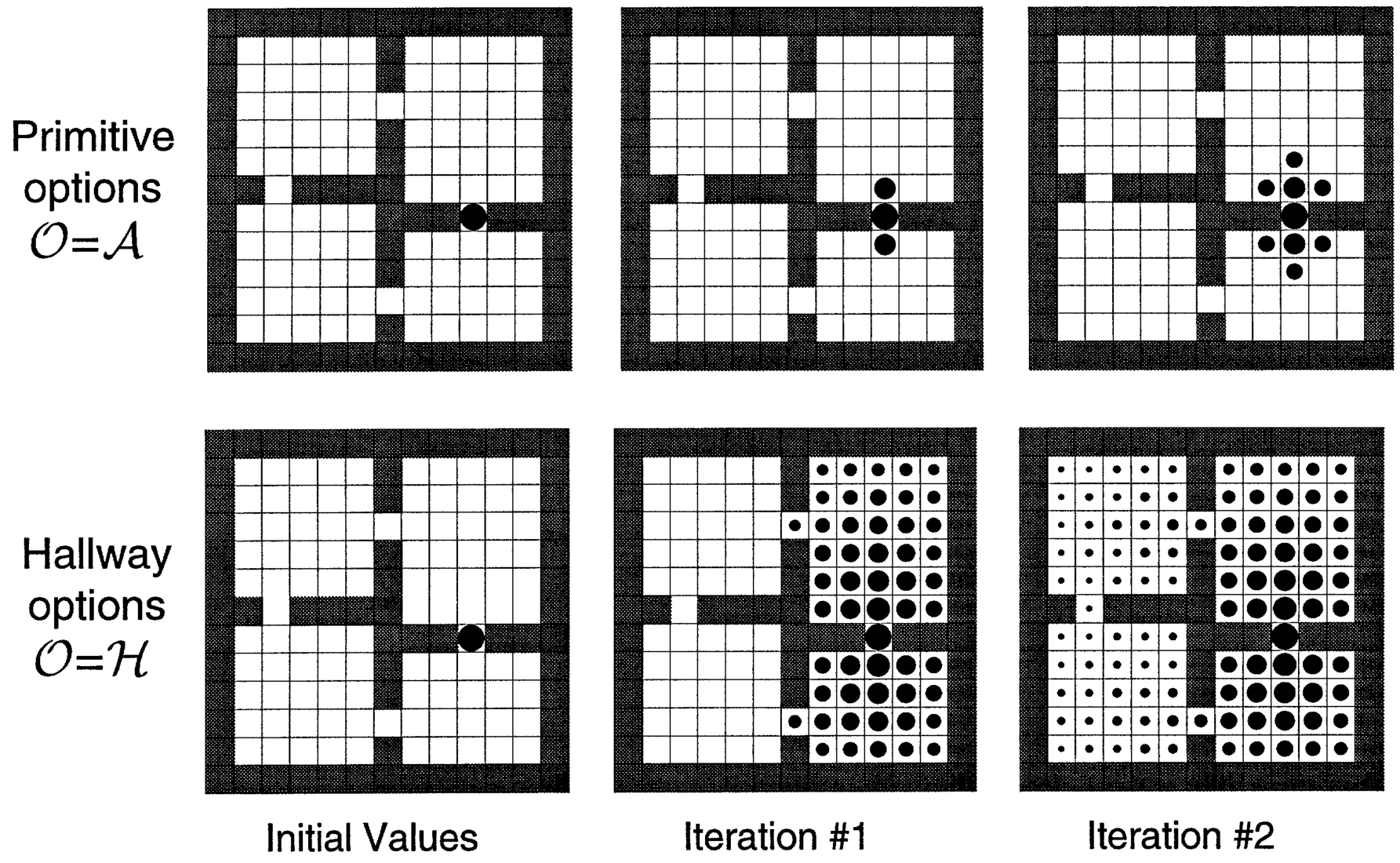

Figure 6.3. Value Functions for the Four Room Problem. Value functions are generated over several iterations of planning through synchronous value iteration using primitive options (top) and multi-step hallway options (bottom). Utilizing hallway options allows planning to advance on a room-by-room basis instead of cell-by-cell. The size of the disk in each cell represents the state's estimated value. A disk completely filling a cell signifies a value of 1. Source: [87].

Options have been shown to facilitate knowledge transfer across different tasks within RL environments [75], [143]–[146]. This transfer is crucial for enhancing the learning efficiency of agents when they encounter similar tasks or environments, allowing them to effectively apply previously learned strategies. Additionally, these action abstractions can significantly enhance the exploration strategies of an agent [147]–[150]. By structuring the action space with high-level strategies, agents can avoid exhaustive exploration of unproductive actions and instead focus on actions that are more likely to achieve significant states or goals. This targeted exploration is critical in complex environments where random exploration would be computationally expensive or less effective [151].

While the concept of action abstraction has traditionally been viewed as a broad category of operations that simply alter an agent's action space, Abel [75], defines action abstraction more precisely as the substitution of an agent's basic actions with a set of options, denoted as $O$. He defines *action abstraction* as a function $\omega : A \rightarrow O$ that replaces actions $A$ with a set of options $O$. Abel [75] notes, however, that with $O$ replacing the primitive action space, every policy over the set of states $S$ and actions $A$ may not necessarily be represented. Thus, it is the case that action abstractions may prevent the agent from ever discovering an optimal or near-optimal policy [75].

When an agent selects from available options, it commits to following the policy associated with that option until it reaches a state where the sampled termination condition is met. Once reaching the option's termination condition, the agent ceases to follow the option's policy and will decide on its next action or option [75]. This selection process essentially prunes other potential actions while the option is active, as any action not included in the option's policy is disregarded during its execution. Consequently, the decision-making process introduces a dependency on the option being followed, since the policy an agent follows while an option is active varies based on the specific option [75]. This process is illustrated in Figure 6.4.

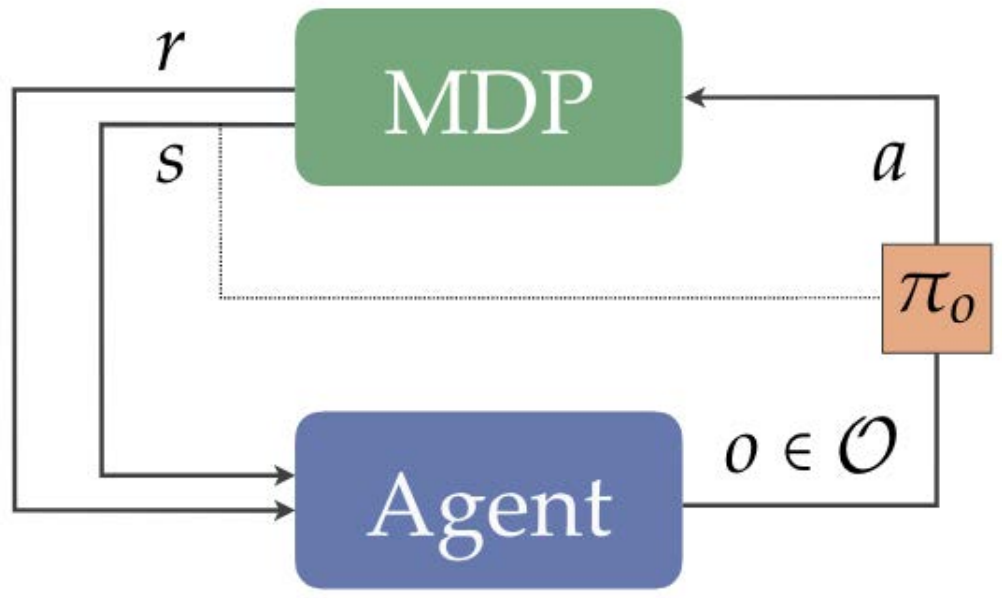

Figure 6.4. Reinforcement Learning with Action Abstraction. Instead of a primitive action, the agent chooses an option $o \in O$ per policy $\pi_O$. Source: [75].

Next, by combining the state abstraction process (discussed in Section 3.2 and shown in Figure 3.3) with the action abstraction process (shown in Figure 6.4), Abel [75] presents the resultant state-action abstraction depicted in Figure 6.5. This figure illustrates how after an agent receives the abstract state $s_\phi$ it then selects an option $o \in O$ and follows the policy $\pi_o$ until reaching a termination condition, where the process repeats.

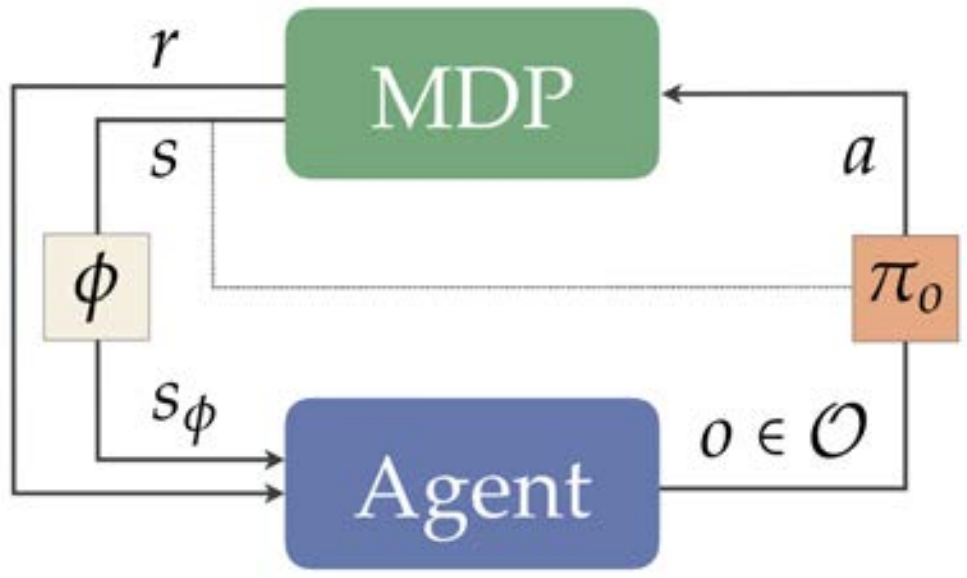

Figure 6.5. State-Action Abstraction in Reinforcement Learning. The state output $s$ by a Markov decision process is fed through a function $\phi$ prior to becoming an input $s_\phi$ to the agent. The agent then chooses an option $o$ and follows the option's policy $\pi_o$ until termination. Source: [75].

Finally, Abel [75] formalizes the notion of hierarchical abstractions as illustrated in Figure 6.6. He first defines a *hierarchy* as $n$ sets of $(\phi, O_\phi)$ pairs consisting of a state abstraction $\phi$ and associated options $O_\phi$. He then defines $\pi_n : S_{\phi,n} \rightarrow O_{\phi,n}$ which describes the relationship between policies, state spaces, and options within the context of RL [75]. In this model, $\pi_n$ represents the policy at the $n$-th level of the hierarchy. This policy maps abstracted states to options based on the state of the system as abstracted by the function $\phi$. The term $S_{\phi,n}$ denotes the abstracted state space at the $n$-th level, simplified by $\phi$ to make decision-making more manageable at this specific level of complexity [75]. The policy $\pi_n$ maps these abstracted states to options, denoted by $O_{\phi,n}$. These options are essentially higher-level actions or sequences of actions aimed at achieving specific sub-goals, tailored to operate effectively within the simplified context provided by $S_{\phi,n}$ [75].

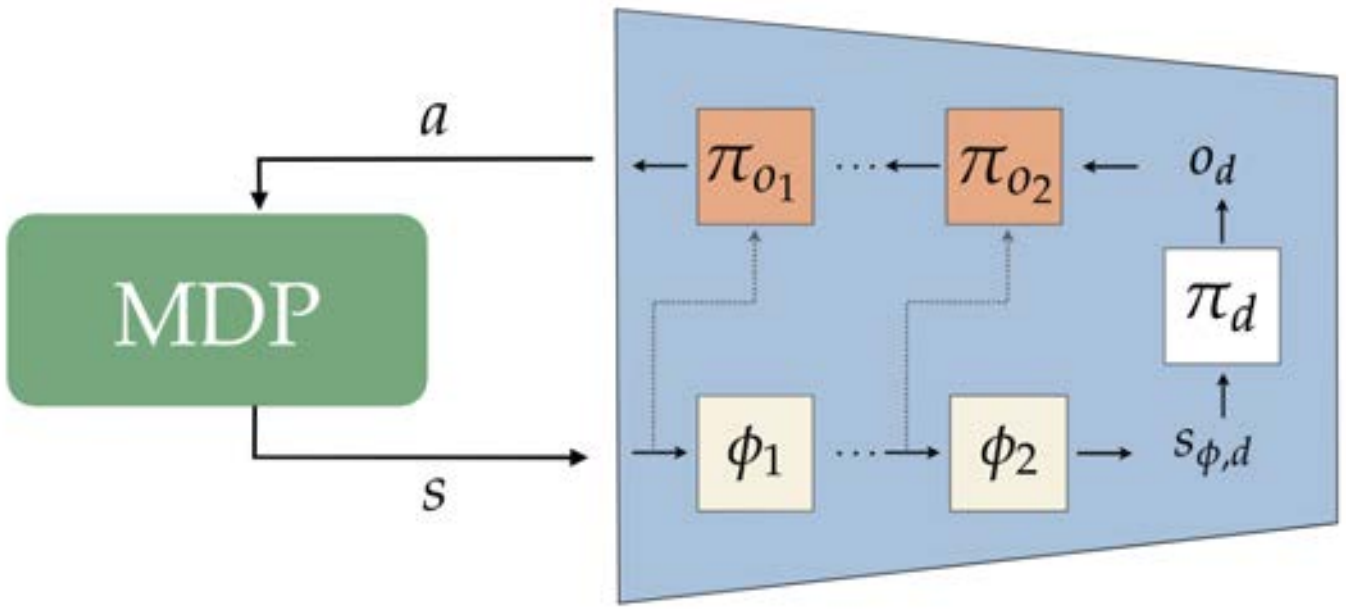

Figure 6.6. Hierarchical Abstraction for Reinforcement Learning. Construction of a hierarchy from $(\phi, O)$ pairs. Source: [75].

In summary, this section highlights how HRL, through different levels and forms of abstraction, may be able to decompose a problem into hierarchical sub-problems to better manage the inherent complexities involved in combat simulations. By incorporating the notion of options and action abstractions, HRL not only simplifies the decision-making process but may also significantly enhance the efficiency and adaptability of RL agents in complex environments where the vastness of the state and action spaces often overwhelms traditional RL approaches.

## 6.2 Related Works

In this section, we review key works in HRL, focusing on its development from foundational theories to advanced applications across domains relevant to combat modeling and simulation. Beginning with Dayan and Hinton's [152] introduction of feudal reinforcement learning, which organized learning through a hierarchy of Managers and Sub-Managers, we explore how HRL has evolved to address challenges in managing large state and action spaces, long task horizons, and sparse rewards. We examine significant contributions to the field to understand and highlight the current state and ongoing challenges in HRL.

In their seminal work, Dayan and Hinton [152] introduced the concept of *Feudal Reinforcement Learning* as a method to accelerate RL training, where they create a Q-learning hierarchy in which *Managers* learn to set tasks for their *Sub-Managers*. The Sub-Managers then learn to satisfy these higher-level tasks. The idea of feudal control mirrors aspects of feudal fiefdom, where Managers are given absolute power over Sub-Managers, and these same Managers ultimately must satisfy their own Super-Managers. Each layer within the hierarchy simply seeks to maximize its own expected reward, leading to a recursive reinforcement and selection process until the entire hierarchical system satisfies the goal of the highest-level Manager [152]. This approach overcomes the three bottlenecks identified by Singh [153] of *temporal resolution*, *divide and conquer*, and *exploration*.

Under this control hierarchy, there may exist Super-Managers, Intermediate-Managers, and Sub-Managers, where each Manager level only controls the level immediately below, and only the bottom level of Managers can actually act on the world [152]. Higher-level Managers have a larger grain of temporal resolution as they hand tasks downwards; intermediate-level Managers choose amongst Sub-Managers and set their sub-tasks; while the lowest level of Managers performs actions in the environment [152]. This type of hierarchy allows for the exploration of actions that maximize rewards to be more efficient since high-level Managers can send their agents directly toward the region where a higher reward can be found rather than letting the agents explore the entire space in detail [152].

For feudal reinforcement learning to be effective, Dayan and Hinton [152] identify two key principles: *reward hiding* and *information hiding*. *Reward hiding* consists of rewarding Sub-Managers for accomplishing their tasks regardless if it satisfies the goals of the Super-Managers (i.e., Sub-Managers should just obey their immediate Managers regardless of how it satisfies levels further up the hierarchy) [152]. This allows Sub-Managers to learn to achieve their assigned sub-goals even if the Manager is mistaken in assigning these specific sub-goals, which in turn allows low-level Managers to become competent even if the higher-level Managers are still learning. *Information hiding*, on the other hand, consists of allowing Managers to only be able to observe the state space at the appropriate granularity based on their own choices of tasks [152]. This allows some decision-making to happen at a coarser grain, which is one of the major benefits of a hierarchical decomposition. Additionally, information hiding occurs in both directions—Super-Managers are not aware of tasks set by Managers, and Sub-Managers are not aware of tasks set by Super-Managers [152]. This principle includes both coarsening the state representation and selectively hiding actions, and it is implemented as needed to support the algorithm's effectiveness [152].

For their experiments, Dayan and Hinton [152] borrowed a standard maze task (shown in Figure 6.7) from Barto et al. [154] where an agent is tasked to find an initially unknown goal. Figure 6.8 then shows how the environment can be split into successive hierarchical layers of increasingly finer grain where Managers are assigned to different parts of the maze at each level. Each Manager maintains Q-values over the actions it can instruct its Sub-Managers to perform (e.g., move to the north, south, east, or west) based on the location of the agent at the next subordinate level and the command it has received from its own Super-Manager [152].

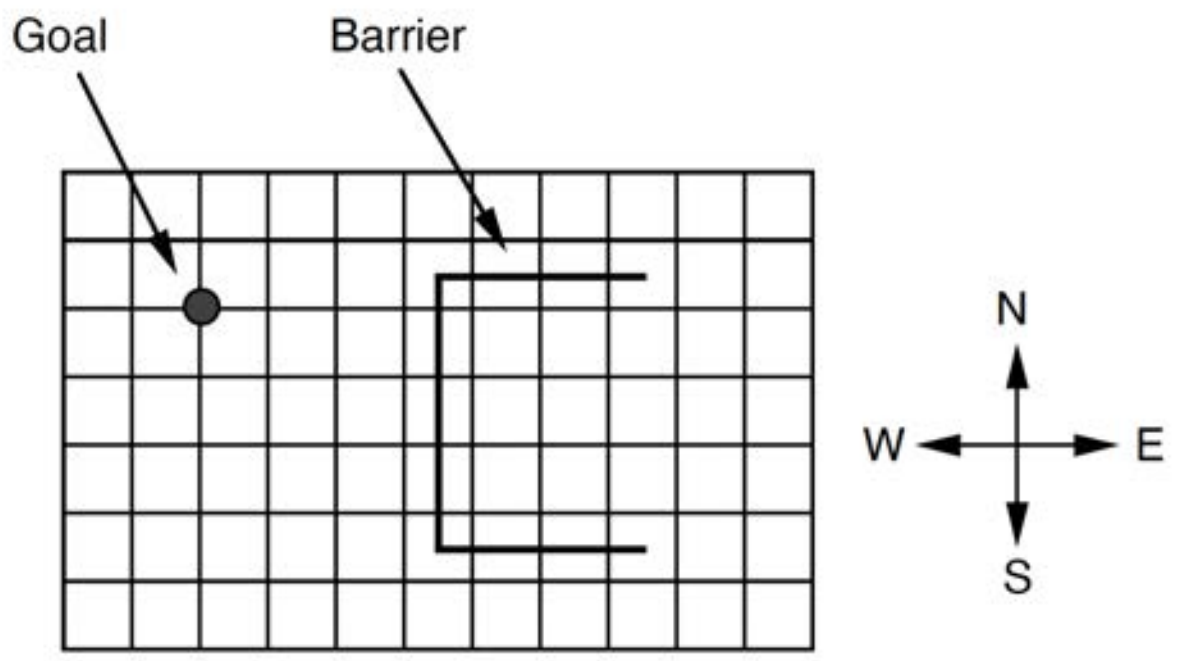

Figure 6.7. Maze Task. Maze task used in [154]. Line intersection are locations an agent can move to. The barrier and goal location are labeled. Source: [154].

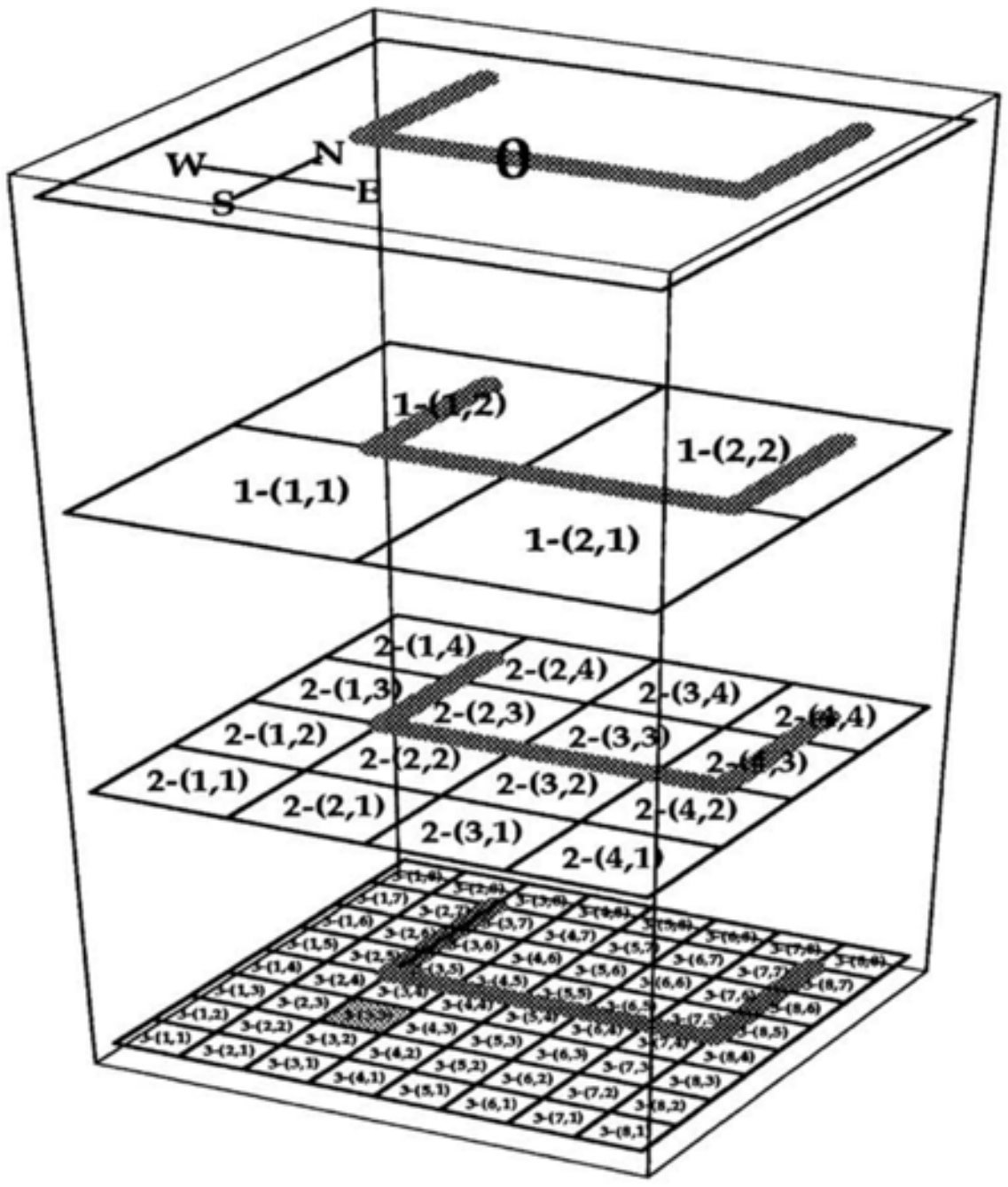

Figure 6.8. Grid Task. This shows how the maze used in Figure 6.7 can be divided up into different hierarchical levels. The ‘u’ shape represents the same barrier, and the shaded square represents the goal. Source: [152].

In this construct, action choices flow from top to bottom. Figure 6.9 shows the probabilities (represented by the area of the shaded rectangles) for taking a specific action (the direction the line is pointed) learned for each location [152]. As one can see, actions are generally correct for each location with the exception of the right side of the lowest-level grid. Here it shows the agent choosing north when it should instead choose south. This sub-optimal behavior is caused by this specific decomposition of the grid space [152].

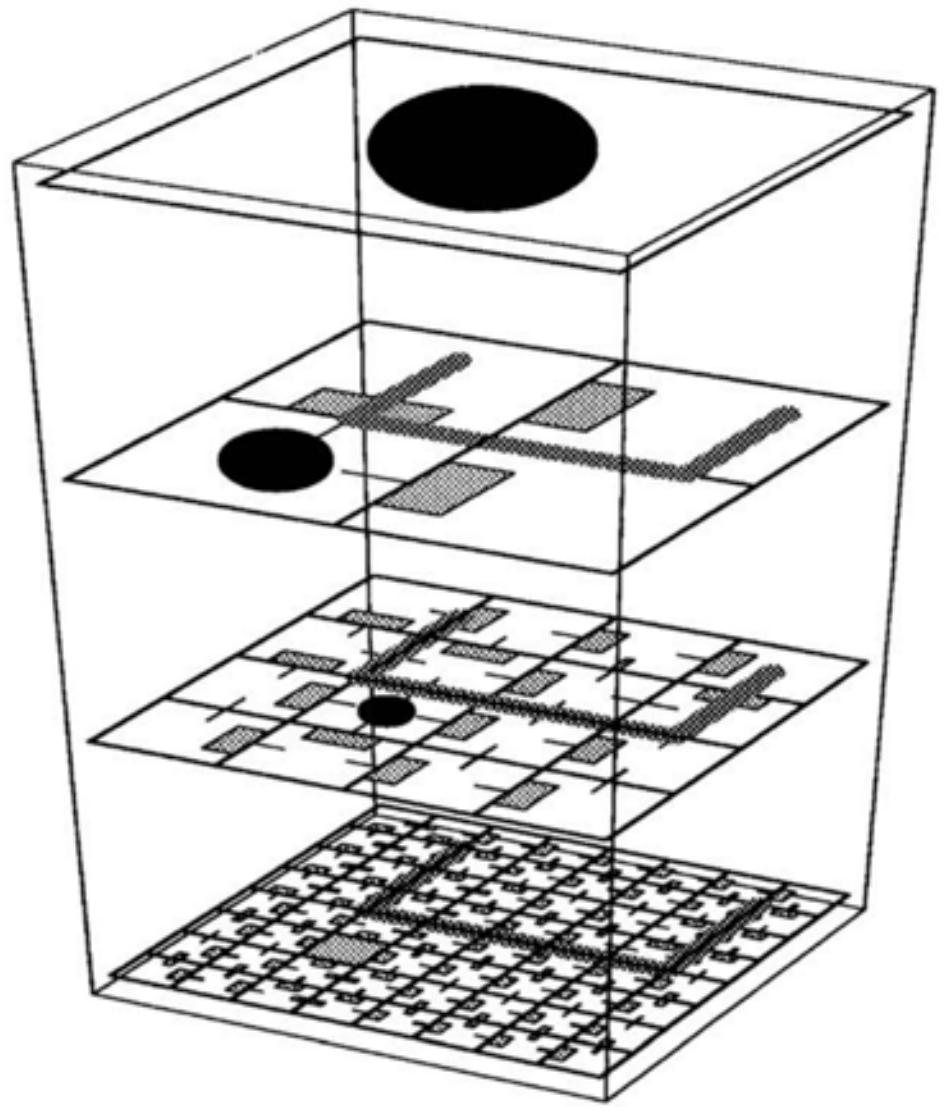

Figure 6.9. Grid Task Learned Actions. The area of the shaded boxes gives the probability of taking a step in the direction indicated by each line. Source: [152].

Of note, in their experiments, Dayan and Hinton [152] showed that while learning initially occurs from a bottom-up manner, after a while, the agents begin to learn at the highest levels resulting in additional top-down learning. Moreover, although the information hiding results in smaller-sized state spaces that must be searched, it also introduces some inefficiencies. For example, there are scenarios where if a Sub-Manager knew the super-task assigned to its Super-Manager, then it could circumvent its immediate Manager and eliminate an entire layer completely. However, this reduction would then require the Sub-Manager to observe as large of a state space as the entire problem itself, thus countering one of the benefits of a hierarchical approach. Lastly, for this specific task, Dayan and Hinton [152] showed that the feudal system learned much more quickly than a standard Q-learning system would. Whereas a standard Q-learning system is only concerned with the target, the feudal system learns to efficiently execute higher-level moves—not only those necessary for achieving the current goal but also those that could be useful for future objectives—thereby making exploration more efficient [152].

Vezhenevets et al. [155] built on Dayan and Hinton's work [152] and introduced Feudal Networks (FuNs), an HRL architecture that formulates sub-goals as directions in latent space which result in meaningful primitive actions. FuNs decouple end-to-end learning across multiple hierarchical levels allowing it to use different resolutions of time [155]. Key insights in this approach include how goals can be generated and promulgated in a top-down hierarchical manner; how goal setting and goal achievement can be decoupled; how each level in the hierarchy can tell the level immediately below which goal to achieve (without specifying how to do so); and how making use of a lower temporal resolution for high-level goals naturally structures the lower-level agents' behavior into temporally extended sub-policies [155].

The FuN framework is essentially a modular neural network that uses a *Manager* module and a *Worker* module. This leads to a natural hierarchical approach that is stable and results in both modules learning desired behaviors in complementary ways. Under this framework, first, the Manager uses a lower temporal resolution and assigns abstract goals to its Workers. However, the Manager and Worker still share a perceptual module that takes in an observation and computes a shared intermediate representation of the environment [155]. The Workers then train to accomplish the Manager's goals using intrinsic rewards and an approximate transition policy gradient, which is a more efficient form of policy gradient training in that it assumes that the Worker's behavior will ultimately align with the Manager's goals. In other words, because the Worker is learning to achieve the Manager's direction, over time, its transitions should converge to a distribution around the Manager's direction. FuN then computes a latent space representation and outputs a goal vector. Based on its current state, its observations, and the assigned goals from its Manager, the Workers then accomplish these goals by generating the needed primitive actions at every time step. Overall, this decoupled structure enables better long-timescale credit assignments and it also results in the emergence of sub-policies associated with different goals or tasks set by the Manager. Vezhnevets et al. [155] demonstrate this through their experiment where FuNs dramatically outperformed a strong baseline agent on tasks involving long-term credit assignments.

While conventional RL would suggest training the FuN architecture monolithically (i.e., concurrently training both levels of the hierarchy), Vezhnevets et al. [155] argued that this approach would make the Manager's goals just become internal latent variables, depriving them of any semantic meaning. Therefore, Vezhnevets et al. [155] proposed to instead independently train the Manager and the Worker (i.e., train the Worker first, then freeze its parameters to train the Manager), where the Manager is trained to predict the best transitions in the state space and to intrinsically reward its Worker to realize these advantageous state transitions [155]. Also of note, Vezhenevets et al. [155] took a modified approach to Dayan and Hinton's [152] principle of completely concealing the environment rewards from lower levels in the hierarchy. Instead, they added an intrinsic reward for the Worker to follow the Manager's goals while also retaining the environment rewards. Additionally, their approach involved using different discount factors when computing returns which allows, for example, for the Manager to have a more long-term perspective while the Worker can be greedier for immediate rewards.

In the end, in their experiments, Vezhenevets et al. [155] showed that the FuN architecture—using a representation learned by a CNN— outperformed a traditional recurrent Long Short-Term Memory (LSTM) network in several games, including Montezuma's Revenge, ATARI games (i.e., Ms. Pacman, Amidar, Gravitar, Space Invaders, Breakout, etc.), and a variety of mazes from Deepmind Lab [156] (a first-person 3D game platform). These experiments demonstrated that the FuN architecture can make long-term credit assignment and memorization problems more tractable. Furthermore, constructing deeper hierarchies with multiple time scales also opens the door to scaling to large, partially-observable environments containing sparse rewards. Lastly, this modular FuN architecture may also lend itself to transfer learning and multi-task learning (i.e., reusing learned primitives to acquire new complex skills) [155], a potentially valuable method to deal with the complexity of training combat agents for large-scale scenarios.

Pope et al. [157] also used a hierarchical approach to RL to develop an AI capable of piloting an F-16 in simulated air-to-air combat as part of the Defense Advanced Research Projects Agency (DARPA)'s AlphaDogfight Trials (ADT). The Lockheed Martin team [157] achieved a 2nd place finish, defeating an Air Force F-16 Weapons Instructor in match play. Pope et al. [157] used an approach that combined a hierarchical architecture, maximum entropy RL, expert knowledge integration through reward shaping, and support for the modularity of policies. The environment used in this competition was an OpenAI Gym environment where the F-16 aircraft were simulated using JSBSim [158], an open-source, high-fidelity flight dynamics model.

Using this environment, Pope et al. [157] developed an agent Policy Hierarchy for Adaptive Novel Generation of MANeuvers (PHANG-MAN) composed of a two-layer hierarchy of policies, shown in Figure 6.10. The high-level policy is a single policy that chooses which low-level policy to use given the current state or context of the engagement [157]. The low-level policies are thus trained to excel in only a specific region of the state space.

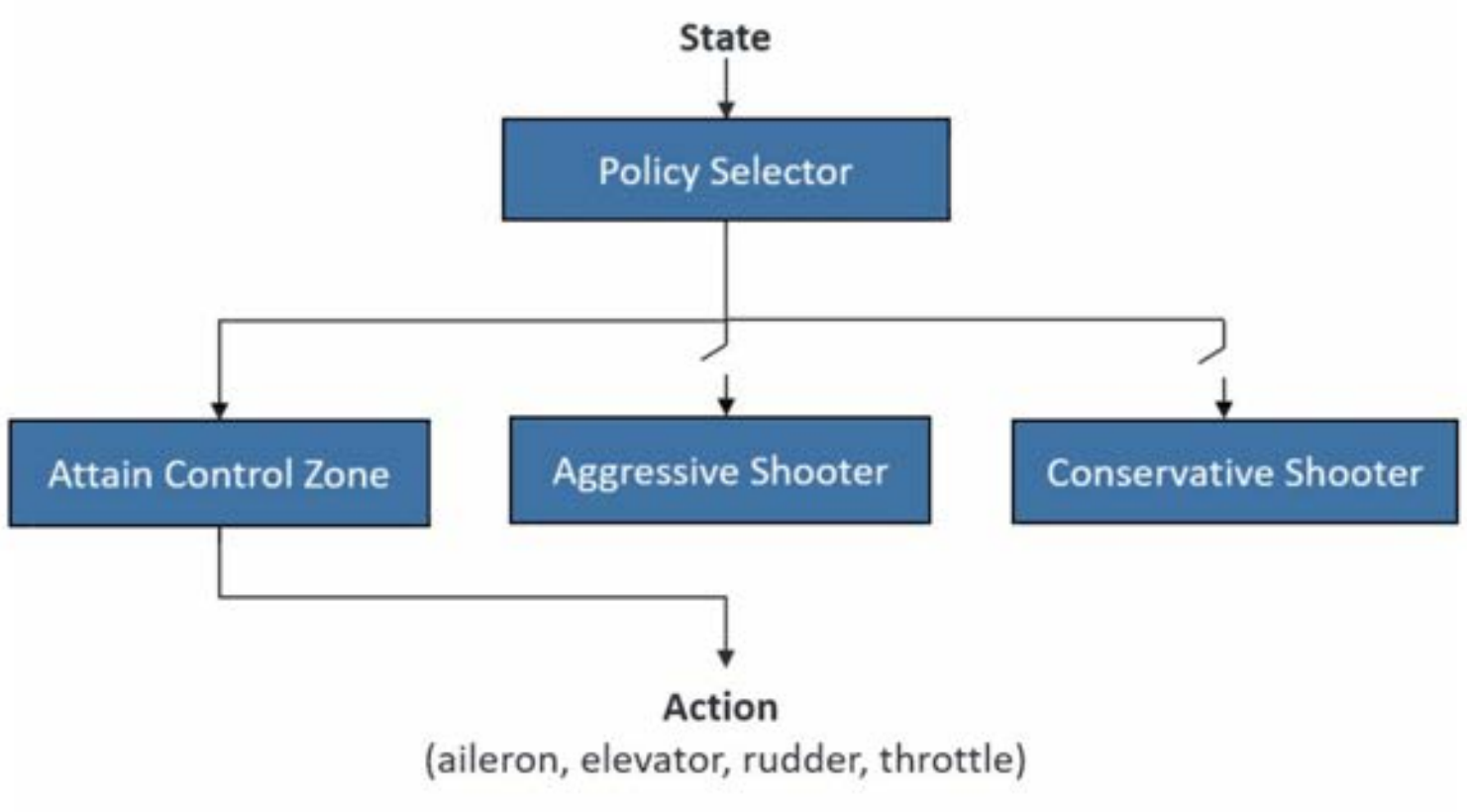

Figure 6.10. High-level Architecture of PHANG-MAN Agent. Source: [157].

The low-level policies were trained with the Soft Actor Critic (SAC) algorithm. SAC is an off-policy RL algorithm that, in addition to maximizing expected rewards, also maximizes entropy—resulting in both maximizing performance and exploration [157]. Each of the low-level policies in Figure 6.10 ingests the same state values and is composed of the exact same three-layer multi-layer perceptron (MLP) architecture. The difference between each low-level policy is a range of initial conditions used in training their individual and distinct reward functions. These individual reward functions are designed to encourage specific behaviors. For example, the control zone policy would try to attain a pursuit position aft of the opponent within their "control zone;" the active shooter policy would try to take shots, even if they were low-probability aggressive shots from the side and from head-on; and the conservative shooter policy would attempt to maintain an offensive position at the cost of low scores [157].

The high-level portion of the hierarchy is composed of a policy-selector network that chooses the best low-level policy based on the current state. Their approach is similar to Comanici and Precup [159], however, instead of estimating the terminal conditions directly, a new selection is made periodically at a rate of 10 Hz (versus the low-level policy update rate of 50 Hz) [157]. Unlike the standard option-critic architecture, however, the low-level policies in this architecture are trained separately and then their parameters are frozen. This allowed the high-level policy selector to be trained without the added complexity associated with non-stationary policies. This approach helped simplify the learning process and allowed for the re-use of agents in a modular manner [157].

Figure 6.11 depicts an example of how different agents (or policies) were dynamically selected based on the current state. The y-axis shows the track angle that represents what is happening in the episode. A smaller track angle usually indicates a more offensive position. As shown in the top pane in the match against the conservative shooter policy, the policy selector agent exploited its initial positional advantage (i.e., smaller track angle) by closing the distance while reducing the track angle, as can be observed by the decrease in track angle. This eventually resulted in an overshoot, as shown in step 500 in the bottom pane where the track angle begins to increase rapidly. However, when the high-level policy switched back to a conservative shooter policy, it resulted in the aircraft arriving back into an offensive position and attaining persistent gun snaps (shown in steps 1100 to 1300 of the bottom pane where the track angle remains steady at near 0) and ultimately ending with a win [157].

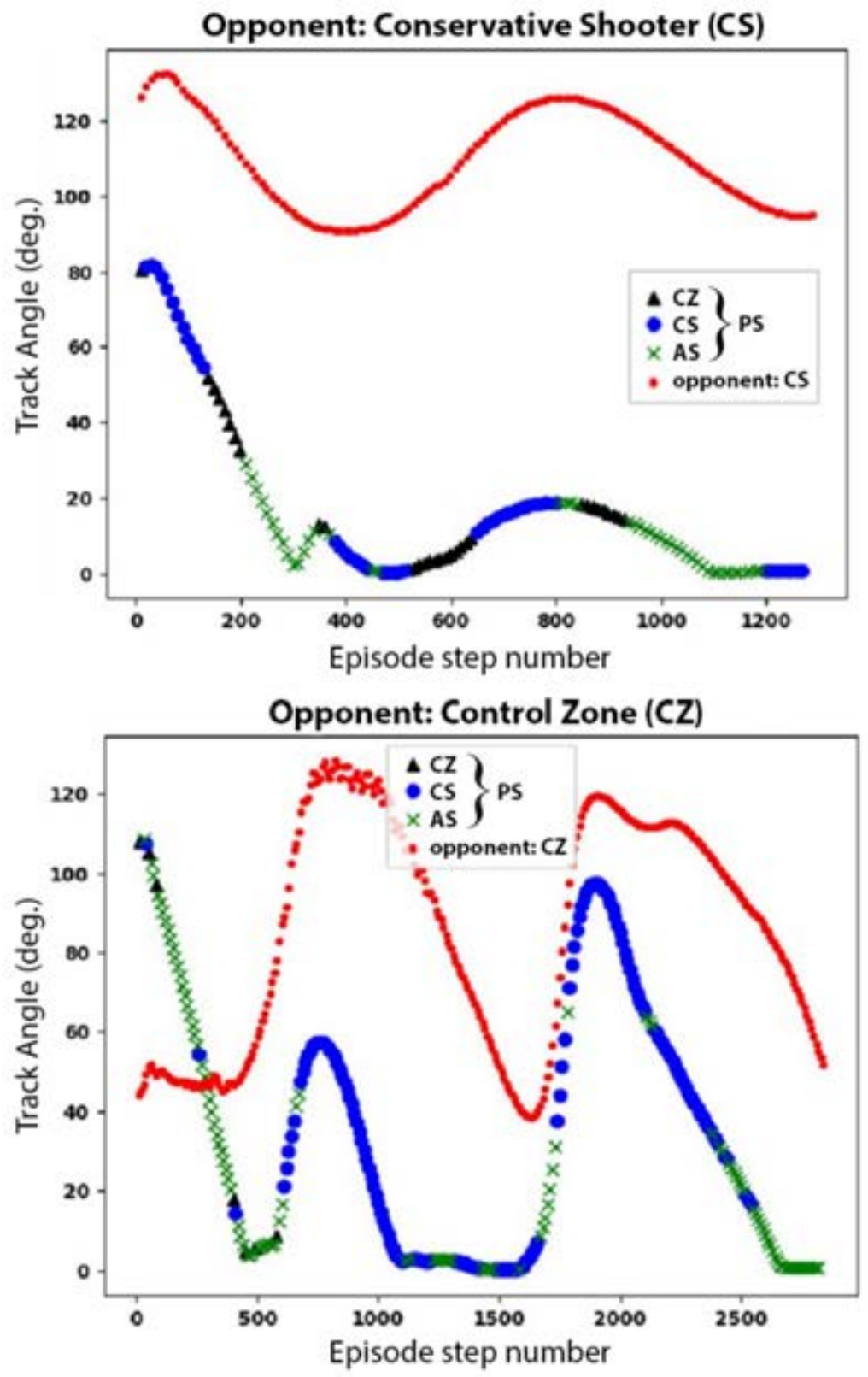

Figure 6.11. PHANG-MAN Agent Selection During Single Episode. Illustration of PHANG-MAN agent selection during a single episode. The policy selector selects from a team of three agents: control zone (CZ), conservative shooter (CS), and aggressive shooter (AS). Opponents are a CS (top) and a CZ (bottom). Source: [157].

In their experiments, Pope et al. [157] observed that the policy selector performed at least as well as the highest-performing low-level policy against a given opponent. Moreover, when used across several opponents, the policy selector performed significantly better than that of any individual low-level policy [157]. This suggests that the policy selector was able to leverage the strengths of its low-level policies while generating unique complementary strategies [157], an approach that could prove valuable for training agents in even more complex environments.

Similarly, Wang et al. [160] proposed an HRL method for multi-unmanned combat aerial vehicle (UCAV) air combat in simulation. Their hierarchy also consisted of two levels. The top level was composed of general basic fighter maneuvers (BFM) strategies of attack, retreat, or defend. The bottom level was composed of micro-actions represented by the BFM control parameters. In their experiments, Wang et al. [160] showed that the hierarchical approach outperforms their current state-of-the-art of using a Q-Mix RL algorithm. Although this study did not provide scores and only compared reward outcomes, the rewards attained by the HRL approach were approximately 50% greater (950 for HRL vs 650 for RL) than that of the Q-Mix RL approach. However, also of note from their experiment, Wang et al. [160] found that the loss value actually increased with more episodes and pointed out that the core reason for this is that the scale of the reward was not properly set.

Zhang et al. [161] presented an HRL model for mastering Multiplayer Online Battle Arena (MOBA) games that addressed the need for real-time strategy (RTS) games to require both macro strategies as well as low-level manipulation to achieve satisfactory performance. In their hierarchical framework, agents learned macro strategies through imitation learning and learned micro-manipulations through RL. Figure 6.12 shows the hierarchical architecture used in this study. It depicts the four macro-strategies: attack, move, purchase, and learning skills. The appropriate macro-strategy was selected using imitation learning while the low-level micro-management actions were chosen using an RL algorithm. As previous research shows, this architecture eliminates the need for an agent to deal with a massive state and action space directly and the associated complexity of exploration in a sparse reward environment [161].

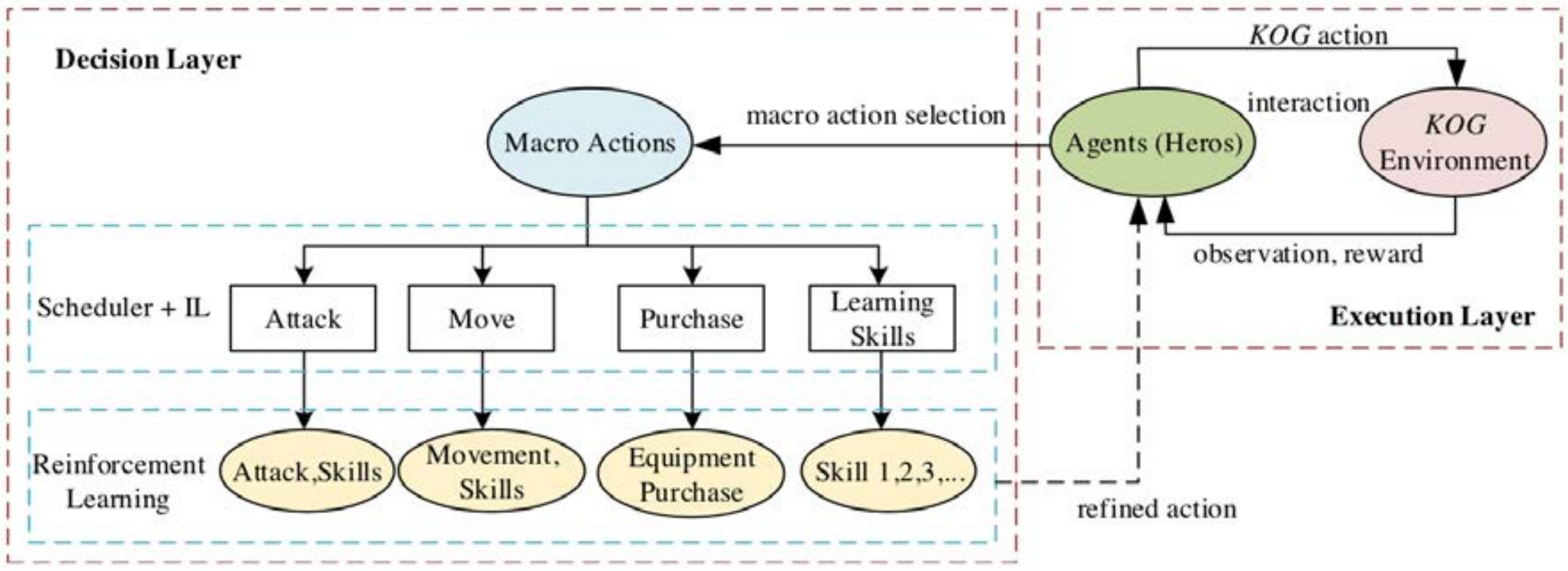

Figure 6.12. MOBA Agent Hierarchical Architecture. The hierarchical architecture consists of macro-strategies and micro-manipulations. Source: [161].

In their experiments, Zhang et al. [161] showed promise for their hierarchical approach using their proposed network architecture (shown in Figure 6.13) in that their agents outperformed a PPO algorithm (that does not use macro-strategies) in a 5 vs. 5 mode of the game King of Glory (KOG). Against an easy KOG entry-level internal AI, the hierarchical approach achieved a 100% win rate vs. the 22% win rate achieved by the PPO algorithm. This research continues to show the potential of HRL to scale to more complex scenarios over conventional RL approaches.

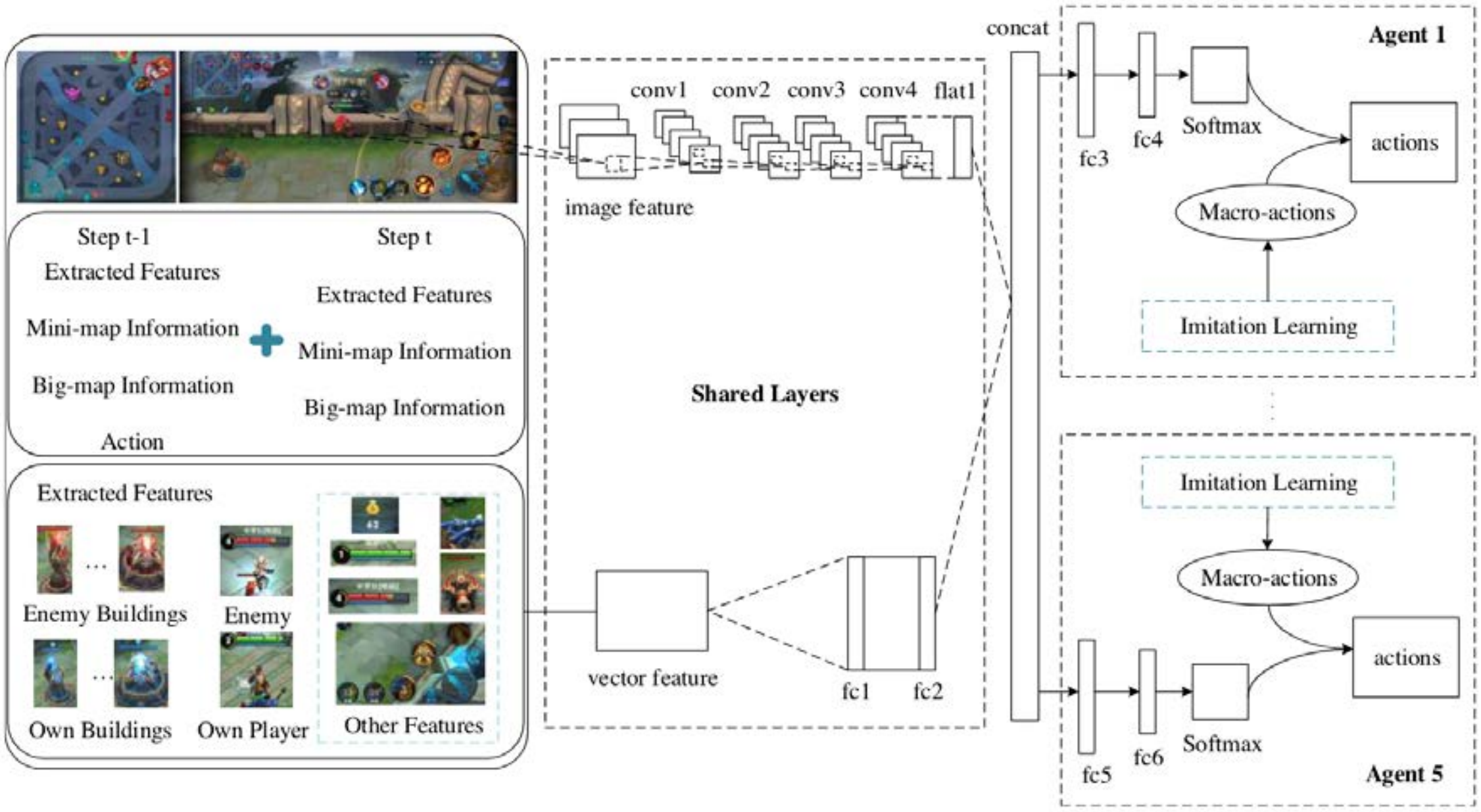

Figure 6.13. MOBA Network Architecture of Hierarchical Reinforcement Learning Model. Source: [161].

Li et al. [162] suggested that, although HRL is able to explore the environment by decomposing tasks and re-using low-level policies, existing HRL methods rely heavily on external rewards for high-level decision making. Unlike traditional RL, where agents directly learn from dense rewards, HRL faces the challenge of sparse rewards at the higher levels of the hierarchy. This reliance on sparse external rewards makes it difficult for high-level policies to optimize effectively, especially in long-horizon tasks. In contrast, low-level policies can still be optimized with intrinsic rewards. The dynamic nature of the sub-goal space further complicates high-level exploration, as the high-level policy must adapt to changes in the sub-goal representation while learning [162].

To address these limitations, Li et al. [162] improved upon conventional goal-conditioned hierarchical reinforcement learning (GCHRL) approaches by proposing a regularization that results in more stable and efficient subgoal representation learning. Furthermore, by developing Hierarchical Exploration approach with Stable Subgoal representation learning (HESS), Li et al. [162] designed measures of both novelty and potential for the sub-goals, as well as developed an active hierarchical exploration strategy that looked for new sub-goals and states without the need for intrinsic rewards.

This method starts with a regularization parameter designed to limit changes in representation during each update by linking the current representation to the previous subgoal representation. In this approach, the expected value was calculated over a replay buffer, with a regularization weight that varied for different states. Li et al. [162] introduced two measures for evaluating sub-goals. The first measure assessed novelty using visit counts, encouraging the agent to explore distant, novel states by considering both immediate and future state visitations. The second measure, aimed at guiding the agent towards less explored areas, involved modifying the novelty measure by adding perturbations to a selected subgoal from the replay buffer to create a new, imagined subgoal, which is then used to direct the low-level policy [162].

Additionally, Li et al. [162] proposed an active exploration strategy that does not use intrinsic rewards, as they argue that using intrinsic rewards could introduce additional non-stationarity problems associated with HRL. Instead, to appropriately balance exploration and exploitation, HESS used a probabilistic approach in its exploration strategy with the probability decaying over the course of learning [162]. They then evaluated HESS using a suite of MuJoCo tasks [163] and compared them against a hierarchical Intrinsic Curiosity Module (ICM) [41], hierarchical Successor Representation (SR) [164], Dynamics-Aware Discovery of Skills (DADS) [165], LEarns the Subgoal representation with SlOw dyNamics (LESSON) [166], and SAC [167]. Overall, HESS outperformed these other algorithms and showed that this approach could better balance traditional exploration and exploitation strategies.

Wang et al. [168] introduced using a hierarchical multi-agent RL framework to train agents in a discrete, hexagon-based wargame. Their higher-level network addressed task decisions by solving the credit assignment problem through cooperative training between agents. The lower-level network, on the other hand, was primarily used for route planning and task-associated behaviors. Essentially, the higher-level network selected an available task for the agent to perform, while the lower-level network selected the action the agent needed to perform to accomplish the task. Wang et al. [168] argued that this results in the higher-level network focused more on the cooperation aspect of multiple agents while the lower-level network remained more independent, which allowed the lower-level network to be replaced with methods that leveraged prior knowledge, such as heuristics.

Wang et al. [168] used a Q-Mix architecture to conduct the multi-agent training for the higher-level network. This architecture allowed agents to train their individual Q-networks in a way that improved coordination. For the lower-level network training, they used a standard DQN algorithm of a reward function specifically designed for route planning. When training the higher-level and lower-level networks, Wang et al. [168] did so alternately by freezing the parameters of the network not being trained. Of note, to train the higher-level network initially, they used a heuristics-based lower-level network until this lower-level policy was considered well-trained.

To build robustness in their agents, Wang et al. [168] employed grouped self-play. They initially conducted "warm-up" training by training their agents against a rule-based AI. After creating several AI models from this "warm-up" training, the models were then trained by playing against each other in groups. This was followed by a round-robin competition where the average win rate was calculated. Models with higher win rates were then duplicated to form a new generation, and a new iteration was performed. Finally, the top policies were combined via policy distillation [169].

In the end, Wang et al. [168] demonstrated that their HRL model surpassed both rule-based and DQN models in performance. Similar to earlier discussed studies, they attributed this improvement over the DQN model to the drawbacks of using a flat action space. Specifically, they noted that blending various candidate actions, each with distinct meanings, in a flat action space leads to decreased sample efficiency. This observation underscores the advantage of hierarchical approaches in managing complex action spaces more effectively.

Levy et al. [170] introduced a new HRL framework they term Hierarchical Actor-Critic (HAC) that overcomes the instability issues inherent in jointly training several policy levels, which in turn accelerates the training process. Learning several different levels of policies in parallel is difficult due to non-stationary state transition functions; thus, modifications in a policy at one level may result in adjustments to the transition and reward functions at subsequent levels. Under these conditions, RL struggles to learn since they typically require that the distribution of states that map to actions be stable [170].

This HAC framework consists of a hierarchical architecture and a technique for simultaneously learning in scenarios with sparse rewards. This architecture consists of a set of goal-conditioned policies that are nested. These break down tasks into subtasks by using the state space [170]. The highest-level policy uses the current state and goal state as inputs and outputs a subgoal state, which in turn serves as the goal state for the policy in the next hierarchical layer. This policy utilizes the current state and goal state provided by the higher-level policy to generate its own subgoal state for the subsequent layer of the hierarchy to pursue. This sequence persists until it reaches the base level, which receives similar types of input but produces a primitive action as its output. At each level in this hierarchy, there is a defined limit of attempts or time steps to attain its goal state. If the goal state is not achieved in this time—or when the goal state is achieved—the higher-level outputs a new subgoal [170].

Additionally, HAC leverages two types of hindsight transitions (*hindsight action* and *hindsight goal* transitions) in order to train multiple policies in parallel using only sparse rewards. *Hindsight action* transitions simulate an optimal lower-level policy hierarchy to train high-level policies. *Hindsight action* transitions use the hindsight-achieved subgoal state rather than the original, similar to Hindsight Experience Replay (HER) [171]. Put simply, if an agent is given subgoal state *A* but is unsuccessful at reaching this state, and reaches subgoal state *B* instead, the subgoal level receives a transition as if state *B* were the desired subgoal state [170]. This allows the higher-level policy to focus on learning a sequence of subgoals required to achieve a goal state while the lower-level policies concentrate on the sequence of actions necessary to achieve each subgoal. This results in always using an optimal lower-level policy and allows training multiple levels of the hierarchy simultaneously.

In their experiments, Levy et al. [170] evaluated their framework in both discrete and continuous state-action environments. Gridworld environments were used for discrete tasks while simulated MuJoCo [163] robotics environments were used for continuous tasks. Levy et al. [170] evaluated agent performance using policy hierarchies with one (a flat agent), two, and three levels. The flat agent used Q-learning [74] with HER for the discrete environment, and Deep Deterministic Policy Gradient (DDPG) with HER in the continuous environment. Overall, they found that the hierarchical approach outperformed the flat agent in each and every task; moreover, the three-level hierarchical agent outperformed the two-level agent. Levy et al. [170] also found that their HAC approach also outperformed another leading HRL technique called HIerarchical Reinforcement learning with Off-policy correction (HIRO) [172]. This research is promising as it shows that more levels in the hierarchy—as will likely be needed for intelligent combat agents—can potentially produce better results.

In his thesis, Rood [11] also investigated the scalability of HRL for use in combat simulations, directly extending the research conducted by Boron [80] and Cannon and Goericke [81] using the Atlatl simulation environment. Rood [11] specifically focused his research on using feudal multi-agent hierarchy (FMH) as introduced by Vezhnevets et al. [155] discussed above.

Rood [11] used the entire gameboard as the Manager's observation space, and the hexagons on the gameboard as the Manager's action space. These hexagons are then translated into goals by the Workers. Of note, due to the limitations of Q-learning algorithms to predict a single value, Rood [11] instead used the PPO algorithm and implemented the action space as an array of row and column coordinates for each Worker's goal.

On the other hand, the Worker's observation space used was a local observation of the $5 \times 5$ surrounding hexagons around the agent (or a radius of two hexagons from the agent). The Worker's action space consisted of the hexagons reachable by the unit in one turn as well as taking no action, as described in Section 3.4.3. The algorithm used for the Worker was DQN since their action space was made up of discrete values of either zero to six for infantry (capable of moving only one hexagon in one turn), or zero to 18 for all other units (capable of moving two hexagons in one turn).

In his experiments, Rood [11] compared the FMH approach against two different agents (rules-based and RL-based) in increasingly complex environments. These scenarios included a 2 vs. 1, 3 vs. 2, 6 vs. 4, 12 vs. 6, and a three-city maze. The results indicated that using a local observation space for the Workers resulted in poor performance compared to using a global observation space (converging on 20 reward points as opposed to 140 reward points using the global observation space). Using a global observation space, however, means that agents need to be trained for each scenario that uses a different map size independently, whereas agents trained on the same local observation space dimensions could be re-used for any scenario [11].

In the analysis of FMH performance compared to the rule-based and RL-based agents in the five scenarios, Rood [11] found that the FMH failed to demonstrate any learning and underperformed the two other agents. Rood [11] attributed this result to the Worker's inability to provide predictable actions to the Manager's goals, which prevented the FMH agent from developing the optimal policy. Although these results were not positive, they did offer valuable insights for future HRL research in this domain [11].

To accelerate training, Florensa et al. [173] leveraged the combined strengths of hierarchical methods and intrinsic motivation. Their approach involved first learning a broad range of skills in a pre-training environment using only proxy rewards and then applying these skills to enhance learning efficiency in downstream tasks. This two-part framework consisted of an unsupervised procedure for skill acquisition and a hierarchical structure to encapsulate and reuse these skills for future tasks. The learning process utilized intrinsic motivation through a single proxy reward, minimizing the need for domain-specific knowledge about the downstream tasks [173]. This setup encouraged the agent to independently explore its capabilities, fostering a diverse skill set that could later be utilized across various tasks by training a separate high-level policy to orchestrate these skills.

To generate and learn a diverse set of skills, Florensa et al. [173] employed stochastic neural networks (SNNs) [174], which are trained with minimal supervision. Recognizing that SNNs alone might not ensure broad enough skill acquisition, they incorporated an information-theoretic regularizer during the pre-training phase to promote a wider range of behavioral patterns within the SNN policy. Their experiments demonstrated that this method of layering high-level policies over learned skills significantly enhanced performance in complex, long-horizon tasks with sparse rewards, effectively boosting the agent's capability to explore new environments [173]. This hierarchical approach not only improved learning efficiency but also provided a robust framework for applying learned skills to new and challenging scenarios.

Frans et al. [85] developed a meta-learning approach termed metalearning shared hierarchies (MLSH) for hierarchically structured policies that leveraged shared primitives. Their proposed algorithm iteratively learned a set of sub-policies that maximized a reward over a distribution of tasks. Although they sought to ensure the sub-policies were fined-tuned enough to achieve high performance, they also attempted to ensure that these sub-policies were robust enough to work over a wide range of tasks [85].

The architecture for this approach is shown in Figure 6.14. Frans et al. [85] showed that by discovering a set of strong sub-policies $\phi$, new tasks could be more easily learned simply by updating the master policy $\theta$. Moreover, because the master policy was operating on a different time scale than its sub-policies, its time horizon was much shorter. Hence it could more quickly adapt to new tasks [85].

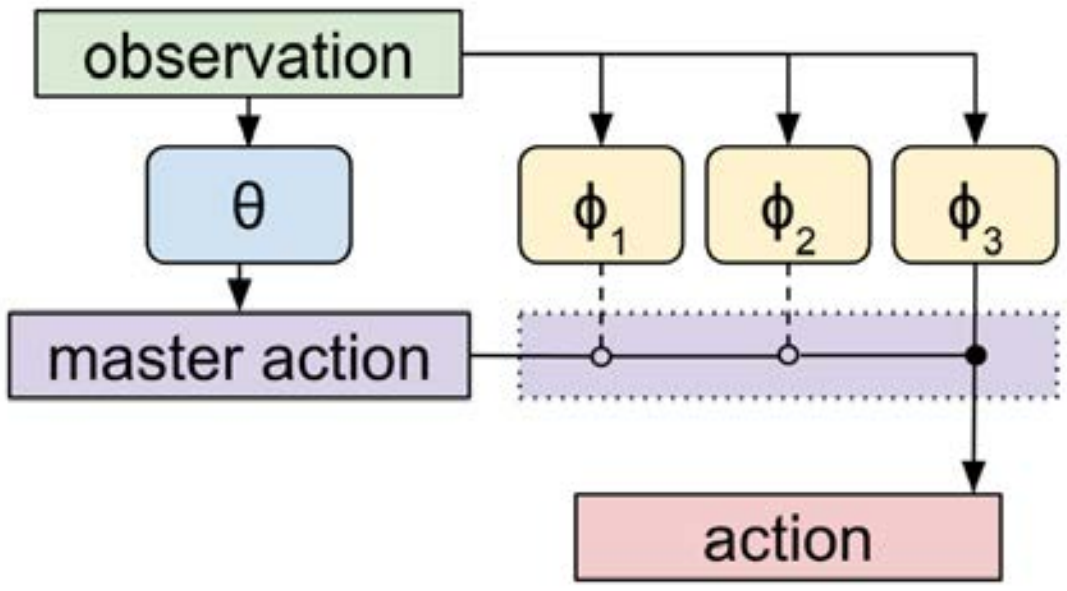

Figure 6.14. Structure of Hierarchical Sub-Policy Agent for Metalearning. The master policy is represented by $\theta$ and the sub-policies are represented by $\phi_i$. The current active sub-policy is $\phi_3$ which is producing actions as output. Source: [85].

In their experiments, Frans et al. [85] demonstrated the effectiveness of their approach in learning meaningful sub-policies across a range of tasks. They successfully increased the complexity of these tasks, showing that the MLSH framework could handle complex physics domains. In one study, they found that MLSH, by learning sub-policies over a distribution of tasks, significantly outperformed both shared and single policy agents. In their experiment, Frans et al. [85] tackled a complex task previously unsolvable with a naive PPO algorithm. Due to the sparse rewards and vast exploration space, the PPO algorithm could not effectively learn the task using only primitive action sequences. However, the MLSH approach, which navigated through the space of sub-policies, allowed for the discovery of a reward sequence, demonstrating its advantage in complex scenarios [85].

Overall, the literature suggests that the benefits of an HRL approach over the more standard flat RL approaches are numerous. Other examples showing this include: Vezhnevets et al. [175] developed a deep recurrent neural network (RNN) architecture and created a model, STRategic Attentive Writer (STRAW), that was able to learn high-level, temporally abstracted macro-actions; Morimoto and Doya [176] used an upper-level where the agent coarsely explored a low-dimensional state space and a lower-level, where the agent finely explored a high-dimensional state space, effectively learning to set up sub-goals for tasks in the upper-level and achieve those goals in the lower-level; and Tobi [177] showed that an HRL, Max-Q, method outperformed a flat Q-learning method in an RTS game on a military unit transportation task.

In summary, HRL has shown the potential to more efficiently solve RL problems that may otherwise be intractable for standard RL algorithms to solve on their own. However, although this analysis shows the potential of HRL methods to be used to scale RL to work in more complex environments, no work has yet accomplished this in a generalized, self-similar manner as applied to combat modeling and simulation to train intelligent agents capable of performing in large scenarios.

## 6.3 Research Objective

We expand on the research by Rood [11] and incorporate our previous work presented in Chapter 3 (localized observation abstraction), Chapter 4 (multi-model framework), and Chapter 5 (hierarchical hybrid AI approach) into a new HRL architecture and training framework that we propose in this chapter. Also implemented in our overall HRL approach are concepts borrowed from the literature discussed above. For example, we employ some elements of reward hiding and information hiding as proposed by Dayan and Hinton [152]; we decouple the training of different layers of the hierarchy while freezing parameters as proposed by Vezhnevets et al. [155] and Pope et al. [157]; we use a heuristics-based approach to train higher-level networks to stabilize learning as suggested by Wang et al. [168]; and we use the hierarchical abstraction theoretical framework proposed by Abel [75] to guide the development of our approach.

Ultimately, this study aims to develop an HRL architecture and training framework that integrates different levels of observation abstractions and a multi-model framework consisting of various AI methodologies to explore the potential benefits and limitations of hierarchical approaches in complex decision-making environments. By integrating key insights from the literature and our previous studies, we seek to gain further insights into the dynamics and challenges of architecting and training HRL systems. Our goal is to evaluate how these techniques influence learning processes and decision-making efficacy as compared to conventional scripted and RL approaches. This exploration will help identify key areas for further improvement and innovation in HRL methodologies.

## 6.4 Methodology

As in our previous studies, we employ the following methodology using the Atlatl simulation environment described in Section 3.4. We design, develop, and implement the agent, decision, and policy-selection architectures as well as the HRL training framework. We then conduct an experiment and limited ablation study to evaluate the performance of this framework as compared to scripted and RL agents alone. We then provide our results and conclusion.

### 6.4.1 Hierarchical Reinforcement Learning Agent Architecture

Instead of simply implementing a single hierarchy of agents, this research explores multiple levels of embedded hierarchies. This includes a hierarchy of agents (i.e., how agents are decomposed and organized), a hierarchy of decisions (i.e., the corresponding decision level of each agent), and a hierarchy of policies (i.e., how each decision is chosen based on several decision models).

Figure 6.15 depicts the three architectures. The *Agent Hierarchy* depicts three levels of a hierarchy. For illustration and discussion purposes, we have labeled them *Commander*, *Manager*, and *Operator*; however, this hierarchy can extend from one to $n$ levels, with level 1 at the bottom, or base, of the hierarchy, and level $n$ at the top, or root, of the hierarchy.

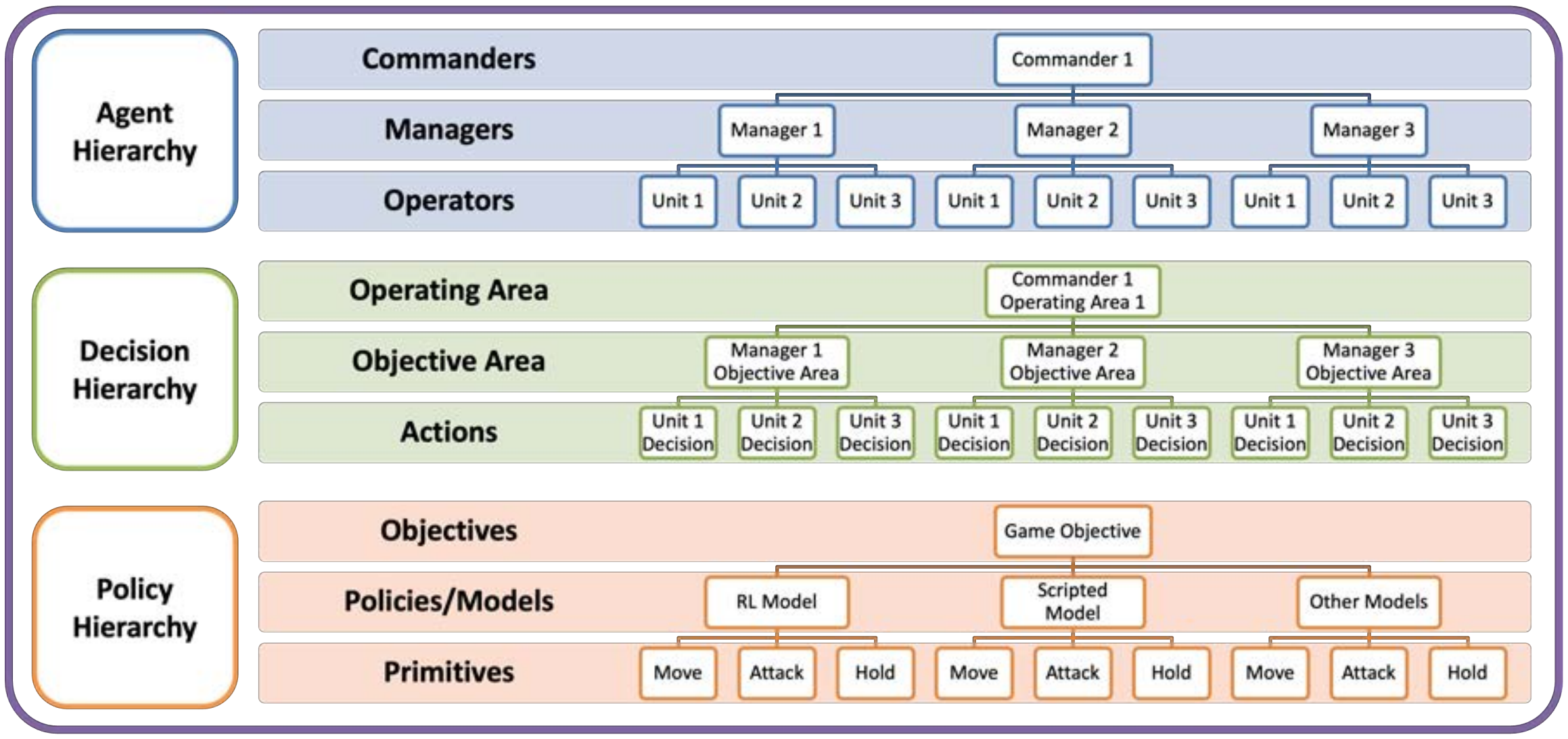

Figure 6.15. Hierarchical Reinforcement Learning Architecture of Units, Decisions, and Policies. The first level is the Agent Hierarchy which controls the different levels of agents. The second level is the Decision Hierarchy which results in decisions for each agent to take. The last layer is the Policy Hierarchy which turns decisions into actions at the entity level.

**Agent Hierarchy**

The *Agent Hierarchy* is depicted at the top of Figure 6.15. In this study, we implement three levels for the *Agent Hierarchy*. Smaller games with few units may require only two levels, whereas larger games involving more units could benefit from three or more levels.

*Operators* exist at the base or leaf level. These are the actual entities on the gameboard. The *Unit* is the actual entity represented on the gameboard, depicted as its respective operational terms and graphics. We then wrap an *Operator* around this *Unit*, which simply contains additional information such as details regarding their parent agents, goals, etc. Up one level is the *Manager* level. These agents are not actual entities on the gameboard, but rather abstract or cognitive agents. Each *Manager* is assigned up to three *Operators*, takes in an observation, and passes its decisions down to its *Operators*. Likewise, the *Commander* represented at the top level or root of the hierarchy is also an abstract entity not physically represented in the simulation that also takes in its own observation, makes a decision, and then passes this decision down to its *Managers*.

An example of this organization of units is shown in Figure 6.16. This figure depicts a $20 \times 20$ gameboard with red forces in the north and blue forces in the south. Units under Commander 1 are shaded in green; units under Commander 2 are shaded in yellow; and units under Commander 3 are shaded in blue. We further depict the breakdown of Commander 1's units into three Managers, each owning three units (or Operators). The scenario generator used in Atlatl places units on the board based on a Gaussian distribution to ensure units of the same hierarchy level are placed near each other. Thus, our logic simply takes these labeled units and assigns them accordingly to their respective Commanders and Managers.

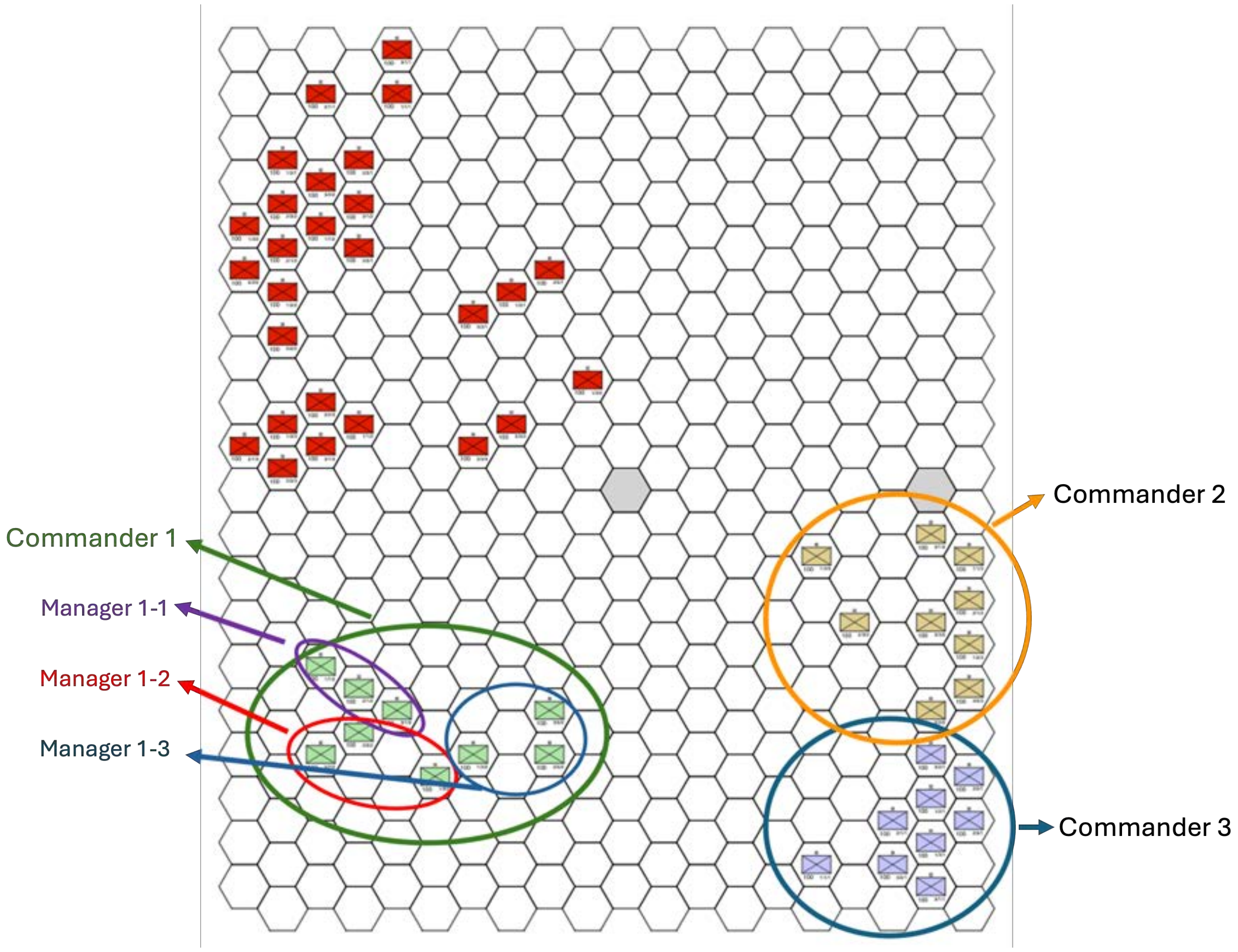

Figure 6.16. Example of Hierarchical Agent Decomposition. Units are grouped first by Commanders and then by Managers. The example gameboard shows three Commanders as circled. Commander 1 is further decomposed into its three Managers, each of which controls three Operators. Although we do not show the decomposition of Commander 2 and Commander 3 in this figure, those Commanders are also likewise decomposed.

**Decision Hierarchy**

The middle of Figure 6.15 illustrates the *Decision Hierarchy*. This hierarchy depicts the type of decisions made by agents at each level of the hierarchy. This hierarchy can be thought of as being directly mapped to the *Agent Hierarchy*. The base of the hierarchy, labeled *Actions*, is the Operator level, where units make decisions that actually change the state of the game.

The next level up, *Objective Area*, represents the decisions being made by the Manager agents. These decisions are a form of *options* [87] in that they are macro-actions that persist for longer timespans than the lower-level decisions of the Operator. The *Objective Areas* produced are $5 \times 5$ square areas within which its subordinate Operator units are instructed to operate. The top-most level is the Commander's decision space consisting of *Operating Areas*. These *Operating Areas* are $10 \times 10$ square areas within which the Commander's Managers (i.e., the Operator or Units belonging to those Managers) are to operate. If we were to extend this hierarchy, the next level would involve a Commander taking in a $40 \times 40$ gameboard and passing down a $20 \times 20$ area to the next level for its subordinate agents to operate within. This combination of *Agent Hierarchy* and *Decision Hierarchy* can be seen to be self-similar in that it can continually scale up in this manner, where the same logic is used to further break down the gameboard to pass down to its subordinate agents.

Of note, in Chapter 5, our Manager agents created super "hexagons" as their Objective Areas to pass down to their subordinate agents. However, constructing a super "hexagon" involved identifying all adjacent hexagons up to a specific layer, resulting in exponentially increasing computations with each added layer. Thus, a Commander would have to compute four layers of hexagons and a Super-Commander would have to compute eight layers of hexagons. To enable greater efficiency and scalability, therefore, in this experiment, we simplify the calculations and create square areas that are more computationally efficient to derive.

An example is shown in Figure 6.17. Here the green box represents Commander 1's Operating Area. This Operating Area is passed down to each of the Commander 1's Managers. Within the Operating Area, each Manager then selects an Objective Area, depicted by the purple, red, and blue outlines. These Objective Areas are then passed down to each of its Operators. The Operators then proceed to and operate within this Objective Area taking game actions such as moving to a specific hexagon or attacking an enemy.

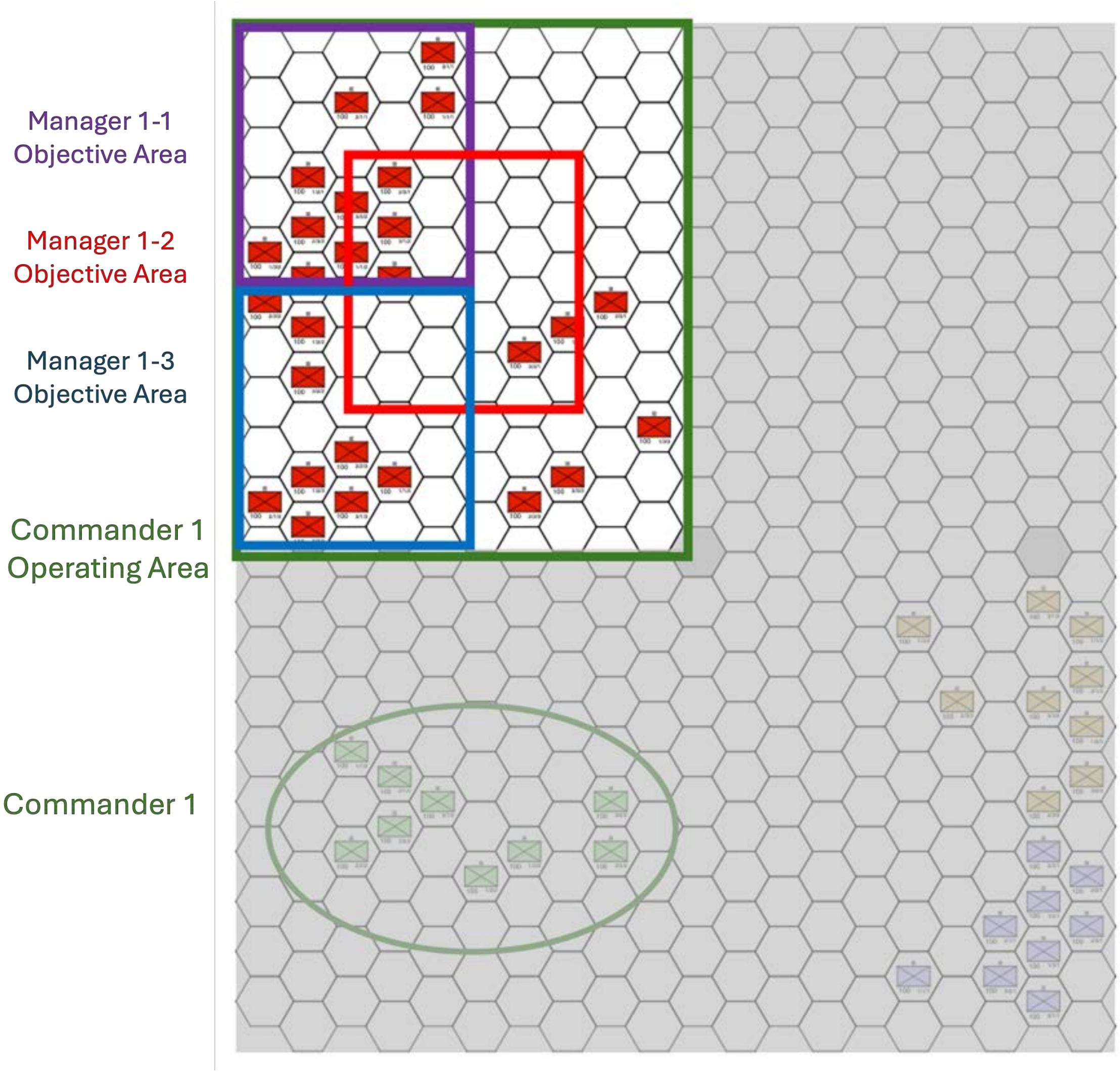

Figure 6.17. Operating and Objective Area Breakdown. The Commander assigns an Operating Area to its Managers, depicted in green in this example. Next, each Manager assigns an Objective Area to each of its Operators, depicted in purple, red, and blue.

Commanders construct their respective Operating Areas as depicted in Figure 6.18. First, the Commander's observation space is overlaid with a $3 \times 3$ grid as shown in the second pane. In our research, since the Commander is at the highest level of the hierarchy, the entire gameboard becomes its observation space. The Commander then selects, per its policy, one of the nine entries in the $3 \times 3$ grid (as highlighted in green in the third pane) and builds a $10 \times 10$ Operating Area around the center of the selected entry. This $10 \times 10$ Operating Area is then shifted so that it falls entirely within the gameboard. Of note, each Commander independently selects its Operating Area which may coincide or overlap with the other Commanders' Operating Areas

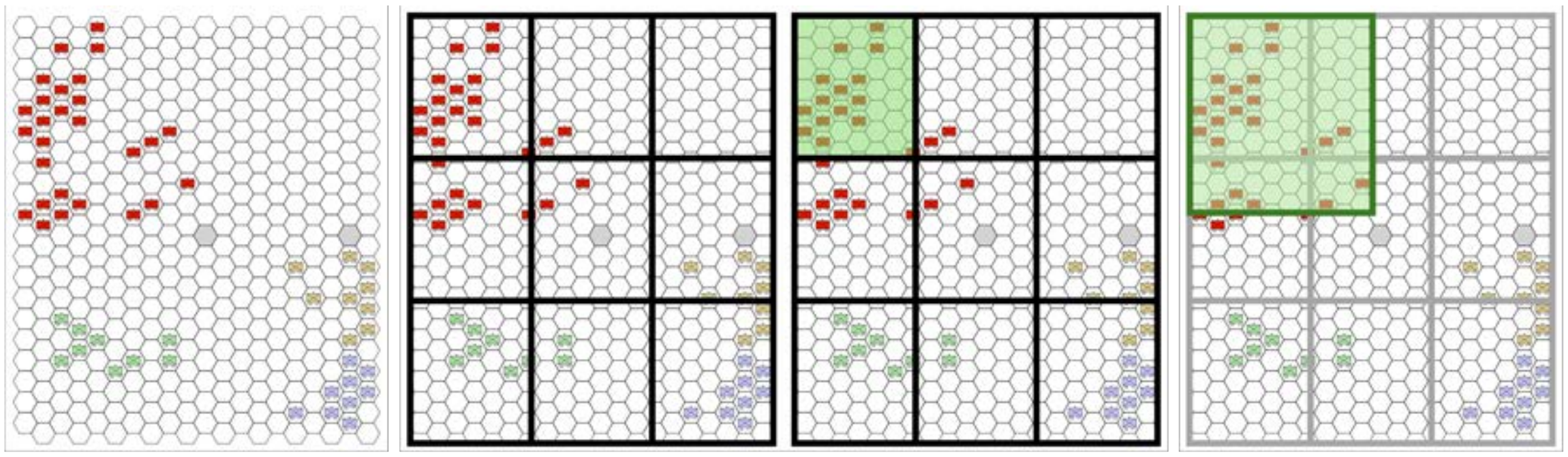

Figure 6.18. Assignment of Commander Operating Area. An example $20 \times 20$ gameboard is shown in the first pane. The second pane depicts how the Commander first divides their observation space into a $3 \times 3$ grid. The third pane depicts how the Commander selects one of the nine entries, shaded in green, per its policy. The last pane depicts how the Commander constructs a $10 \times 10$ Operating Area and shifts this area to fall within the bounds of its observation space. This Operating Area is then passed to each of its Managers.

Figure 6.19 depicts a similar process for the Manager to take the Operating Area passed by its Commander and generate an Objective Area to pass to each of its Operators. As shown in the first pane, we use the same example $20 \times 20$ gameboard. The Commander then passes a $10 \times 10$ Objective Area to its Managers. The second pane depicts this Operating Area which now becomes the Manager's observation space. The third pane depicts how, like with the Commander, this Operating Area is divided into a $3 \times 3$ grid and an entry is selected (shaded in purple) per the Manager's policy. Finally, as shown in the last pane, this resulting $5 \times 5$ Objective Area is shifted to fall within the Manager's Operating Area. This Objective Area is then passed to each of the Manager's Operators. Each of these Operators then either moves to or operates within its assigned Objective Area based on its policy. Of note, like with the Commanders, each Manager independently selects its Objective Area that may coincide or overlap with the other Managers' Objective Areas

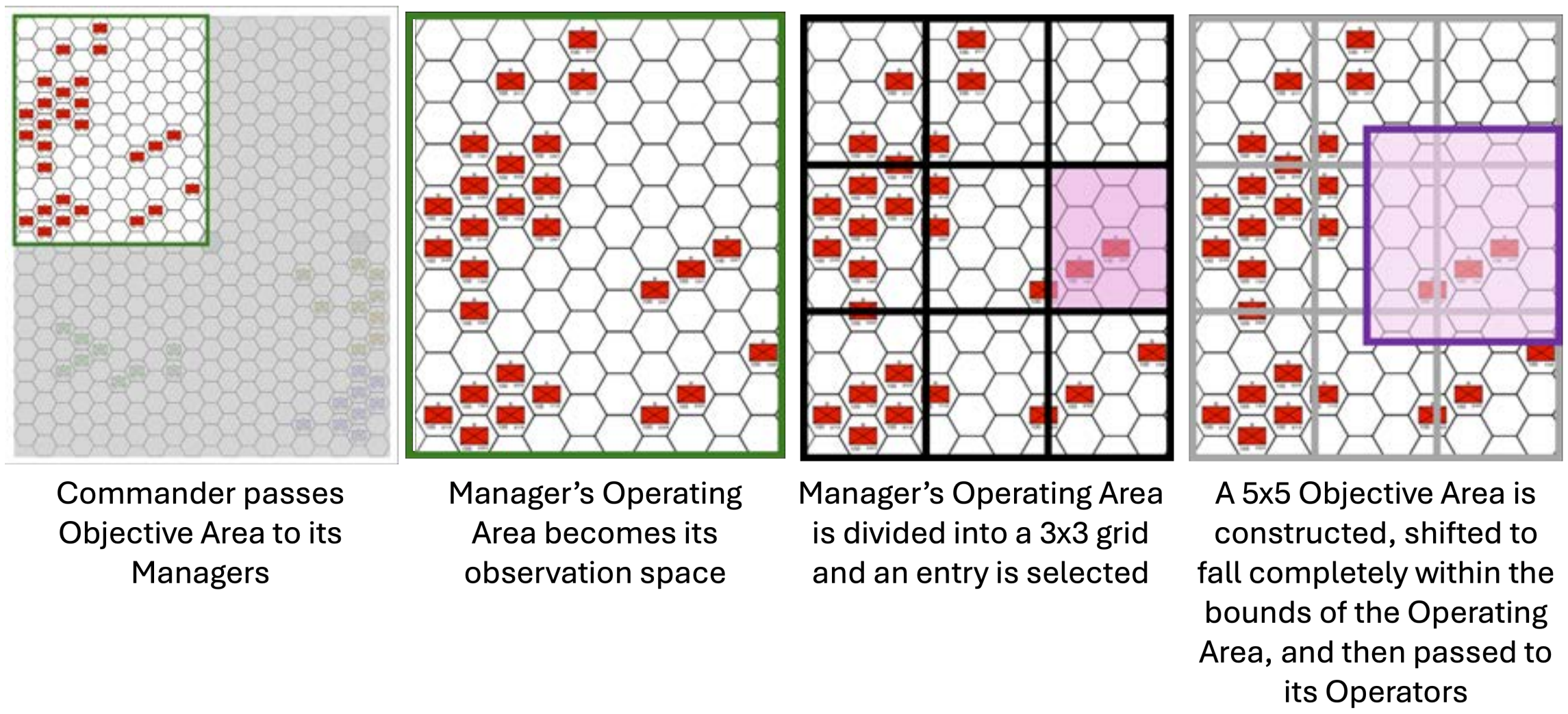

Figure 6.19. Assignment of Manager's Objective Area. In the first pane, an example $20 \times 20$ gameboard is shown which is reduced to a $10 \times 10$ Objective Area passed from the Commander to the Manager. The second pane depicts this Operating Area which becomes the Manager's observation space. The third pane depicts how the Operating Area is divided into a $3 \times 3$ grid and the right middle entry, shaded in purple, is selected by the Manager as its Objective Area. The last pane depicts the resulting $5 \times 5$ Objective Area being shifted to fall within the Manager's Operating Area. This Objective Area is then passed to each of the Manager's Operators.

**Policy Hierarchy**

The bottom part of Figure 6.15 depicts the *Policy Hierarchy*. The top level, labeled *Objective*, of this hierarchy represents the game's goal. For example, for our experiments in Atlatl, we use the default scoring function as described by Equation 3.4.4 which assigns positive points to the blue faction for holding key terrain (urban hexagons) and for inflicting damage on the adversary. The game objective for this research is to maximize this score. However, this can be changed to maximize a different metric, such as avoiding the adversary, or simply inflicting damage without concern for damage received. The second level, labeled *Policies/Models*, depicts different models that can be chosen. In this diagram, they are generally labeled, however, in implementation, these are specifically trained or scripted models and not simply a class of models. For example, this level could contain a collection of models such as RL-Attack, RL-Defense, RL-Hold, Scripted-Attack, Scripted-Maneuver, etc. Lastly, at the base of this hierarchy, labeled *Primitives*, is the specific output or action of the respective model given the observation or state of the game. In other words, the unit or entity on the board would receive an observation and use the selected *Policy* or *Model* to define an action.

The hierarchy of policies we use for this implementation is developed and implemented per our Multi-Model Framework as described in Chapter 4. This policy hierarchy exists for each level of the Decision Hierarchy. In other words, the Commanders will employ a Policy Hierarchy (i.e., our Multi-Model Framework) to select amongst models to use to determine which Operating Area to assign to its Managers. The Managers will employ their own Policy Hierarchy that will select an Objective Area for each of its Operators. Lastly, each Operator will employ its own level of Policy Hierarchy to select a specific action to take in the game. Of note, although we list "Other Models" in our example hierarchy in the bottom right corner of Figure 6.15, in this research we only use RL and scripted models to ensure the rapid execution times needed for training. However, it would be possible to implement game-solving agents such as MCTS-based agents, though these often require extensive computing time to execute properly.

### Hierarchical Reinforcement Learning Training Framework

Our HRL Training Framework consists of three levels illustrated in Figure 6.20. The base level, Level 1, is contained within the box shaded yellow. At this level, the Operator Multi-Model serves as the agent who takes in an abstracted observation and produces a primitive action, which directly affects the environment. This change in the environment then produces an engineered reward that is provided back to the agent, and the standard RL feedback loop resumes.

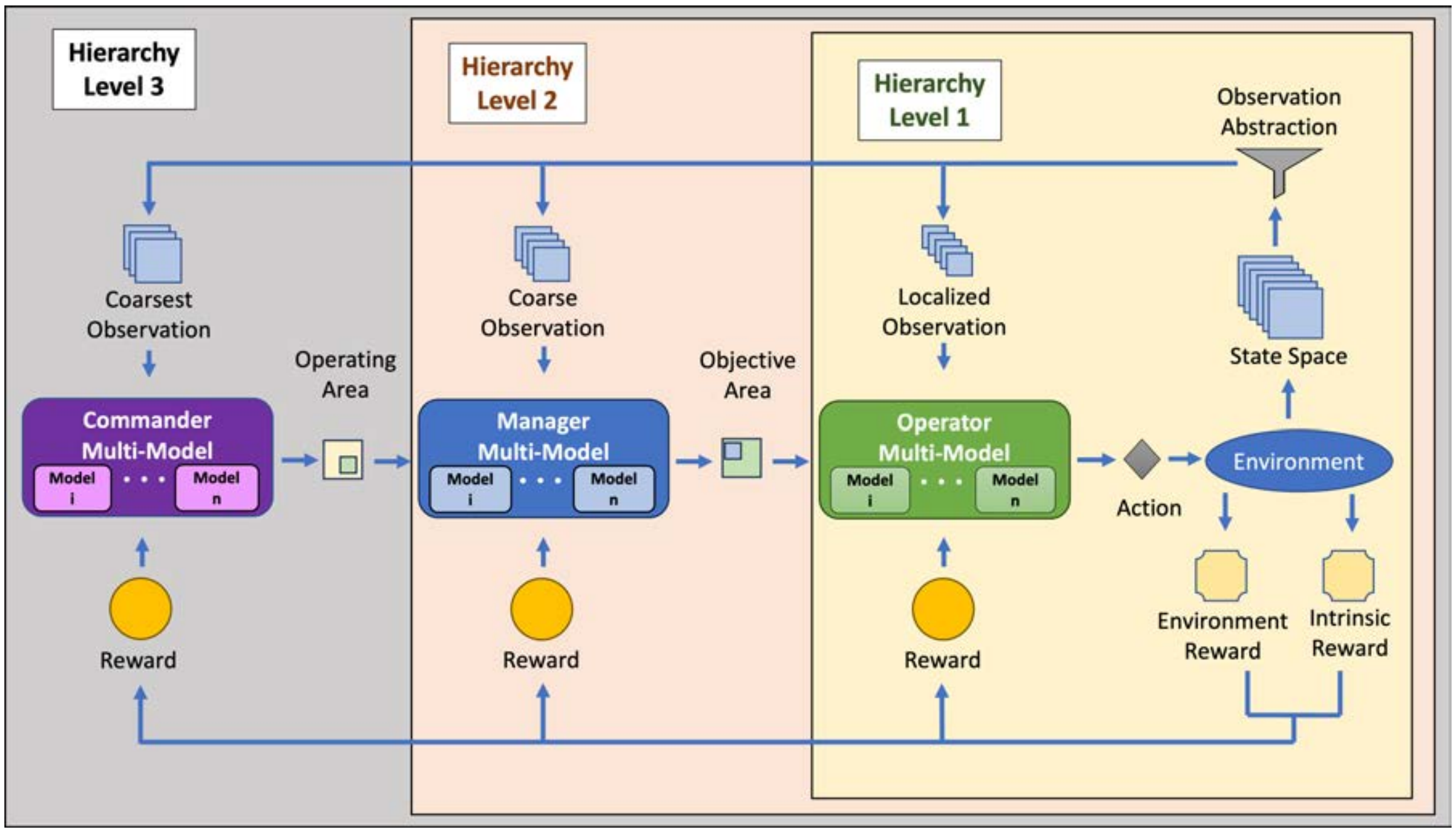

Figure 6.20. Hierarchical Reinforcement Learning Training Framework. The HRL framework consists of three levels. The Operator level, Level 1, is contained within the box shaded yellow. The Manager level, Level 2, is contained within the red box. The Commander level, Level 3, is contained within the grey box.

The middle level of the hierarchy, Level 2, consists of the Manager Multi-Model as the agent. This Manager takes in its own abstracted observation of the environment and produces an option. In this case, the option is the Objective Area which is passed to its Operators. The option persists until a termination condition is reached. At this point, an engineered reward is passed to the Manager and the cycle resumes.

The highest level, Level 3, consists of the Commander Multi-Model as the agent who takes in its own abstracted observation of the state space. The Commander then produces an option, which is an Operating Area that is passed to its Managers. This option also persists until a termination condition is reached, at which point an engineered reward is passed to the Commander, and the cycle is restarted.

As suggested by the HRL literature [11], [155], [157], [168], we begin by training only one hierarchical layer at a time so as to avoid the RL's known problems with non-stationarity, beginning with the base level, or the Operator level. We then freeze these parameters and train the Manager level. Once trained, we freeze the Manager's parameters and train the Commander Level. Of note, to maximize training efficiency, we train each level of the hierarchy based on a single agent rather than attempting to train using the Multi-Model agent. For the Manager level, we use the Pass-Agg behavior model as its Operator. For the Commander level, we develop and use a scripted Manager model for the Manager agent, and the Pass-Agg model for the Operator agent.

### 6.4.2 Operator Multi-Model

We use a total of 11 individual behavior models for our Operator Multi-Model. We use the following scripted models:

- Operator Pass-Agg: as described in Section 3.4.5.
- Operator Pass: as described in Section 4.4.1.
- Operator Agg: as described in Section 4.4.1.
- Operator Burt-Plus: as described in Section 4.4.1.
- Operator City: this model proceeds directly to the nearest urban hexagon (calculated via Euclidean distance) unless it has a fire target (i.e., an enemy unit in an adjacent hexagon). Once on the urban hexagon, the agent stays in place only engaging enemy units within its attack range. Of note, like in previous experiments, we only use infantry units in this experiment, which have an attack range of one hexagon (i.e., can only attack an enemy unit in an adjacent hexagon.
- Operator Killer: this model proceeds directly to the nearest enemy unit (also calculated via Euclidean distance). Once an enemy unit is within attack range, the model Operator will attack it.
- Operator Shootback: this model remains in place and only attacks units that come within its attack range.

We use the following RL-trained models:

- Operator RL-Boron: as described in Section 4.4.1.
- Operator RL-Scotty: as described in Section 4.4.1.
- Operator RL-Killer: as described in Section 4.4.1.
- Operator RL-City: as described in Section 4.4.1.

**Operator Observation Abstraction**

For this experiment, we re-train all of our RL agents from scratch using the localized observation abstraction with piecewise linear spatial decay described in Chapter 3. We use the same parameter values described in Section 3.5.1 for the spatial decay rate. In summary, the Operator agents use an observation abstraction consisting of a $7 \times 7$ localized abstraction of each channel from the global observation (as was shown in Figure 3.9) for a final observation space of $18 \times 7 \times 7$. Information within the inner $5 \times 5$ is kept to scale, while information in the outer perimeter of the $7 \times 7$ is a weighted sum by radial sectors of the information outside of the inner $5 \times 5$, described in Chapter 3 and shown in Figure 3.11 and in Figure 3.12. Of note, we cull the observation space of the Operator so that it only includes information within its assigned Objective Area. In other words, any information that lies outside of the agent's Objective Area is set to zero.

**Operator Gymnasium Environment**

We use the same Gymnasium Environment described in Section 3.5.2 consisting of an action space of seven discrete actions, one for each adjacent hexagon, plus the option to "pass."

The state space for our RL agent is the resultant abstracted $18 \times 7 \times 7$ observation space obtained by taking our original global state space $s$ and passing it through $\phi_{\text{operator}}$ to produce the abstracted observation space $s_{\phi_{\text{operator}}}$.

**Operator Reinforcement Learning Neural Network Architecture**

We use the same residual CNN described in Section 3.5.2 consisting of seven layers of 64 channels each (with a ReLU activation function between layers), which is then flattened into a 512-dimensional feature vector and passed through a final ReLU activation function.

**Operator Reinforcement Learning Algorithm**

We use the same DQN algorithm and hyperparameter values described in Section 3.5.2.

**Operator Training Scenarios**

To train our Operator models, we use Atlatl to generate random scenarios during training that are intended to represent the different configurations of the $5 \times 5$ Objective Areas that the Operator may encounter during gameplay. We modify the scenario generator to produce $5 \times 5$ square gameboards with a random number of blue units from one to nine and a random number of red units from zero to nine. We use zero to two urban hexagons and, rather than placing the urban hexagon(s) based on force ratio as described in Section 3.5.2, we randomly place the urban hexagon(s) anywhere on the gameboard. This is intended to create more diverse scenarios rather than attempting to even the playing field. We set a maximum of 25 phases. The objective of this scenario setup is to train the Operator agent to be as generalized as possible within the scope of possible gameboard configurations it might encounter in its assigned Objective Area.

**Operator Training**

We use Pass-Agg as the adversary behavior model and train each of our models for 10 million steps using `scenarioCycle` = 0. Based on the model we are training, we use the respective engineered rewards.

The learning curve graphs for each of our RL-trained models are depicted in Figure 6.21. These graphs show the mean rewards obtained through evaluation over the course of the training process. Model evaluation during training was conducted every 10,000 training steps for 100 evaluation episodes each.

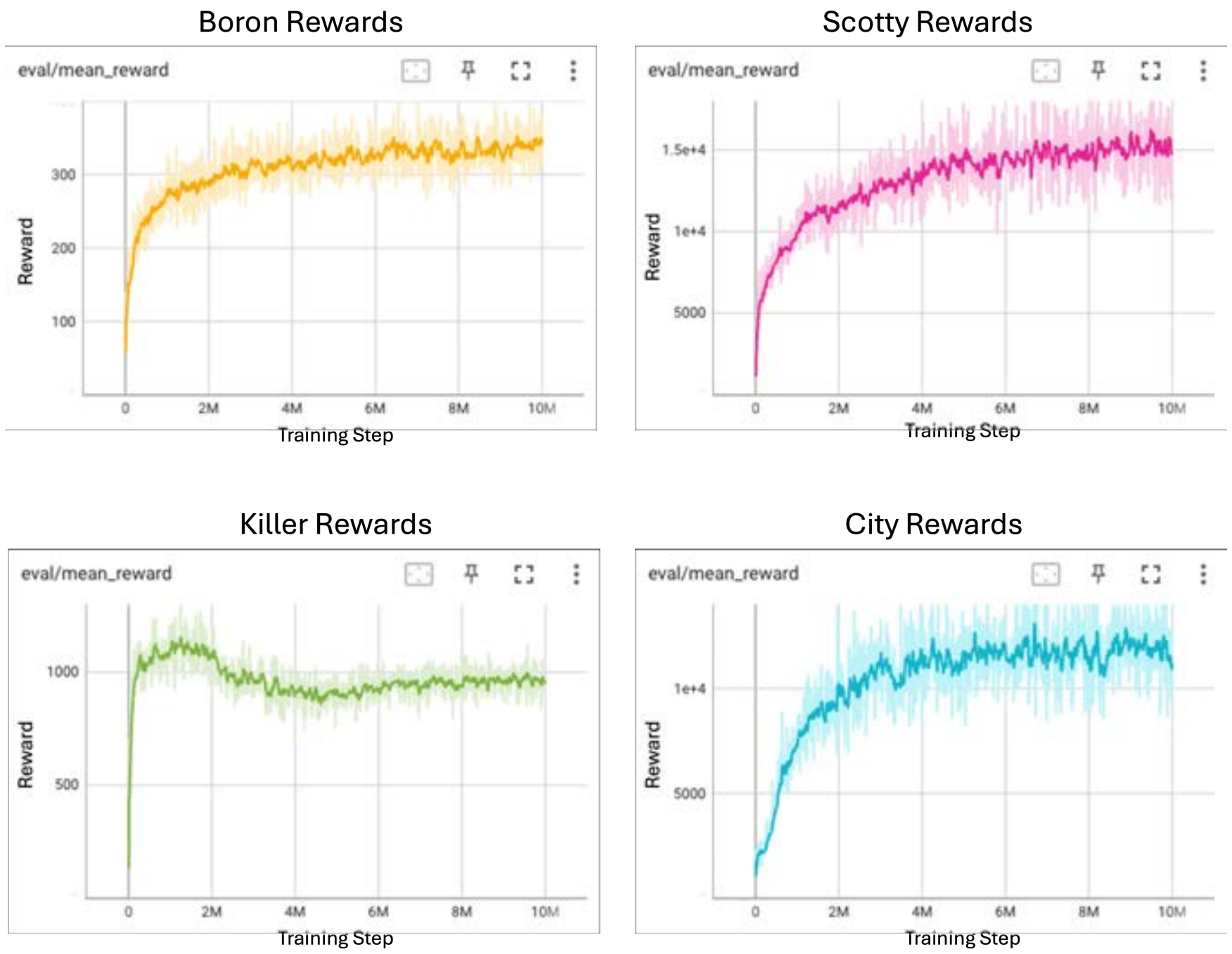

Figure 6.21. Operator Learning Curves. The training curves for the four different Operators are shown. The x-axis depicts the training step, and the y-axis depicts the mean reward achieved during evaluation. Of interest should be the shape of the training rather than the mean reward values attained, as each engineered reward function produced a different scale of reward values.

As noted in Chapter 3, agents trained with the localized observation abstraction tend to learn quickly initially, and then begin to level off. These training curves show either convergence or near-convergence in learning across the four agents.

### Operator Model Evaluation

To gain a better understanding of how these models perform independently, we evaluate each of our models, including our scripted models, against the Pass-Agg behavior model using `scenarioCycle` = 0. We perform 10,000 evaluation runs for each model. A summary of the scores is depicted in Table 6.1. A box and whisker plot is shown in Figure 6.22

Table 6.1. Manager Individual Model Evaluation

| | Model | Mean | SEM |
|---|---|---|---|
| Scripted | Pass-Agg | 113.309 | 4.402 |
| | Pass | 122.136 | 4.408 |
| | Agg | 81.006 | 4.465 |
| | Burt-Plus | 112.395 | 4.458 |
| | City | 123.648 | 4.411 |
| | Kill | -119.445 | 3.874 |
| | Shootback | -138.789 | 3.357 |
| RL | RL-Boron | 144.477 | 4.437 |
| | RL-Scotty | 60.277 | 4.138 |
| | RL-Killer | -232.732 | 4.366 |
| | RL-City | 27.951 | 4.637 |

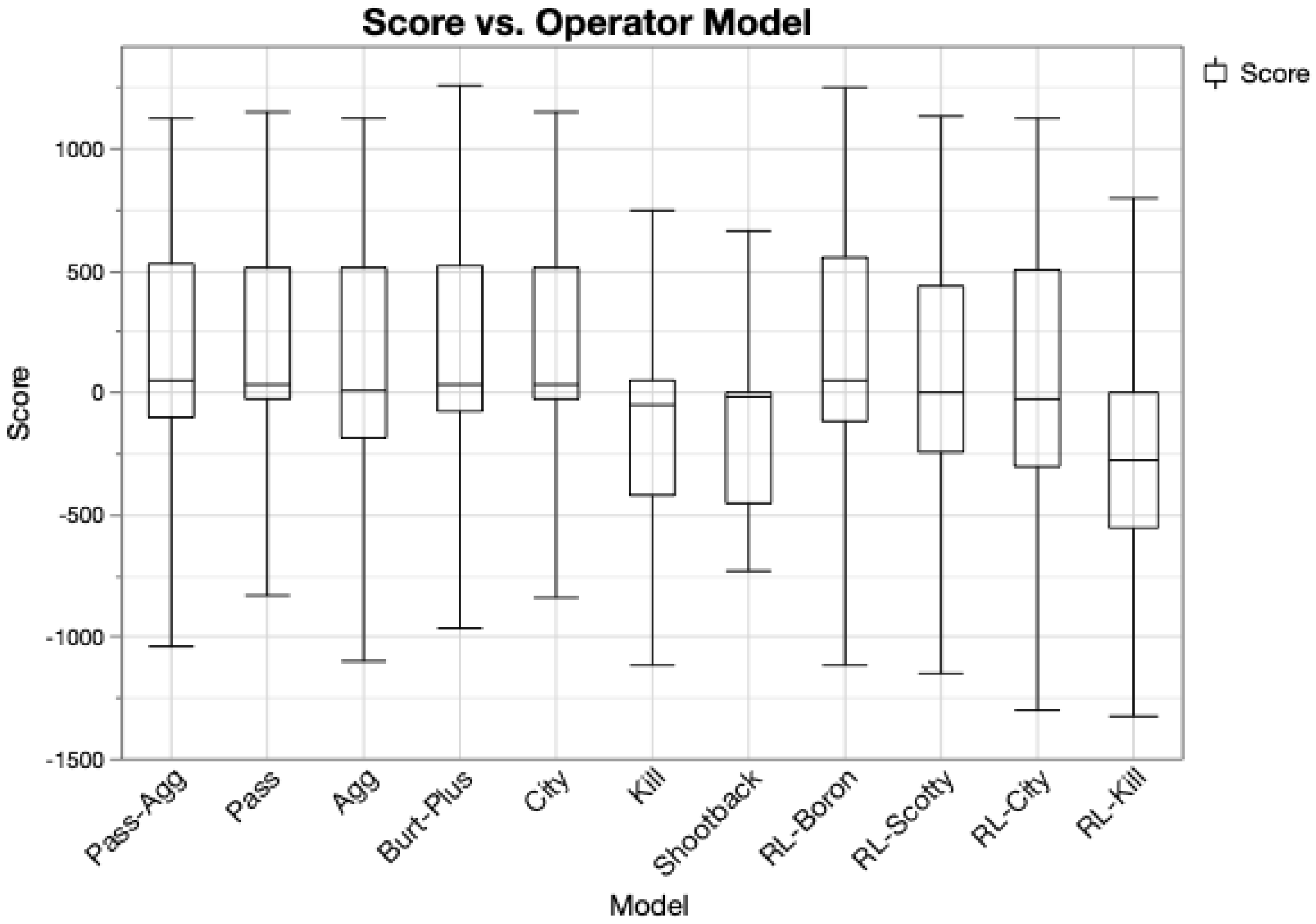

Figure 6.22. Box and Whisker Plot for Operator Individual Model Evaluation.

Of the models examined, RL-Boron performs the best with a mean score of 144.477. City performs the second best with a mean score of 123.648 and Pass performs the third best with a mean score of 122.136. Our baseline Pass-Agg model comes in next with a mean score of 113.309. Of note, Kill, Shootback, and RL-Killer all come in with negative mean scores. However, based on the results of our study in Chapter 3, we still include these in our Multi-Model as there may be specific occasions when these models might excel.

**Operator Score Prediction Models**

For each of our individual behavior models, we train a respective Score Prediction Model. We follow the same methodology described in Chapter 4. In essence, each score prediction model takes in the state observation and outputs a predicted game score with the same assumptions described in Section 4.4.2. The Score Prediction Model for the Operator Multi-Model is called each time a blue unit needs to take an action (i.e., at each action-selection step).

The Score Prediction Model's neural network architecture is as described in Section 4.4.2 using six residual hexagonal convolutional layers of 128 convolution units each, which are then flattened and passed through two fully-connected linear layers of 8,000 units each. Between each convolutional and linear layer, we use a ReLU activation function.

For the dataset to train each Score Prediction Model for the Operator, we employ Atlatl to generate 18 million samples of synthetic data using the respective single model and the scenario described above. Each exemplar consists of a tensor representation of the observation as the input and a final game score as the label. We repeat this process to train a separate Score Prediction Model for each of the 11 Operator single-models, for a total of 198 million samples.

From the 18 million samples for each single model, we use 12 million for training, 6 million for training validation, and 6 million for final testing prior to incorporating each Score Prediction Model into our Operator Multi-Model. Due to the volume of data, we find training our model over just one epoch was sufficient. During testing, for each of the 11 Operator models for which we train a Score Prediction Model, we observe Mean Squared Errors (MSEs), Mean Absolute Errors (MAEs), and $R^2$ as depicted in Table 6.2.

MSE measures the average squared difference between the predicted values and the actual values. Better accuracy and precision of the model are suggested by a lower MSE, indicating that the predictions are closer to the actual values. However, MSE is sensitive to outliers, as it squares the errors. MAE, on the other hand, calculates the average absolute difference between predicted and actual values. Like MSE, a lower MAE indicates better accuracy, but MAE is less sensitive to outliers since it does not square the errors. Therefore, MAE provides a more straightforward interpretation of the model's predictive performance. Lastly, $R^2$ is a statistical measure representing the proportion of the variance in the dependent variable explained by the model's independent variables. This measure usually ranges from 0 to 1, where 1 represents a perfect fit and 0 implies no improvement over using the mean of the dependent variable. $R^2$ helps describe how well a given model fits the observed data points. A higher $R^2$ suggests that the model provides a better fit to the data. These measures collectively provide crucial insights into the predictive accuracy, precision, and overall performance of our models, allowing us to assess their effectiveness in capturing the complexities of the underlying data and making accurate predictions

Table 6.2. Operator Score Prediction Model Evaluation

| Model | MSE | MAE | $R^2$ |
|---|---|---|---|
| Pass-Agg | 9999.844 | 45.734 | 0.929 |
| Pass | 7247.048 | 35.822 | 0.949 |
| Agg | 10711.516 | 44.681 | 0.928 |
| Burt-Plus | 7848.246 | 39.750 | 0.933 |
| City | 7459.899 | 33.359 | 0.947 |
| Kill | 14743.190 | 44.993 | 0.893 |
| Shootback | 2222.010 | 19.149 | 0.983 |
| RL-Boron | 9545.669 | 45.474 | 0.944 |
| RL-Scotty | 13543.909 | 60.921 | 0.916 |
| RL-Killer | 17234.470 | 67.364 | 0.931 |
| RL-City | 14376.016 | 61.548 | 0.927 |

We note that Shootback exhibits relatively low MSE and MAE values, indicating higher predictive accuracy compared to others like RL-Killer and RL-City with higher error metrics. As a point of comparison for these MAEs, the full range of scores for each model is depicted in Figure 6.22.

### 6.4.3 Manager Multi-Model

We use a total of 12 individual behavior models for our Manager Multi-Model. We use the following scripted models:

- Manager Balanced: this model uses the following logic priority to select its Objective Area within its Operating Area. If the Manager has a unit on an urban hexagon, it selects the same Objective Area (attempts to hold the city), otherwise, it selects an Objective Area that encompasses the nearest unoccupied urban hexagon. If no unoccupied urban hexagons exist within its Operating Area, it will select an Objective

Area that encompasses the nearest urban hexagon occupied by the red faction. If this condition is not met, the Manager will select the Objective Area where the nearest enemy unit is located. If no enemy units exist within its Operating Area, the Manager will select an Objective Area that encompasses the nearest urban hexagon (within its Operating Area) occupied by the blue faction. Lastly, if none of these conditions are met, the Manager will assign the middle of the Operating Area as its Objective Area.

- Manager Prioritized-City: this model uses the same logic above with the exception that it removes the step for selecting an Objective Area where the nearest enemy unit is located. In summary, the priorities for urban hexagon selection are unoccupied, red-occupied, and blue-occupied.
- Manager Seize-Red-City: this model simply selects the Objective Area of the nearest red-occupied urban hexagon. If none exist, it selects the center of its Operating Area.
- Manager Killer: this model always selects the Objective Area encompassing the nearest enemy unit. If none exist, it selects the center of its Operating Area.
- Manager Hold: this model simply selects its current location as its Objective Area.

We use the following RL-trained models:

- Manager RL-Boron: uses the same engineered reward system as described in Section 5.3.8 and detailed in Equation 5.1.
- RL-Defeat-and-Capture: this reward system's objective is to completely eliminate the adversary in combat and occupy all urban hexagons. In essence, it overvalues damaging opponents and occupying cities beyond the default scoring system within Atlatl. It accomplishes this by tracking accumulated game score rewards for damaging adversary units and occupying urban hexagons. This reward system, as well as all of the following Manager RL reward systems below, penalizes Managers for duplicate objectives to discourage redundancy, as well as adds bonuses at game termination for complete enemy elimination and full city occupation. Moreover, the reward calculation considers the Manager and overall faction performance by combining base rewards with kill and capture bonuses, slightly scaled to emphasize the preservation of the Manager's and the entire faction's strength relative to their original strength.
- Manager RL-Occupy-City: this reward system primarily emphasizes control of urban hexagons, assigning a greater value to capturing urban hexagons than the standard scoring system in Atlatl. Additionally, at game termination, it awards additional bonuses for the complete occupation of all urban hexagons. The reward calculations consider both the individual performance of Managers and the overall faction by integrating base rewards with specific bonuses for successful city captures.

- Manager RL-Seize-Red-City: this system is designed to prioritize capturing urban hexagons that are owned by the opposing faction by providing significant rewards for capturing and holding these locations. Additional bonuses are added at the end of the game if key locations remain under the Manager's control.
- Manager RL-Defensive: this reward system is designed to prioritize the preservation of the Manager's unit strength in combat. It assigns a higher value to maintaining high unit survival rates and minimizing losses. At game termination, additional bonuses are awarded for maintaining a high percentage of the original Manager and overall force strength, effectively rewarding defensive actions.
- Manager RL-Kill: this reward system is designed to prioritize the complete defeat of the adversary in combat. It emphasizes inflicting damage on opposing units rather than occupying urban hexagons, assigning a higher value to defeating the enemy than the default scoring system used within Atlatl. Additional bonuses are added at game termination for the total elimination of the enemy scaled by the original force ratio.
- Manager RL-Kamikaze: this system rewards aggression by focusing on the complete defeat of enemy forces without regard for the preservation of its own force. It significantly values actions that involve aggressive actions. The system tracks and accumulates game score rewards specifically for eliminating adversary units and rewarding managers for damaging and defeating the opponent regardless of force ratio.

Of note, if a Manager is not within its Operating Area, as passed down by its Commander, it will select the nearest Objective Area within its Operating Area. Once the Manager's Operators arrive within this Objective Area, the Manager will be called to select one of the above-described behavior models.

**Manager Observation Abstraction**

We use a similar logic to abstract the Manager's observation as described in Section 5.3.3, however, we implement two significant changes in this implementation in an attempt to make the RL problem easier to learn on the Manager as well as more computationally efficient. In our study in Chapter 5, we used a `scenarioCycle` = 10 as we were not able to attain the generalization needed for a `scenarioCycle` = 0. In order to implement a Manager agent within our overall HRL architecture, it is necessary for our Manager agent to be able to further generalize beyond a `scenarioCycle` = 10 as we cannot guarantee any specific configuration of scenarios at this level. Thus, we attempt to simplify the training of our Manager agent by reducing its action space to nine (as opposed to 49 in the previous study), and reducing its observation space to a $5 \times 5$ (as opposed to a $7 \times 7$ in the previous study).

Additionally, whereas the Manager abstraction described in 5.3.3 involved reducing the full tensor into a smaller grid by comparing the original and reduced grid sizes and accounting for overlapping areas in a proportional manner, the method implemented in this experiment removes the proportional calculations for overlapping areas and instead assigns the value in its entirety to the closest grid. For example, if one of the $5 \times 5$ grids overlaps a hexagon containing a unit of health 1.0 by $2/3$, rather than assigning one hexagon a value of 0.67 and the other hexagon a value of 0.33, the function assigns the nearest grid the full value of 1.0. We deliberately implement this design choice in an effort to enhance computational efficiency, as the previous method necessitated the inclusion of two additional embedded `for` loops.

Figure 6.24 shows an example red forces channel where the Manager takes in the observation of its assigned Operating Area, divides the area into a $5 \times 5$ grid, and then sums the information within each grid. When the grid overlaps a hexagon, the entire value is assigned to the nearest grid. We use the same observation channels illustrated in Figure 5.6. Thus, we take Objective Area ($s$) of $18 \times 10 \times 10$, pass it through our manager abstraction function ($\phi_{\text{manager}}$) to obtain our abstracted observation ($s_{\phi_{\text{manager}}}$) $17 \times 5 \times 5$.

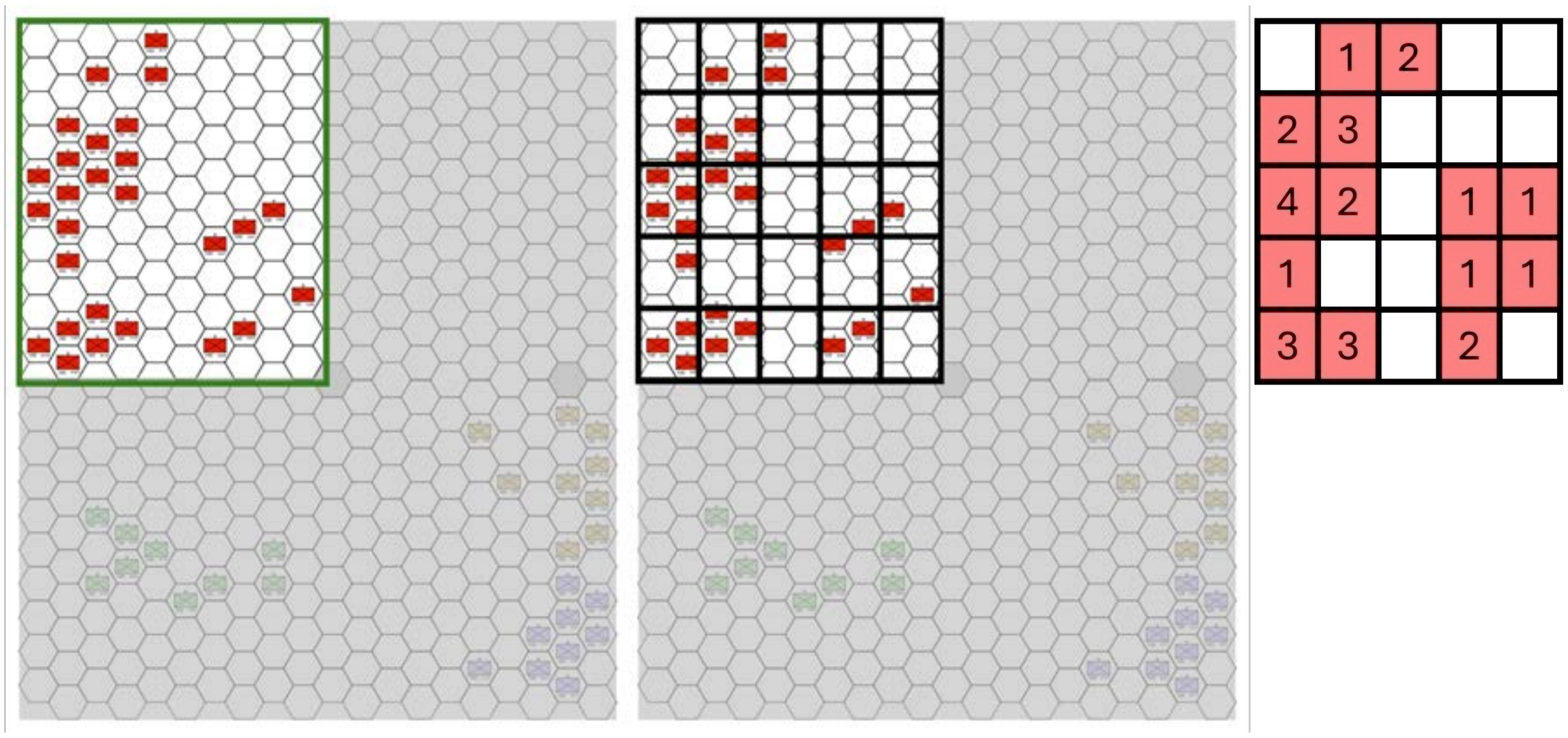

Figure 6.23. Observation Abstraction for Manager. An example of the red force channel of the Manager's entire $18 \times 10 \times 10$ observation space. The Manager takes in the Operating Area as passed by its Commander. It then overlaps a $5 \times 5$ grid over this observation space and sums the information based on channels. When a hexagon overlaps multiple different grids, the value is assigned to the nearest grid in its entirety.

**Manager Gymnasium Environment**

We modify our Manager Gymnasium Environment described in Section 5.3.4 to accept an observation space of $17 \times 5 \times 5$ and reduce the action space from 49 to nine (based on our $3 \times 3$ selection). As discussed above, this is intended to simplify the RL problem for the Manager.

**Manager Reinforcement Learning Neural Network Architecture**

We use the same residual CNN architecture described in 5.3.5 consisting of 7 layers of 64 channels each, flattened to a 512-dimensional feature vector, employing ReLU activation functions after each layer.

### Manager Reinforcement Learning Algorithm

We employ the DQN algorithm, as it has shown better performance over the PPO algorithm for this task based on our exploratory experiments. We use the same hyperparameter values as described in Section 3.5.2 with the exception of the exploration hyperparameters. After experimenting with different values we find that, for the Managers, the original values that included an epsilon ($\epsilon_i$) of 1.0, decaying linearly to a final epsilon ($\epsilon_f$) of 0.01, with an exploration fraction of 1.0, resulted in a model that exhibited a very shallow learning curve and that was still learning at 5 million steps. Setting $\epsilon_i$ to 1.0, decaying linearly to a final epsilon ($\epsilon_f$) of 0.05, with an exploration fraction of 0.5 resulted in overall steeper learning curves and convergence or near convergence for all models.

### Manager Training Scenarios

We generate random training scenarios for our Managers that are intended to be representative of the different configurations of the $10 \times 10$ Operating Area that the Manager may encounter during gameplay. We modify the scenario generator to produce $10 \times 10$ gameboards with a random number of Managers between one and nine (resulting in three to 18 blue units) and a random number of red units between zero to 18. We use zero to two urban hexagons, with the urban hexagon(s) being randomly placed anywhere on the gameboard, regardless of force ratio. We set a maximum of 40 phases. Like with the Operators, the objective of this scenario setup is to train the Manager agent to be as generalized as possible within the scope of possible gameboard configurations it might encounter in its assigned Operating Area.

### Manager Training

To train each of our RL Manager models, we use scripted Operators as suggested by the literature to ensure stability and faster convergence in learning. We acknowledge this method has its limitations in that the Manager model is being optimized to use only the Pass-Agg behavior model despite it having many different models to choose from during execution. However, we decided that to maintain training efficiency for this study, Pass-Agg serves as an acceptable training model for this task.

Figure 6.24 depicts the RL training loop for the Manager training process. As noted, the full observation space undergoes an abstraction ($\phi_{\text{manager}}$) to produce an abstracted observation ($s_{\phi_{\text{manager}}}$). This observation is then passed to the Manager. An action (option $o_{\text{manager}}$) from zero to eight is selected representing one of the $3 \times 3$ center grid locations, which is then converted to a $5 \times 5$ Objective Area. This Objective Area is passed to the Scripted Operators. As described in Section 5.3.2, the Manager is only called to make a decision when all of its Operators are within their assigned Objective Area.

Also described in detail in Section 5.3.2, the Scripted Operator makes decisions at each time step using one of two behavior modules described in Figure 5.4. The Move Model is selected if the Operator is not within its Objective Area, and the Local Pass-Agg Model is selected when the Operator detects it is within its Objective Area. Like in the study in Chapter 5, we restrict the information available to the Scripted Operator to that of its Objective Area.

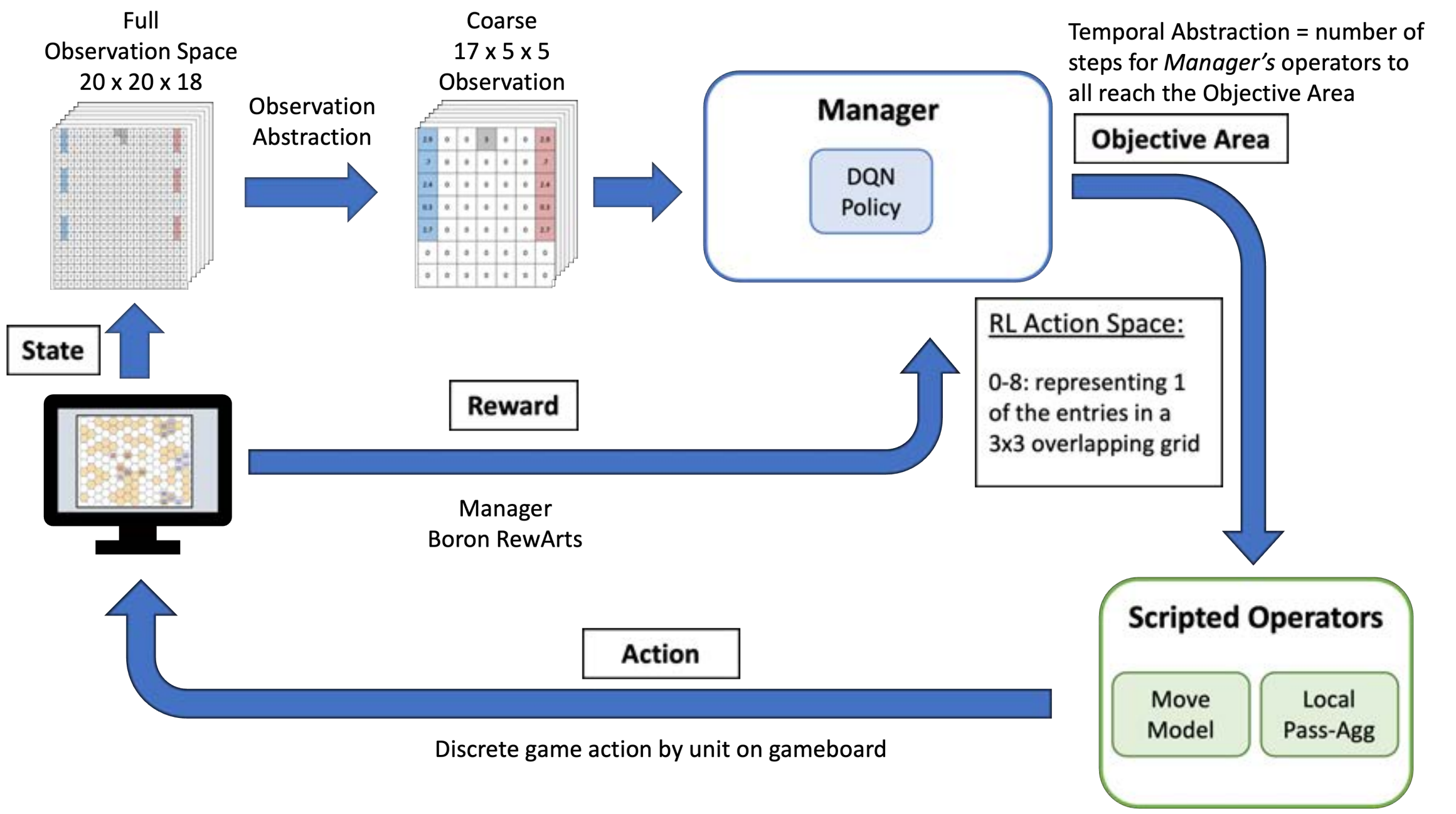

Figure 6.24. Manager Reinforcement Learning Training Meta-Model. A Manager DQN policy is trained using Scripted Operators.

With the exception of the Boron Manager model, we train each model (using their respective engineered rewards) for 5 million steps using `scenarioCycle` = 0 against Pass-Agg as the adversary models. Due to the generalized nature of the behaviors learned by the Boron model, it is trained for 10 million steps to ensure convergence. The learning curve graphs for each of our RL-trained Manager models are depicted in Figure 6.25. These graphs show the mean rewards obtained through evaluation over the course of the training process. Model evaluation during training was conducted every 10,000 training steps for 100 evaluation episodes each.

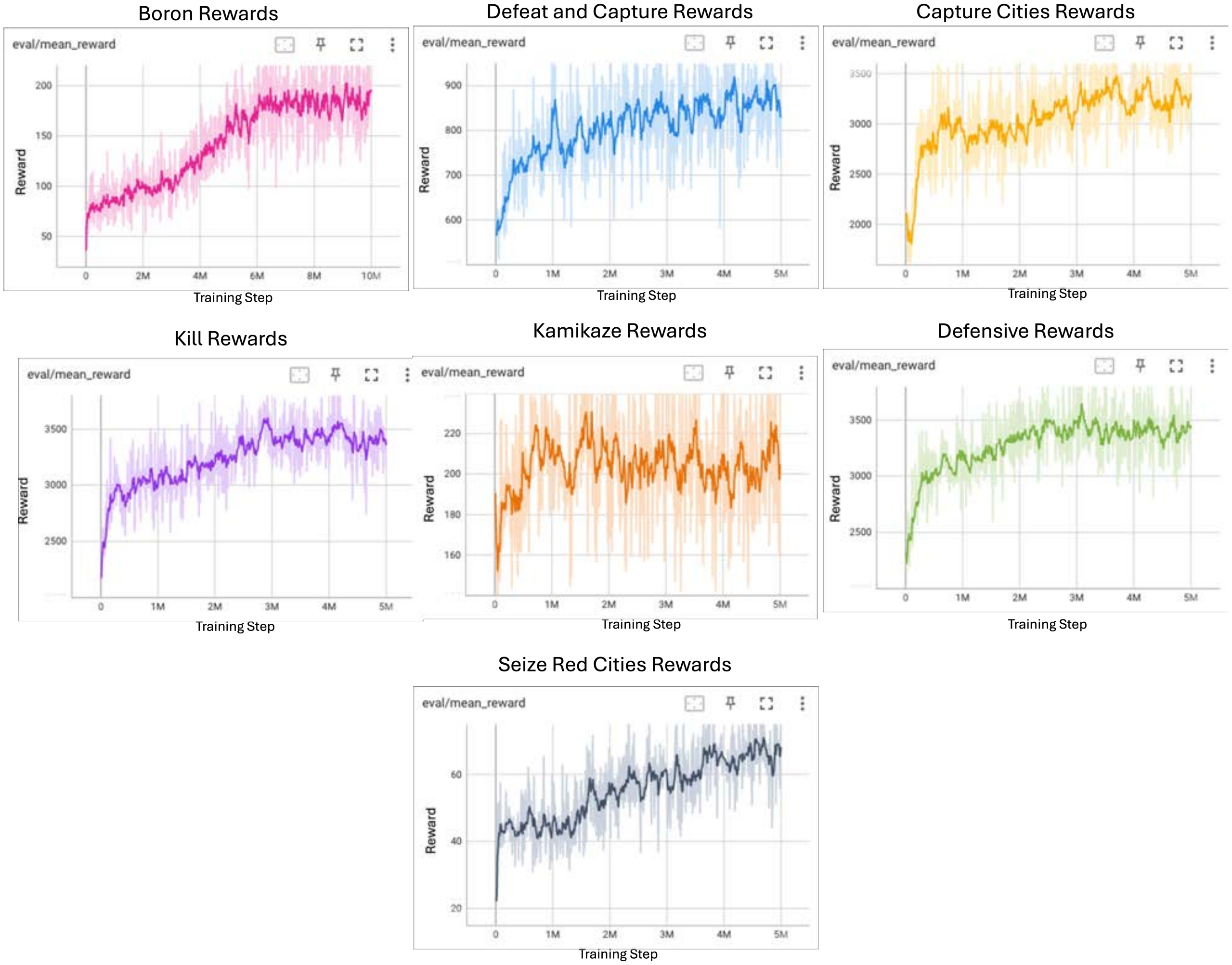

Figure 6.25. Operator Training Curves. The training curves for the seven different Managers are shown. The x-axis depicts the training step, and the y-axis depicts the mean reward achieved during evaluation. Of interest should be the shape of the training rather than the mean reward values attained, as each engineered reward function produced a different scale of reward values.

## Manager Model Evaluation

As with the Operators, to gain a better understanding of how these models perform independently, we evaluate each of our models against the Pass-Agg behavior model using

`scenarioCycle` = 0. We perform 10,000 evaluation runs for each model. A summary of the scores is depicted in Table 6.3. As noted, the scripted models appear to perform better overall, with the Prioritized-City model performing best with a mean score of 195.210 and the RL-Kill model performing worse with −86.239. A box and whisker plot is shown in Figure 6.26.

Table 6.3. Manager Individual Model Evaluation

| | Model | Mean | SEM |
|---|---|---|---|
| Scripted | Balanced | 167.453 | 6.573 |
| | Prioritized-City | 195.210 | 6.427 |
| | Seize-Red-City | -52.166 | 6.103 |
| | Killer | -54.146 | 5.825 |
| | Hold | -50.682 | 6.107 |
| RL | RL-Boron | 0.417 | 6.383 |
| | RL-Defeat-and-Capture | -30.387 | 6.329 |
| | RL-Occupy-City | 52.967 | 6.357 |
| | RL-Seize-Red-City | 23.144 | 6.403 |
| | RL-Defensive | -52.095 | 6.224 |
| | RL-Kill | -86.239 | 5.992 |
| | RL-Kamikaze | -62.984 | 6.009 |

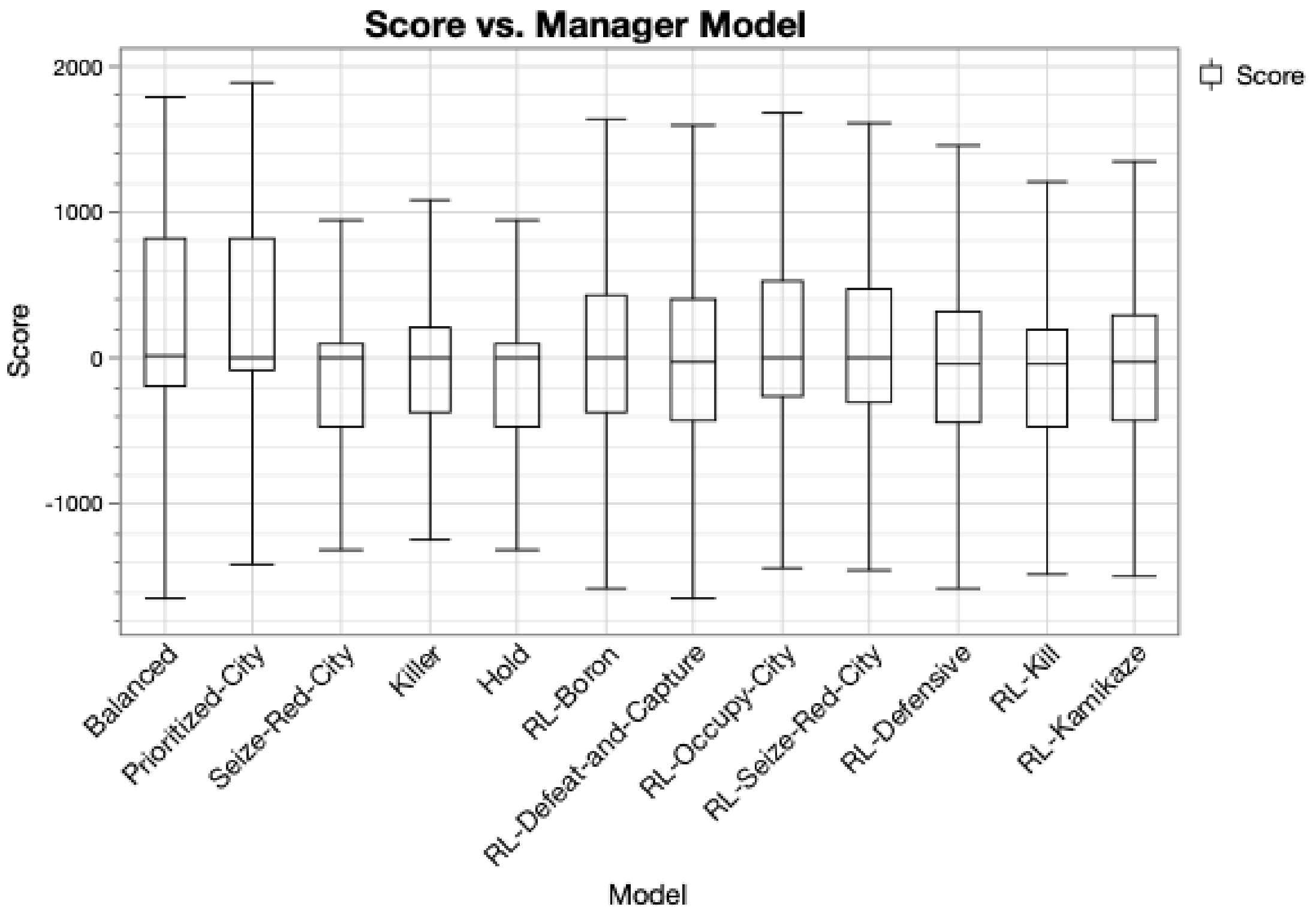

Figure 6.26. Box and Whisker Plot for Manager Individual Model Evaluation.

**Manager Score Prediction Models**

For each of our individual Manager behavior models, we train a respective Score Prediction Model in the same way as previously described. These Manager Score Prediction Models are called each time a Manager is called to take an action (i.e., when all of its Operators are within its assigned Objective Area).

The Score Prediction Model's neural network architecture is as described above for the Operator. Like with the Operator, for the dataset to train each Manager's Score Prediction Model, we generate 18 million samples of data using the respective Manager model and the scenario generator described above. We train each of the 12 Manager models using a total of 216 million samples.

Table 6.4. Manager Score Prediction Model Evaluation

| Model | MSE | MAE | $R^2$ |
|---|---|---|---|
| Balanced | 27125.763 | 70.467 | 0.910 |
| Prioritized-City | 20301.272 | 56.815 | 0.925 |
| Seize-Red-City | 22112.072 | 58.423 | 0.938 |
| Killer | 34522.740 | 87.423 | 0.864 |
| Hold | 26049.011 | 76.665 | 0.927 |
| RL-Boron | 35966.614 | 92.764 | 0.897 |
| RL-Defeat-and-Capture | 31951.817 | 81.001 | 0.898 |
| RL-Occupy-City | 31951.817 | 81.001 | 0.898 |
| RL-Seize-Red-City | 35785.529 | 92.878 | 0.906 |
| RL-Defensive | 42230.478 | 104.937 | 0.885 |
| RL-Kill | 40330.141 | 105.852 | 0.865 |
| RL-Kamikaze | 40106.697 | 102.189 | 0.863 |

From the 18 million samples for each Manager model, we use 12 million for training, 6 million for training validation, and 6 million for final testing of our model prior to incorporating each into the Manager Multi-Model. We find that one epoch is also sufficient for the Manager's Score Prediction Model training. For each of these models, we observe MSEs, MAEs, and $R^2$ as depicted in Table 6.4.

We note increased MSEs and MAEs as compared to our Operator Score Prediction Models. However, this can be expected as the scores produced in these evaluation scenarios are of a much wider range than those of our Operator scenarios. Nevertheless, we observe smaller $R^2$ values, indicating the increased challenges of predicting a score when using a higher level of abstraction.

### 6.4.4 Commander Multi-Model

We use a total of six individual behavior models for our Commander Multi-Model. We develop and integrate the following scripted models:

- Commander Balanced: this model uses the same logic of priorities as the Scripted Manager Balanced model; however, instead of assigning an Objective Area, it assigns an Operating Area to its Managers.

- Commander City: this model assigns Operating Areas using the same priority logic as the Scripted Manager Prioritized-City model.
- Kill: this model assigns Operating Areas using the same priority logic as the Scripted Manager Killer model.
- Hold: this model simply selects its current location as its Operating Area.

We train and implement the following RL-trained models:

- Commander RL-Boron: uses the same engineered reward system as described in Section 5.3.8 and detailed in Equation 5.1 but aggregating rewards from all units under the respective Commander (as opposed to just at the Manager level).
- Commander RL-City: uses a similar reward system as the Manager RL-Capture-City model but at the Commander level of aggregation.
- Commander RL-Kill: uses a similar reward system as the Manager RL-Aggressive model but, again, at the Commander level of aggregation.

**Commander Observation Abstraction**

The process to obtain the Commander's observation abstraction is identical to that of the Manager's observation abstraction; however, it is abstracting the full $20 \times 20$ gameboard into a $5 \times 5$ representation. We use the same observation channels as the Manager uses, illustrated in Figure 5.6, with two exceptions. The first channel depicts the units owned by the Commander on-move (rather than just the Manager on-move) and the second channel depicts other Commanders' goal locations (rather than other Managers' goal locations). In essence, we take the original state space ($s$) of $18 \times 20 \times 20$ and pass it through our Commander's abstraction function ($\phi_{\text{commander}}$) to obtain our $17 \times 5 \times 5$ abstracted observation ($s_{\phi_{\text{commander}}}$), as illustrated in 6.27.

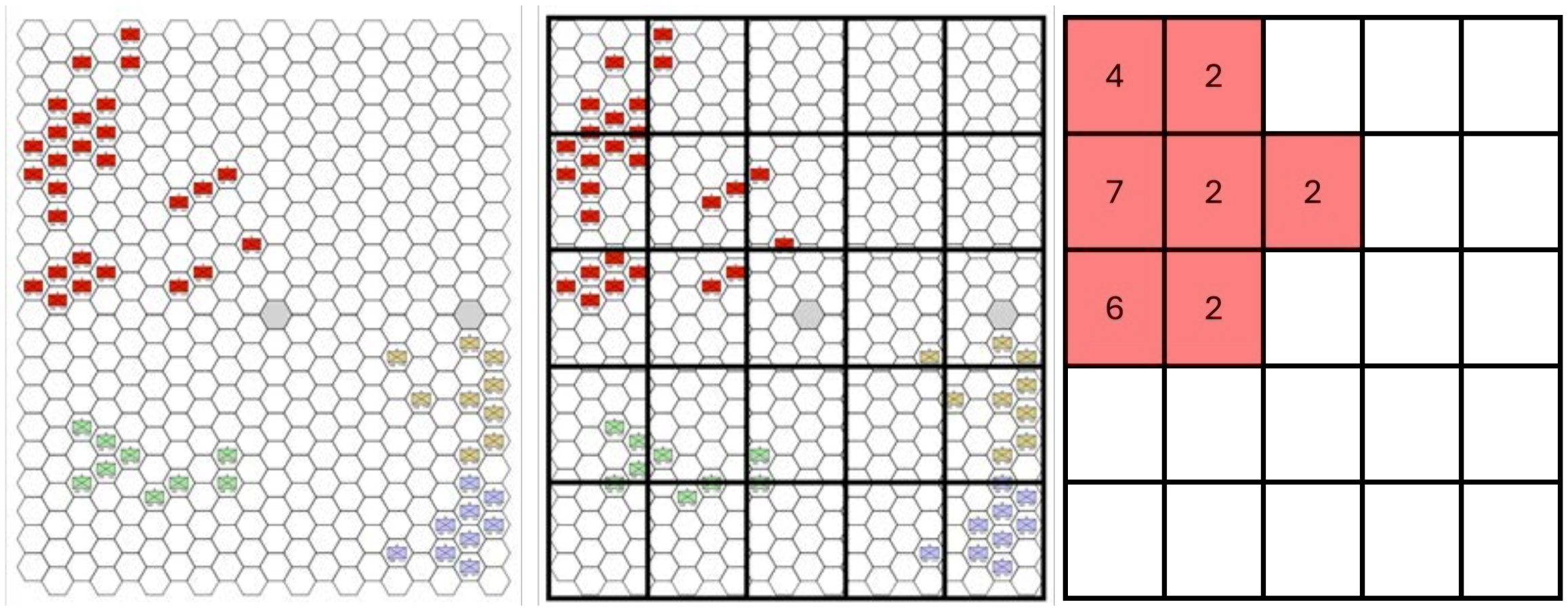

Figure 6.27. Observation Abstraction for Commander. An example of the red force channel of the Commander's entire 17 × 20 × 20 observation space. The Commander takes in the entire gameboard, then overlaps a 5×5 grid over this observation space, and finally sums the information based on channels. When a hexagon overlaps multiple different grids, the value is assigned to the nearest grid in its entirety

**Commander Gymnasium Environment**

We create a Commander Gymnasium Environment that is similar to the Manager's Gymnasium Environment, except that it accepts the Commander's observation space of $17 \times 5 \times 5$. Like the Manager, it uses an action space of nine (based on our $3 \times 3$ selection).

**Commander Neural Network Architecture**

We use the same residual CNN architecture described in 5.3.5 and as used by the Manager models.

**Commander Reinforcement Learning Algorithm**

Although we find success using the PPO algorithm for `scenarioCycles` of 10 or less in exploratory Commander experiments, we find that the DQN algorithm shows slightly better performance in `scenarioCycle` = 0. We use the same hyperparameter values as described above for our RL Manager models.

**Commander Training Scenarios**

We use the Atlatl scenario generator to create random starting scenarios that consist of $20 \times 20$ gameboards with up to three Commanders, each of which is composed of three Managers of three Operators each, for a total range between nine to 27 units per faction. Up to two urban hexagons are randomly placed on the gameboard according to force ratio as described in Section 3.5.2, where urban hexagons are placed in locations more favorable to the weaker faction. We set the maximum number of phases to 80.

**Commander Training**

As with our RL Manager models, we use scripted subordinate agents to train our RL Commander models. Figure 6.28 depicts the RL training loop for the Commander training process. As noted, the full observation space undergoes an abstraction ($\phi_{\text{commander}}$) to produce an abstracted observation ($s_{\phi_{\text{commander}}}$). This observation is then passed to the Commander. Every 40 phases, the Commander chooses an action (option $o_{\text{commander}}$) from 0 to 8. This is then mapped to one of the $3 \times 3$ center grid locations, which is then converted to a $10 \times 10$ Operating Area. This Operating Area is passed to its Scripted Managers. We use the Scripted Manager Balanced behavior model described in Section 6.4.3. Each Scripted Manager then selects an Objective Area to pass to its Scripted Operator models, as described in the Manager section above. The Scripted Operator model then selects an action to take on the environment according to one of its two embedded behavior modules, described in the Operator section above. A reward (based on the specific Commander model being trained) and next-state is provided to the Commander and the RL training loop is restarted.

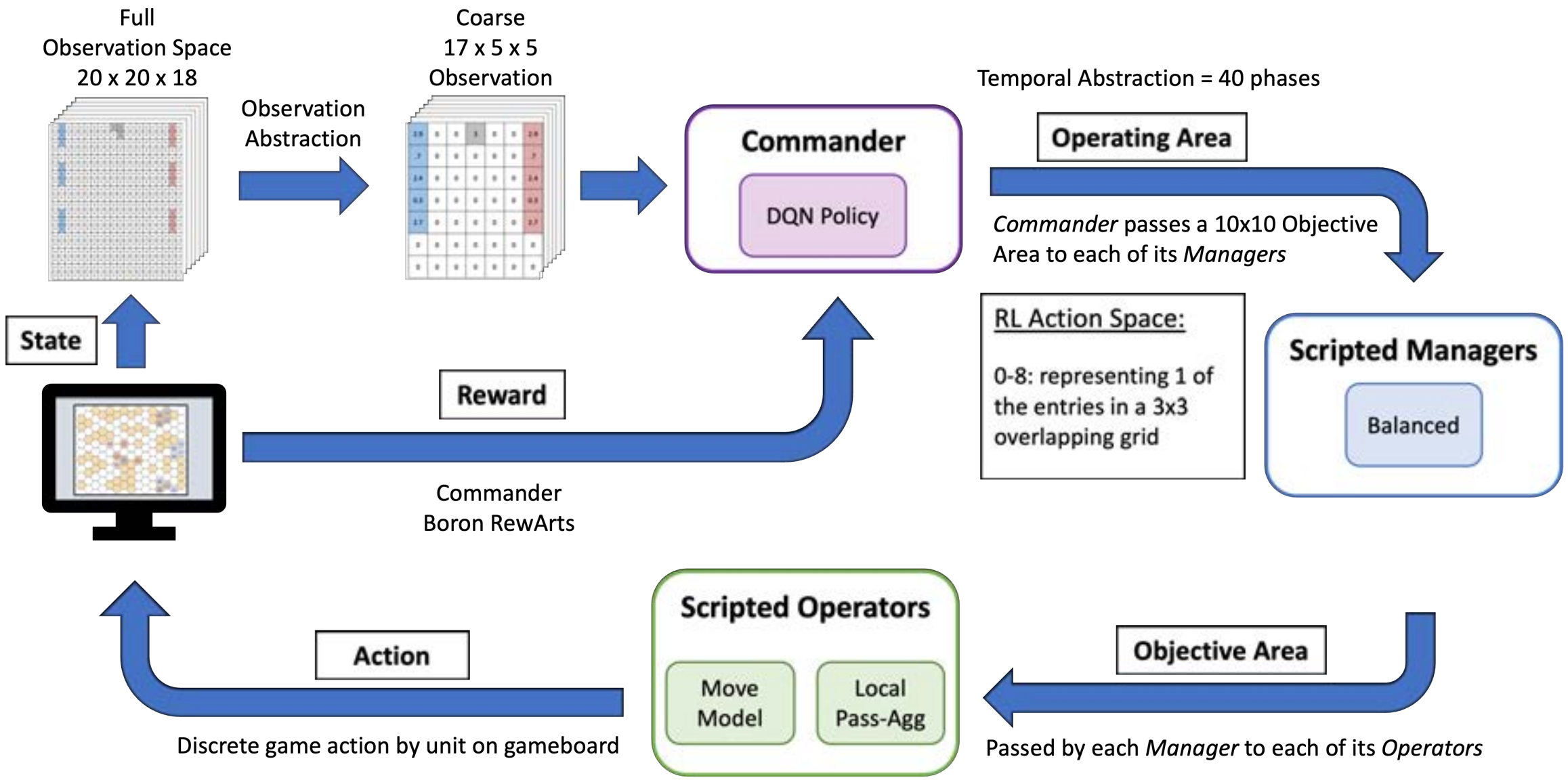

Figure 6.28. Commander Reinforcement Learning Training Meta-Model. The Commander's DQN policy is trained using Scripted Managers and Scripted Operators.

Due to the amount of time required to train the Commander level models, we train each model (using their respective engineered rewards) for 1 million steps using `scenarioCycle` = 0 against Pass-Agg as the adversary models. The learning curve graphs for each of our RL-trained models are depicted in Figure 6.29. These graphs show the mean rewards obtained through evaluation over the course of the training process. Model evaluation during training was conducted every 10,000 training steps for 100 evaluation episodes each.

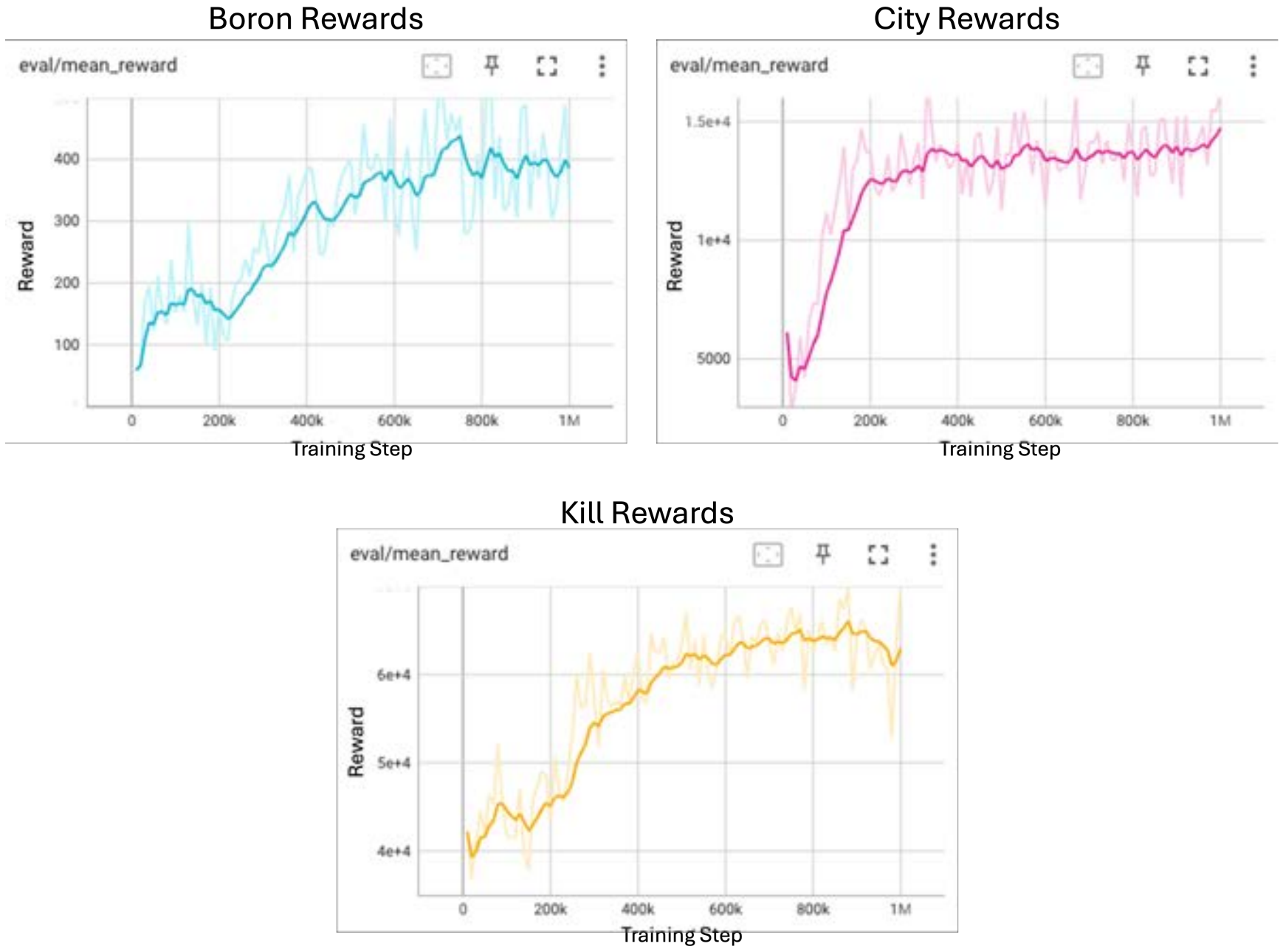

Figure 6.29. Commander Training Curves. The training curves for the three different Commanders are shown. The x-axis depicts the training step, and the y-axis depicts the mean reward achieved during evaluation. Of interest should be the shape of the training rather than the mean reward values attained, as each engineered reward function produced a different scale of reward values.

**Commander Model Evaluation**

We evaluate each of our Commander models against the Pass-Agg behavior model using `scenarioCycle` = 0. We perform 10,000 evaluation runs for each model. A summary of the scores is depicted in Table 6.5. A box and whisker plot of the scores is shown in Figure 6.31. Of note, all of the mean scores are substantially negative, even for the scripted agents.

Table 6.5. Commander Individual Model Evaluation

| | Model | Mean | SEM |
|---|---|---|---|
| Scripted | Balanced | -792.337 | 12.009 |
| | City | -784.283 | 11.990 |
| | Kill | -1065.823 | 10.863 |
| | Hold | -1029.968 | 10.628 |
| RL | RL-Boron | -554.461 | 13.564 |
| | RL-City | -391.112 | 13.327 |
| | RL-Kill | -444.042 | 13.457 |

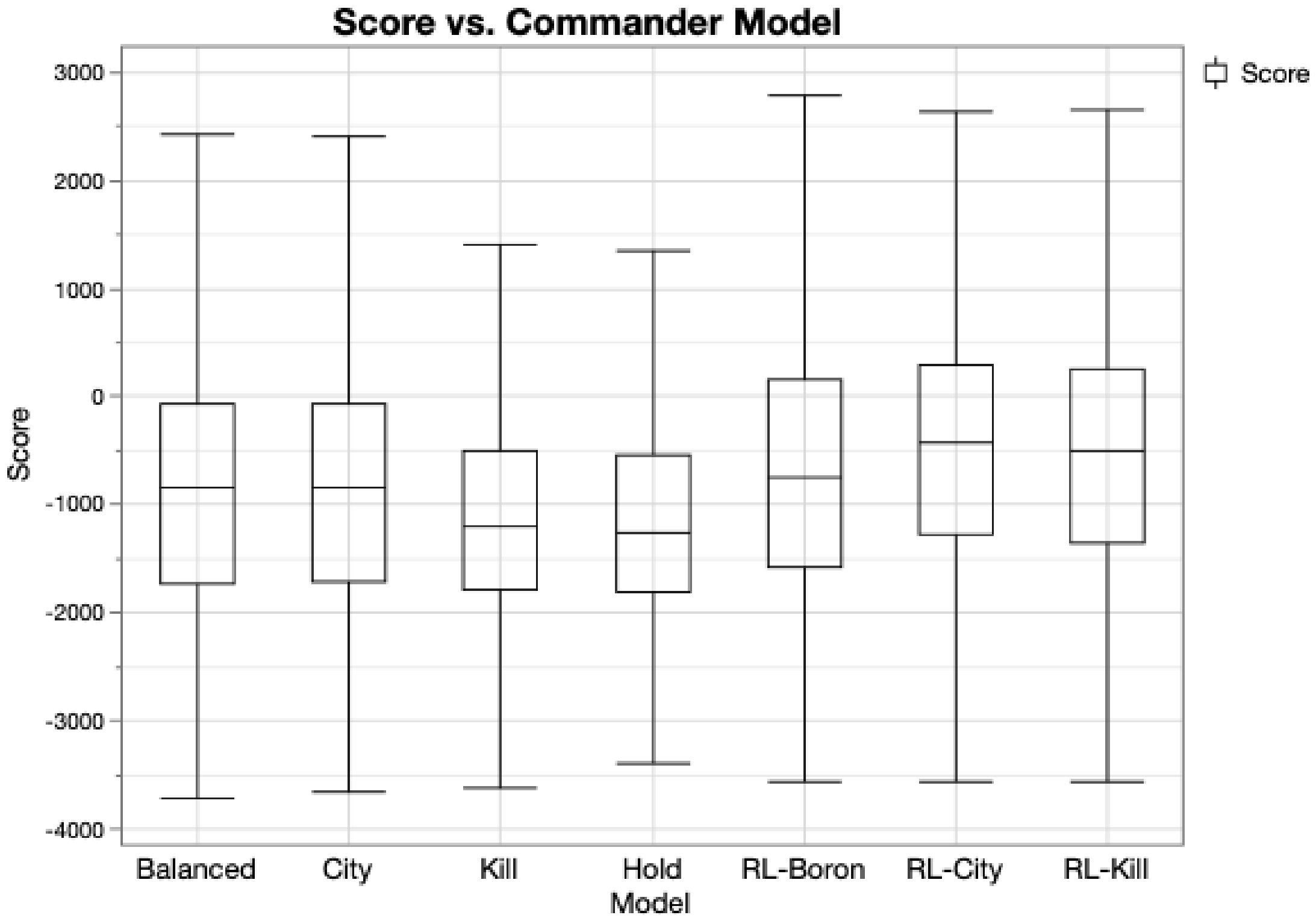

Figure 6.30. Box and Whisker Plot for Commander Individual Model Evaluation.

This outcome suggests that the Commander models are not effective in achieving desirable outcomes based on the current design. Although visual replays show sensible choices being made, the evaluation scores indicate that this implementation needs further refinement and improvement. Of interest, however, is that unlike the Manager agent where the RL models underperformed as compared to the scripted models, we find that the RL Commander models outperformed their scripted counterparts in this evaluation. Further analysis is needed to understand the underlying factors contributing to this divergence in performance and to leverage these insights for refining both RL and scripted Commander models in future iterations.

**Commander Score Prediction Models**

We train a Score Prediction Model for each of our Commander models using the same method as previously described. Each Commander Score Prediction Model is called to make an inference during each Commander's turn (every 40 phases).

The Score Prediction Model's neural network architecture is as described above for the Manager. Because each Commander model takes an extended time to run and generate data, we generate 6 million samples of data using the respective Commander model and the scenario generator described above, for a total of 42 million samples across all Commander models.

From the 6 million samples for each Commander model, we use 4 million for training, 1 million for training validation, and 1 million for final testing prior to incorporation into our Commander Multi-Model. We find training for three epochs minimized the loss while preventing overfitting. For each of the seven Commander models for which we train a Score Prediction Model, we observe MSEs, MAEs, and $R^2$ as depicted in Table 6.6 and illustrated in Figure 6.31.

Table 6.6. Manager Score Prediction Model Evaluation

| Model | MSE | MAE | $R^2$ |
|---|---|---|---|
| Balanced | 785423.524 | 579.188 | 0.582 |
| City | 860055.238 | 689.700 | 0.544 |
| Kill | 739054.602 | 602.356 | 0.500 |
| Hold | 489458.484 | 408.002 | 0.528 |
| RL-Boron | 651647.183 | 548.061 | 0.681 |
| RL-City | 678099.332 | 573.638 | 0.643 |
| RL-Kill | 734339.129 | 603.483 | 0.531 |

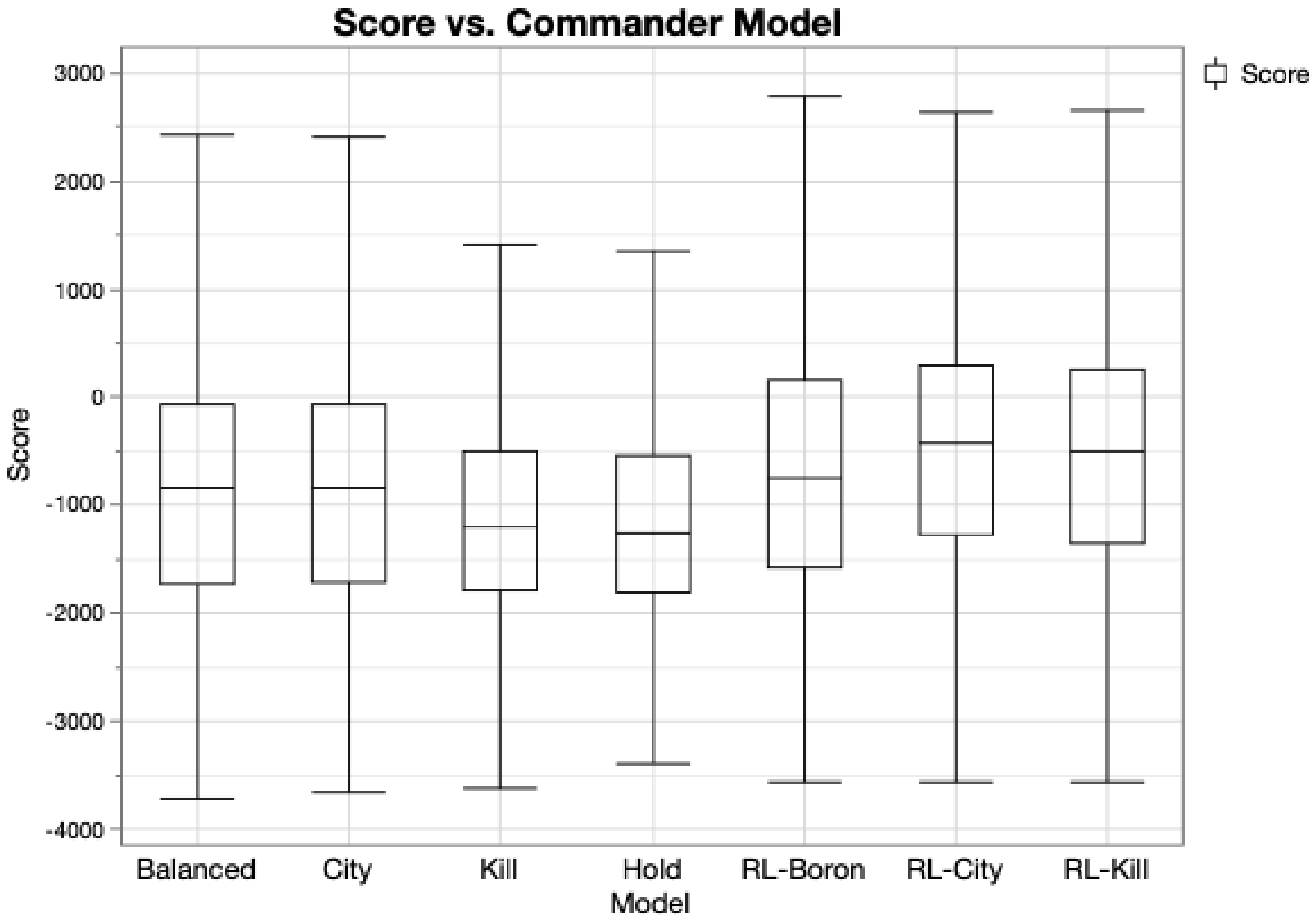

Figure 6.31. Box and Whisker Plot for Commander Individual Model Evaluation.

As with the Manager Score Prediction Model evaluation, we note increased MSEs and MAEs for our Commander Score Prediction Models. Likewise, a slight increase in errors can be expected as the scores produced in these evaluation scenarios are of a much wider range than those of our Operator and Manager scenarios. However, these errors also appear to come from observed significantly smaller $R^2$ values, indicating even more difficulties in predicting the final game score at the Commander level of abstraction.

### 6.4.5 Multi-Model Summary

We take each scripted and RL model described above and implement each into their respective Multi-Model. We end up with an Operator Multi-Model, a Manager Multi-Model, and a Commander Multi-Model. As a summary, the composition of each Multi-Model is depicted in Table 6.7

Table 6.7. Composition of Multi-Models

| Operator Multi-Model | Manager Multi-Model | Commander Multi-Model |
|---|---|---|
| Pass-Agg | Balanced | Balanced |
| Pass | Prioritized-City | City |
| Agg | Seize-Red-City | Kill |
| Burt-Plus | Killer | Hold |
| City | Hold | RL-Boron |
| Kill | RL-Boron | RL-City |
| Shootback | RL-Defeat-and-Capture | RL-Kill |
| RL-Boron | RL-Occupy-City | |
| RL-Scotty | RL-Seize-Red-City | |
| RL-Killer | RL-Defensive | |
| RL-City | RL-Kill | |
| | RL-Kamikaze | |

## 6.4.6 Hierarchical Reinforcement Learning Scenarios

For our final HRL experiment, we use randomly generated scenarios consisting of $20 \times 20$ gameboards. Each game initiates with a random number between one and three Commanders; each Commander created consists initially of three Managers, with each Manger consisting of three Operators. This results in a random number between nine and 27 units per faction being initiated per scenario. Each scenario is also initialized with either one or two urban hexagons randomly placed according to force ratio as described in Section 3.5.2. The number of phases in the game is set to 80. Three examples of scenario initial starting conditions are shown in Figure 6.32.

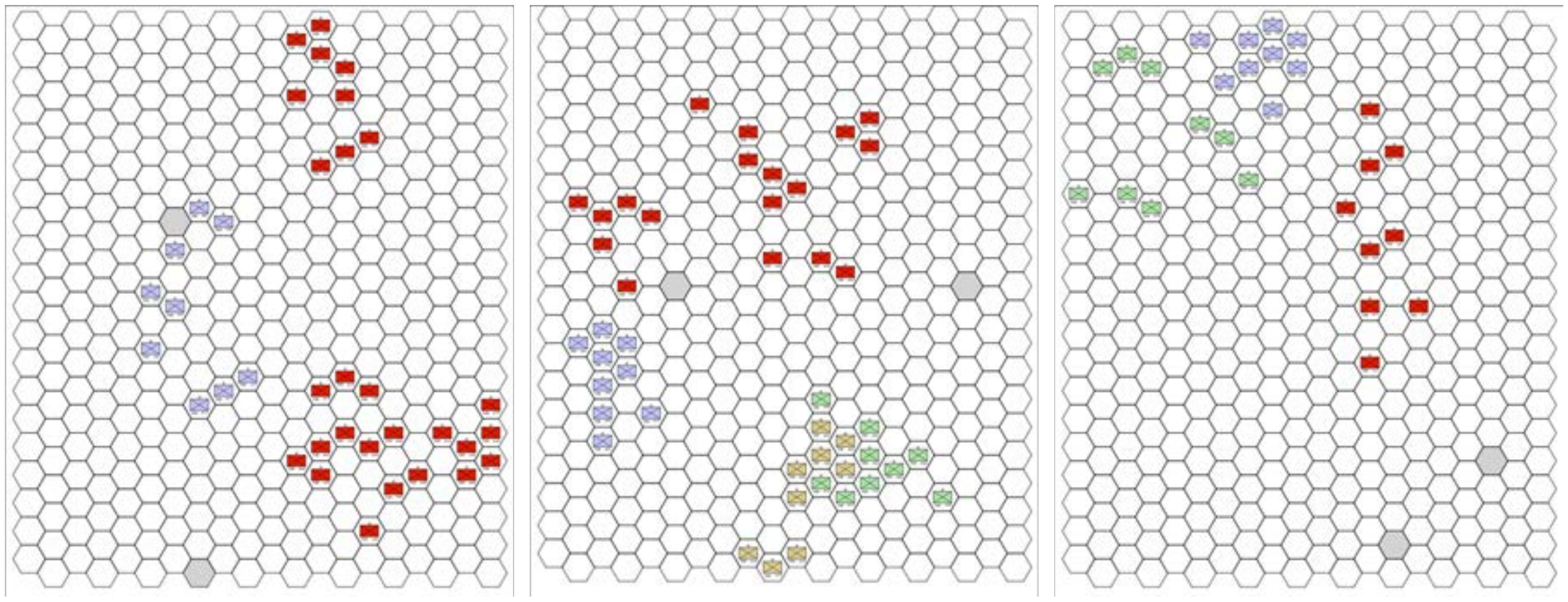

Figure 6.32. Example Evaluation Scenarios. Three examples of randomly generated scenarios. The blue faction is color-coded blue, green, and yellow to indicate they belong to separate Commanders. The red faction is color-coded red. Urban hexagons are color-coded grey.

## 6.4.7 Hierarchical Reinforcement Learning Model Evaluation

We run 10,000 evaluation games to assess the efficacy of our overall HRL framework consisting of multi-models at each level of the hierarchy. We use Pass-Agg as the adversary model and set `scenarioCycle` = 0. We then conduct a limited ablation study where we selectively investigate using different combinations of scripted, RL, and multi-model agents at each level of the hierarchy. Due to the combinatorial complexity of possible agent configurations and the computational resources required, we only focus on a subset of potentially influential variations. This approach allows us to efficiently explore the impact of key components while managing the practical constraints of this experiment.

## 6.5 Results and Discussion

The results of our evaluation are summarized in Table 6.8 and visualized in Figure 6.33. Table 6.8 depicts the mean scores, standard deviation, SEM, and 95% upper and lower confidence intervals for each model's mean score across the $10,000$ evaluation games. Figure 6.33 is the plot of scores depicting mean scores, box plots, and standard deviation. When multiple models are listed separated by a hyphen (e.g., **Balanced - Multi - Multi**), it is in the order of **[Commander Model] - [Manager Model] - [Operator Model]**. The term "Multi" is used for short as that hierarchy level's Multi-Model consisting of all of its embedded individual models. For example, the model **Balanced - Multi - PassAgg** means that, for their decision-making, the Commander was assigned the Scripted Balanced model, the Manager was assigned the Manager Multi-Model, and the Operator was assigned the Pass-Agg model.

Table 6.8. Final Experiment Results.

| Model | Mean | Std Dev | SEM | Lower 95% | Upper 95% |
|---|---|---|---|---|---|
| PassAgg | 94.473 | 1201.913 | 12.019 | 70.913 | 118.033 |
| Full HRL Model | -948.571 | 866.679 | 8.667 | -965.559 | -931.582 |
| Balanced - Multi - Multi | -915.580 | 894.076 | 8.941 | -933.106 | -898.054 |
| Balanced - Multi - PassAgg | -277.776 | 1168.952 | 11.690 | -300.690 | -254.863 |
| Balanced - Balanced - Multi | -920.844 | 859.318 | 8.593 | -937.689 | -904.000 |
| City - CityPri - Multi | -908.022 | 845.619 | 8.456 | -924.598 | -891.447 |
| Balanced - Balanced - PassAgg | -151.737 | 1178.598 | 11.786 | -174.840 | -128.634 |
| City - CityPri - City | -8.755 | 1205.852 | 12.059 | -32.392 | 14.881 |
| Balanced - CityPri - RLBoron | -1699.297 | 926.587 | 9.266 | -1717.460 | -1681.134 |
| RLBoron - RLBoron - RLBoron | -1482.423 | 836.484 | 8.365 | -1498.820 | -1466.027 |
| RLCity - RLOccupyCity - RLCity | -1669.867 | 823.036 | 8.230 | -1686.000 | -1653.734 |

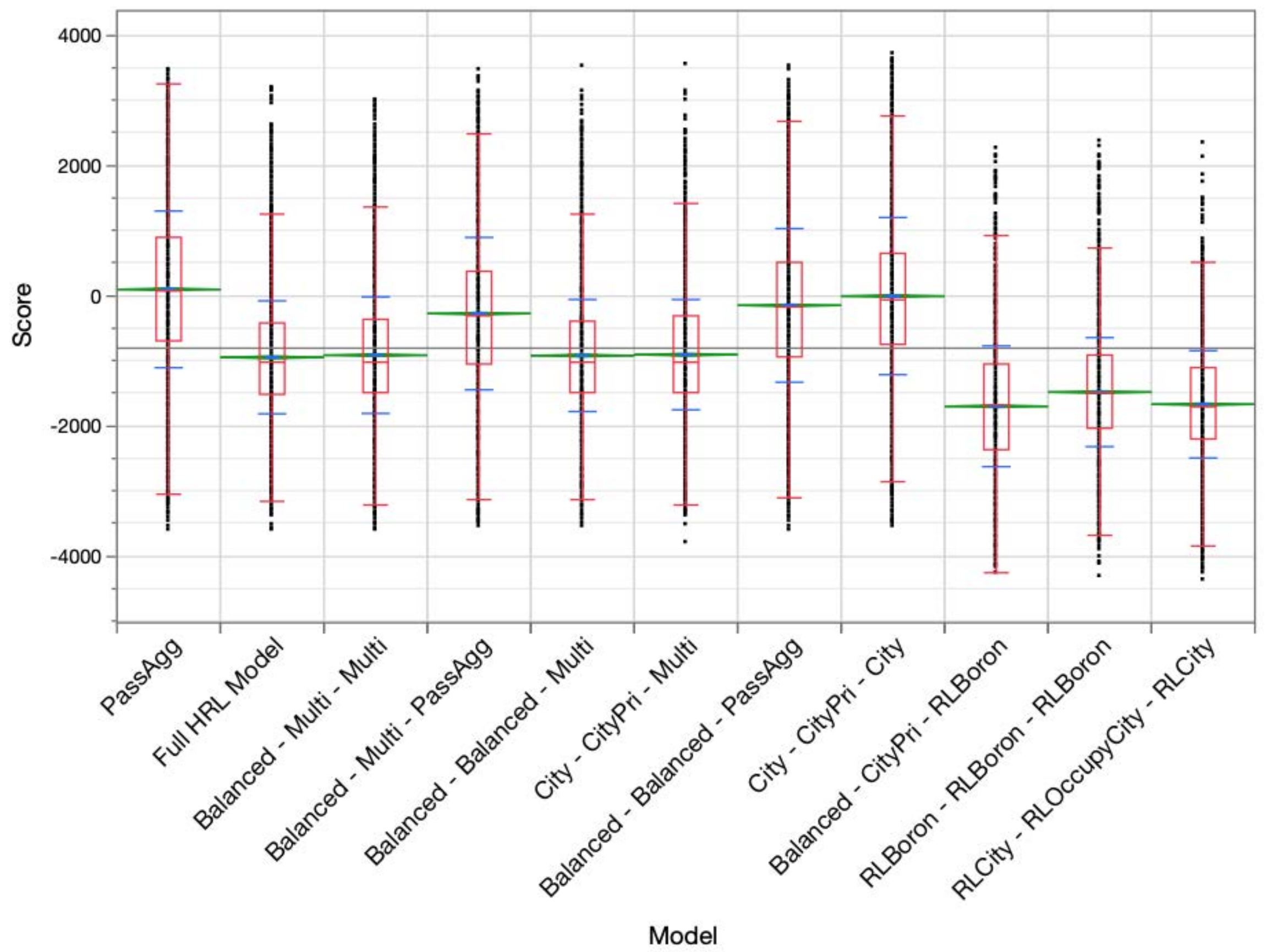

Figure 6.33. Comparison of Scores. Box plots are shown in red, mean scores are shown with the green line, and standard deviation is shown by the blue segments.

Because of the varied random scenarios in our evaluation runs, the notably high standard deviation values are expected due to the broad diversity of scenario initial starting conditions under which the models were tested. The scenario generator creates a wide random array of games, including very offensive and defensive setups, contributing to this variability. While a high standard deviation typically suggests a greater spread in data points around the mean, indicating inconsistent model performance across different scenarios, it should not be viewed in isolation as a measure of a model's reliability in this instance. Instead, combining the insights provided by both the standard deviation and the SEM offers a more comprehensive understanding of each model's consistency and variability. The SEM, in particular, helps in estimating the accuracy of the mean scores reported, providing a clearer picture of how the models might perform in general conditions, not just under the specific conditions of each individual game played.

Although the difference in performance is clear for our main comparison between our **Full HRL Model** and **Pass-Agg**, we set a $\alpha = 0.05$ and run a Tukey-Kramer HSD test to gain further insights into potential differences among the different model configurations. The results of this test is summarized by the Connecting Letters Report in Figure 6.34.

**Connecting Letters Report**

| Level | | | | | | | | Mean |
|---|---|---|---|---|---|---|---|---|
| Pass-Agg | A | | | | | | | 94.473 |
| City-CityPri-City | | B | | | | | | -8.756 |
| Balanced-Balanced-PassAgg | | | C | | | | | -151.738 |
| Balanced-Multi-PassAgg | | | | D | | | | -277.777 |
| City-CityPri-Multi | | | | | E | | | -908.023 |
| Balanced-Multi-Multi | | | | | E | | | -915.580 |
| Balanced-Balanced-Multi | | | | | E | | | -920.845 |
| Full HRL Model | | | | | E | | | -948.571 |
| RLBoron-RLBoron-RLBoron | | | | | | F | | -1482.423 |
| RLCity-RLOccupyCity-RLCity | | | | | | | G | -1669.867 |
| Balanced-CityPri-RLBoron | | | | | | | G | -1699.297 |

Levels not connected by same letter are significantly different.

Figure 6.34. Tukey-Kramer HSD Connecting Letters Report. Statistical differences in means are shown via the Connecting Letters Report.

Overall, **Pass-Agg** performed significantly higher than any of the other model combinations with a mean score of 94.473. All of the other models demonstrated negative mean scores, suggesting challenges under the tested conditions using the respective frameworks. **Balanced - CityPri - RLBoron** recorded the lowest mean score of -1699.29.

The best model utilizing three levels of hierarchy was **City - CityPri - City** with a mean score of -8.755, suggesting a strong strategic advantage in models that prioritize urban control, likely due to the defensive benefits and score advantages associated with urban hexagons. The next best model combination is **Balanced - Balanced - PassAgg** with a mean score of -151.737. This also helps reinforce the idea that taking the urban hexagon in the scenarios we presented might be the best strategy in that the Balanced Commander and the Balanced Manager both have logic that prioritizes unoccupied cities, followed by cities occupied by red, and only if these two conditions are not met does it prioritize attacking the red force. Lastly the next model that does not have substantially negative performance is **Balanced - Multi - PassAgg**, with a mean score of -277.776. This is the only model within the top three performing hierarchical models that utilizes the Multi-Model, possibly suggesting the Multi-Models efficacy under certain conditions.

Conversely, hierarchical models exclusively employing an RL model in at least one level of the hierarchy underperformed significantly. This could indicate potential limitations in the RL model's ability to generalize from its training environment to the diverse conditions encountered during testing. While we attempted to build training scenarios that would be representative of scenarios it would encounter in gameplay, we may not have accounted for enough variability in the environments or the complexity of the tasks. Additionally, the ability of hierarchical RL models to decompose the gameboard for lower-level agents may also introduce challenges in coordination and optimization across different levels of the hierarchy. This could be a reason for the Commander's individual behavior models negative scores in Section 6.4.4, all showing negative mean scores ranging from -391.112 to -1065.823.

The comparative analysis between scripted and RL models also revealed notable differences in performance stability and adaptability. Scripted models, such as **PassAgg** and **City - CityPri - City**, showed less variability in their performance across different scenarios, suggesting that these models are more predictable and possibly more reliable under known conditions. However, their lack of adaptability could be a limitation in environments that deviate significantly from their programmed assumptions.

These results also provide insights into the strategic implications of different hierarchical setups. Models that integrated Multi-Models (e.g., **Balanced - Multi - Multi**) did not perform as expected, which could suggest that the increased complexity of managing decisions across multiple levels may introduce inefficiencies or decision conflicts that degrade overall performance. This observation is particularly important for the design of hierarchical models, as it highlights the potential trade-offs between model complexity and overall model effectiveness. Lastly, the superior performance of simpler, more focused strategies (e.g., urban control) over more complex, generalized approaches suggests that in environments characterized by high uncertainty or complexity, strategies that simplify decision-making by focusing on key objectives might outperform those that attempt to optimize across a broader range of variables.

## 6.6 Conclusion

This chapter presents our development, implementation, and evaluation of a comprehensive HRL framework. Our study expanded on existing literature by implementing a novel approach that combined scripted and RL behavior models, all integrated across different hierarchical levels using our previously validated observation abstraction, multi-model framework, and hierarchical hybrid AI approaches.

The experiments conducted were designed to test the efficacy of different model configurations in a generalized manner. Notably, the model **City - CityPri - City**, which emphasized urban control, demonstrated the least negative impact across the scenarios tested. This underscores not just the importance of urban hexagons within the default Atlatl scoring mechanism, but also the importance of accurately modeling a combat environment and scoring system to ensure that the system aligns effectively with the desired behaviors.

We find that models incorporating RL, particularly **Balanced - CityPri - RLBoron**, tended to underperform, suggesting potential misalignments between the RL models' training and the testing environment or perhaps limitations in the current state of RL algorithms to handle the complexity and variability of more complex scenarios.

One of the critical insights from this research is the potential trade-off between system complexity and overall agent effectiveness. While hierarchical models may offer a nuanced approach to managing complex decision-making tasks, our findings suggest that increasing complexity does not always correlate with better performance. This could be due to the overhead of managing multiple decision layers or conflicts between the decisions made at different hierarchical levels.

Furthermore, this study indicated that simpler, more focused strategies could be more effective in environments marked by high uncertainty or complexity. This principle can be broadly applied to the development of intelligent models across various applications. The success of the combined models, which prioritized urban control, supports this observation.

By simplifying the decision-making process and focusing on a primary objective closely tied to success, these models may have enhanced their performance.

Nevertheless, we acknowledge that the presented HRL framework is notably intricate and complex, consisting of many distinct components. While we conducted thorough hyperparameter tuning for each individual component in previous chapters, we did not revisit hyperparameter tuning after integrating these elements within the overall HRL framework. This oversight could have resulted in suboptimal interactions between components that were individually optimized but not collectively harmonized. Consequently, the potential for suboptimal parameter settings across different levels of the hierarchy might have compounded these inefficiencies, adversely affecting the overall performance and efficacy of the framework. Addressing this issue in future studies will be crucial for enhancing the robustness and effectiveness of the integrated system.

In conclusion, while the HRL framework developed and tested in this study shows promise, the findings suggest that careful consideration must be given to the design and integration of different components and model types within a hierarchical system. Future work should focus on refining these models, particularly in enhancing the robustness and generalizability of RL components. Further research could also explore alternative configurations and training strategies that might mitigate the observed issues with complexity and possible inter-level conflicts. Ultimately, this study contributes to the broader field of AI and intelligent agent decision-making in complex environments by highlighting both the potential and the challenges of applying hierarchical learning models in these domains.

# CHAPTER 7:
# Conclusion

## 7.1 Introduction

This dissertation sought to build on existing research by employing machine learning to develop intelligent agent behaviors within combat simulations. It aimed to scale AI to manage more extensive and intricate scenarios within reasonable computational and training constraints. Building on prior research, our focus centered on a Hierarchical Reinforcement Learning (HRL) architecture and training framework specifically designed to handle the complexities inherent in larger combat simulation scenarios.

### 7.1.1 Research Questions

At the outset of our investigation, we posed the following research questions:

- Can an HRL approach enable agents to perform intelligently in large, complex scenarios?
- How do we develop an HRL architecture that can allow for scalability to larger scenarios than has previously been possible?
  - How can HRL be applied to enable scaling to complex state-action spaces that currently exceed our ability to compute with reasonable computing power available today via DOD HPCs?
  - How can we build a scalable HRL method that can easily grow with complexity?
  - How do we best train each level of the hierarchy in a way that enables scalability but still provides for performance efficacy?
  - How can we abstract the observation space in a dimension-invariant manner that can work with any size scenario without the need to re-train the agents?
  - How can we abstract the observation space to balance training efficiency and performance efficacy?

- How can we leverage a diverse collection of models, consisting of specialized models, generalized models, machine learning models, and scripted models, to produce agents that maintain high performance in diverse scenarios?

To answer these questions, we pursued four research areas intended to enable the scaling of AI both individually and as a collective. After investigating these new approaches, we validated each independently through experimentation and analysis and then integrated all four into our overall HRL architecture and training framework. We investigated the following research areas:

1. A localized observation abstraction that uses piecewise linear spatial decay to reduce the complexity of the observation space while improving agent efficacy.
2. A multi-model approach that employs a collection of unique and specialized AI models, designed to enhance performance by dynamically selecting the most effective model for the current state of the game.
3. A hybrid hierarchical AI framework that integrates RL models for high-level decision-making and scripted models for low-level decision-making, outperforming either approach alone.
4. An HRL approach that integrates each of these methodologies into a scalable, self-similar architecture and training framework.

## 7.2 Summary of Key Findings

Throughout this dissertation research, we developed and integrated several new approaches to tackle the challenges posed by the complexity and scale of combat simulations in support of wargaming. In this section, we summarize key findings from each of our chapters.

### 7.2.1 Chapter 1: Introduction

We motivated the urgent need to modernize our wargaming processes with advanced technologies, such as AI. Our analysis highlighted the proactive steps taken by our potential adversaries, who are already leveraging AI in their wargaming strategies. Drawing parallels from AI's triumphs in strategic games like chess, Go, and poker, we advocated for a "centaur" approach to wargaming. By uniting human expert intuition with AI intuition and computational power, we believe we can improve both the quality and the speed of our decision-making process, which may surpass what either humans or machines could achieve on their own.

### 7.2.2 Chapter 2: Background

We presented common approaches to developing intelligent agents in games and simulations and argued for the need for more advanced adaptive, learning-based methodologies that may be able to scale to more complexity. We discussed the strengths and limitations of traditional game tree search algorithms, game-theoretic strategies, and commonly employed scripted (or rule-based) methodologies when faced with the high-dimensional state spaces typical of combat simulations in support of wargaming. We noted that ML-based approaches, such as RL, could result in performance that exceeds these traditional methodologies while allowing us to scale to greater complexity. However, we also argued that further research is needed in this domain.

### 7.2.3 Chapter 3: Localized Observation Abstraction Using Piecewise Linear Spatial Decay

We introduced a novel approach to address the challenges posed by large and complex environments in RL through localized observation abstraction using piecewise linear spatial decay. This method significantly improved computational efficiency and agent efficacy by simplifying the observation space while retaining critical spatial details necessary for effective decision-making. This abstraction proved superior to the more traditional global observation approaches common in RL. By abstracting away irrelevant information and focusing on crucial environmental aspects, we enabled RL agents to train more efficiently, thus addressing one of the significant hurdles in applying RL to complex simulations in support of wargaming.

Our experiments demonstrated that the localized observation abstraction consistently outperformed traditional global observation methods, even in the smallest scenarios tested. Our experiments also highlighted the tradeoff based on scenario complexity (as defined by the size of the state space) between RL and scripted agents. Given a limited computational budget, we found that RL-trained agents were superior to scripted agents up until a scenario size of $6 \times 6$, after which the scripted agent performed better. This influenced our design choices in how we decomposed the gameboard for the lowest level of our HRL hierarchy.

### 7.2.4 Chapter 4: A Multi-Model Approach

We introduced a novel multi-model framework and validated the effectiveness of employing a diverse set of models, showing that a combination of scripted and RL models could dynamically adapt to changing game states and significantly enhance performance. The primary finding from this study was that the multi-model framework significantly outperformed both individual scripted behavior models and individual RL behavior models.

Our experiments clearly demonstrated that the integration of diverse AI methodologies—such as RL, supervised learning, and scripted behaviors—into a cohesive multi-model framework enhanced the overall effectiveness of our AI agents in the scenarios tested. This integration not only capitalized on the strengths of each model type but also mitigated their individual weaknesses. Specifically, the multi-model approach suggested an advantage in performance by dynamically selecting the optimal model for each decision step based on the current state of the game. This adaptability may be crucial for success in the unpredictable and complex decision-making environments typical of wargaming where there may exist validated models that must be used. We also showed that, within the context of this specific study, a multi-model consisting of more individual behavior models outperformed a multi-model consisting of fewer individual behavior models, even when these behavior models performed worse in the overall sense.

### 7.2.5 Chapter 5: A Hierarchical Hybrid Approach

We developed a hierarchical framework consisting of a high-level RL manager agent with a low-level scripted agent. The structure allowed RL to manage higher-level strategic decisions, thereby enhancing adaptability and long-term decision-making. In contrast, scripted agents handled lower-level, routine tasks, ensuring reliability and consistent behavior in well-defined scenarios. We demonstrated that such hierarchical integration of different AI methodologies could surpass the performance of either RL or scripted approaches when employed independently.

The use of RL in the manager level allowed the system to learn from the environment, adapting its policy based on the scenarios presented. This was a crucial advancement over traditional methods, providing a pathway to address the inherent limitations of scripted agents in novel or unexpected scenarios. The integration of RL at this level of decision-making enabled a more nuanced response to complex challenges, which is essential in military simulations where adaptability is as crucial as tactical precision.

Although a promising approach, we noted that we were not successful in training a manager agent to surpass a scripted agent using a `scenarioCycle` = 0. The experiment we ran in this particular study involved a `scenarioCycle` = 10, though we perform independent training runs and evaluations using different `scenarioSeeds` to introduce a certain level of variability and unpredictability into the training process without overwhelming the RL manager agent with infinite possibilities.

### 7.2.6 Chapter 6: Hierarchical Reinforcement Learning Framework

This chapter detailed the development and implementation of an HRL architecture and training framework that integrated the previously described and validated observation abstraction, multi-model framework, and hierarchical hybrid AI approach. This scalable, self-similar HRL architecture and training framework was aimed at decomposing large scenarios into manageable subproblems that align with military decision-making structures.

This overall HRL approach combined abstractions across three dimensions: space, actions, and time. Spatial abstraction was accomplished through different observation abstractions at each level of the hierarchy. Action and temporal abstractions were simultaneously accomplished through the employment of options or macro-actions that persisted through longer timescales than individual unit actions. These abstractions arguably facilitated more efficient exploration and decision-making. This could be particularly advantageous in settings where the rewards may be sparse and significant states or actions may be few.

Despite the comprehensiveness of our HRL architecture and training framework and the demonstrated success of each individual component, we did not find improved agent performance using the current proposed HRL framework. Several factors could contribute to this lack of improved performance. Although we validated and optimized the hyperparameters of individual components of the overall HRL architecture, we did re-optimize or tune these hyperparameters based on the integration of these different components. Integration issues may also arise because, although individual components performed well on their own, their combined operation might not be as effective due to unforeseen interactions that were not apparent in testing or evaluation. Additionally, the individual components might have been optimized for specific scenarios that may not generalize well across different types of scenarios that the HRL framework encountered.

The complexity of the HRL framework itself could also be a contributing factor, as complex systems often suffer from issues such as overfitting, where the model demonstrates strong performance on the training data but performs poorly on unseen or test data. Moreover, as systems scale in size and complexity, minor problems can become significant, affecting everything from stability to performance. The reward structure in the HRL setup might also not be effectively encouraging desired behaviors across different levels of the hierarchy, leading to suboptimal overall performance. Lastly, insufficient exploration in RL can lead to suboptimal policy formulation. If the agents do not explore the environment adequately due to constraints or poor policy initialization, they might not learn the optimal actions. Addressing these issues might involve revisiting the integration process, enhancing the generalization capabilities of the model, simplifying the system where possible, re-evaluating the scaling strategy, adjusting the reward structures, or promoting more effective exploration strategies.

## 7.3 Discussion

The overall findings of this dissertation suggest that the concept of simplifying the RL problem through the use of different levels of abstraction may be promising in the domain of combat simulations. This research lays a solid foundation for future exploration into more advanced abstraction techniques within RL, potentially opening doors to even greater computational efficiencies and enhanced agent performance. Furthermore, the insights gained from this study can be applied to other domains that involve complex problems and decision-making, such as autonomous vehicle navigation and real-time strategic decision systems.

This research underscores the need for more intricate and adaptable architectures and frameworks. As AI technologies evolve, their practical application necessitates not only advanced computational resources but also robust frameworks designed for continuous learning and adaptation to evolving problems and increasing complexities.

Most importantly, this study illuminates the complexities involved in integrating multiple AI models across various hierarchical decision-making levels. These complexities can introduce challenges that might counteract the efficiency and performance gains observed at the individual model levels. Therefore, integrating multiple models demands careful balancing to preserve the overall system effectiveness and prevent the dilution of the benefits achieved by individual components.

For example, while in Chapter 4 we demonstrated that a multi-model approach significantly outperformed any individual model approach when applied within a singular hierarchical level, this advantage was not replicated in Chapter 6 when employing the multi-model approach within a hierarchical framework. We posit several factors that may contribute to this discrepancy:

- Increased System Complexity: the addition of multiple hierarchical levels introduces greater complexity in the decision-making process. Each layer in the hierarchy potentially may dilute the effectiveness of the decisions made by the multi-model at other levels. For instance, a sub-optimal decision at the Commander level could result in an unusual state space for the Manager level, resulting in another sub-optimal decision that gets propagated down to the Operator level—ultimately leading to compounded errors and less coherent overall strategy execution.
- Inter-Level Dependency: in a hierarchical system, the outcomes at one level are heavily dependent on the inputs from an upper level. This interdependency can create situations where even if a multi-model at one level is performing optimally, its outputs might not be effectively utilized by the next level. This misalignment can be exacerbated in non-repeating, unpredictable scenarios where adaptability and immediate response to new conditions are crucial.
- Action Space Constraints: the simplification of the action and observation space at each level of the hierarchy, intended to make the RL problem more manageable, so as to be able to generalize to a `scenarioCycle` = 0, might have inadvertently restricted the flexibility or control needed by the multi-model to fully leverage its potential to adapt and select among diverse options. A narrower action space reduces the granularity of control, which can be particularly limiting in a complex environment where the nuances of different actions can have significant impacts.
- Training and Optimization Challenges: a hierarchical setup also imposes unique training challenges. Optimizing a multi-model within this framework requires not only tuning it to perform well under the specific conditions of its own level but also ensuring that it synchronizes effectively with other levels. This synchronization is difficult to achieve, especially without extensive cross-level tuning, which was not performed in this study.

- Scenario Unpredictability: creating a set of training scenarios for each level of the hierarchy in a way that is representative of the conditions the agent may encounter during execution in larger scenarios can be challenging. Although ML approaches are able to generalize to unseen conditions, they still require seeing similar conditions during training. It becomes crucial, therefore, to ensure a representative distribution of possible state spaces during the training of each level of the hierarchy.

Likewise, whereas in Chapter 5 we find significant advantages in using a hierarchical hybrid AI framework that integrated an RL manager with scripted agents, this advantage was also not replicated when adding an additional hierarchical layer (i.e., the Commander) in Chapter 6. We attribute this discrepancy to several potential factors:

- Exponential Growth of Complexity: the expansion from a $10 \times 10$ gameboard to a $20 \times 20$ gameboard significantly increases the complexity, not merely doubling the complexity but expanding it exponentially. This increase, combined with the use of a `scenarioCycle` = 0 (as opposed to a `scenarioCycle` = 10 in the experiment conducted in Chapter 5), greatly complicates the problem the RL agent is attempting to solve. These changes were aimed at testing the robustness and adaptability of our hierarchical framework under more dynamic and less predictable conditions. However, a more gradual approach to these expansions might have allowed for better adaptation and learning.
- Action Space Considerations: the broader action space of 49 (using a $7 \times 7$ grid) in the Chapter 5 experiment as opposed to the more restricted action space of nine (using a $3 \times 3$ grid) in the Chapter 6 experiment, likely restricted the RL manager's ability to fine-tune decisions. This reduction in granularity potentially limited the model's effectiveness in navigating the more complex environments introduced by the larger gameboard.
- Observation Space Considerations: the observation space of the Manager in the Chapter 5 experiment was $17 \times 7 \times 7$ as opposed to $17 \times 5 \times 5$ in the Chapter 6 experiment, likely impacting the model's ability to accurately perceive and react to the environment, reducing the fidelity with which the gameboard was represented and possibly affecting decision quality.

These findings call for an even more gradual approach to the development and deployment of hierarchical AI systems, emphasizing the need for systematic enhancements and careful consideration of the complexities introduced by expanded systems and environments.

## 7.4 Theoretical and Practical Implications

This dissertation contributes significantly to the field of AI by developing and critically analyzing hierarchical and multi-model frameworks within complex simulation environments. Although the overall HRL framework did not achieve the anticipated performance outcomes, the exploration of various abstraction methodologies and the integration of diverse AI models provide substantial theoretical and practical insights.

Theoretically, the application of HRL in this study sheds light on the complexities and challenges associated with scaling AI systems through complex hierarchical frameworks. By breaking down decision-making into hierarchical layers, this research has highlighted the difficulties in synchronizing and optimizing performance across different levels of the hierarchy, contributing to a nuanced understanding of HRL's limitations and potential. These findings underscore the importance of further theoretical exploration into HRL frameworks, particularly focusing on the integration and synchronization across hierarchical layers, and how these can impact overall system efficacy.

Moreover, this study's exploration of localized and coarse observation abstractions has highlighted theoretical implications regarding the processing of spatial data by AI systems. While promising, the practical application revealed that reducing the dimensionality of the input space could sometimes improve performance but, at other times, potentially obscure essential information, impacting the decision-making accuracy of RL agents depending on their level in the hierarchy. This insight is crucial for developing AI models that can dynamically balance abstraction granularity and performance, particularly in higher-fidelity simulation environments where abstraction methods will likely be needed.

On the practical front, the findings from the multi-model approach, despite not achieving superior outcomes in the hierarchical setup, have demonstrated the potential and challenges of integrating multiple AI models within complex decision-making frameworks. This has important implications for combat simulations and can be extended to other domains requiring robust decision-making under uncertainty, such as cybersecurity and emergency management. These sectors could benefit from the insights on managing complexity and ensuring the effective integration and utilization of diverse AI strategies within a unified system.

Lastly, this research highlights significant implications for the design and implementation of AI systems in training and operational planning. The ability to configure and test various AI frameworks within simulated environments provides valuable lessons for AI applications in strategic contexts, where adaptability and reliability are paramount. Moreover, the challenges encountered in achieving effective model integration and system scalability underscore the need for sophisticated development and testing environments to better prepare AI systems for real-world complexities.

## 7.5 Strengths and Limitations

This dissertation has advanced our understanding of HRL and its practical application in combat simulations, revealing both significant strengths and notable limitations. The research contributes new insights by exploring the integration of multiple AI models within a hierarchical framework, enhancing the broader AI and RL domains, particularly those involving complex, multi-level problem-solving environments.

The methodologies developed in this research, particularly the approach of localized observation abstraction using piecewise linear spatial decay, represent a significant step forward. This approach markedly enhances computational efficiency and improves model performance in more complex combat simulation scenarios than have previously been possible using RL in Atatl specifically. By focusing on localized observation, the system not only reduces the complexity of the observation space, but also increases the relevance and accuracy of the information provided. This method could lead to more robust and adaptable agent behaviors, as it tailors the learning process to specific, spatially significant aspects of the environment. Therefore, this approach could further enable RL systems to excel in a wider range of environments, thereby advancing the generalization capabilities of RL approaches.

Additionally, the multi-model framework introduced in this dissertation underscores another significant advancement by demonstrating that a diverse set of models, including scripted and RL models, can dynamically adapt to changing game states and significantly enhance performance. This integration effectively leverages the strengths and mitigates the weaknesses of each model type. Particularly noteworthy is the framework's ability to dynamically select the optimal model for each decision step based on the current game state, offering essential adaptability in the complex and unpredictable environments involved in wargaming. Furthermore, the potential for the use of individual validated models within this framework aids in enhancing explainability. By incorporating established models that are already well-understood and validated, the framework not only increases its effectiveness but also improves transparency and interpretability, which are crucial for practical applications and user trust in AI-driven decision-making.

While these strengths are substantial, transitioning to the limitations reveals the challenges that accompany the complexity of integrating multiple AI models within a hierarchical framework. Such integration led to performance inconsistencies, especially as the scale of the environment expanded and the action space was simplified. These complexities often resulted in integration challenges that were not fully overcome, marking a primary limitation in applying HRL in larger, more dynamic settings.

While the abstraction techniques used in higher levels of our hierarchy were intended to reduce computational demands and make for an easier RL problem, there was a potential risk of oversimplifying the environment. This could have led to the loss of critical information or further context, adversely affecting the decision-making quality of the models, particularly in scenarios where additional details may have been crucial. Furthermore, our experience in previous chapters highlighted the critical aspect of hyperparameter tuning in optimizing performance at the component level. However, we may have underestimated the significant impact that precise parameter tuning would have when integrating and stacking different models within a hierarchical framework. This potential oversight revealed that optimal performance in complex AI systems may be heavily dependent on finely tuned parameters, especially when various models interact in a layered structure.

Another notable limitation of our study was that, while we initially aimed to increase complexity all around, we focused on gameboard size and the associated unit counts rather than increasing the number of unit types or terrain types in the game. This decision was driven by the desire to analyze the effects of spatial dynamics on decision-making processes without introducing excessive variability from diverse unit capabilities and terrain interactions. However, by not diversifying unit and terrain types, we may have overlooked critical factors that affect strategic complexity and decision-making in more nuanced combat simulation scenarios.

In summary, while this research has highlighted several strengths in applying advanced AI methodologies to combat simulations, it also faces limitations that should be addressed in future research. These include handling the complexity of HRL systems, ensuring robust performance across varied scenarios, increasing the complexity of scenarios beyond gameboard size and unit counts, and refining abstraction techniques to ensure relevant environmental details are maintained. Future work should focus on overcoming these limitations to enhance the practicality and effectiveness of HRL frameworks in real-world applications.

## 7.6 Contributions

This dissertation contributes significantly to the field of AI as applied to combat modeling and simulation, with a particular focus on scaling HRL systems for complex scenarios. The specific contributions of this research include:

- Developing an HRL architecture and training framework that supported multiple self-similar hierarchical levels. This framework was designed to facilitate the training of agents across different levels of decision-making and policy implementation, thereby addressing the challenges posed by complex, multi-level operational environments.
- Introducing new methods for the hierarchical decomposition of agents, decisions, and policies. These methods were developed to manage the exponential complexity often encountered in large-scale simulations, enhancing the tractability and efficiency of AI systems.
- Creating new techniques for dimension-invariant observation abstractions that support the scalability of RL. Up until a certain level, these techniques were vital for improving and maintaining agent performance as the complexity of the scenario increased.
- Developing a new framework that combines different individual behavior models within a hierarchy of policies that improve agent efficacy. This method enabled the specialization of agents without compromising the overall ability of a multi-model to respond effectively in dynamic environments.
- Investigating the feasibility and effectiveness of HRL for training intelligent simulation agents in complex environments. This work demonstrated the strengths and lim-itations of hierarchical structures in enhancing the adaptability and decision-making capabilities of AI agents.

These contributions collectively advance our understanding and application of RL and HRL to improve intelligent agent behavior modeling in simulation, particularly within the domain of combat simulations—offering new perspectives, insights, and capabilities for dealing with complex and dynamic decision-making environments.

## 7.7 Conclusion

We revisit and answer our research questions posed in Chapter 1 based on the results and insights gained from our study:

- **Can an HRL approach enable agents to perform intelligently in large, complex scenarios?**
  - Our research has not conclusively demonstrated that an HRL approach can enable agents to perform intelligently in large and complex scenarios. While we observed some promising results in individual components of our overall HRL architecture, the overall performance did not meet or outperform our baseline scripted model.
  - These results suggest that while HRL frameworks have the potential to manage complex scenarios, there are significant challenges that need to be addressed to fully realize this potential. Key challenges in our study likely stemmed from integrating multiple hierarchical levels, potentially leading to suboptimal decision-making. Simplifying observation and action spaces to manage larger scales may have also stripped away crucial details and control, thereby limiting higher-level agents' responsiveness to dynamic conditions. These issues underscore the need for further research to refine HRL approaches, aiming for consistent high performance and adaptability in complex settings.
- **How do we develop an HRL architecture that can allow for scalability to larger scenarios than has previously been possible?**
  - Chapter 6 details our HRL architecture and training framework, while the preceding chapters detail specific components within this HRL architecture. We build our system to be self-similar in that extra levels of the hierarchy can be added without needing to re-engineer the entire architecture. For example, whereas we demonstrated three levels of the hierarchy for a $20 \times 20$-sized gameboard, if we were to be presented with a $40 \times 40$-sized gameboard, we could simply add another level to the hierarchy, which would simply decompose the space in the same manner as the Manager and Commander.

  – To build this overall HRL system, we also leveraged abstractions across space, actions, and time. We use spatial abstractions to simplify the agent's observation; we use action abstractions to decompose a complex problem into small subproblems; and we use a multi-model framework to reduce the complexity of training our agents while instead leveraging a diversity of individual agents.

- **How can HRL be applied to enable scaling to complex state-action spaces that currently exceed our ability to compute with reasonable computing power available today via DOD HPCs?**
  – First, we recommend implementing different forms of state and action abstraction techniques to help reduce the dimensionality of the state-action spaces, making the problem more manageable without losing critical information.
  – Second, we suggest developing modular HRL frameworks that facilitate independent training and optimization across different levels within the hierarchy.
  – Finally, we recommend designing a scalable HRL architecture and training framework that can divide the problem space into manageable components, each handled by different agents or subsystems within the hierarchy. If trained and tuned appropriately, these strategies collectively may enable the effective scaling of HRL to handle complex state-action spaces with the available computational resources.
- **How do we best train each level of the hierarchy in a way that enables scalability but still provides for performance efficacy?**
  – Our research highlights several key strategies to train each level of the hierarchy. Firstly, employing a modular training approach where each level of the hierarchy is optimized independently before integration can significantly enhance performance. This allows for focused refinement of each level's strategies without the compounded complexity of the entire system. Secondly, implementing adaptive learning rates and reward shaping specific to each level can improve training efficiency and effectiveness by tailoring the learning process to the unique challenges and roles of different hierarchical levels.
  - Once this is complete, however, it may be necessary to re-tune or fine-tune the training once all levels are initially trained independently. This re-tuning may involve adjusting parameters and strategies in response to the interactions between levels, ensuring that the overall system functions cohesively. By monitoring the system-wide effects of changes at individual levels, it may be possible to identify areas where adjustments are needed to optimize overall performance.
  - Additionally, although not explored in this study, the use of curriculum learning may prove valuable in scaling training efforts. By progressively introducing more complex scenarios, each level of the hierarchy gradually adapts to the increasing complexity, building competencies in a controlled manner. This not only may enhance the agents' ability to manage more challenging situations but may also ensure that foundational skills are well-established before moving on to more advanced tasks.
- **How can we abstract the observation space in a dimension-invariant manner that can work with any size scenario without the need to re-train the agents?**
  - We answer this question in Chapter 3 in detail.
  - We show the efficacy of localized observation abstractions as compared to the non-abstracted global observation space.
- **How can we leverage a diverse collection of models, consisting of specialized models, generalized models, machine learning models, and scripted models, to produce agents that maintain high performance in diverse scenarios?**
  - We answer this question in Chapter 4 in detail.
  - We show the efficacy of this multi-model approach as compared to individual models when used alone.

In conclusion, this dissertation has advanced the application of AI in combat modeling and simulation in support of wargaming by developing multiple different methodologies that we have shown to scale RL to larger scenarios than have been previously possible using the Altatl simulation environment. By leveraging a hierarchical approach, we have not only addressed specific technical challenges but also opened new avenues for the application of AI in areas requiring nuanced, multi-faceted decision-making capabilities. Moreover, while our overall HRL framework did not meet expected performance benchmarks, it highlighted important limitations and challenges. These challenges include the integration complexity across multiple hierarchical layers and the scalability issues when expanding to larger, more dynamic scenarios. Acknowledging these difficulties has provided valuable insights into the critical areas needing further research and development. This recognition is crucial as it forms a basis for future work aimed at overcoming these hurdles, thus enhancing the effective deployment of AI within the domain of combat simulations in support of wargaming. The potential for this research to influence future developments in defense strategies, particularly in refining and deploying advanced AI systems, is substantial and promising.

## 7.8 Future Work

The exploration into HRL within complex combat simulations has opened several areas for further research and development. While this dissertation has laid a solid foundation, the full potential of HRL systems has not yet been fully realized, necessitating future studies focused on enhancing their robustness and reliability across various operational contexts.

One critical area for future work involves the hyperparameter tuning of the integrated HRL system. Our findings suggest that while individual components of the HRL framework performed adequately in the same environment for which they were trained, the overall system integration did not achieve the expected outcomes. This discrepancy underscores the need for comprehensive hyperparameter tuning post-integration to ensure that the combined system functions optimally. Future research should focus on developing methodologies for systematic tuning that consider the interactions between different hierarchical layers and their cumulative impact on system performance.

Another significant aspect of future work is the optimization of model integration. The current research highlighted the challenges of maintaining consistent performance when different AI models are integrated within a hierarchical framework. Future studies should investigate advanced techniques for model integration, ensuring that the system can seamlessly adapt and respond to diverse scenarios without performance degradation. This may involve enhancing training techniques, aligning reward systems more effectively, or implementing curriculum or adaptive learning strategies and objective functions that can dynamically adjust to the changing scenario context.

In addition to technical enhancements, there is a clear need to develop more sophisticated methods for observation and action abstraction. While the current approaches have provided a good starting point, they potentially discard critical details necessary for nuanced decision-making. Future research should focus on creating dimension-invariant abstraction techniques that ensure the preservation of essential granularity of information based on the level of the hierarchy or the task to be performed, thereby effectively supporting both tactical and strategic decision levels.

Finally, implementing some form of cross-level feedback mechanisms could provide a method for real-time adjustment and optimization of the hierarchy during training. Such mechanisms could allow higher levels of the hierarchy to offer corrective guidance or modified rewards based on the performance feedback from lower levels, thereby ensuring that the actions across all layers are well-aligned with the overall strategic objectives.

By addressing these areas, future research can significantly advance the field of AI as applied to developing intelligent agent behaviors, particularly in the context of complex and dynamic combat scenarios. This work will not only refine the theoretical underpinnings of HRL but also enhance its practical applications, ultimately driving forward the integration of AI into strategic, operational, and tactical real-world decision-making processes.

THIS PAGE INTENTIONALLY LEFT BLANK

# List of References

THIS PAGE INTENTIONALLY LEFT BLANK

# Initial Distribution List

1. Defense Technical Information Center
   Fort Belvoir, Virginia

2. Dudley Knox Library
   Naval Postgraduate School
   Monterey, California